\documentclass{article} %
\usepackage{iclr2025_conference,times}

\usepackage{amsmath,amsfonts,bm}

\def\eqref#1{equation~\ref{#1}}

\def\1{\bm{1}}

\DeclareMathAlphabet{\mathsfit}{\encodingdefault}{\sfdefault}{m}{sl}
\SetMathAlphabet{\mathsfit}{bold}{\encodingdefault}{\sfdefault}{bx}{n}

\usepackage{hyperref}
\usepackage{url}
\usepackage{booktabs}       %
\usepackage{amsfonts}       %
\usepackage{amsmath}
\usepackage{xcolor}         %
\usepackage{natbib}
\usepackage{sidecap}
\usepackage{graphicx}
\usepackage{caption}
\usepackage{ulem}
\usepackage{tabularx}

\usepackage{amsmath}
\usepackage{comment}
\usepackage{multirow,bigdelim}
\usepackage{lipsum}
\usepackage{array}
\usepackage{wrapfig}

\usepackage{adjustbox}
\usepackage[percent]{overpic}
\usepackage{makecell}
\usepackage{xcolor}
\usepackage{soul}
\usepackage{cleveref}
\usepackage{amsfonts}

\makeatletter
\newcommand{\settitle}{\@maketitle}
\makeatother

\newcolumntype{C}[1]{>{\centering\let\newline\\\arraybackslash\hspace{0pt}}m{#1}}

\newif\ifdraft
\drafttrue

\ifdraft
\definecolor{darkpink}{rgb}{0.561, 0.282, 0.427}

\newcommand{\dcc}[1]{{\color{red}[\textbf{DC:} #1]}}
\newcommand{\rgc}[1]{{\color{purple}[\textbf{RG:} #1]}}
\newcommand{\opc}[1]{{\color{blue}[\textbf{OP:} #1]}}

\newcommand{\abc}[1]{{\color{green}[\textbf{AB:} #1]}}

\newcommand{\drop}[1]{}

\else
\newcommand{\dcc}[1]{}
\newcommand{\rgc}[1]{}
\newcommand{\opc}[1]{}
\newcommand{\gcc}[1]{}
\newcommand{\hmc}[1]{}
\newcommand{\abc}[1]{}

\fi

\newcommand{\ourmethod}{ComfyGen}

\def\eg{\textit{e.g.}}

\makeatother

\usepackage[bottom]{footmisc}
\usepackage{arydshln}

\makeatletter
\def\blfootnote{\xdef\@thefnmark{}\@footnotetext}
\makeatother

\title{ComfyGen: Prompt-Adaptive Workflows for Text-to-Image Generation}

\author{Rinon Gal \\
NVIDIA, Tel Aviv University \\
\And
Adi Haviv \\
Tel Aviv University \\
\And
Yuval Alaluf \\
Tel Aviv University \\
\And
Amit H. Bermano \\
Tel Aviv University \\
\And
\hspace{58pt} Daniel Cohen-Or \\
\hspace{58pt} Tel Aviv University \\
\And 
\hspace{19pt} Gal Chechik \\
\hspace{19pt} NVIDIA \\
}

\iclrfinalcopy
\begin{document}

\maketitle

\begin{figure}[!h]
    \vspace{-25pt}
    \centering
    \includegraphics[width=\linewidth,trim={0 35pt 0 35pt},clip]{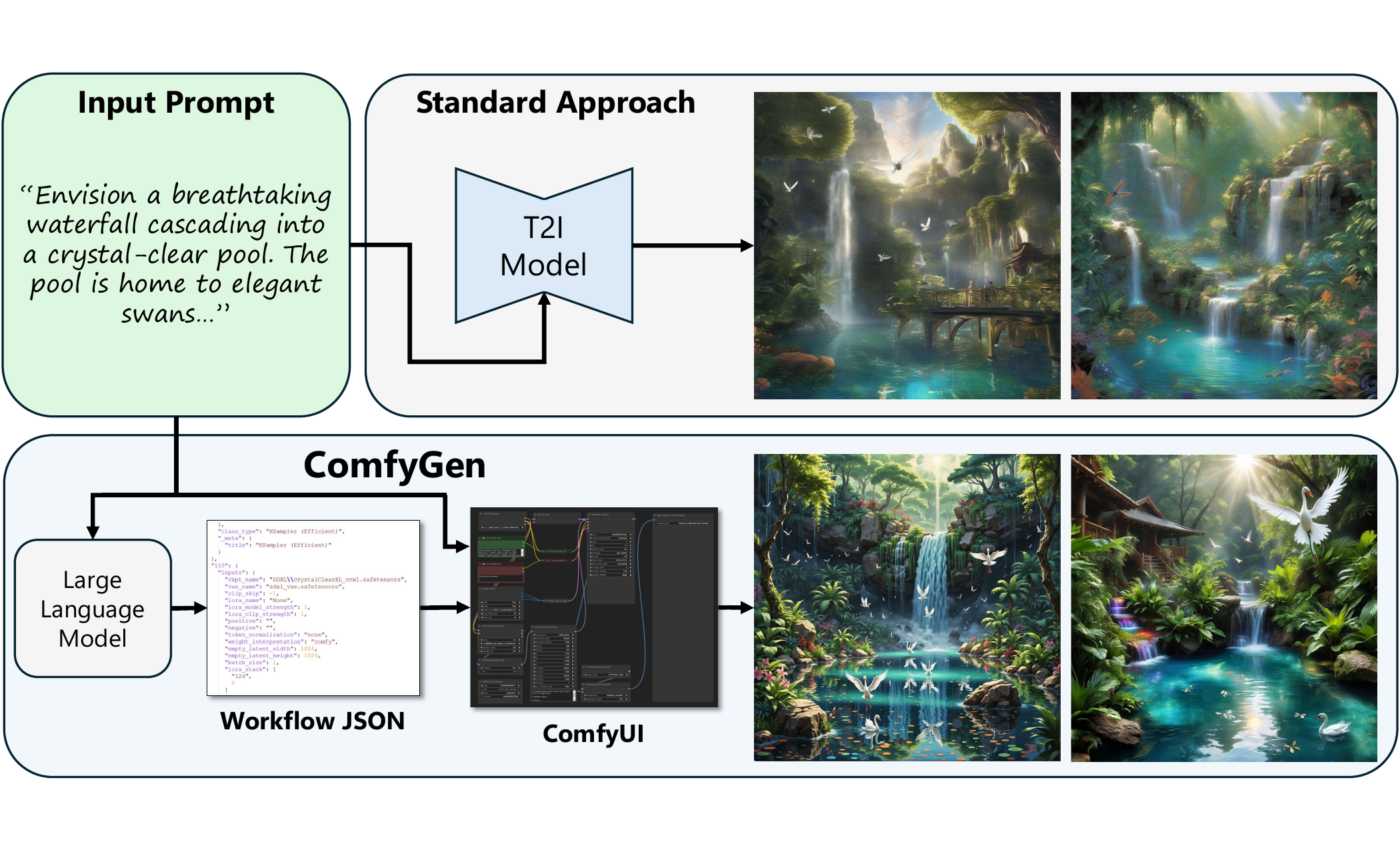}
    \caption{The standard text-to-image generation flow (top) employs a single monolithic model to transform a prompt into an image. 
    However, the user community often relies on complex multi-model workflows, hand-crafted by expert users for different scenarios. 
    We leverage an LLM to automatically synthesize such workflows, conditioned on the user's prompt (bottom). 
    By choosing components that better match the prompt, the LLM improves the quality of the generated image.
    }
    \label{fig:teaser}
\end{figure}

\begin{abstract}
\vspace{-6pt}
The practical use of text-to-image generation has evolved from simple, monolithic models to complex workflows that combine multiple specialized components. While workflow-based approaches can lead to improved image quality, crafting effective workflows requires significant expertise, owing to the large number of available components, their complex inter-dependence, and their dependence on the generation prompt. 
Here, we introduce the novel task of \textit{prompt-adaptive workflow generation}, where the goal is to automatically tailor a workflow to each user prompt. 
We propose two LLM-based approaches to tackle this task: a tuning-based method that learns from user-preference data, and a training-free method that uses the LLM to select existing flows. Both approaches lead to improved image quality when compared to monolithic models or generic, prompt-independent workflows. Our work shows that prompt-dependent flow prediction offers a new pathway to improving text-to-image generation quality, complementing existing research directions in the field.

\end{abstract}

\vspace{-7pt}
\section{Introduction}\label{sec:intro}

As the field of text-to-image generation~\citep{rombach2021highresolutionLDM,ramesh2021zero} matures, researchers and practitioners shift from simple, monolithic workflows to more complex ones. Instead of relying on a single model to produce an image, those advanced workflows combine a variety of components, or blocks, designed to enhance the quality of the generated image~\citep{AUTOMATIC1111WebUI,Zhang2023focus,comfyUI}.
These components may include fine-tuned versions of the generative model, large language models (LLMs) for refining the input prompt, LoRAs~\citep{luo2023lcmlora,simoLoRA2023} trained to correct poorly generated hands or to introduce specific artistic styles, improved latent decoders for creating finer details, super resolution blocks, and more.

Importantly, effective workflows are prompt-dependent. The choice of blocks often depending on the text prompt and the content of the image being created.
For example, a workflow aimed at emulating nature photographs may elect to use a model fine-tuned for photorealism, while workflows focused on generating human images often contain the term ``bad anatomy'' as a negative prompt or leverage specific super-resolution models that also correct distorted facial features, such as the eyes.
Due to the richness of available blocks and complexity of workflows, building a well-designed workflow often requires considerable expertise.

In this work, we propose to \textit{learn} how to build text-to-image generation workflows, conditioned on a user prompt. Specifically, we propose to leverage an LLM to take as input a prompt describing an image, and output a workflow that is specifically tailored to that prompt. Below, we outline two approaches to achieving this goal.
The prompt-specific workflow can then be used to synthesize images for that prompt, resulting in improved quality compared to using fixed base models or popular human-crafted workflows. Importantly, using an LLM enables the model to leverage its extensive prior knowledge to parse the prompt and match its content to the most appropriate blocks.

To represent flows, we build on ComfyUI~\citep{comfyUI}, a widely used tool that stores workflows as JSON files, which can be easily parsed by recent LLMs. The popularity of ComfyUI also provides access to multiple human-created workflows, which we then augment to create a more diverse training set. 
To teach the LLM the link between flow components and image quality, we collect $500$ diverse prompts from human users.\footnote{Sampled from \url{https://civitai.com/} after filtering out NSFW content.} These prompts are used for generating images using each workflow in our training set, and the results are scored by an ensemble of aesthetic predictors and human preference estimators~\citep{kirstain2023pickapic,xu2024imagereward,wu2023human}. This process effectively creates a training set composed of triplets of (prompt, flow, score). 

We then consider two approaches for matching flows to novel prompts. In the first, we leverage a closed-source LLM, and provide it with a table of flows and their scores across a closed-set of categories automatically derived from our training prompts. This table serves as a context for a followup request, where we ask the LLM to select the flow that is most suitable for a novel prompt. In the second approach, we fine-tune an open LLM~\citep{dubey2024llama} so that, given a prompt and an ensemble score, it predicts the flow that achieved that score. During inference, we provide the LLM with an unseen prompt and a target score and ask it to provide us with an appropriate workflow. We name these approaches \ourmethod{}-IC and \ourmethod{}-FT respectively. The design choices behind each approach and their motivations are discussed below. 

We compare our prompt-adaptive approach against several baselines, including: (1) single-model approaches (the baseline SDXL model~\citep{podell2024sdxl}, popular fine-tunes, and a DPO-optimized version~\citep{rafailov2024direct,wallace2024diffusion}), and (2) prompt-independent, popular workflows.
\ourmethod{} outperforms all baselines on both human-preference and prompt-alignment benchmarks, highlighting the benefit of prompt-dependent flows.

Finally, we analyze the workflows selected by our method, demonstrate their relation to the domains represented in the input prompts, and investigate the scaling behaviors of our model.

\vspace{-3pt}
\section{Related work}\label{sec:related}
\vspace{-3pt}

\paragraph{Improving Text-to-image generation quality.}
With the growing popularity of text-to-image diffusion models~\citep{rombach2021highresolutionLDM,nichol2021glide,ramesh2022hierarchical}, a range of works sought to improve the visual quality of their outputs, and their alignment to human preferences.

One approach is to fine-tune pretrained models using curated, high quality datasets or improved captioning techniques~\citep{dai2023emu,betker2023improving,segalis2023picture}. Instead of collecting data, a range of works use reward models~\citep{kirstain2023pickapic,wu2023human,xu2024imagereward,lee2023aligning} to guide text-to-image generation. This can be done using reinforcement-learning~\citep{black2024training,deng2024prdp,fan2024reinforcement,zhang2024large}. However, these methods can be computationally expensive and struggle to generalize effectively. As an alternative, the model can be fine-tuned using differentiable rewards~\citep{clark2024directly,prabhudesai2023aligning,wallace2024diffusion}. Instead of tuning the model directly, one can also use reward models to explore the diffusion input-noise space~\citep{eyring2024reno,qi2024not}, finding seeds for which the output is of higher quality. Finally, some approaches leverage self-guidance~\citep{hong2023improving} or frequency-based feature manipulations~\citep{si2024freeu,luo2024freeenhance} to drive the model towards more detailed and sharper outputs.

Our work proposes a new, orthogonal path to improving image quality. Instead of modifying the diffusion model or intervening in its sampling process, we use reward models to better match workflow components to a given prompt, aligning the entire pipeline towards human preferences. 

\paragraph{LLM-based tool selection and Agents}
Recent advancements in large language models have demonstrated significant improvements in reasoning abilities and their capacity to adapt to novel content and tasks. 
This adaptability can be achieved through efficient fine-tuning methods, but more commonly simply through zero-shot prompting or in-context learning. 

Building on these capabilities, a range of works proposed to leverage LLMs for tasks beyond text generation. A common line of work aims to equip the LLM with external tools \citep{schick2024toolformer}, either through appropriate API tags within the generated text~\citep{schick2024toolformer}, by providing in-context API documentations~\citep{wang2024voyager,suris2023vipergpt}, model descriptions~\citep{shen2024hugginggpt} and code samples~\citep{gupta2023visual}, or by retrieving models from a pre-defined collection.~\citep{wu2023visual}. Such tools are often referred to as LLM agents, and their latest variants are often equipped with components such as memory mechanisms, retrieval modules or self-reflection and reasoning steps, all aimed at improving their overall performance.

Our work can similarly be viewed as an agent, as it employs an LLM to directly select and connect external tools. Here, we focus on the novel task of prompt-adaptive pipeline creation, and on tapping this under-explored path to improving the quality of downstream generations.

\paragraph{Worfklow generation}
An emerging trend in machine learning research is the use of compound systems, where multiple models are used in collaboration to achieve state-of-the-art results. These systems have been successfully used across various domains, ranging from coding competitions~\citep{alpha2024alphacode2} to olympiad-level problem solving~\citep{trinh2024solving}, medical reasoning~\citep{nori2023can} and video generation~\citep{yuan2024mora}. 
Crafting such compound systems can be a daunting task, as the components must be carefully selected and their parameters tuned to perform well in tandem, 
rather than optimized on each individual step of the task.
To address this, recent approaches have proposed optimization-based frameworks that tune pipeline parameters for improved end-to-end performance~\citep{khattab2023dspy}, or even optimize the connections within a graph representing the components of a complex system~\citep{zhuge2024gptswarm}.

Our work similarly tackles the task of pipeline generation. Here, we focus on text-to-image models, and demonstrate that their performance can be enhanced by designing compound pipelines that depend on the user's prompt.

\vspace{-3pt}
\section{Method}\label{sec:method}
\vspace{-3pt}

Given an input prompt describing an image, our goal is to match it with an appropriate text-to-image \textit{workflow}, leading to improved visual quality and prompt alignment. We hypothesize that effective workflows will depend on the specific topics present in the prompt. Therefore, we propose to tackle this task by leveraging an LLM that can concurrently reason over the prompt and identify these topics, while also serving as a means to directly select or synthesize the new flow. 

In the following section, we provide a detailed description of ComfyUI and our method, focusing on our training data, as well as our retrieval-based and score-based tuning approaches.

\begin{figure*}
    \vspace{-6pt}
    \centering
    \setlength{\belowcaptionskip}{-6pt}
    {\scriptsize

    \begin{tabular}{@{}m{0.23\textwidth}@{\hspace{0.02\textwidth}}m{0.322\textwidth}@{\hspace{0.02\textwidth}}m{0.368\textwidth}@{}}
        \includegraphics[width=\linewidth]{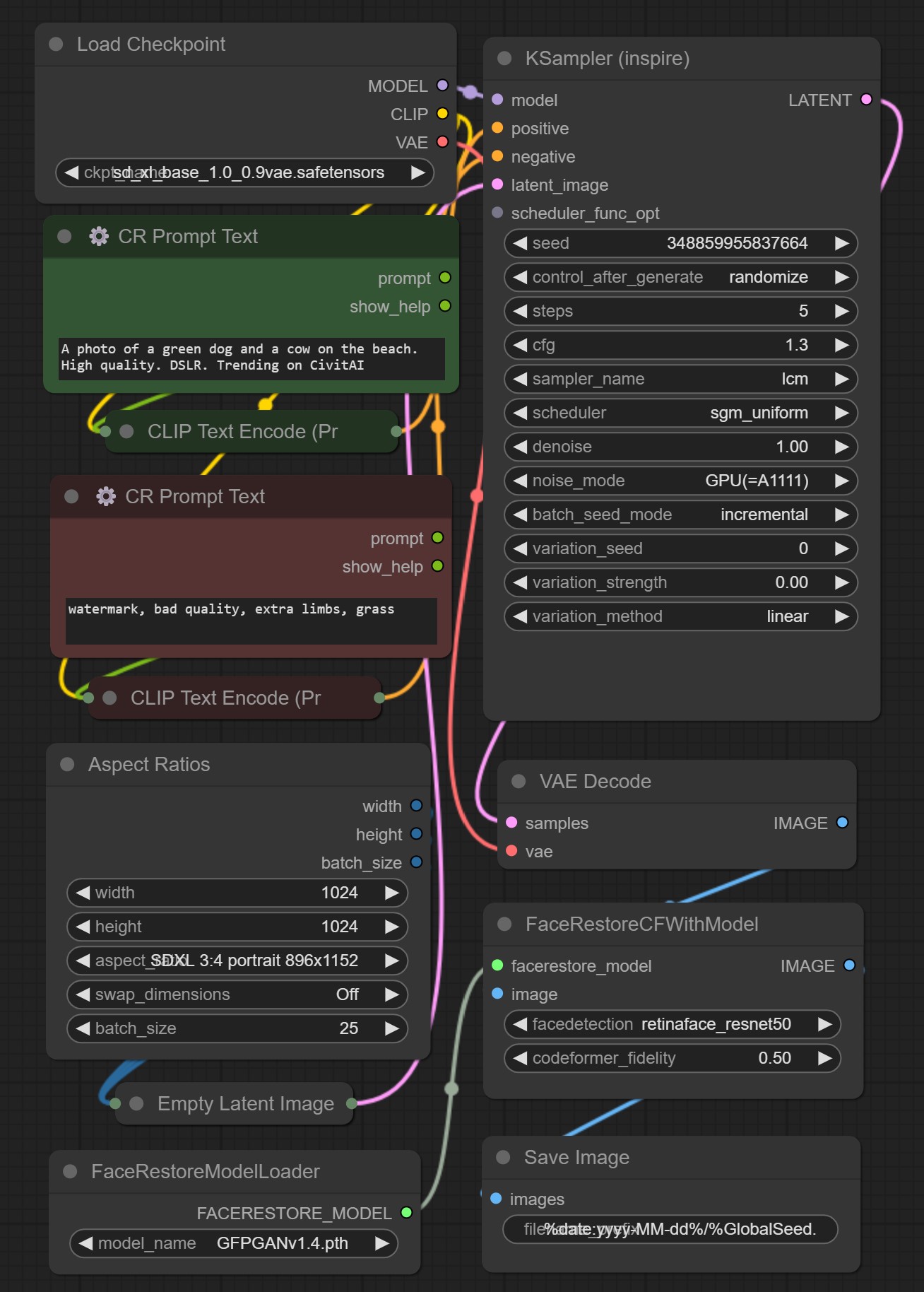} &
        \includegraphics[width=0.96\linewidth]{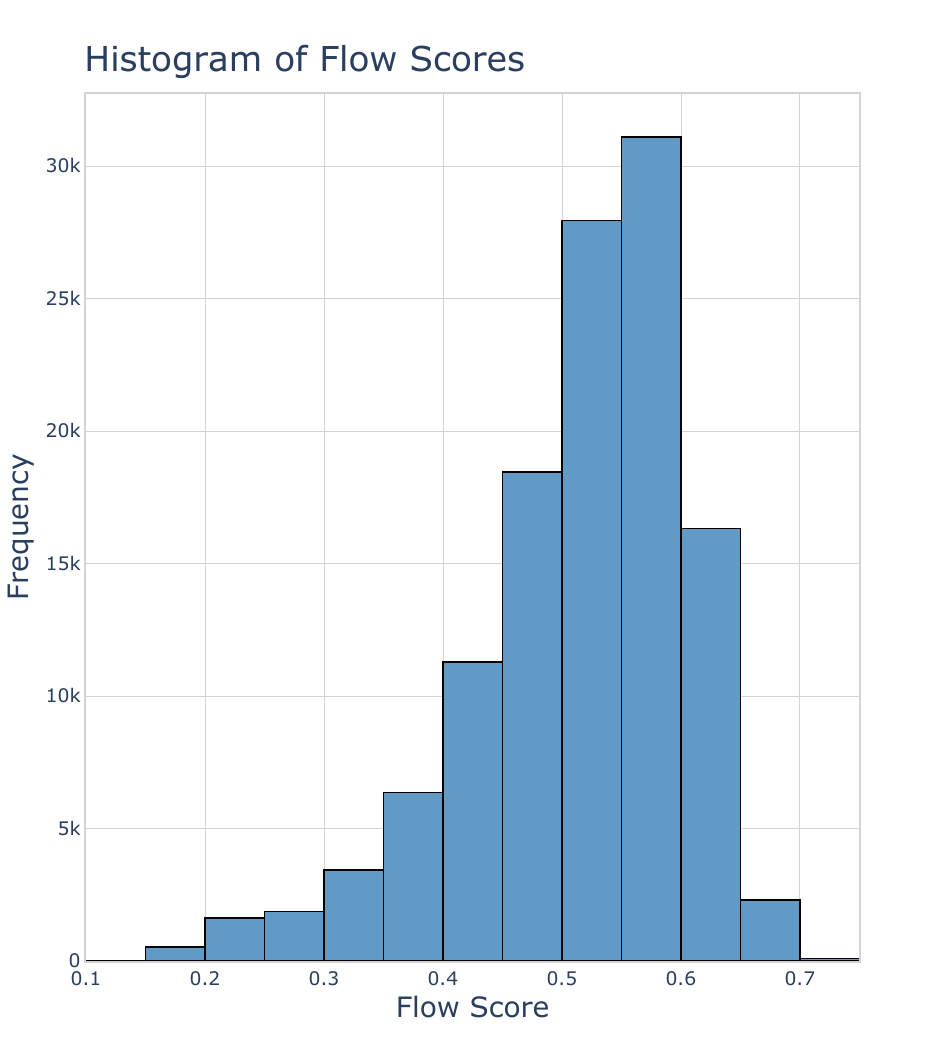} &
        \begin{tabular}{@{}cc@{}}

            \multicolumn{2}{c}{\begin{tabular}{@{}c@{}} \tiny ``a boat in the middle of the cherry blossoms, lush \\ \tiny  scenery, fairycore, cute and dreamy, pink and green..." \end{tabular}  }

            \\
            \includegraphics[width=0.4\linewidth,height=0.4\linewidth]{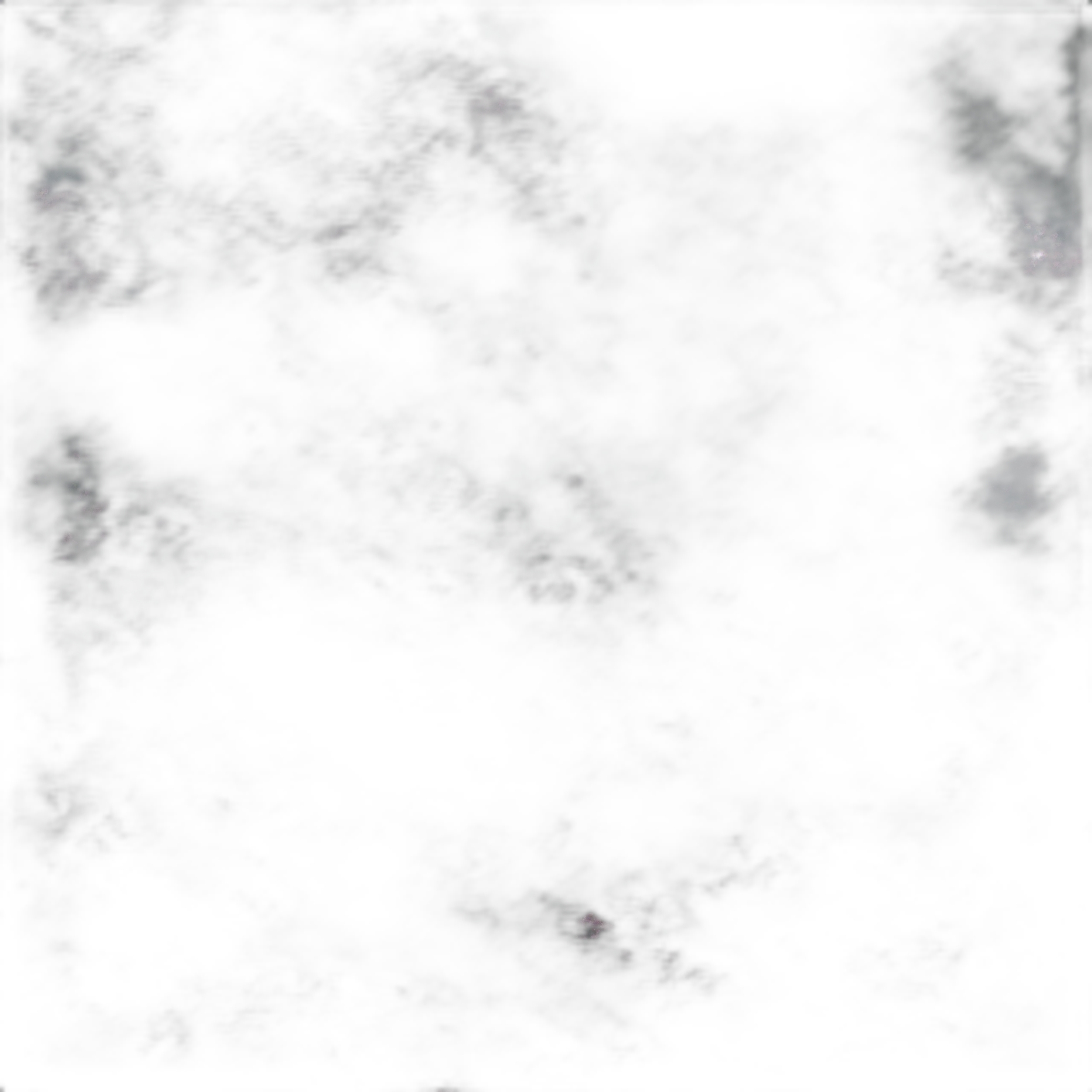} &
            \includegraphics[width=0.4\linewidth,height=0.4\linewidth]{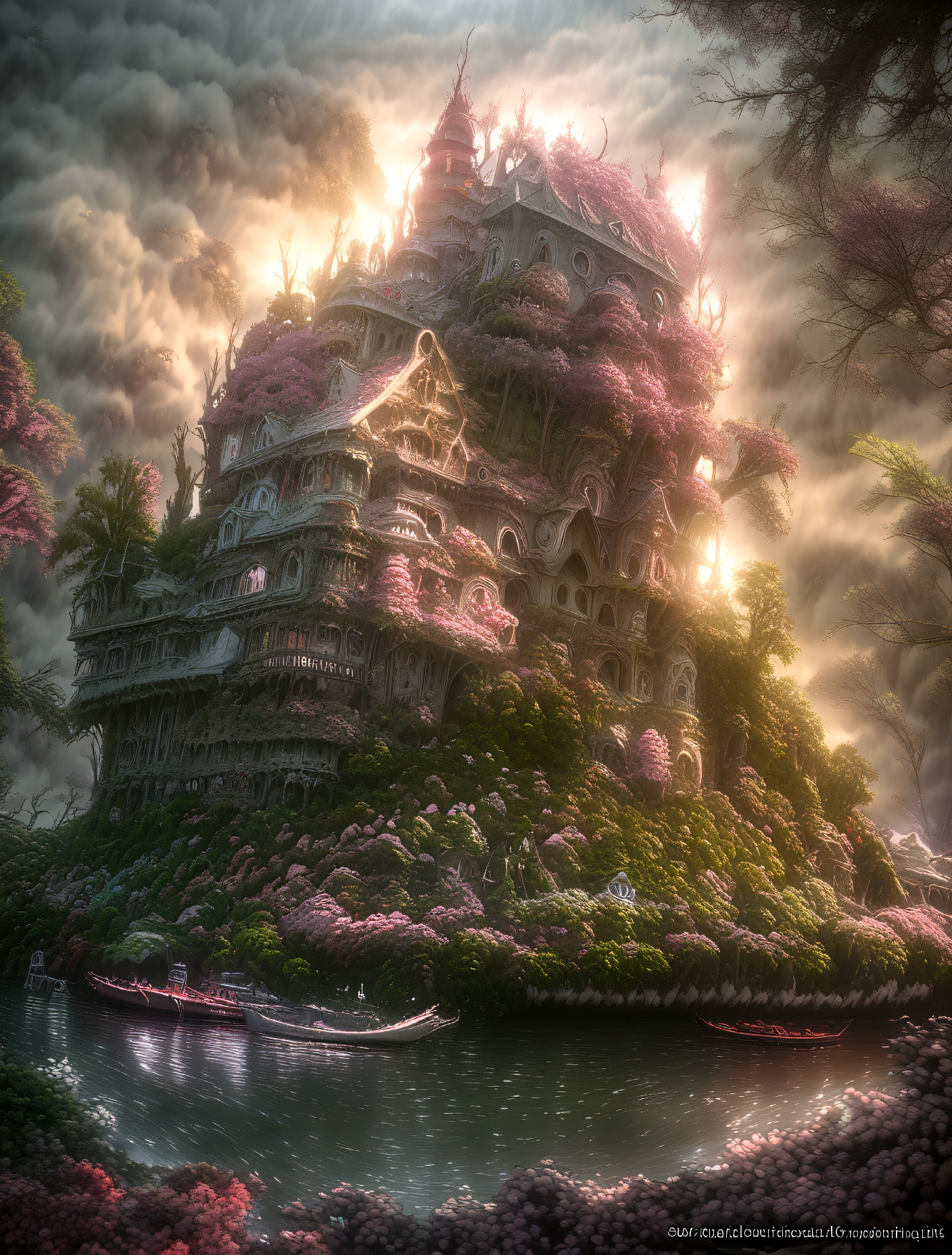} \\
            Score: 0.14 & Score: 0.42 \\
            \includegraphics[width=0.4\linewidth,height=0.4\linewidth]{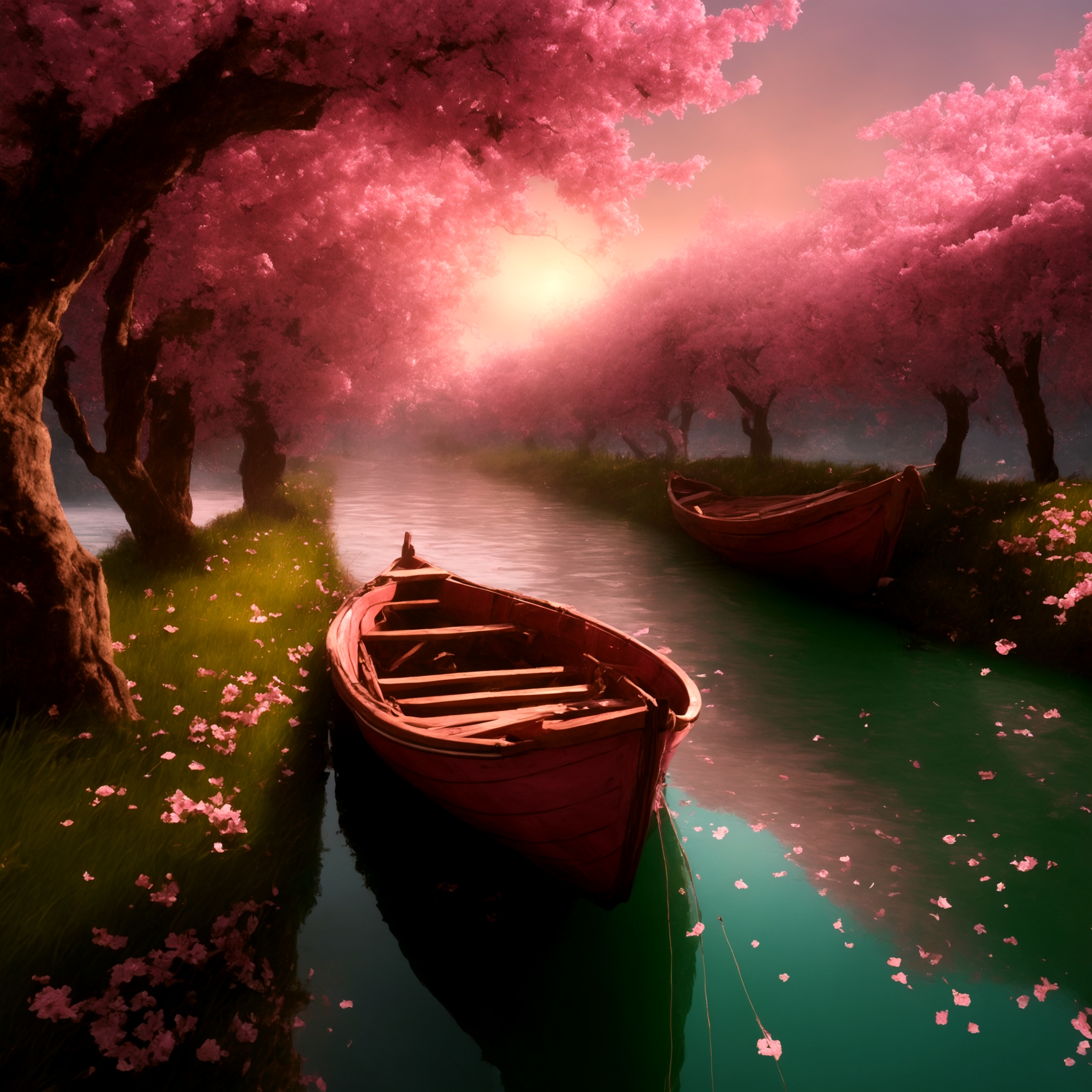} &
            \includegraphics[width=0.4\linewidth,height=0.4\linewidth]{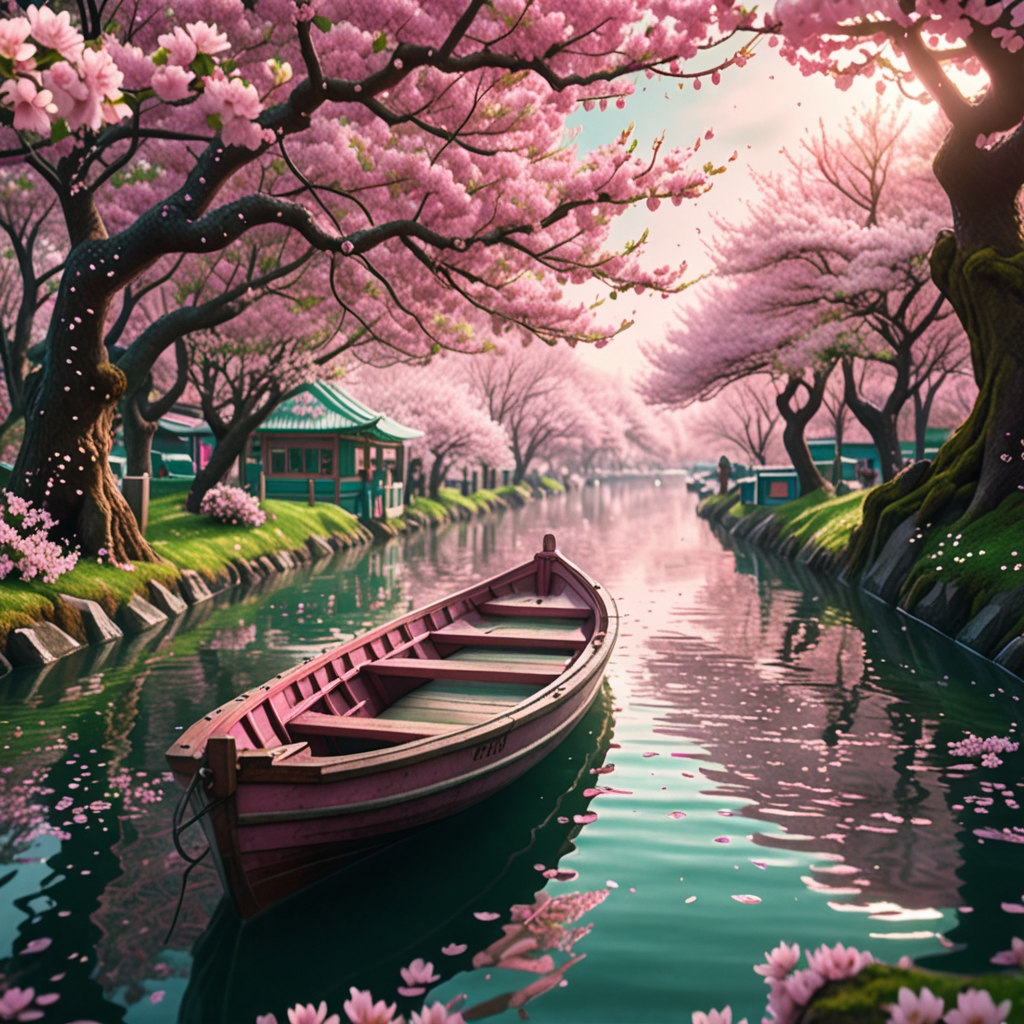} \\
            Score: 0.56 & Score: 0.66
            
        \end{tabular} \\
        \multicolumn{1}{c}{\normalsize (a)} & \multicolumn{1}{c}{\normalsize (b)} & \multicolumn{1}{c}{\normalsize (c)}
    \end{tabular}

    }
\caption{\textbf{(a)} A simple ComfyUI pipeline using a base model and a face restoration block, as well as both a positive and a negative prompt. \textbf{(b)} Distribution of scores for the prompt, flow pairs in our training set. \textbf{(c)} Example images produced for the same prompt by flows with different scores. A higher score typically correlates with more detailed and vibrant results, and fewer artifacts.}\label{fig:flow_and_score_examples}
\end{figure*}

\vspace{-3pt}
\subsection{ComfyUI}
\vspace{-3pt}

Our work focuses on ComfyUI, an open-source software for designing and executing generative pipelines. 
In ComfyUI, users create pipelines by connecting a series of blocks that represent specific models or their parameter choices. In \cref{fig:flow_and_score_examples}a, we show a simple example ComfyUI pipeline. 
This pipeline includes blocks for loading a model, specifying prompts and latent dimensions, a sampler, a VAE decoder, and a face restoration upscaling model. More complex pipelines may involve additional components like LoRAs~\citep{simoLoRA2023} or embeddings~\citep{gal2022image}, ControlNets~\citep{zhang2023adding}, IP-Adapters~\citep{ye2023ipadapter}, blocks that re-write and enhance the input prompt, and more. In many cases, complex blocks are introduced into ComfyUI through user-created extensions, which are then shared across the community.

Importantly, each ComfyUI pipeline can be exported to a JSON file which outlines both the graph nodes and their connectivity. ComfyUI's standard JSON format also contains UI information, such as the position and color of the blocks. We use the simpler API version which excludes this UI-specific information. Moreover, the API format of the flow can be used to trigger novel generations without using the UI, allowing us to automate much of our process.

We note that the concurrent work of \citet{xue2024genagent} also leverages ComfyUI pipelines. However, their work focuses on using ComfyUI as a test bed for exploring the stability of collaborative workflow generation approaches. Hence, their evaluation focuses on examining the rate at which ComfyUI-compliant workflows are created. In contrast, we focus on learning to tailor specific workflows to a user's prompt, with the aim of improving downstream generation quality.

\vspace{-3pt}
\subsection{Training Data}
\vspace{-3pt}

As a starting point, we collect a set of approximately $500$ human-generated ComfyUI workflows from popular generative-resource-sharing websites such as Civitai.com. We limit ourselves to text-to-image workflows, flitering out video generation flows, and flows that take a control image as an input. We further discard highly complex flows, whose JSON representations often span tens of thousands of lines. Finally, we discard flows that use community-written blocks appearing in fewer than three flows. This leaves us with a small set of $33$ flows, which we augment by randomly switching diffusion models (see supplementary for list), LoRAs and samplers, or changing parameters like the guidance scale and number of steps. In total, this process resulted in $310$ distinct workflows.

Recall that our goal is to predict \textit{effective} flows for a given prompt, which will enhance the quality of the generated output. To achieve this, we need a way to assess workflow performance. To do so, we begin by collecting a set of $500$ popular prompts from Civitai.com and using them to synthesize images with each flow in our dataset. These images are then scored using an ensemble of quality prediction models (LAION Aesthetic Score~\citep{schuhmann2022laion}, ImageReward~\citep{xu2024imagereward}, HPS v2.1~\citep{wu2023human}, and Pickscore~\citep{kirstain2023pickapic}). We standardize the outputs of these models so that they are of approximately the same scale, and sum them up, assigning higher weights to models that better align with human preferences. This process yields a single scalar score for each prompt and flow pair, where higher scores typically correlate with better image quality. \Cref{fig:flow_and_score_examples}b,c shows the distribution of scores in our data set, along with visual examples of images created with the same prompt, across a range of scores.

Our final dataset consists of triplets of prompt, workflow, and score. We use these to implement both the in-context and the fine-tuning based approaches detailed below.

\vspace{-3pt}
\subsection{ComfyGen-IC}
\vspace{-3pt}

As a first approach to providing prompt-dependent flows, we look to in-context based solutions that leverage a powerful, closed-source LLM. To do so, we first need to provide the LLM with some knowledge about the quality of results produced by each flow. We thus start by asking the LLM to come up with a list of $20$ labels which will best fit our $500$ training prompts. These include object-categories (``People", ``Wildlife"), scene categories (``Urban", ``Nature") and styles (``Anime", ``Photo-realistic"). A complete list of the labels is provided in the supplementary. 
With these labels in hand, we can now calculate the average quality score of images produced by each flow across all prompts belonging to a specific label. Repeating this for all flows and all labels gives us a table of flows and a measure of their performance across all $20$ categories. 

Ideally, we would have liked to provide the LLM with the full JSON description of the flows, allowing it to learn the relationships between flow components and downstream performance on specific categories. Unfortunately, the flows are too lengthy to fit more than a handful into the context window of most LLMs. Hence, our table contains only flow identities, and we simply ask the LLM to choose the flow that it believes will perform best on a given, unseen prompt. 

All in all, this approach provides us with a classifier capable of parsing new prompts, breaking them down into relevant categories, and selecting the flow that best matches these categories. We name this variation \ourmethod{}-IC.

\vspace{-3pt}
\subsection{ComfyGen-FT}
\vspace{-3pt}

As an alternative approach, we can fine-tune an LLM to predict high-quality workflows for given prompts. One way to approach this problem could be to simply fine-tune the LLM so that, given an input prompt provided in-context, it would need to predict the flow that achieved the highest score for that prompt. However, this approach has several drawbacks: it significantly reduces the number of training tokens, using only one flow per prompt instead of all $310$; it's more sensitive to randomness in our data creation process, such as the seed chosen for each generated image; and it doesn't allow learning from negative examples, which could help the model identify ineffective flow components or combinations.

Instead, we propose an alternative fine-tuning formulation where the LLM takes as its context both the prompt and a score. The model is then tasked with predicting the specific flow that achieved the given score for the corresponding prompt. 
This approach tackles the drawbacks of the best-scoring-flow method. First, it greatly increases the number of tokens available for training by utilizing all available data points, not just the highest-scoring ones. Second, it reduces the impact of random fluctuations by considering a wider range of scores and their associated flows. Third, it allows the model to learn from negative examples, i.e., flows that achieved low scores for a given prompt.
At inference time, we can simply provide the LLM with a prompt and a high score and have it predict an effective flow for our prompt. We name this variation \ourmethod{}-FT.

\vspace{-3pt}
\subsection{Implementation details}
\vspace{-3pt}
We implement ComfyGen-IC using Claude Sonnet 3.5, and ComfyGen-FT on top of pre-trained Meta Llama3.1 8B and 70B checkpoints~\citep{dubey2024llama}. Unless otherwise noted, all ComfyGen-FT results in the paper use the 70B model with a target score of $0.725$. In all cases, we finetune for a single epoch using a LoRA of rank $16$ and a learning rate of $2e-4$. Additional details are in the supplementary.

\vspace{-3pt}
\section{Experiments}\label{sec:experiments}
\vspace{-3pt}

\begin{figure*}
    \centering
    \setlength{\belowcaptionskip}{-8pt}
    \setlength{\abovecaptionskip}{2pt}
    \setlength{\tabcolsep}{0.5pt}
    {\normalsize
    \begin{tabular}{c c c}
        \includegraphics[width=0.33\textwidth,height=0.33\textwidth]{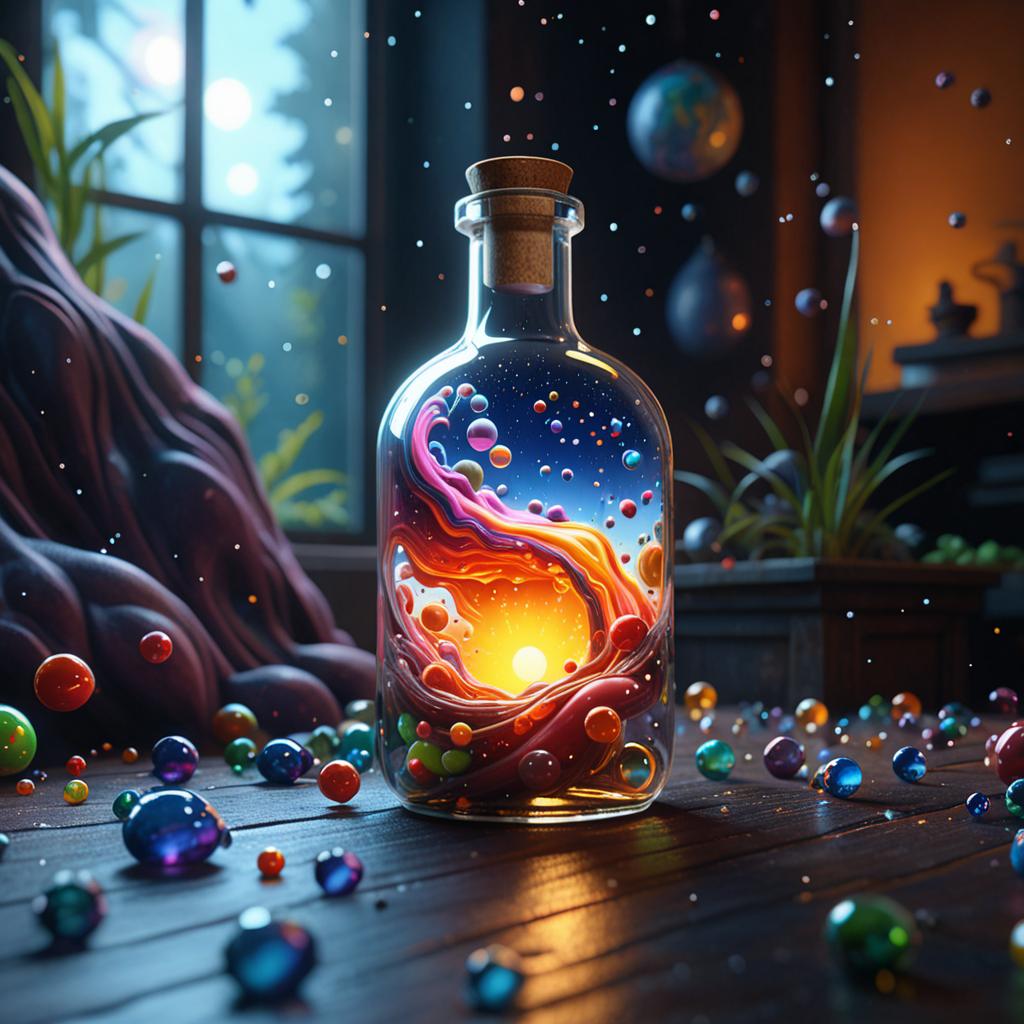} &
        \includegraphics[width=0.33\textwidth,height=0.33\textwidth]{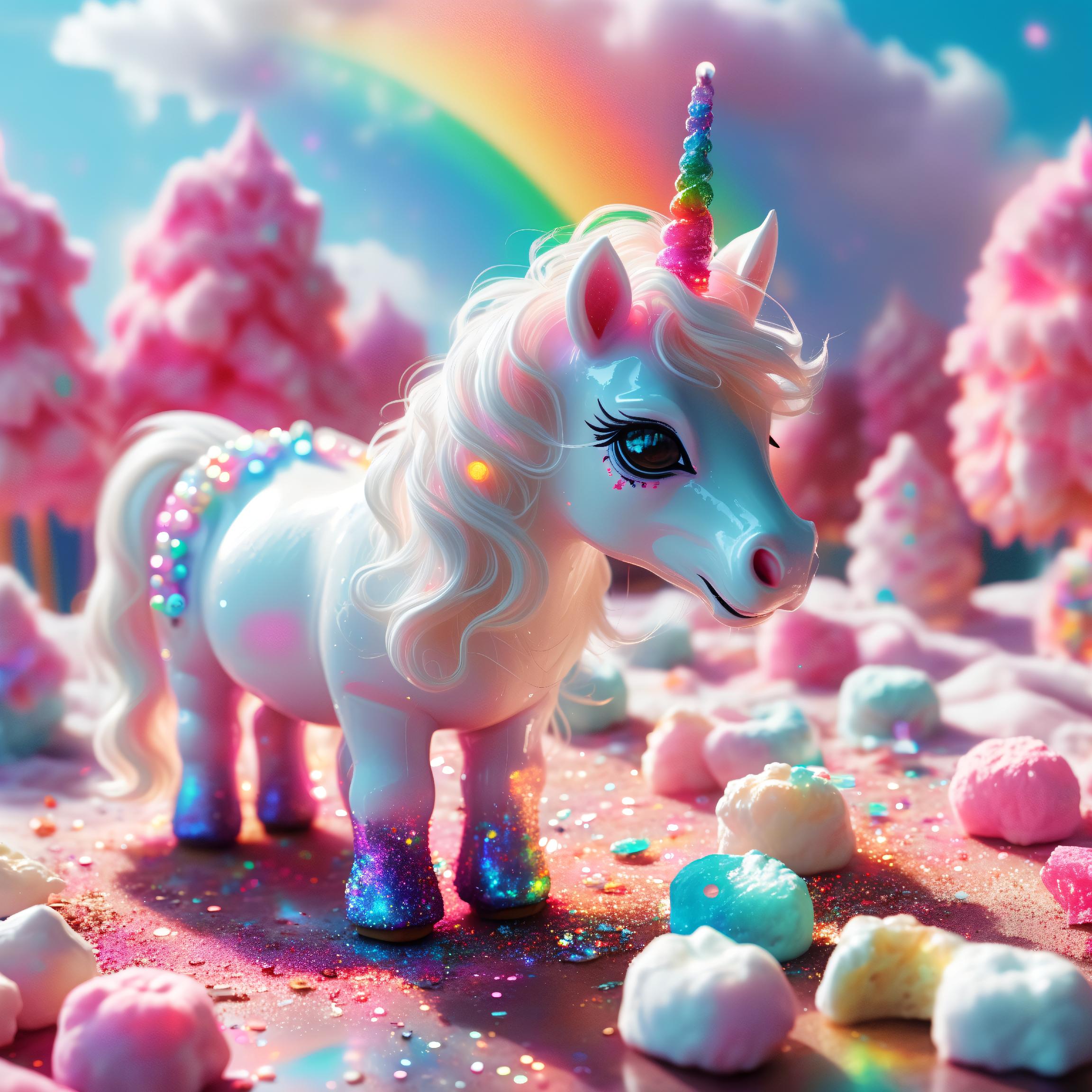} &
        \includegraphics[width=0.33\textwidth,height=0.33\textwidth]{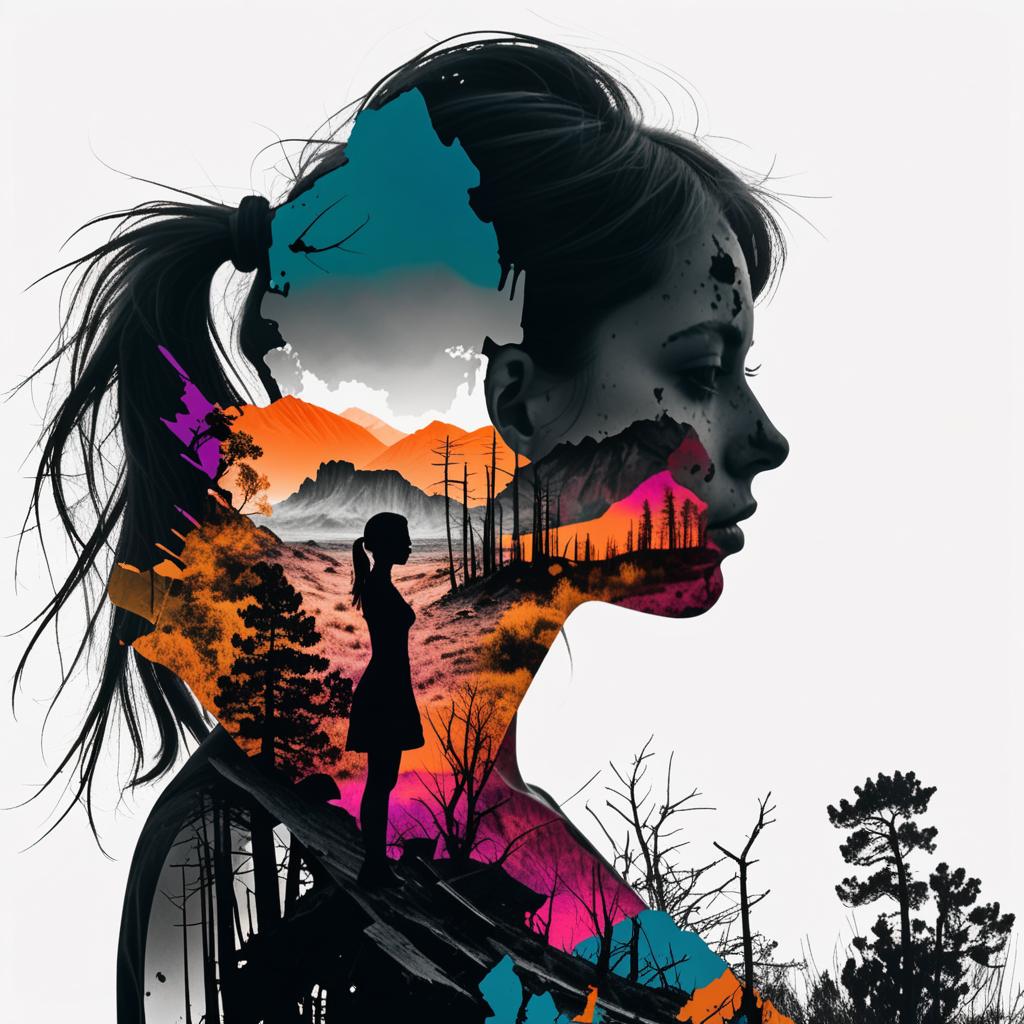} \\

        \includegraphics[width=0.33\textwidth,height=0.33\textwidth]{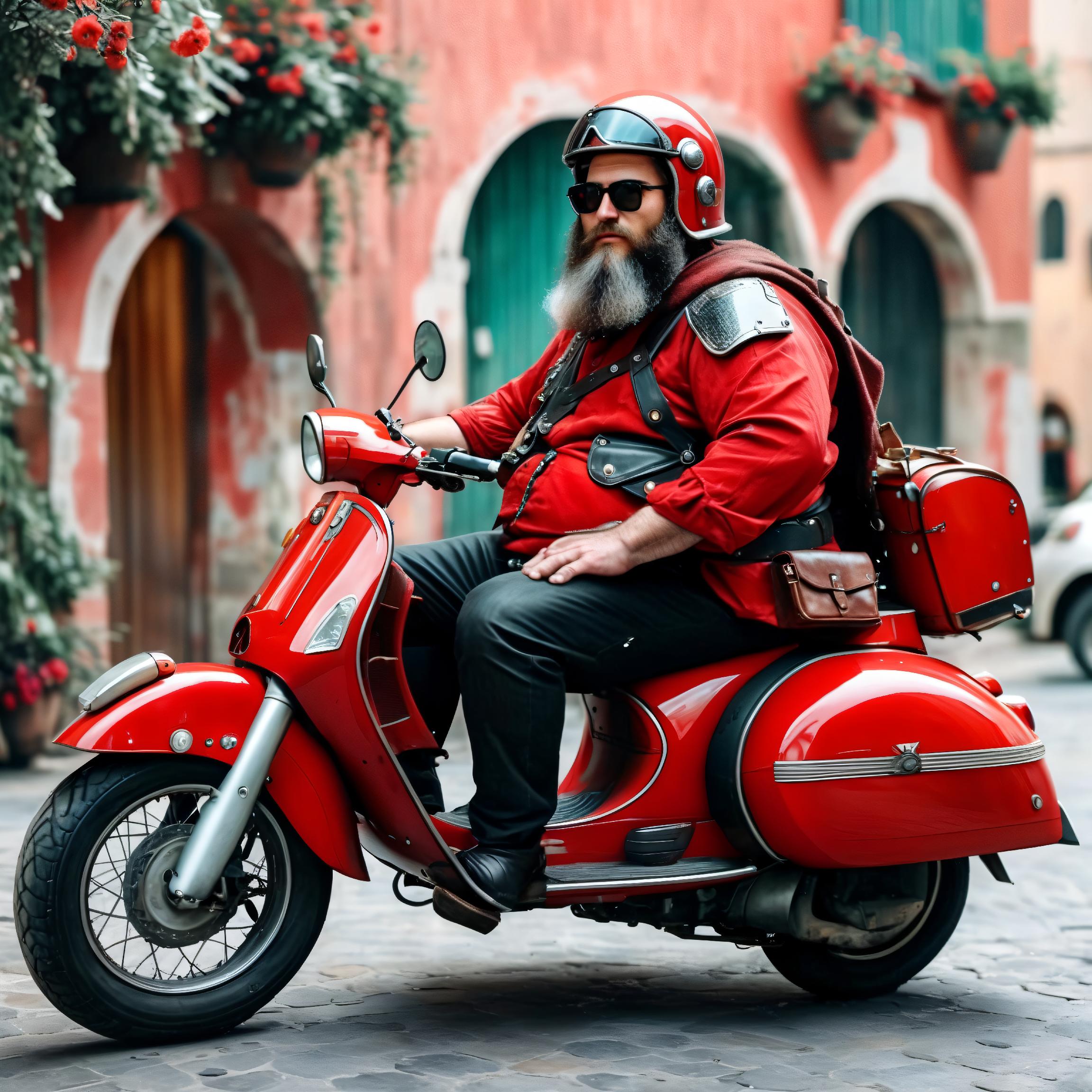} &
        \includegraphics[width=0.33\textwidth,height=0.33\textwidth]{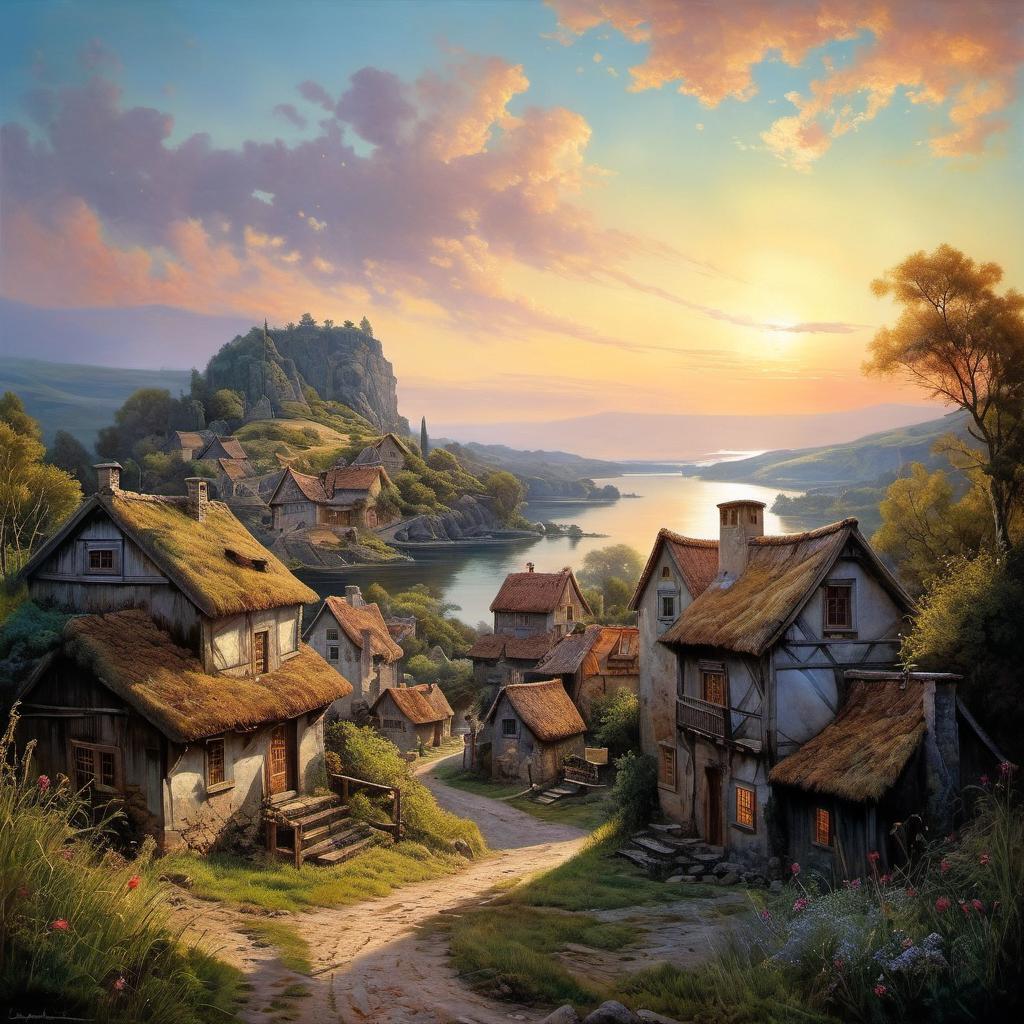} &
        \includegraphics[width=0.33\textwidth,height=0.33\textwidth]{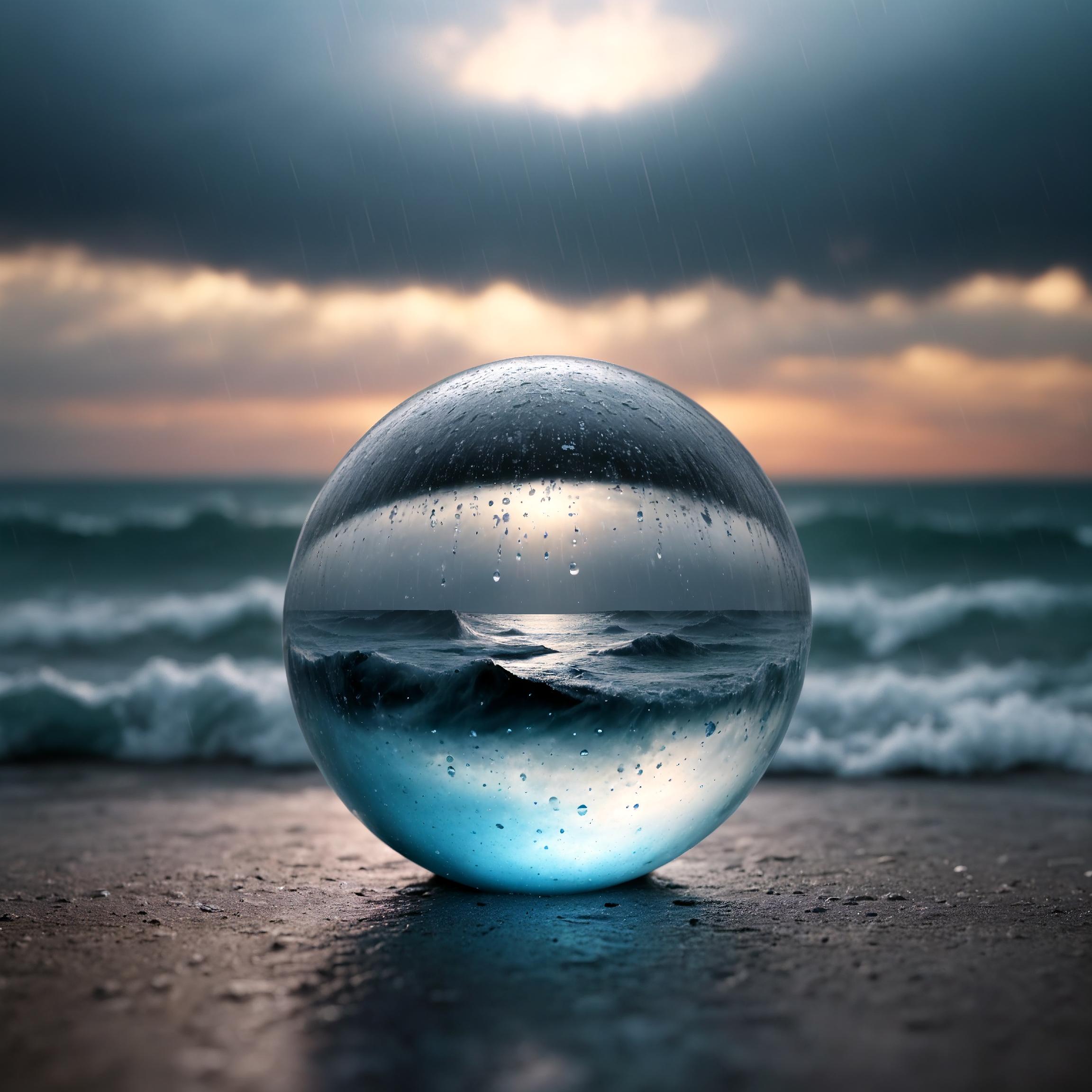} \\
        \includegraphics[width=0.33\textwidth,height=0.33\textwidth]{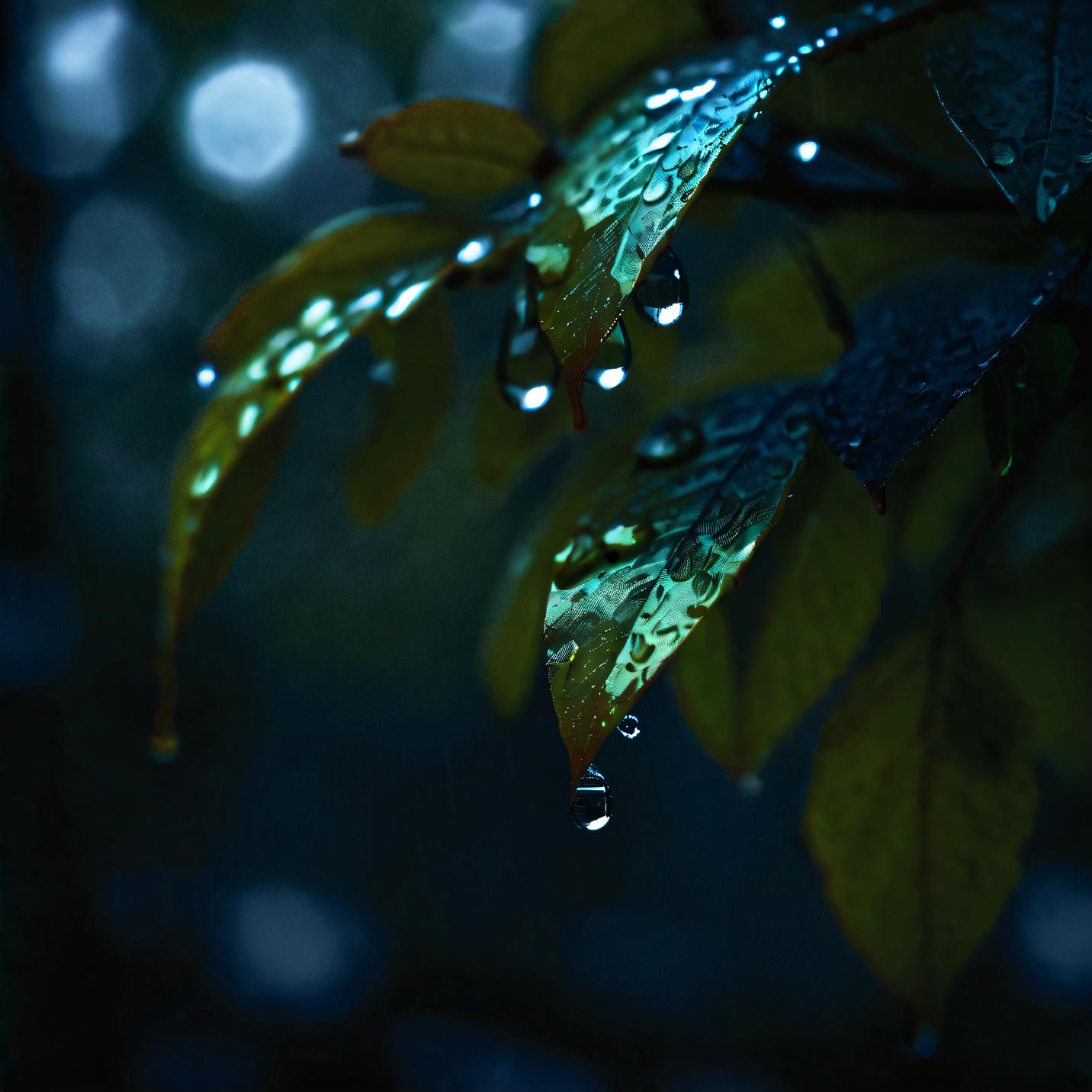} &
        \includegraphics[width=0.33\textwidth,height=0.33\textwidth]{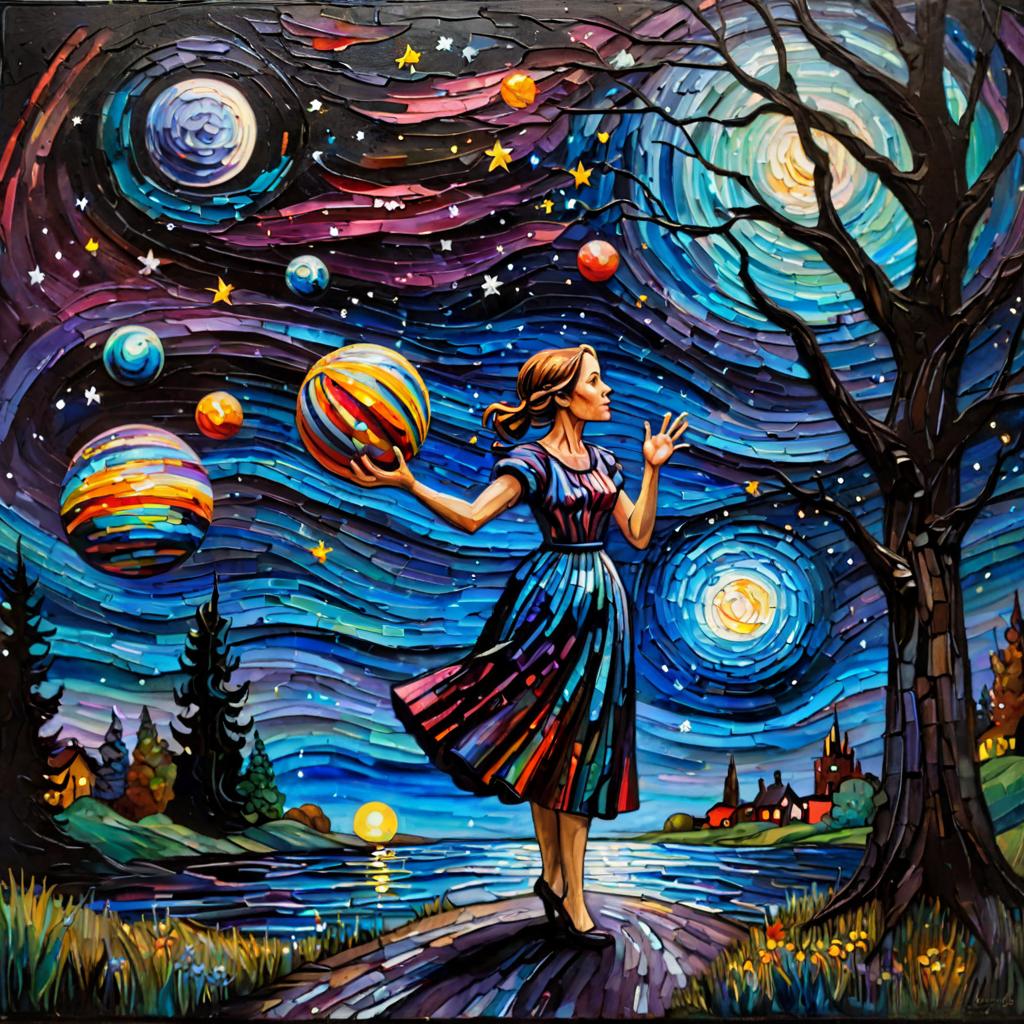} &
        \includegraphics[width=0.33\textwidth,height=0.33\textwidth]{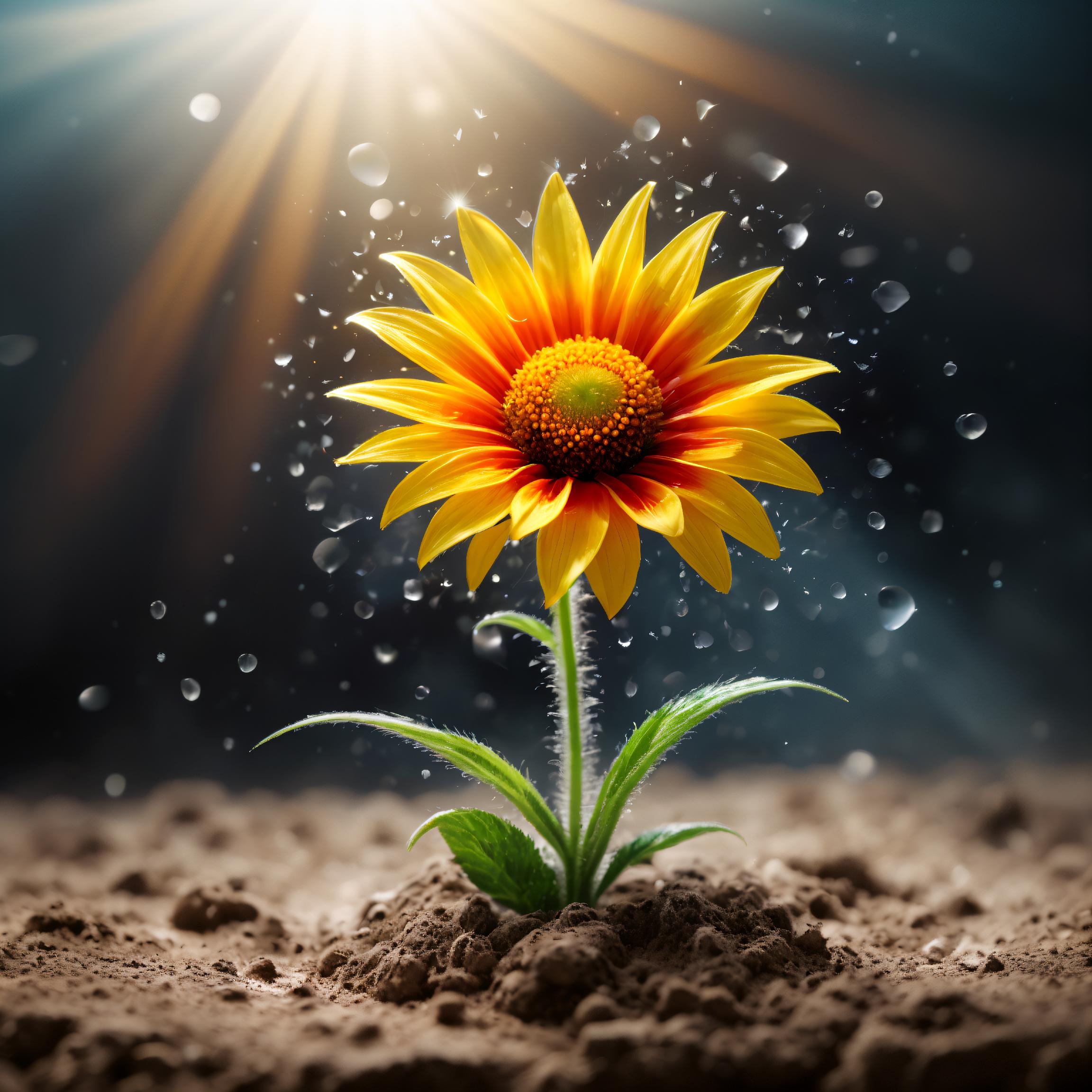} \\

    \end{tabular}
    }
    \caption{Our method can generate higher quality images across diverse domains and styles. Prompts are available in the supplementary.}\label{fig:ours_large}
\end{figure*}

We begin by showcasing images generated with our approach across a range of prompts, including subject-focused, photo-realistic imagery, as well as artistic or abstract creations. These are shown in \cref{fig:ours_large}, with additional large-scale figures in the supplementary.

Next, we compare the images produced by our approach with those generated by a series of baselines. We consider two types of alternative approaches: (1) Fixed, monolithic models, where we simply use the prompts to directly condition a pre-trained diffusion model. (2) Generic workflows, where we use the same workflow to generate all images, regardless of the specific prompt.

For (1), we consider the original SDXL model~\citep{podell2024sdxl} and its two most popular (most downloaded on CivitAI) fine-tuned variations: JuggernautXL and DreamshaperXL. We further consider a version of SDXL fine-tuned with DPO (DPO-SDXL,~\citep{wallace2024diffusion}). For (2), we selected the two most popular flows (based on download counts) from our training corpus. These flows use SSD-1B~\citep{gupta2024progressive} and Pixart-$\Sigma$~\citep{chen2024pixart} respectively.

We evaluate our models and the baselines on two fronts. First, we use the GenEval benchmark~\citep{ghosh2024geneval}, which uses object detection to assess generative models across prompt-alignment tasks like single-object generation, counting, and attribute binding. Quantitative results are reported in \cref{tab:geneval} with qualitative samples shown in \cref{fig:geneval_qualitative}. Our tuning-based model outperforms all baselines, despite only using human preference scores during training. The in-context approach underperforms. We hypothesize that it suffers due to GenEval's short, simplistic prompts, which typically only describe a few objects in one or two words each. This challenges the LLM's ability to match the prompts with labels, and performance degrades.

\begin{table}[t]
    \fontsize{8.5pt}{8.5pt}\selectfont
    \centering
    \setlength{\belowcaptionskip}{-4pt}
    \begin{tabular}{lccccccc}
        \toprule
 & Single & Two &  &  &  & Attribute \\
 Model &  object & object & Counting & Colors & Position & binding & \textbf{Overall} \\
\cmidrule(lr){1-1}
\cmidrule(lr){2-2}
\cmidrule(lr){3-8}
Single Model - SDXL & 0.98 & 0.74 & 0.39 & 0.85 & \underline{0.15} & 0.23 & 0.55 \\
Single Model - JuggernautXL & \textbf{1.00} & 0.73 & 0.48 & \underline{0.89} & 0.11 & 0.19 & 0.57 \\
Single Model - DreamShaperXL & \underline{0.99} & 0.78 & 0.45 & 0.81 & \textbf{0.17} & 0.24 & 0.57 \\
Single Model - DPO-SDXL & \textbf{1.00} & \underline{0.81} & 0.44 & \textbf{0.90} & \underline{0.15} & 0.23 & \underline{0.59} \\
\midrule
Fixed Flow - Most Popular & 0.95 & 0.38 & 0.26 & 0.77 & 0.06 & 0.12 & 0.42 \\
Fixed Flow - 2nd Most Popular & \textbf{1.00} & 0.65 & \textbf{0.56} & 0.86 & 0.13 & \textbf{0.34} & \underline{0.59} \\ 
\midrule
ComfyGen-IC (ours) & \underline{0.99} & 0.78 & 0.38 & 0.84 & 0.13 & 0.25 & 0.56 \\
ComfyGen-FT (ours) & \underline{0.99} & \textbf{0.82} & \underline{0.50} & \textbf{0.90} & 0.13 & \underline{0.29} & \textbf{0.61} \\
\bottomrule
    \end{tabular}
    \vspace{0.1cm}
    \caption{GenEval comparisons. \ourmethod{}-FT outperforms all baseline approaches, despite being tuned with human preference scores, and not strictly for prompt alignment.}
    \label{tab:geneval}
\end{table}

\begin{figure*}
    \centering
    \setlength{\belowcaptionskip}{-8pt}
    \setlength{\abovecaptionskip}{3pt}
    \setlength{\tabcolsep}{0.5pt}
    {\footnotesize
    \begin{tabular}{c c c c c c c}
    
        {\small SDXL} & {\small Juggernaut} & {\small DreamShaper} & {\small Flow 1} & {\small Flow 2} & {\small ComfyGen-IC} & {\small ComfyGen-FT} \\

        \includegraphics[width=0.14\textwidth,height=0.14\textwidth]{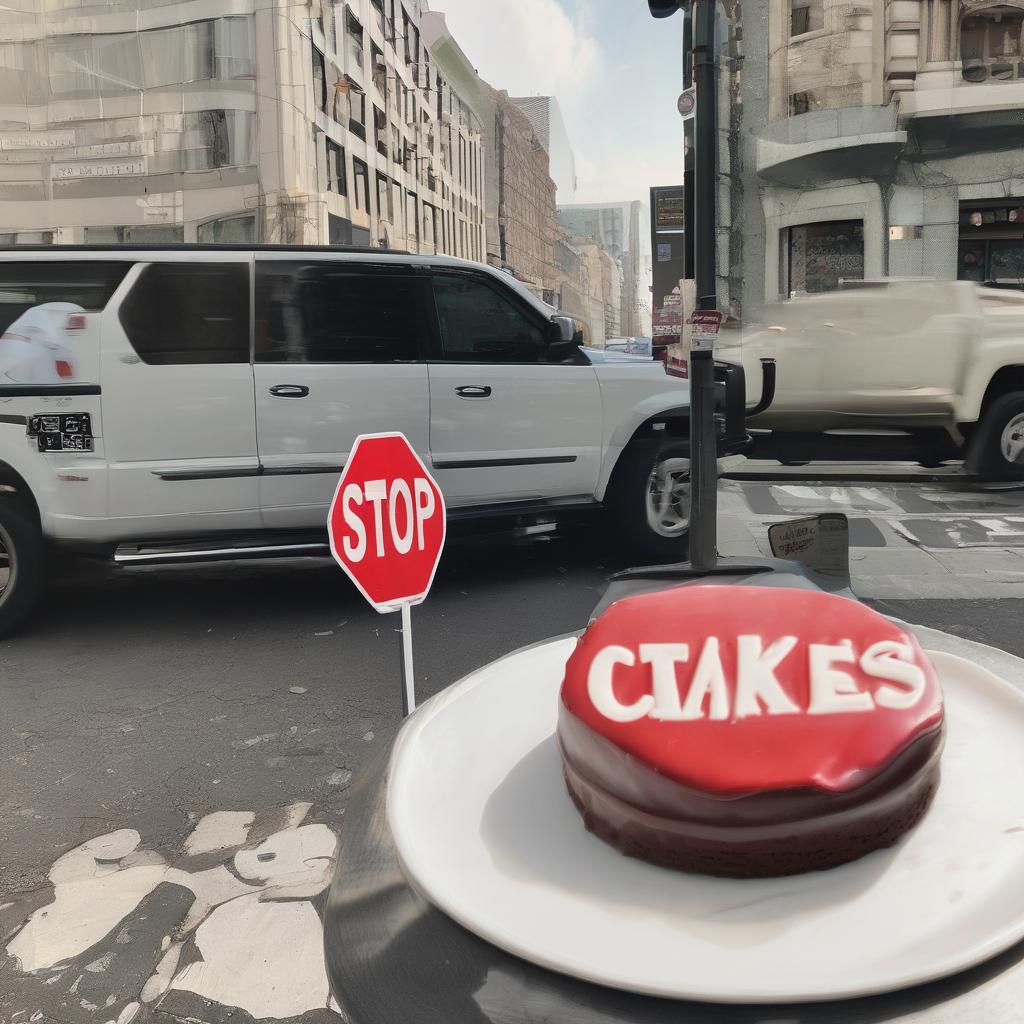} &
        \includegraphics[width=0.14\textwidth,height=0.14\textwidth]{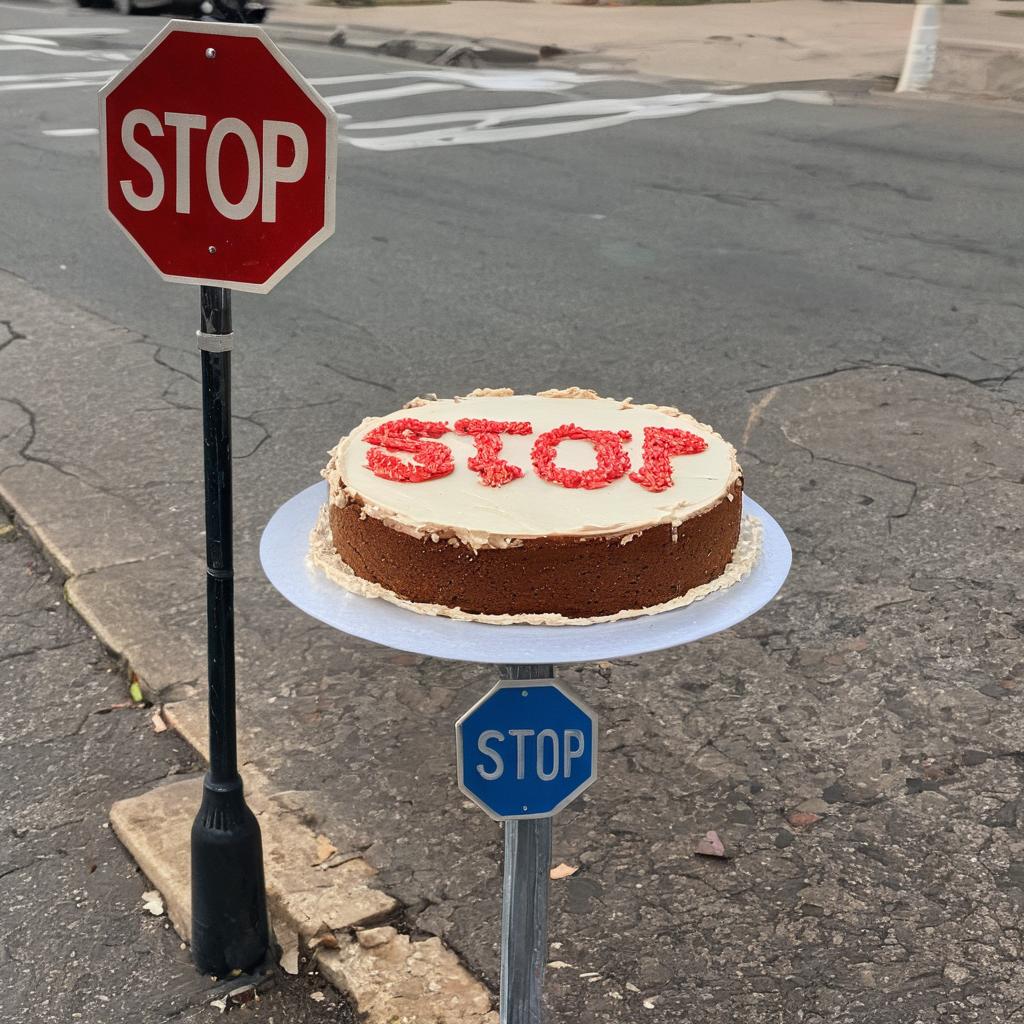} &
        \includegraphics[width=0.14\textwidth,height=0.14\textwidth]{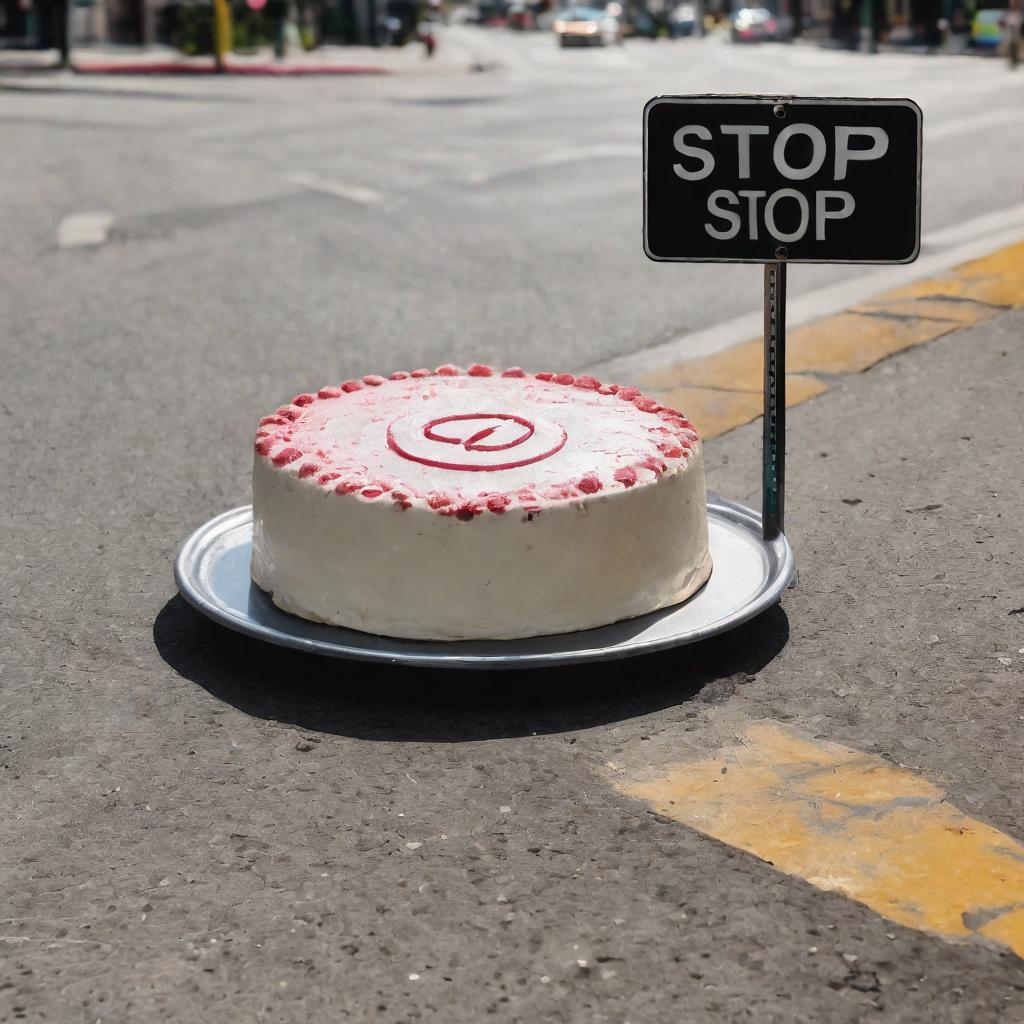} &
        \includegraphics[width=0.14\textwidth,height=0.14\textwidth]{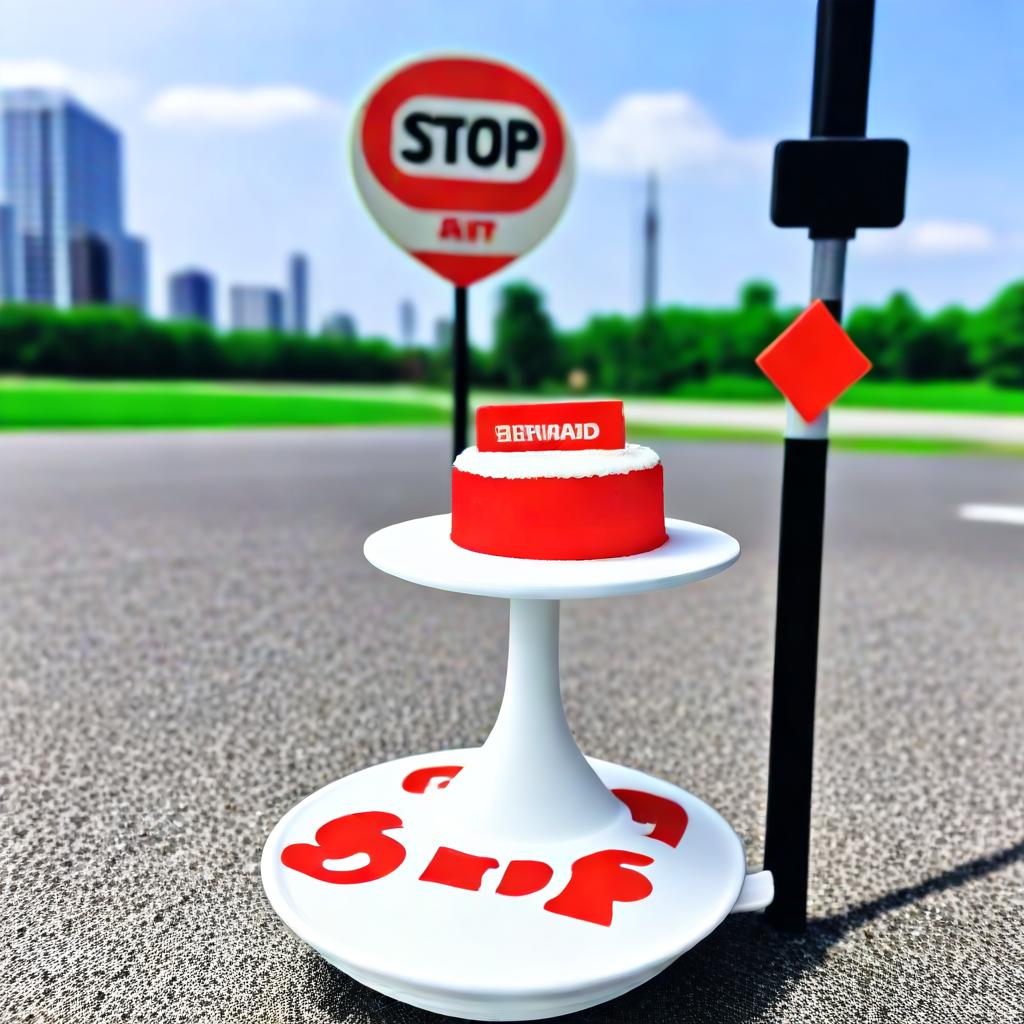} &
        \includegraphics[width=0.14\textwidth,height=0.14\textwidth]{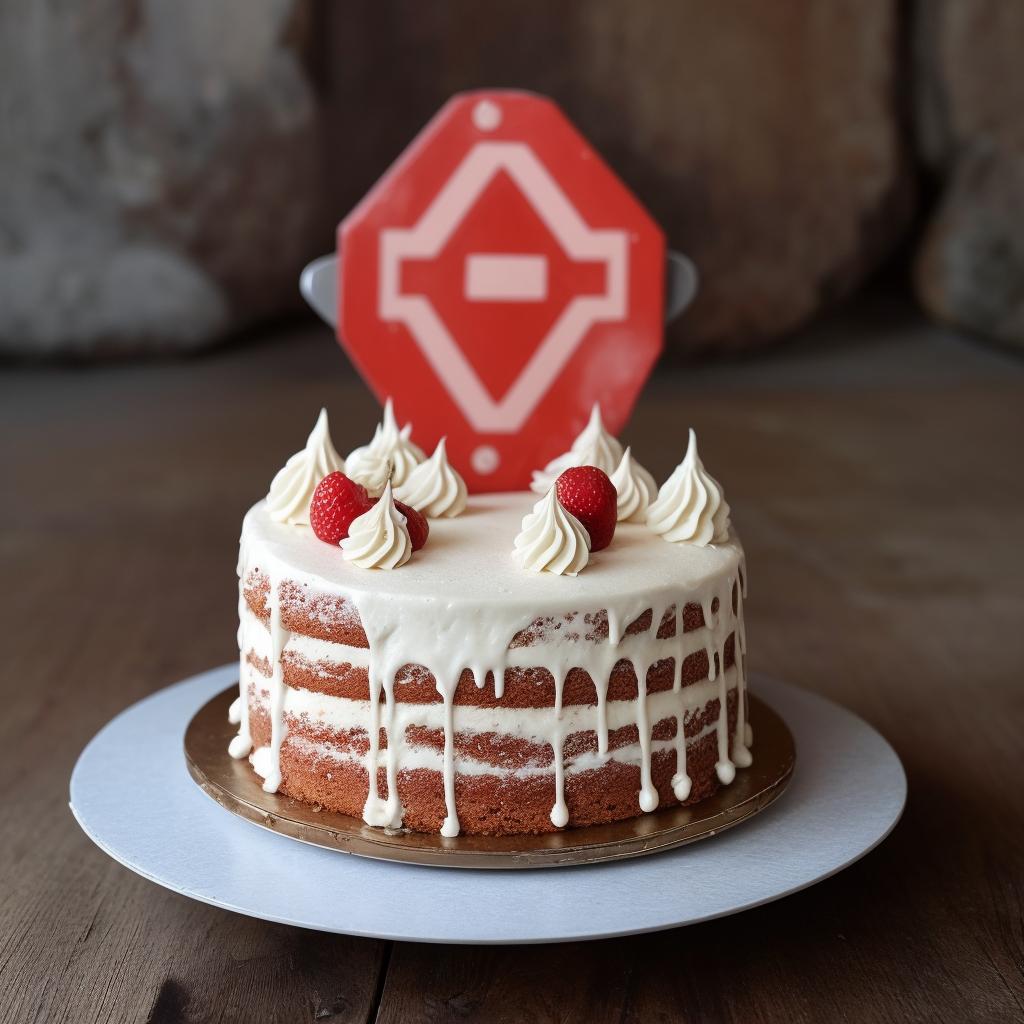} &
        \includegraphics[width=0.14\textwidth,height=0.14\textwidth]{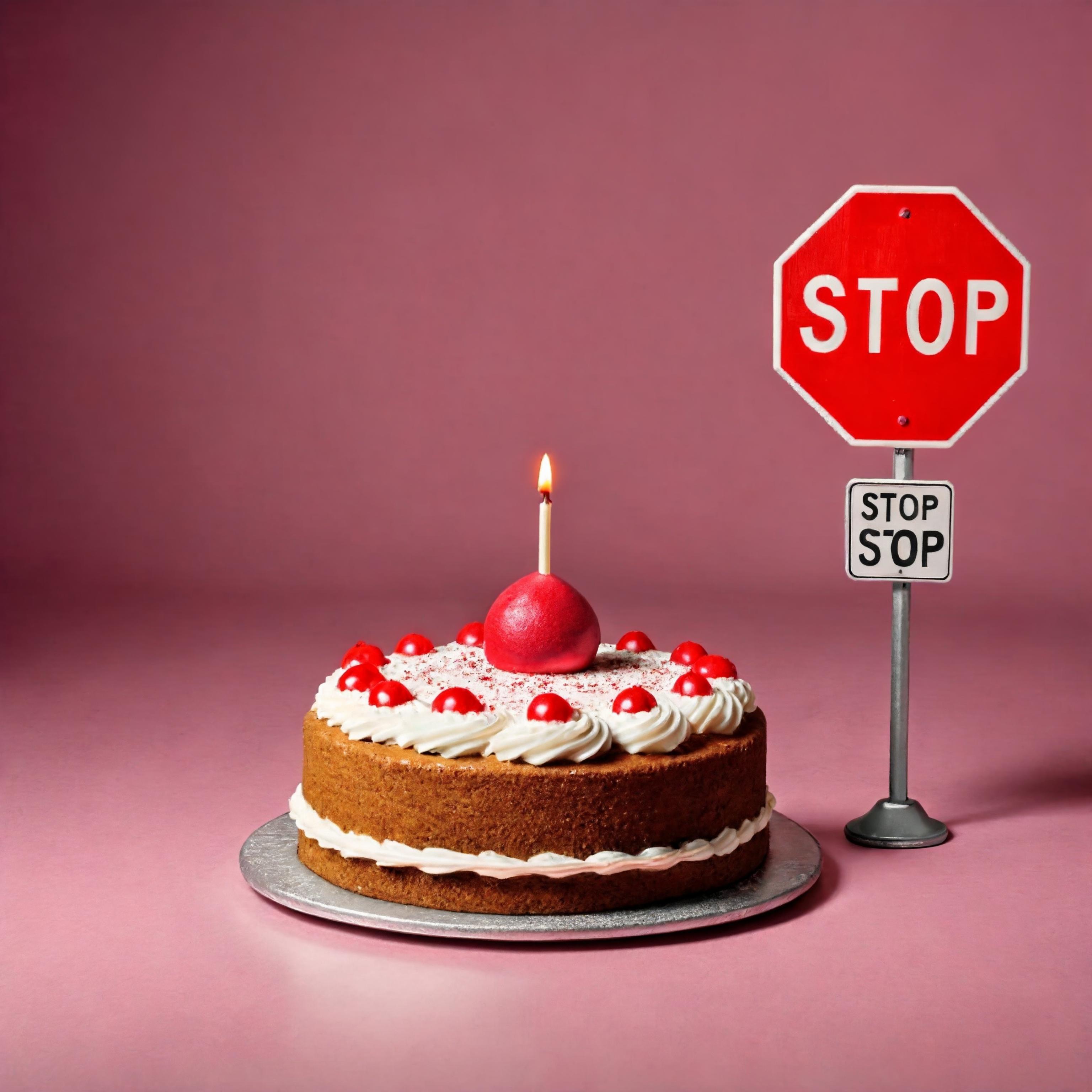} &
        \includegraphics[width=0.14\textwidth,height=0.14\textwidth]{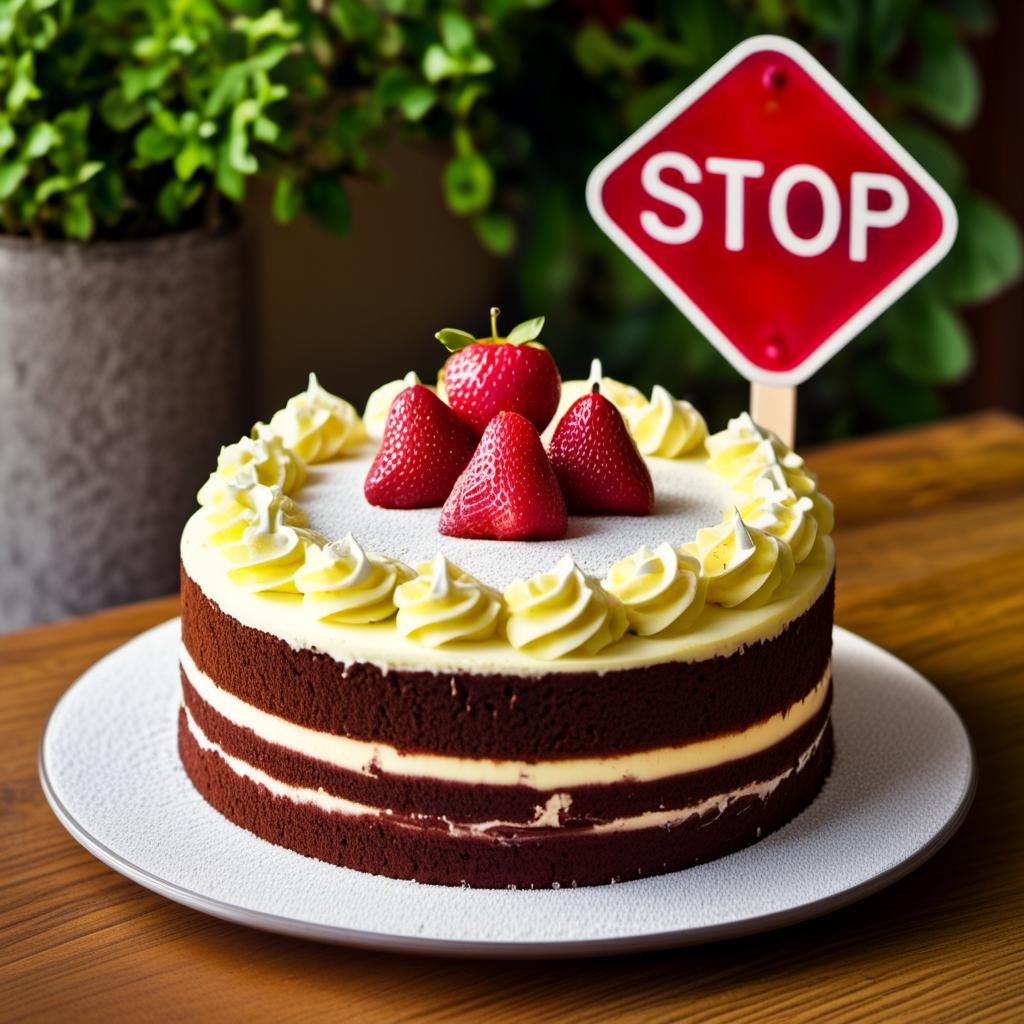}  \\

        \multicolumn{7}{c}{``A photo of a cake and a stop sign" } \\

        \includegraphics[width=0.14\textwidth,height=0.14\textwidth]{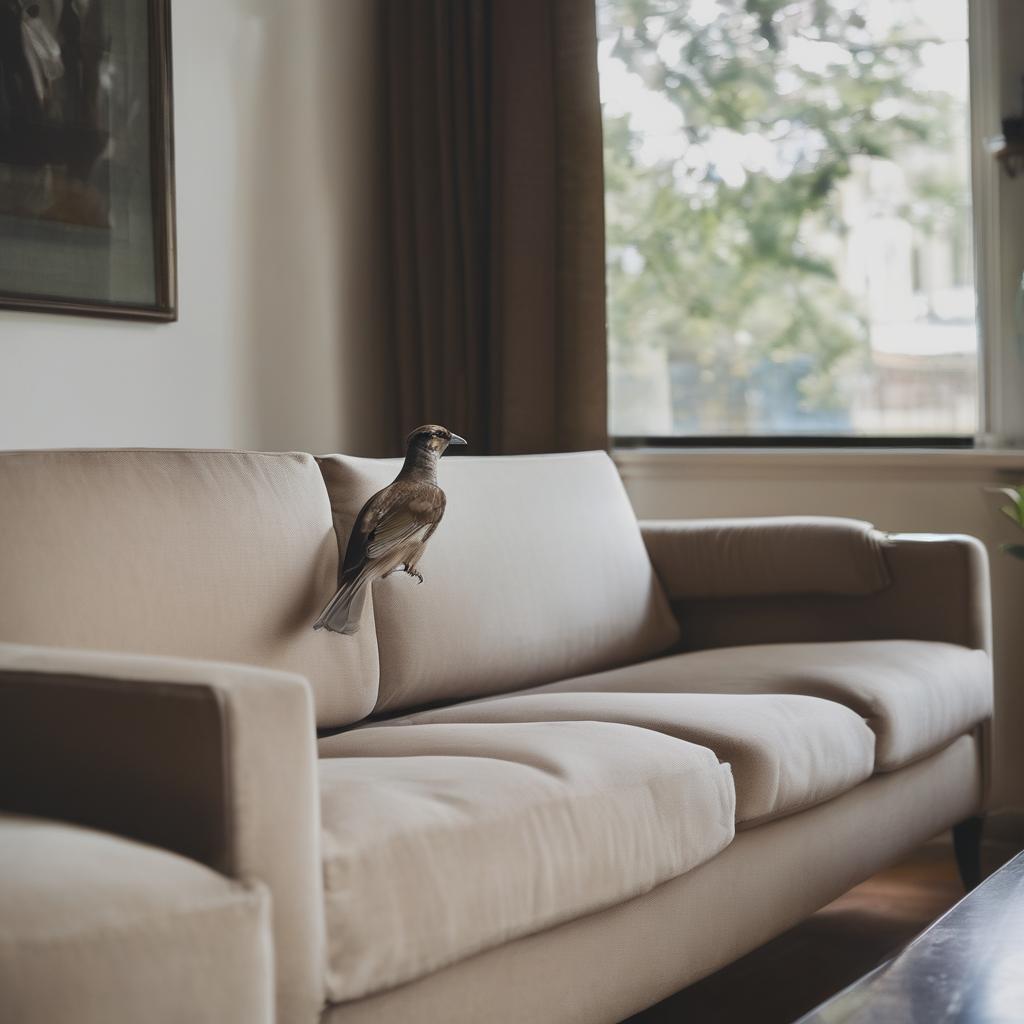} &
        \includegraphics[width=0.14\textwidth,height=0.14\textwidth]{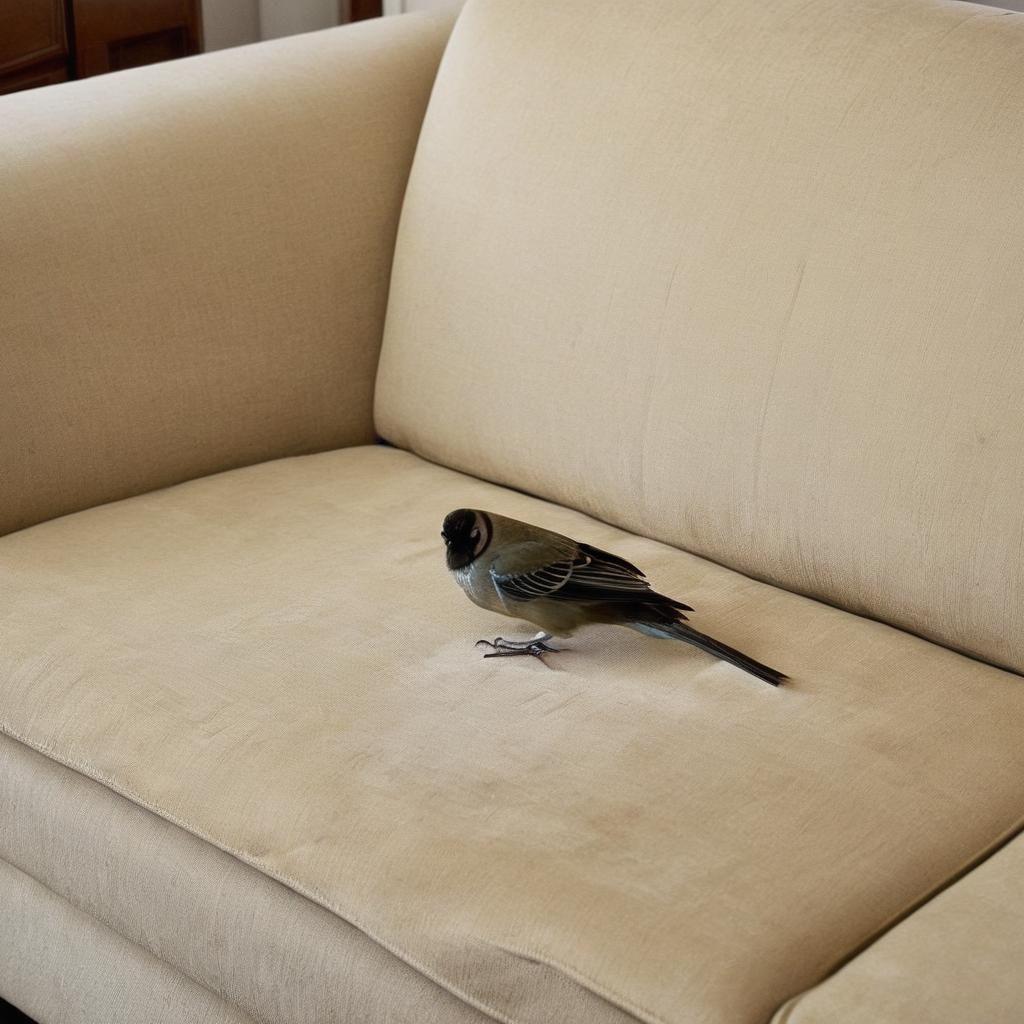} &
        \includegraphics[width=0.14\textwidth,height=0.14\textwidth]{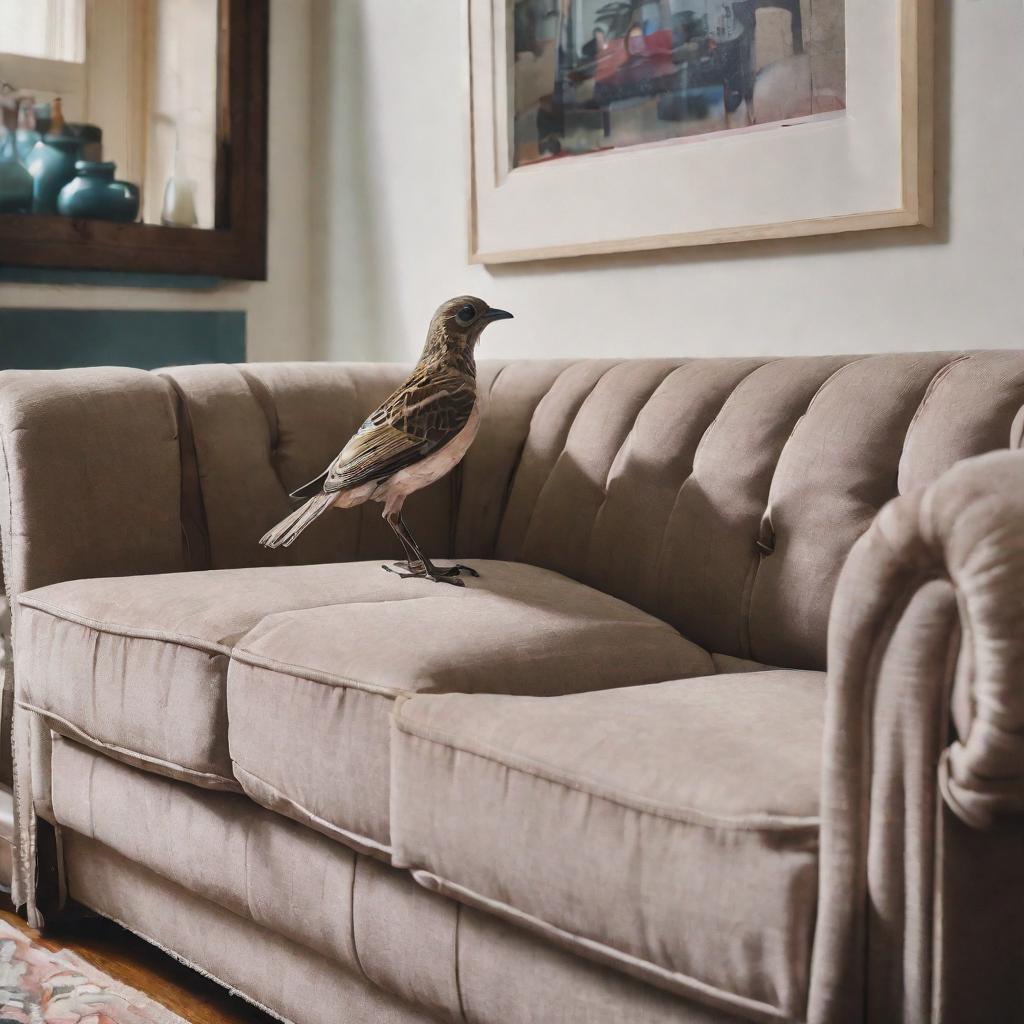} &
        \includegraphics[width=0.14\textwidth,height=0.14\textwidth]{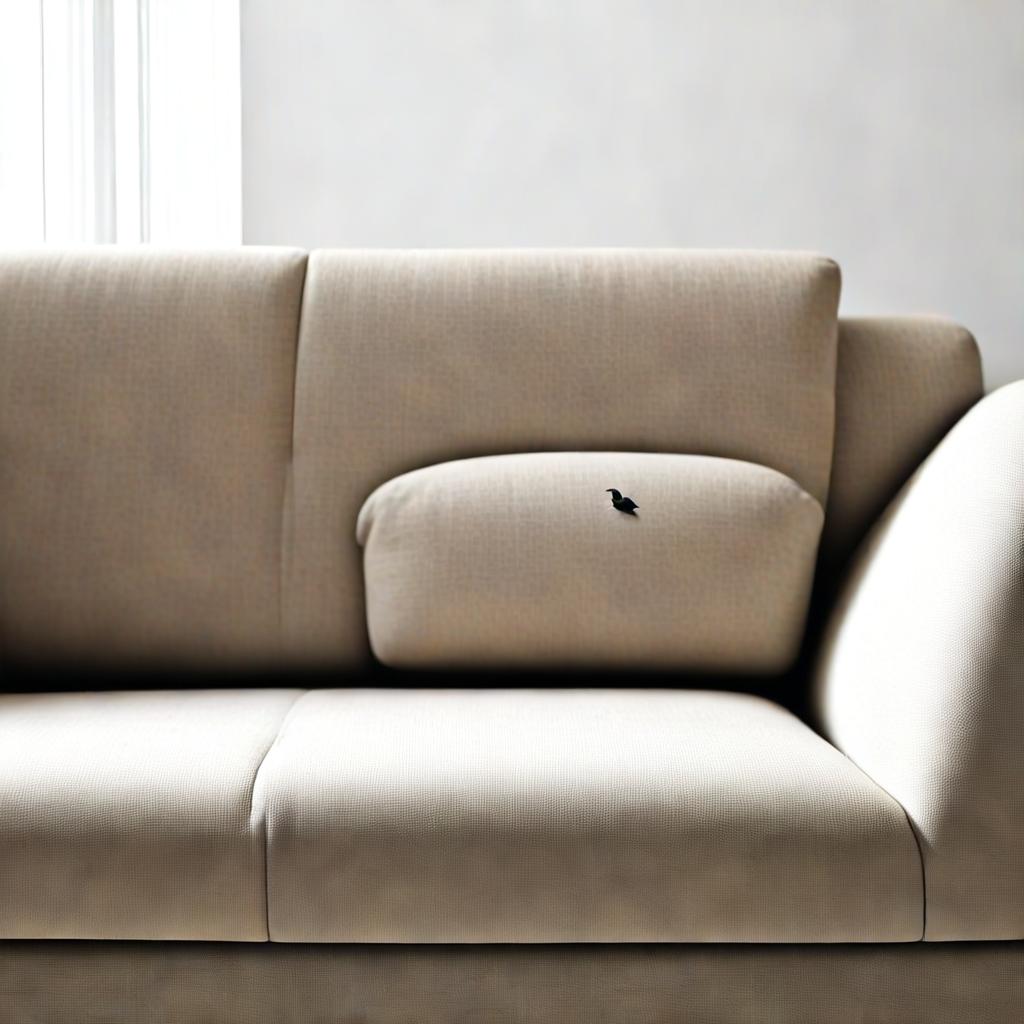} &
        \includegraphics[width=0.14\textwidth,height=0.14\textwidth]{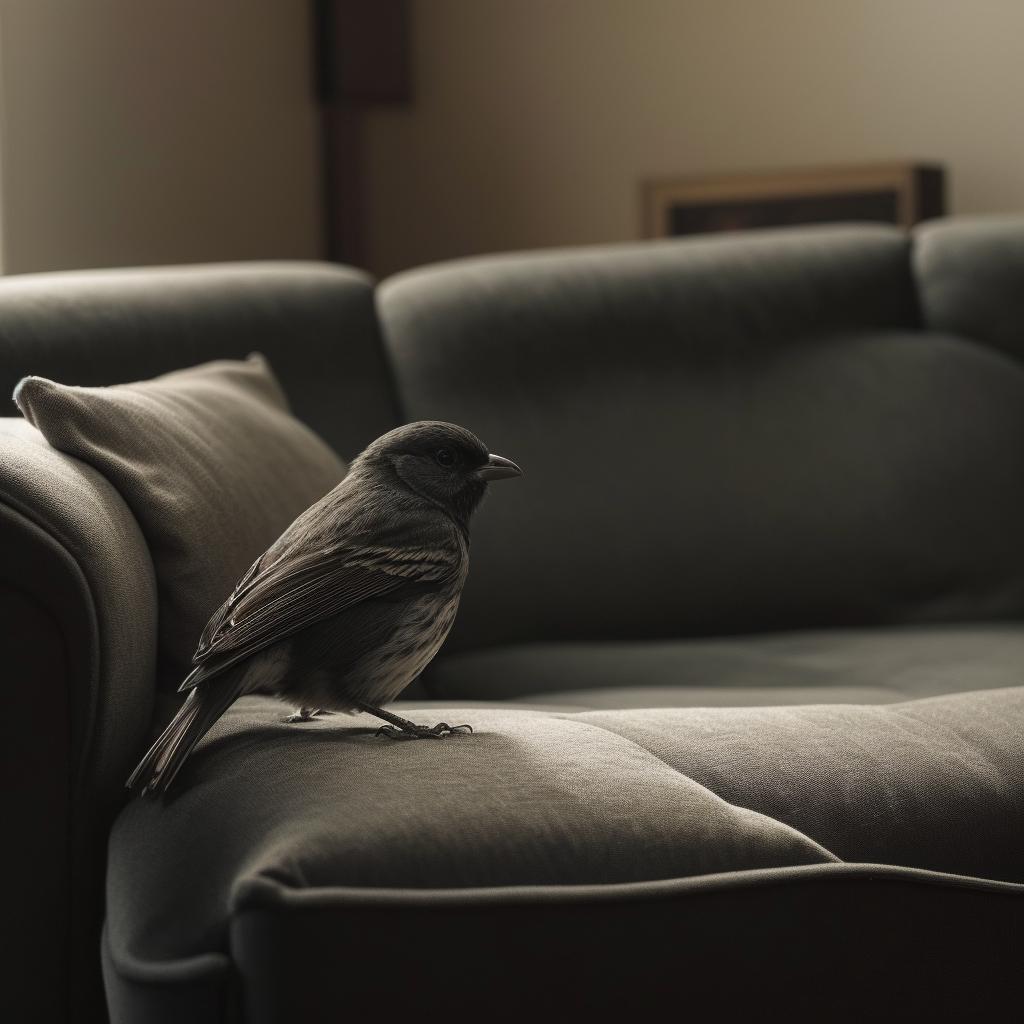} &
        \includegraphics[width=0.14\textwidth,height=0.14\textwidth]{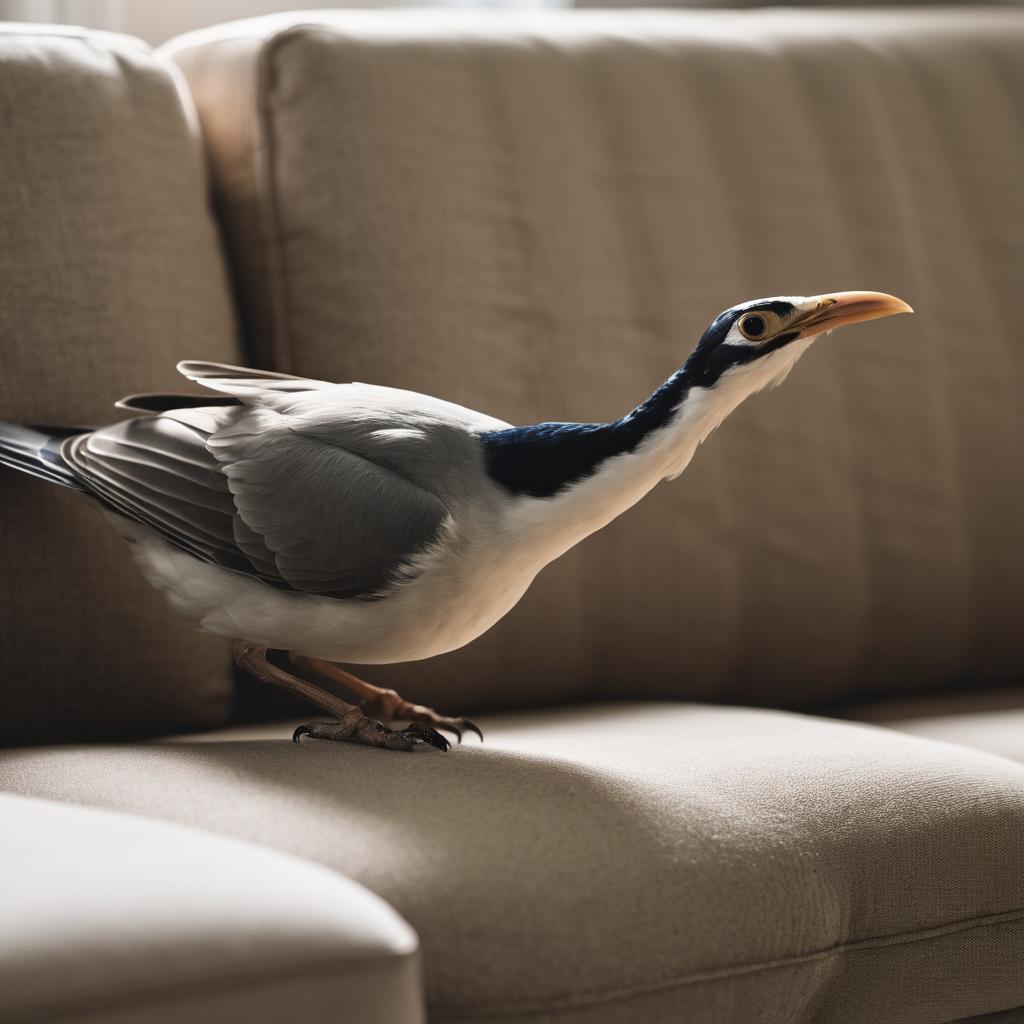} &
        \includegraphics[width=0.14\textwidth,height=0.14\textwidth]{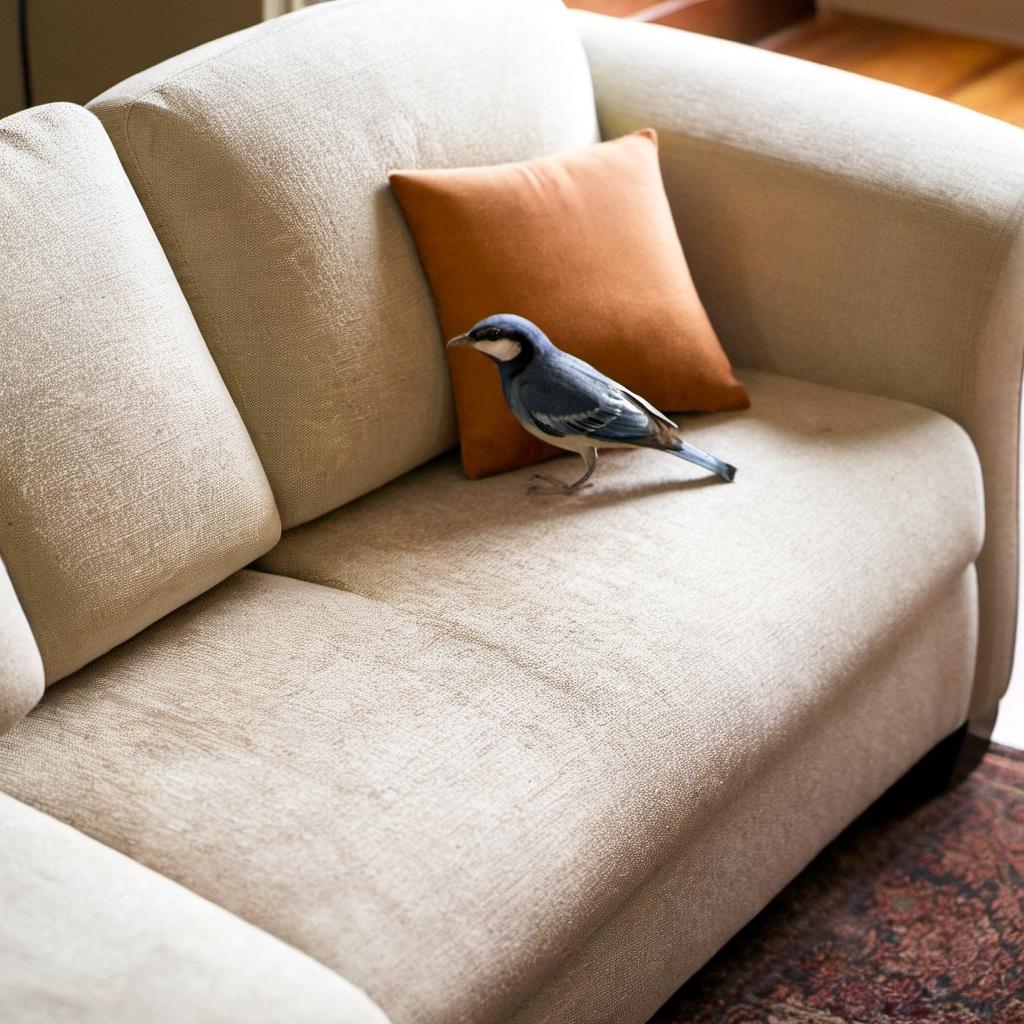} \\

        \multicolumn{7}{c}{ ``A photo of a bird left of a couch" } \\

        \includegraphics[width=0.14\textwidth,height=0.14\textwidth]{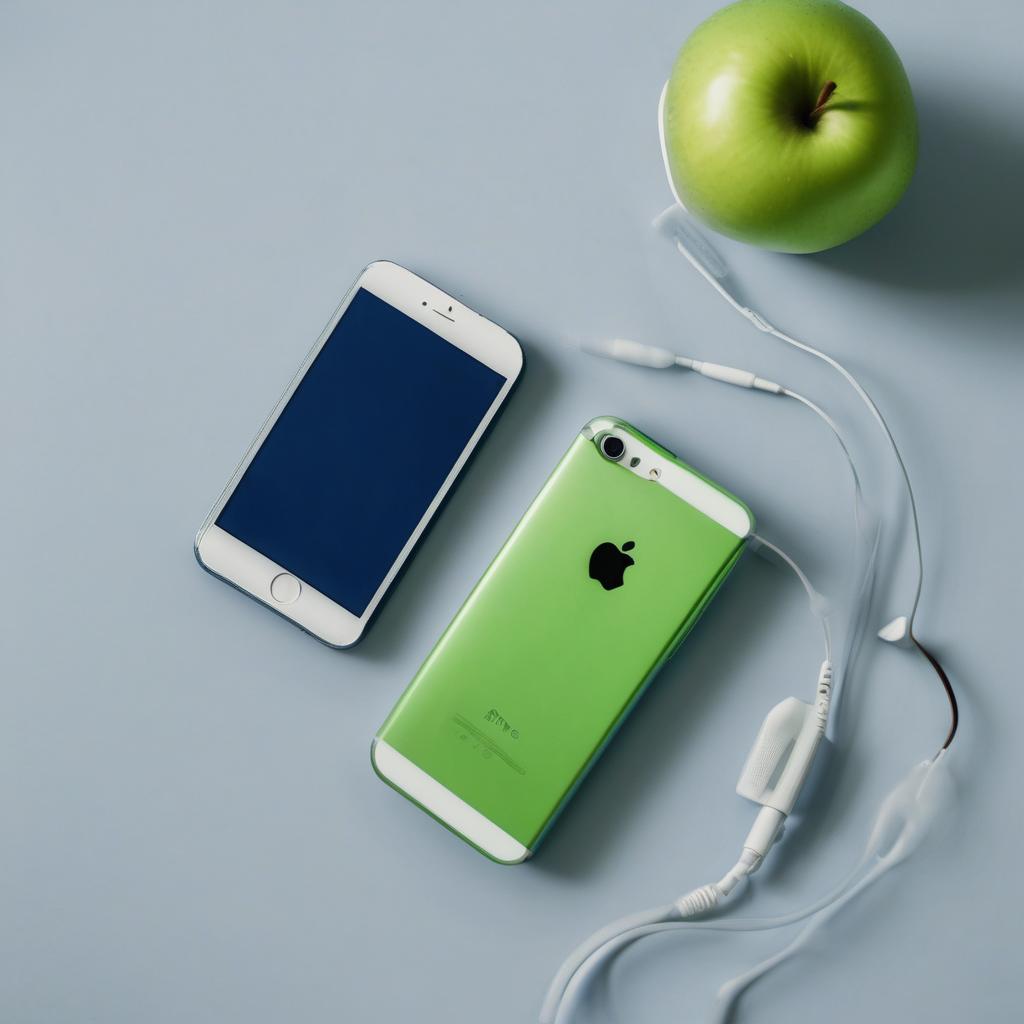} &
        \includegraphics[width=0.14\textwidth,height=0.14\textwidth]{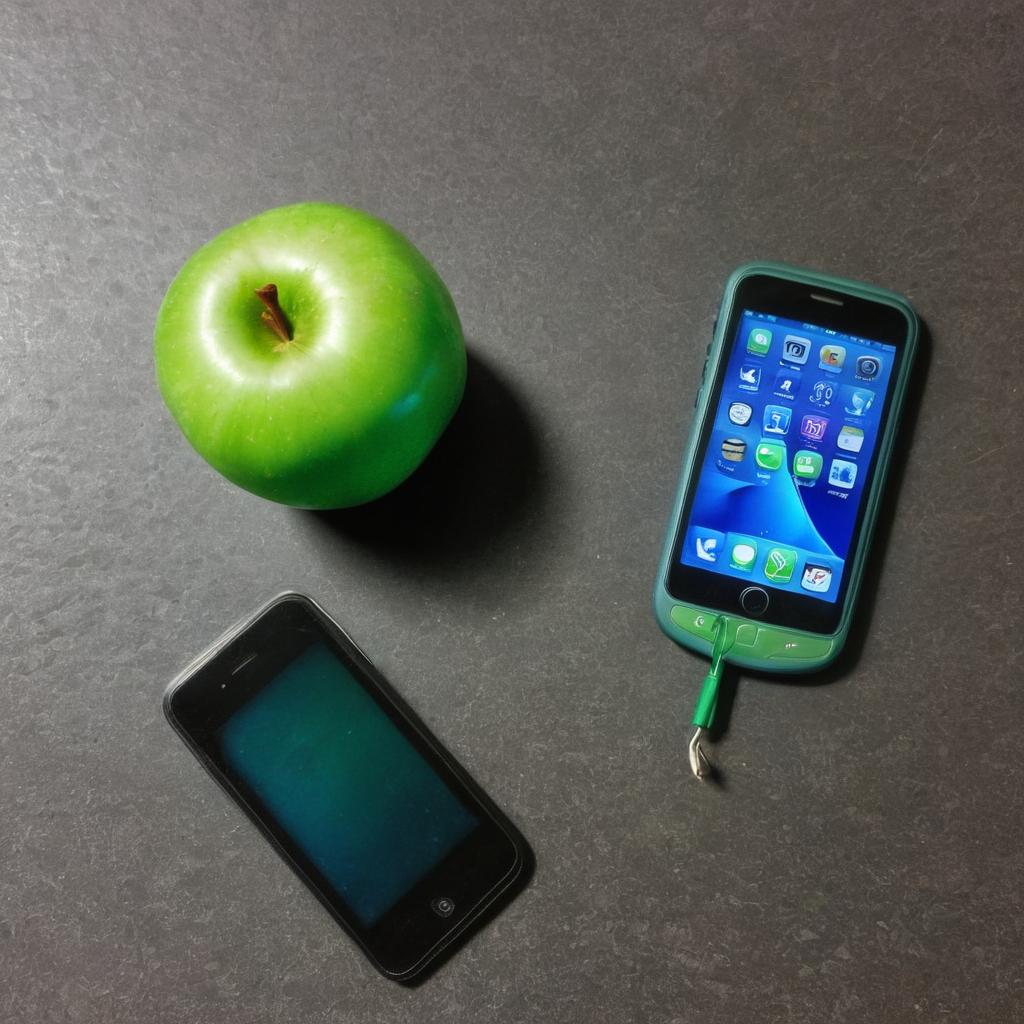} &
        \includegraphics[width=0.14\textwidth,height=0.14\textwidth]{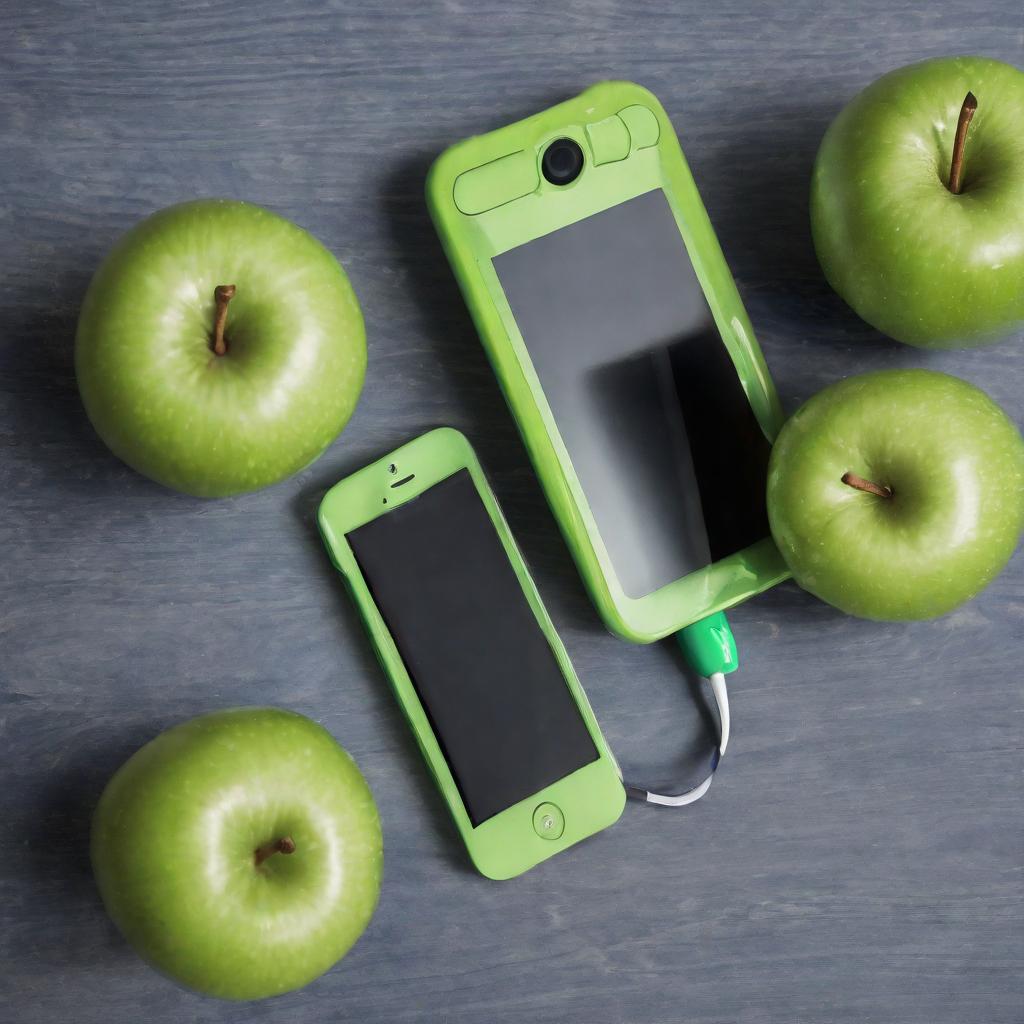} &
        \includegraphics[width=0.14\textwidth,height=0.14\textwidth]{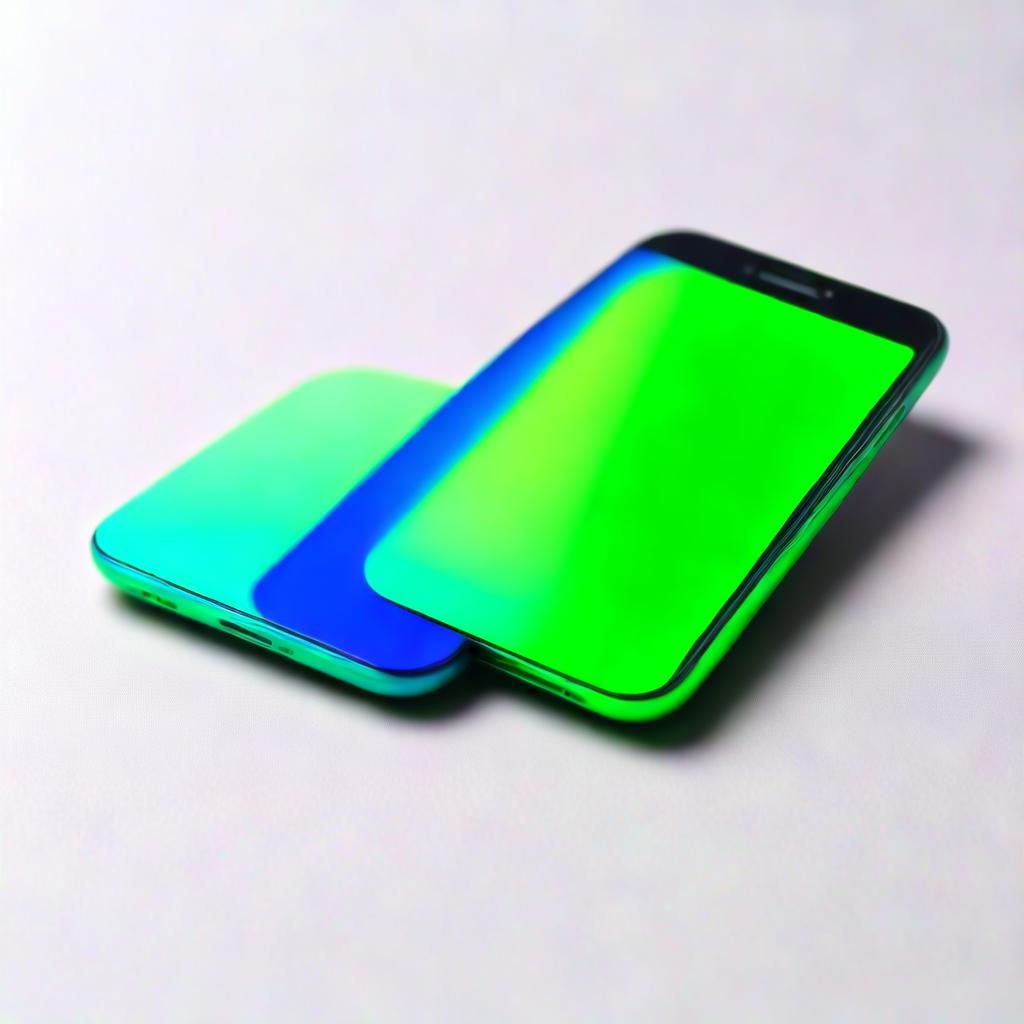} &
        \includegraphics[width=0.14\textwidth,height=0.14\textwidth]{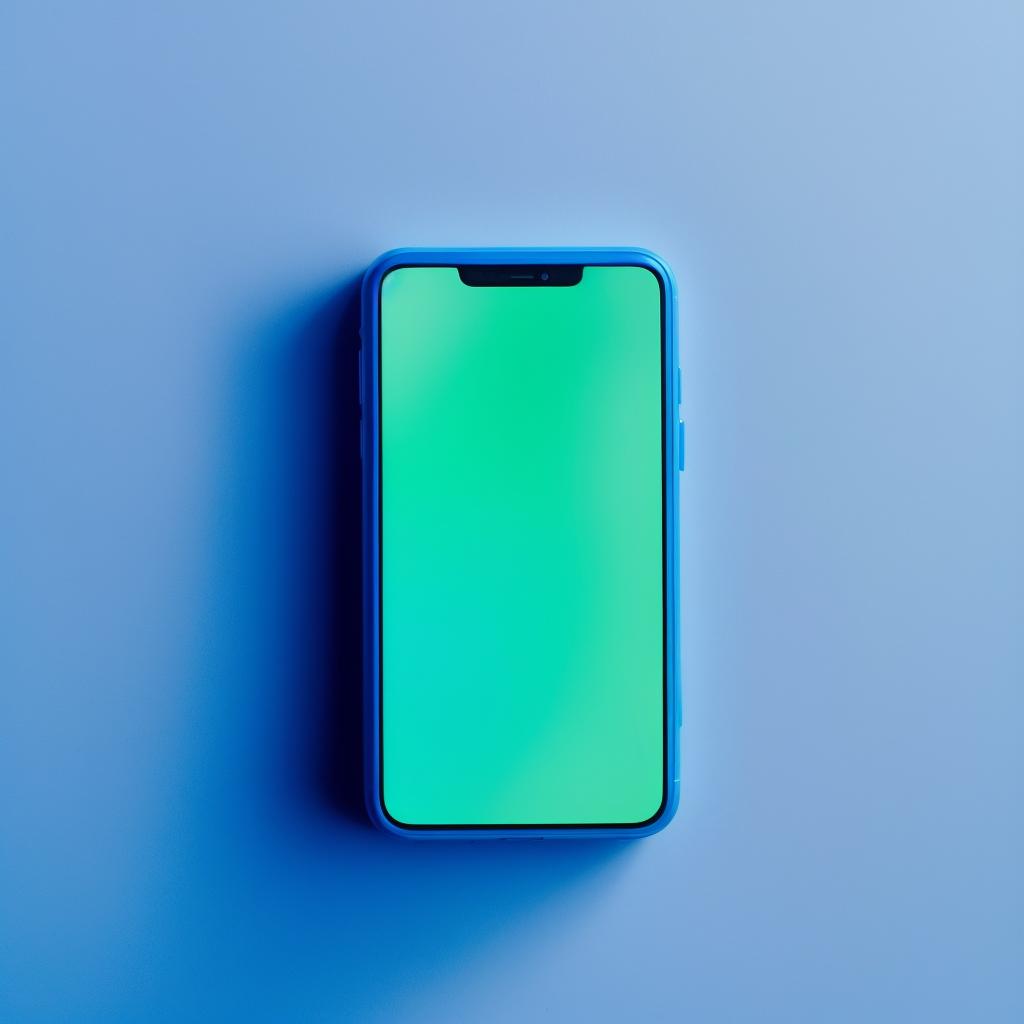} &
        \includegraphics[width=0.14\textwidth,height=0.14\textwidth]{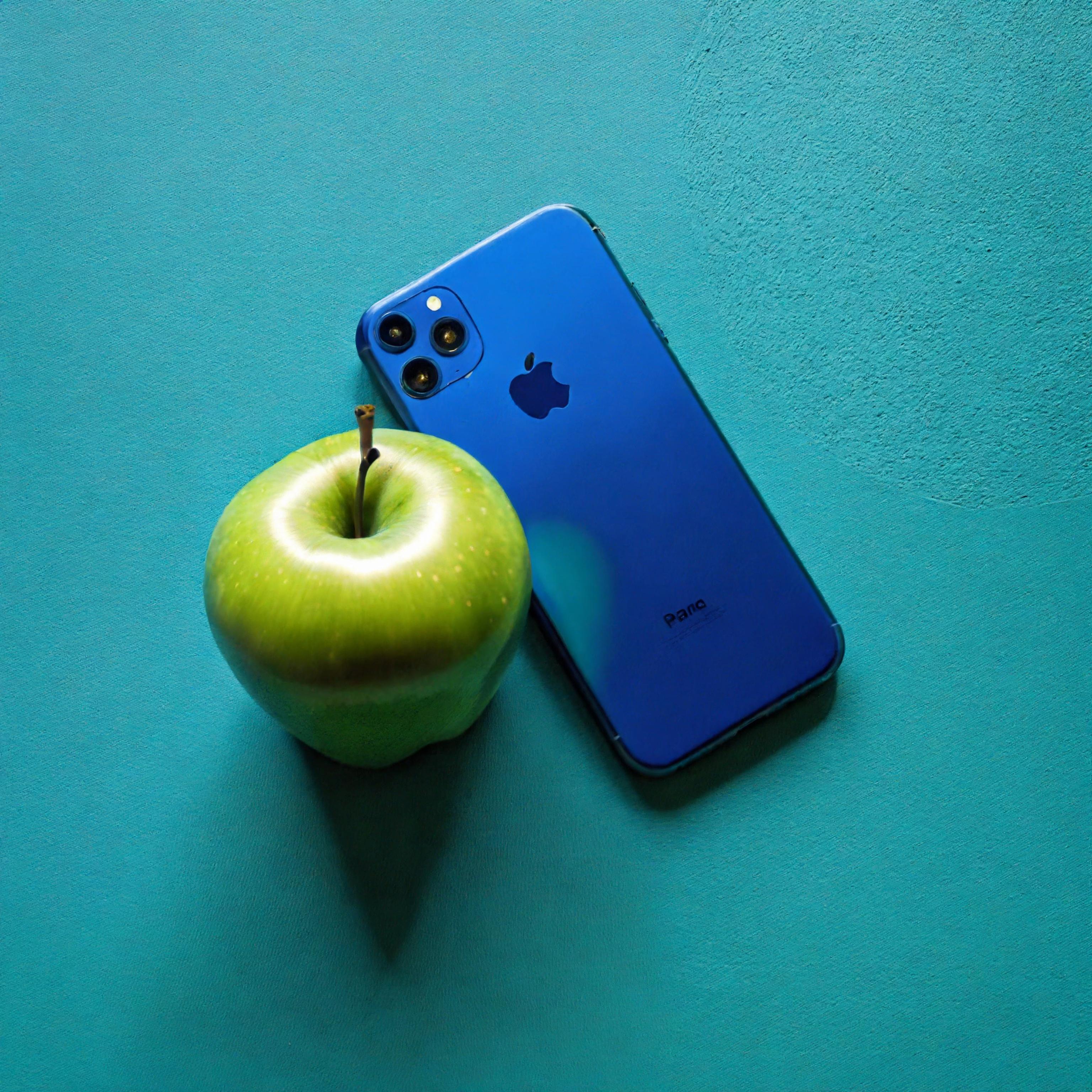} &
        \includegraphics[width=0.14\textwidth,height=0.14\textwidth]{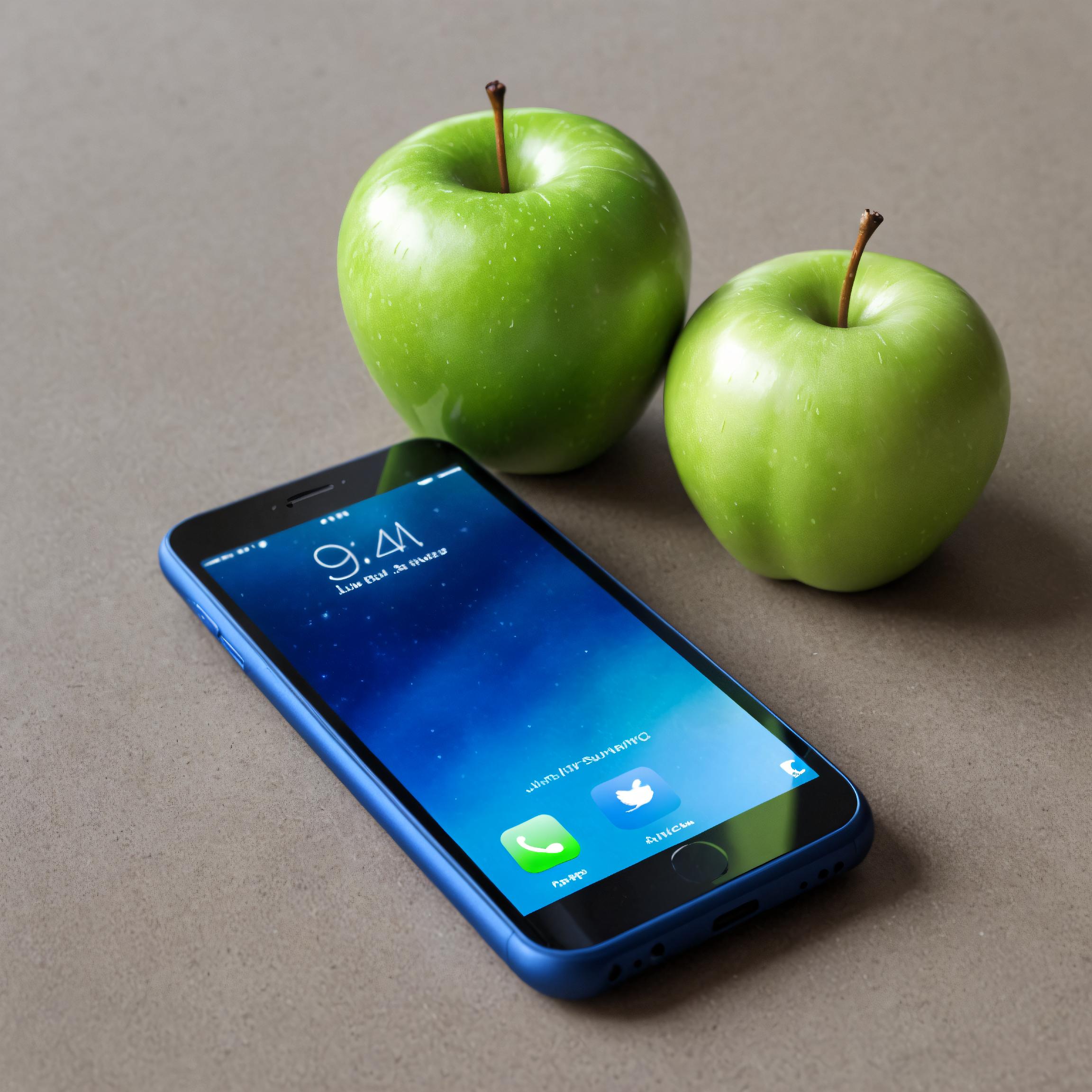} \\

        \multicolumn{7}{c}{ ``A photo of a blue cell phone and a green apple" } \\

    \end{tabular}
    
    }
    \caption{Qualitative results on GenEval prompts. ComfyGen shows better performance on multi-subject prompts, colorization and attribute binding, but may struggle with positioning.}\label{fig:geneval_qualitative}
\end{figure*}

To better evaluate the visual quality of our images, we turn back to CivitAI, and sample $500$ prompts from the $10,000$ highest-ranked images on the website, after filtering out prompts used for training our model and prompts that contain nudity or excessive violence. We assess the results both automatically --- using HPS V2.0, a model not included in the weighted score used during training --- and through a human preference study. 
For HPS, we follow~\citet{wallace2024diffusion,qi2024not} and perform a pair-wise comparison, each time pitting our method against a baseline using the same prompt. We report the fraction of prompts for which our approach received a higher score.

For the user study, we pit each version of our model against each baseline in a two-alternative forced-choice setup and report the fraction of times our model was preferred over each baseline. We sample roughly $20$ prompts for each baseline and ComfyGen version pair, for a total of $231$ questions. We collected a total of $682$ responses from $35$ users. More details are provided in the supplementary.

\begin{figure}[!h]
    \centering
    \setlength{\belowcaptionskip}{-6pt}
    \includegraphics[width=\linewidth]{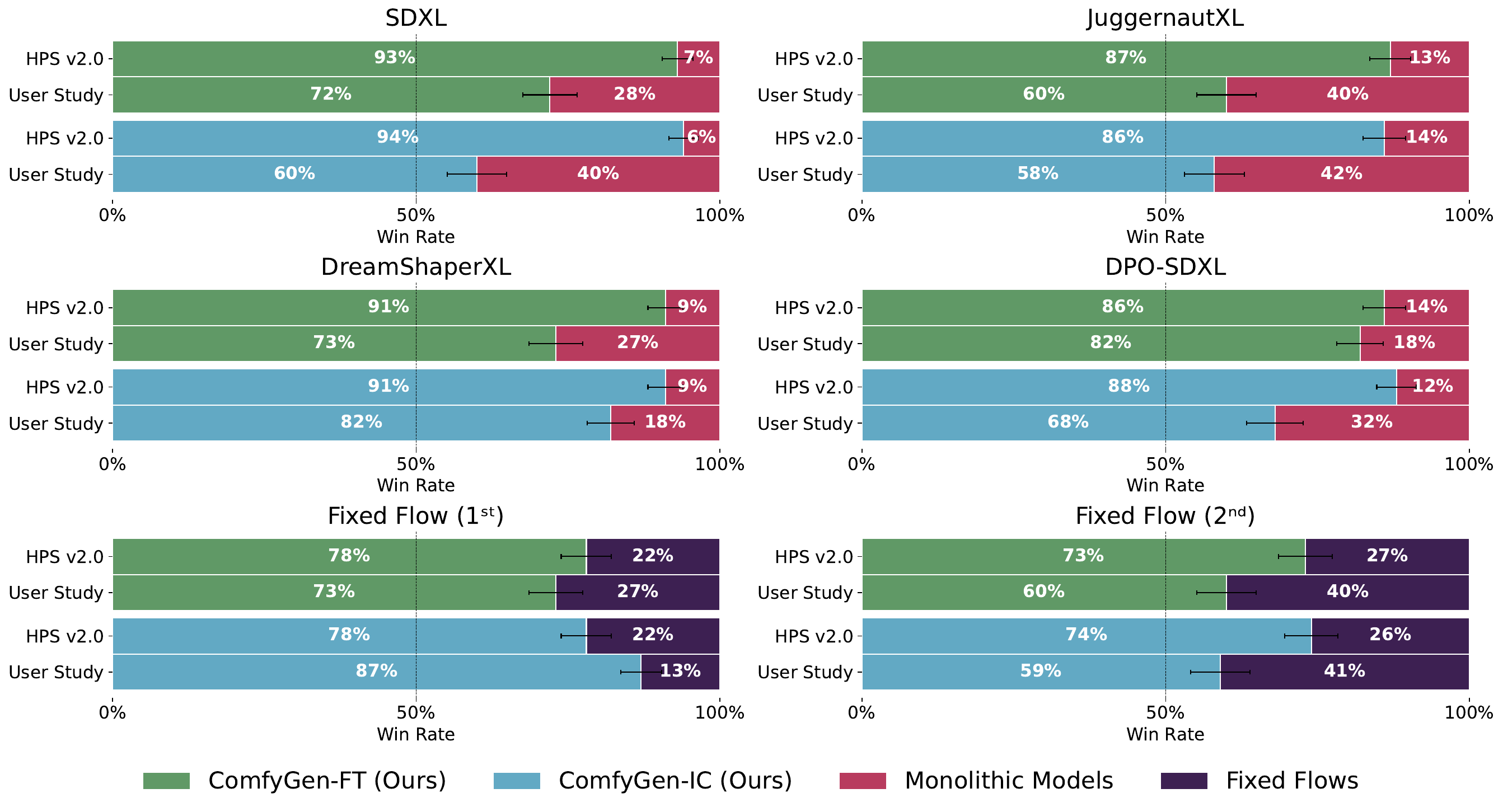}
    \caption{HPS V2.0 and User Study win rates. We compare each baseline against both ComfyGen-FT (green) and ComfyGen-IC (teal). ComfyGen variants are favored over all baselines.}
    \label{fig:usr_study}
\end{figure}

\begin{figure*}
    \centering
    \setlength{\belowcaptionskip}{-8pt}
    \setlength{\abovecaptionskip}{3pt}
    \setlength{\tabcolsep}{0.5pt}
    {\normalsize
    \begin{tabular}{c c c c c c c}
    
        {\small SDXL} & {\small Juggernaut} & {\small DreamShaper} & {\small Flow 1} & {\small Flow 2} & {\small \ourmethod{}-IC} & {\small \ourmethod{}-FT} \\

        \includegraphics[width=0.147\textwidth,height=0.147\textwidth]{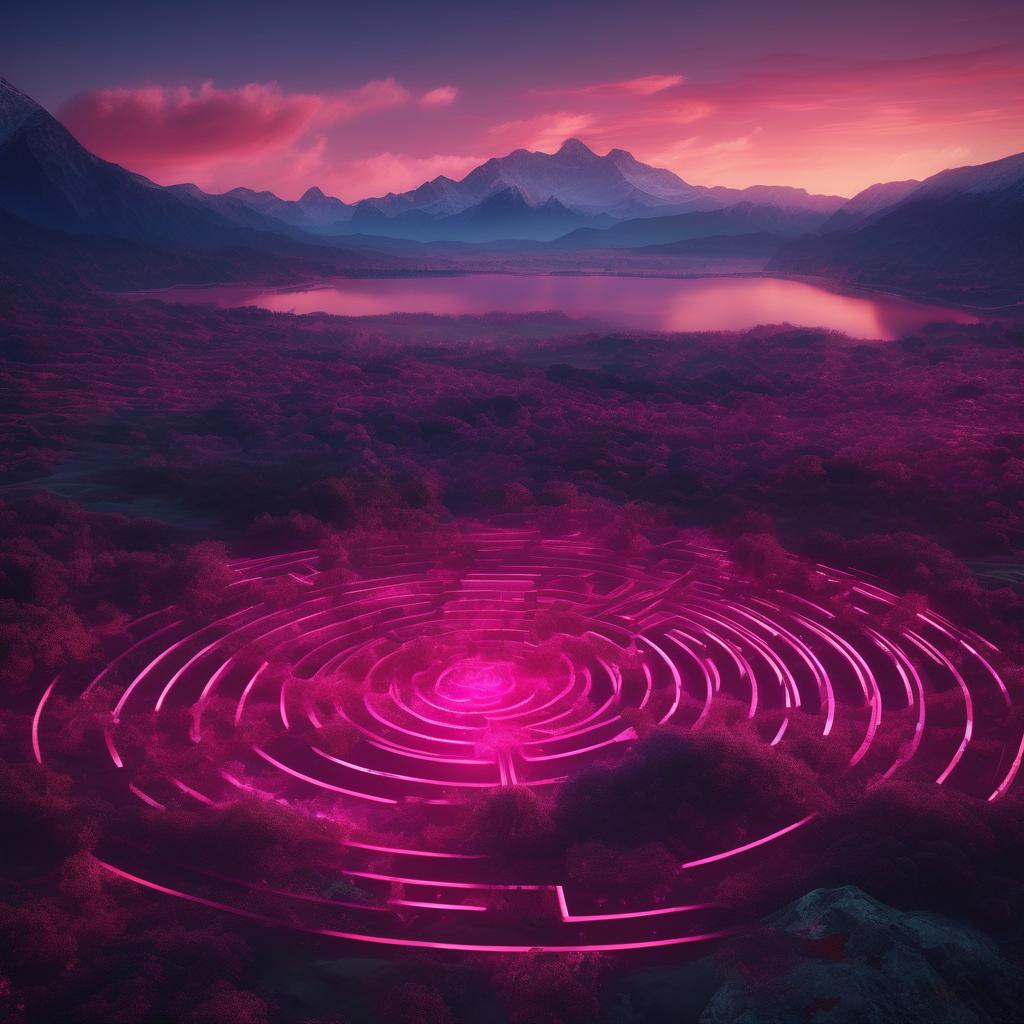} &
        \includegraphics[width=0.147\textwidth,height=0.147\textwidth]{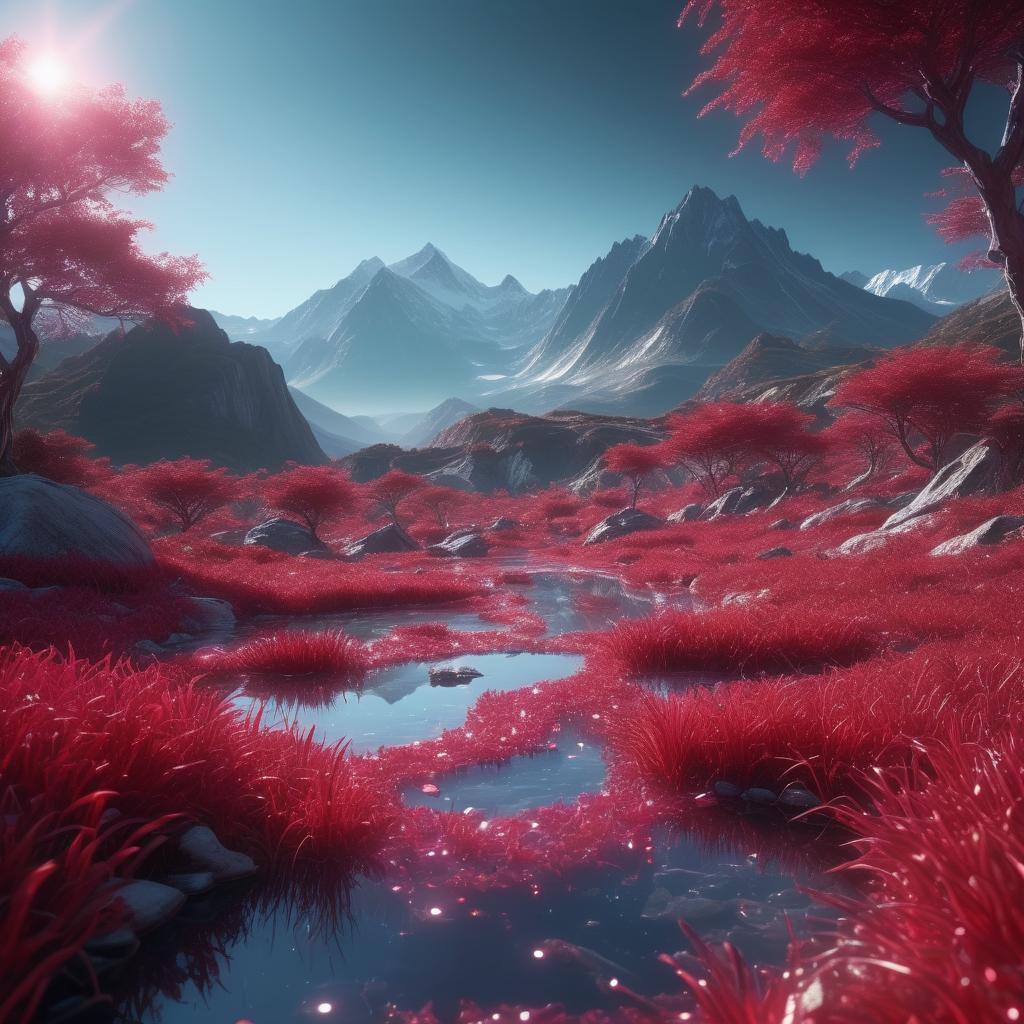} &
        \includegraphics[width=0.147\textwidth,height=0.147\textwidth]{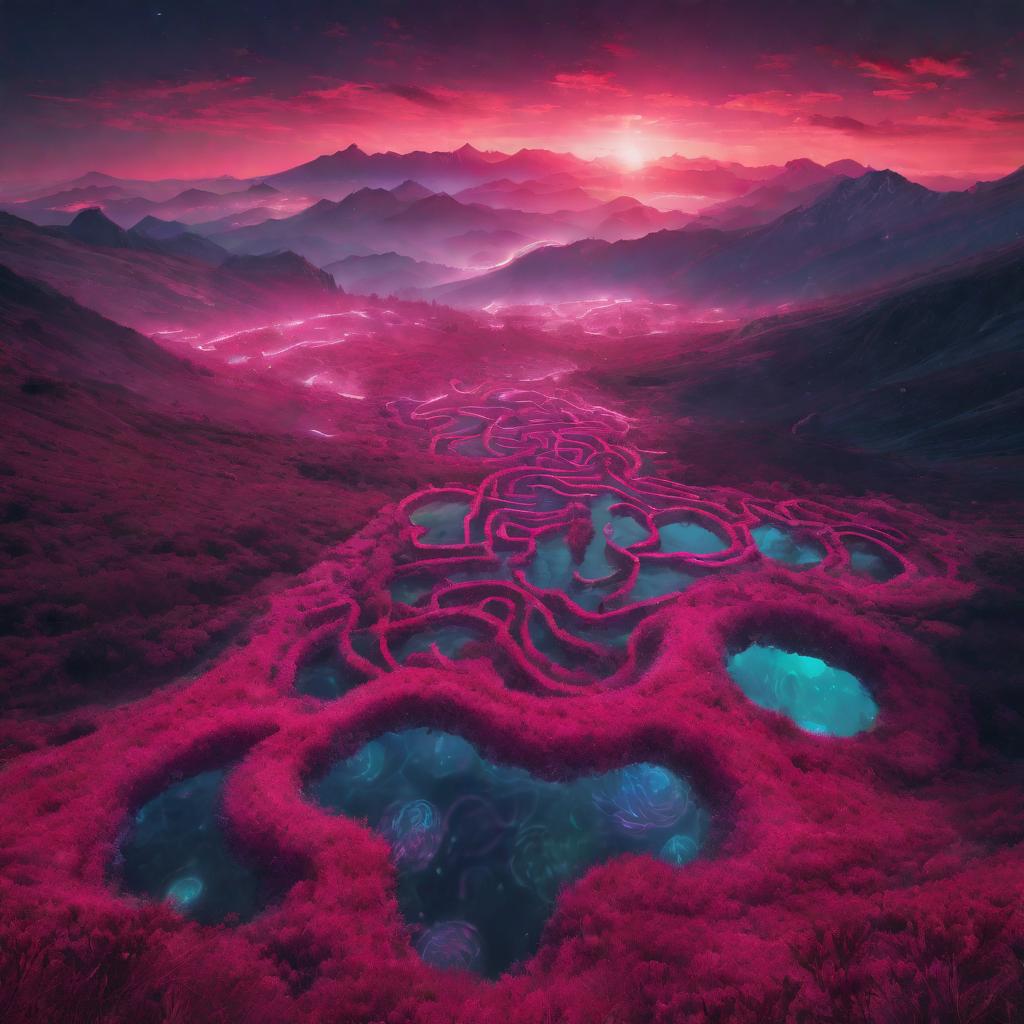} &
        \includegraphics[width=0.147\textwidth,height=0.147\textwidth]{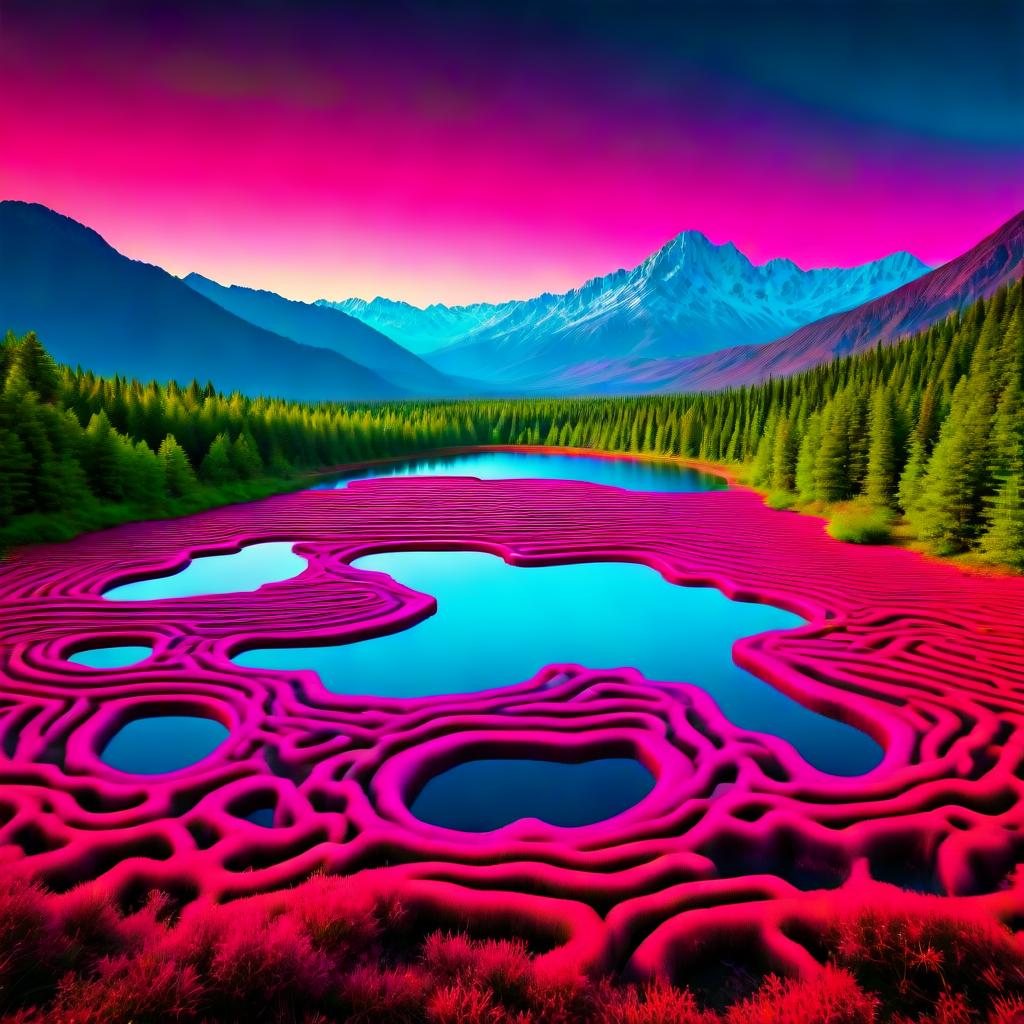} &
        \includegraphics[width=0.147\textwidth,height=0.147\textwidth]{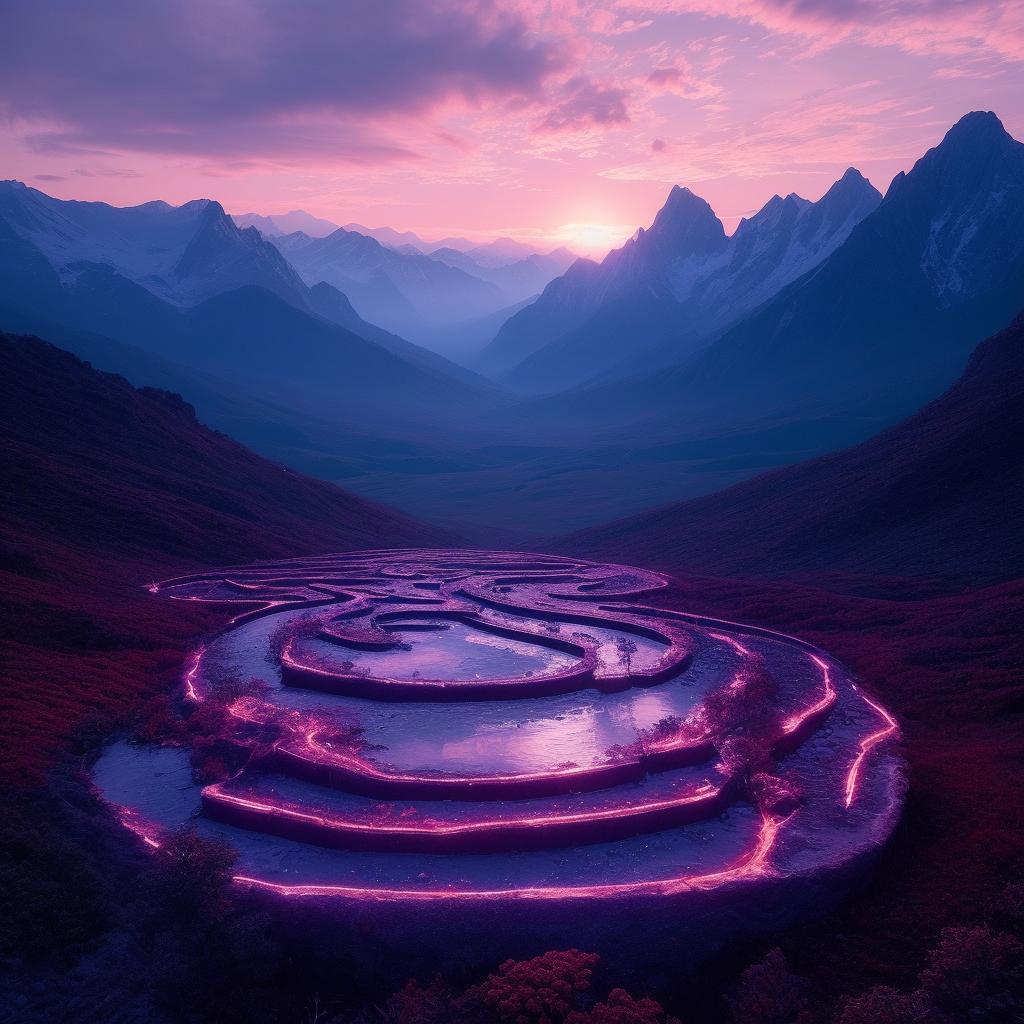} &
        \includegraphics[width=0.147\textwidth,height=0.147\textwidth]{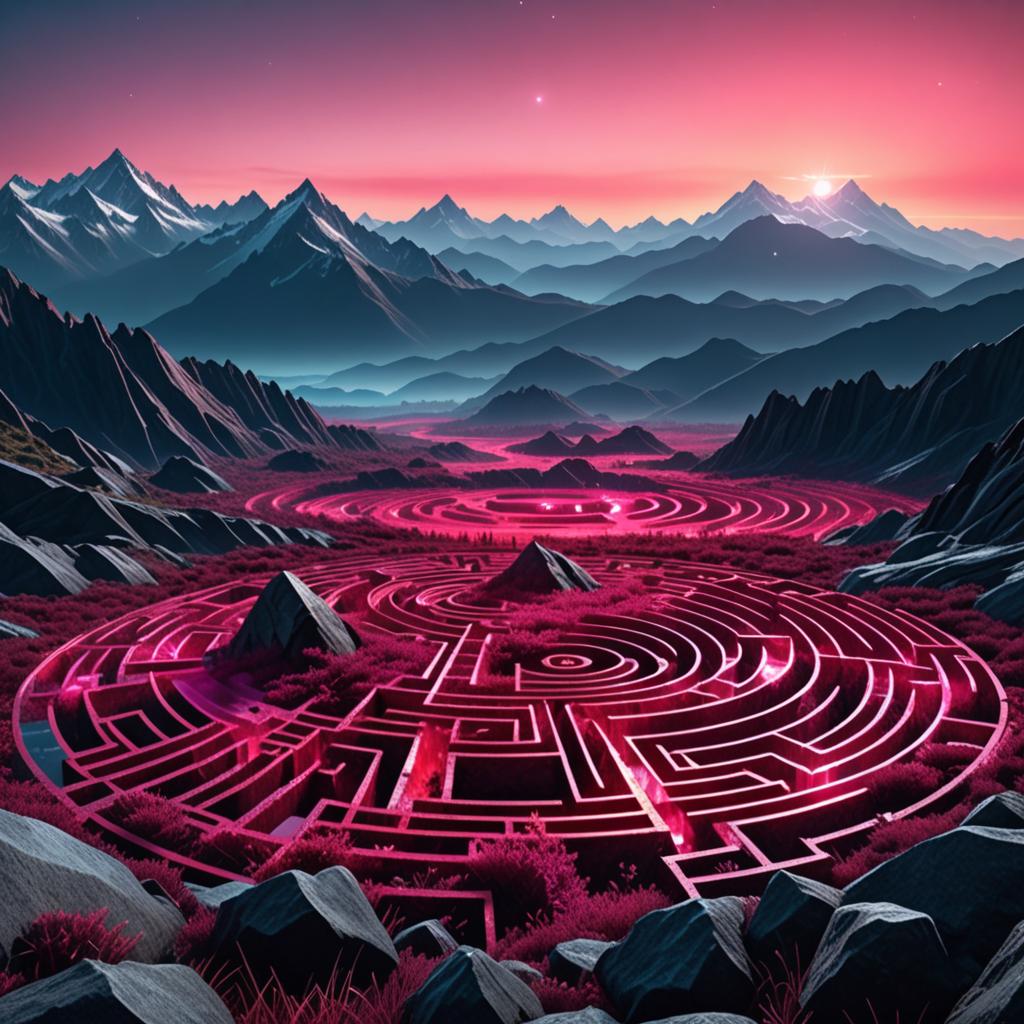} &
        \includegraphics[width=0.147\textwidth,height=0.147\textwidth]{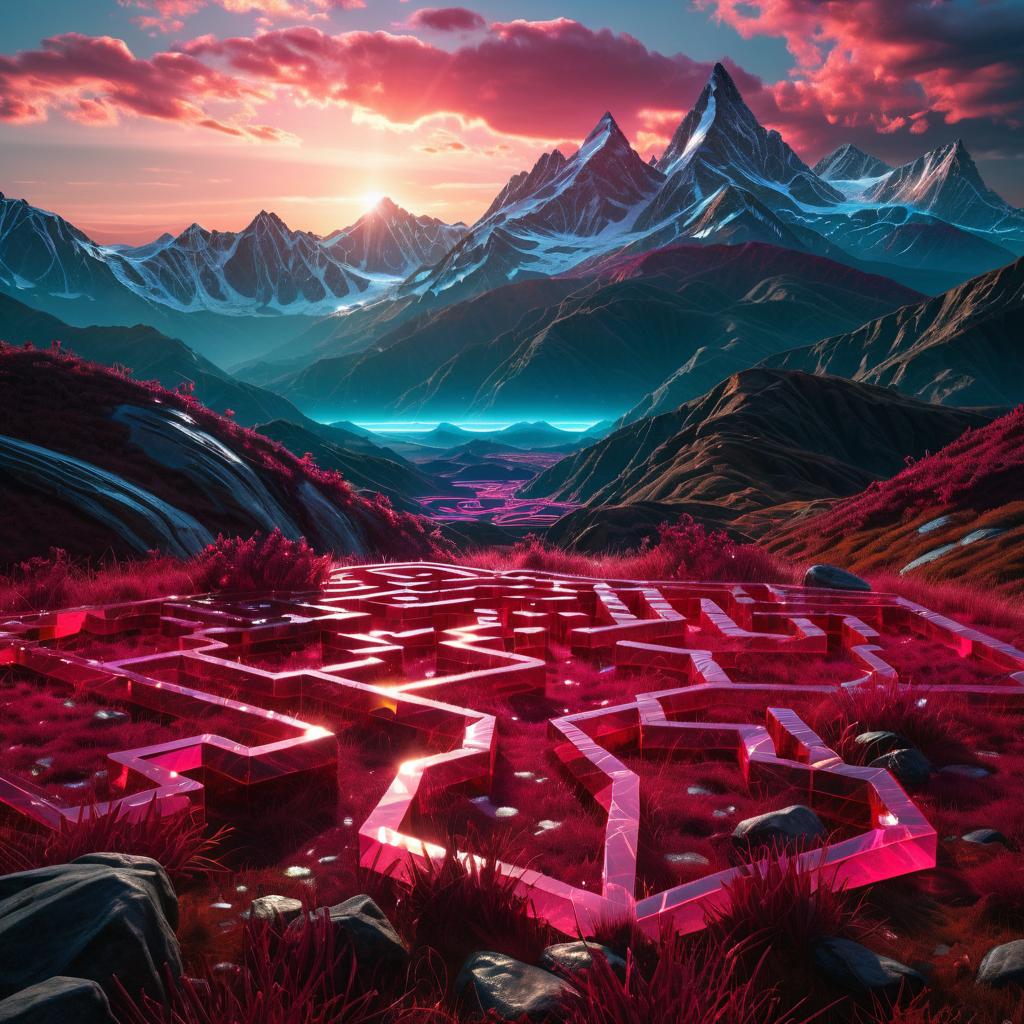}  \\

        \multicolumn{7}{c}{\begin{tabular}{@{}c@{}} \scriptsize ``8k digital nature photography, an idyllic landscape with mountains in the distance, shiny ruby see-through, Sprawling labyrinth, bioluminescent" \end{tabular}  } \\

        \includegraphics[width=0.147\textwidth,height=0.147\textwidth]{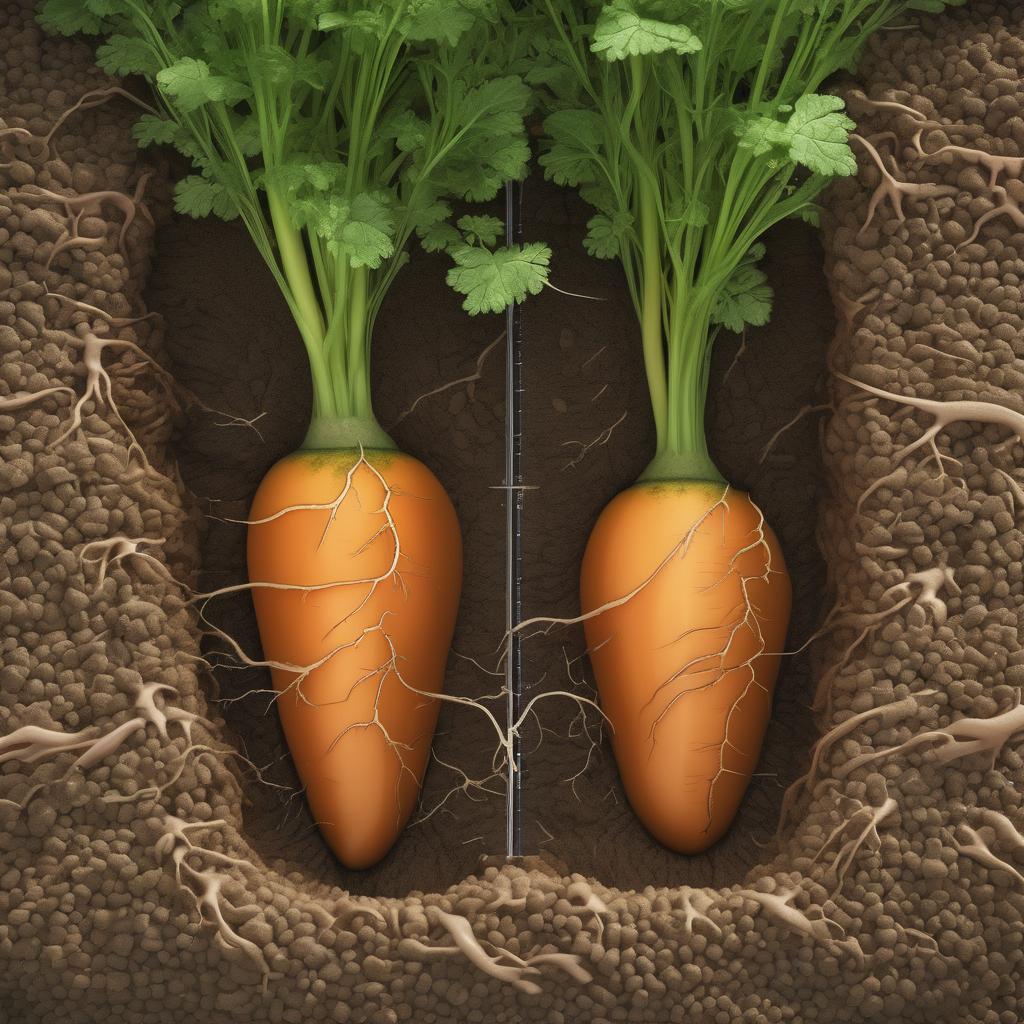} &
        \includegraphics[width=0.147\textwidth,height=0.147\textwidth]{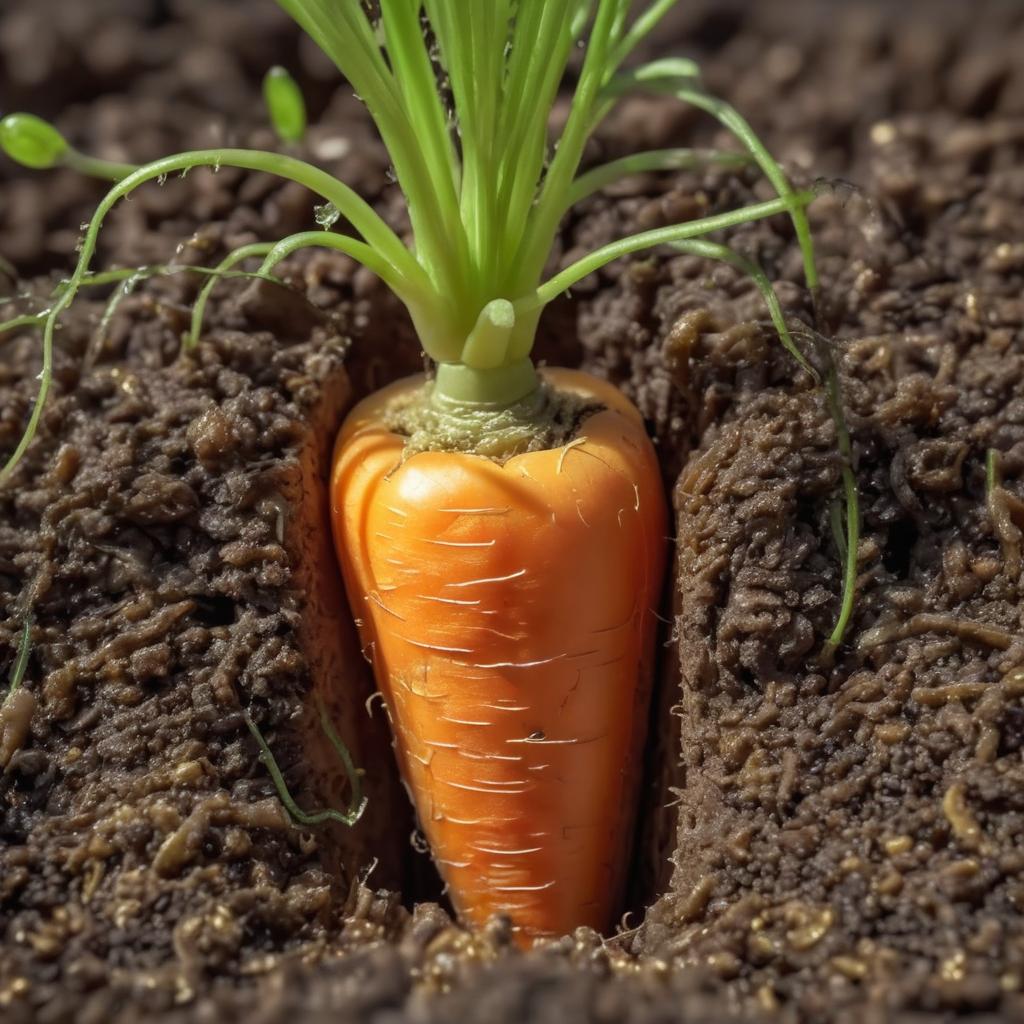} &
        \includegraphics[width=0.147\textwidth,height=0.147\textwidth]{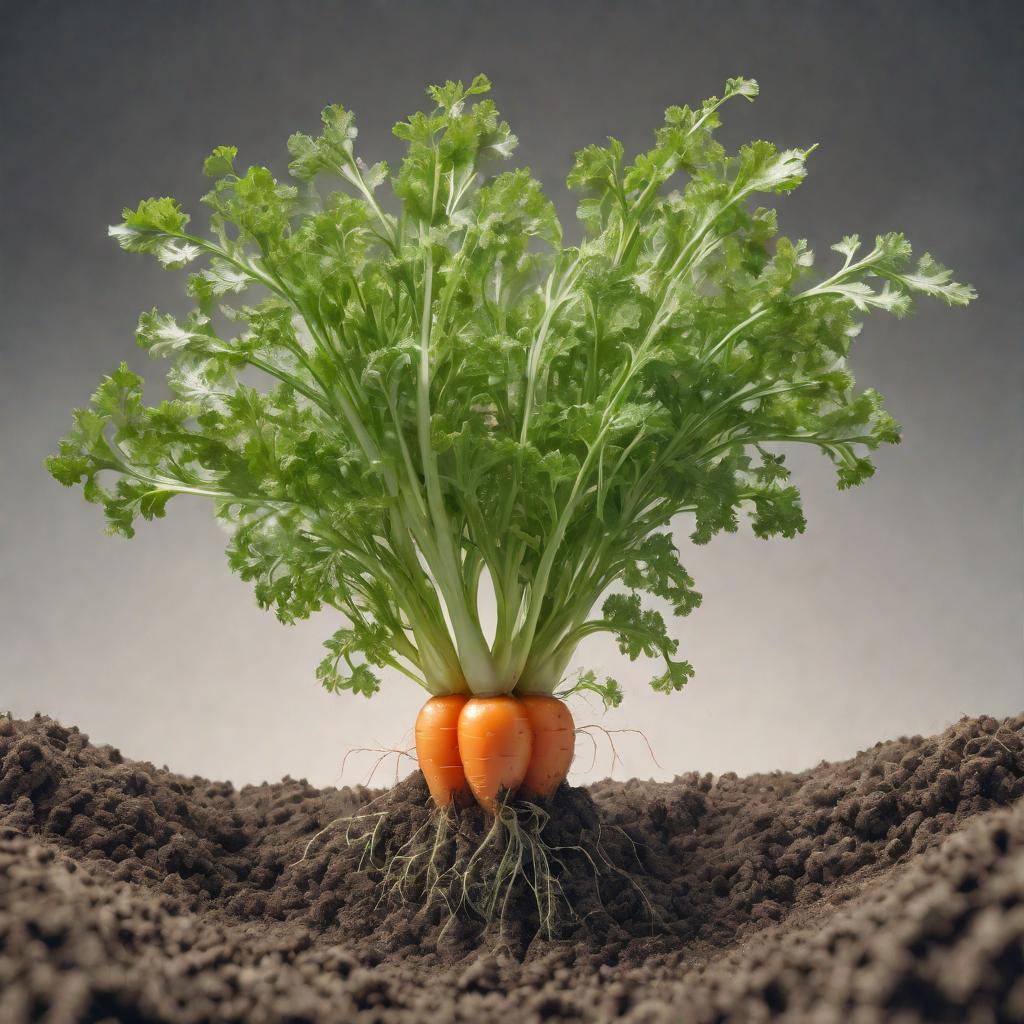} &
        \includegraphics[width=0.147\textwidth,height=0.147\textwidth]{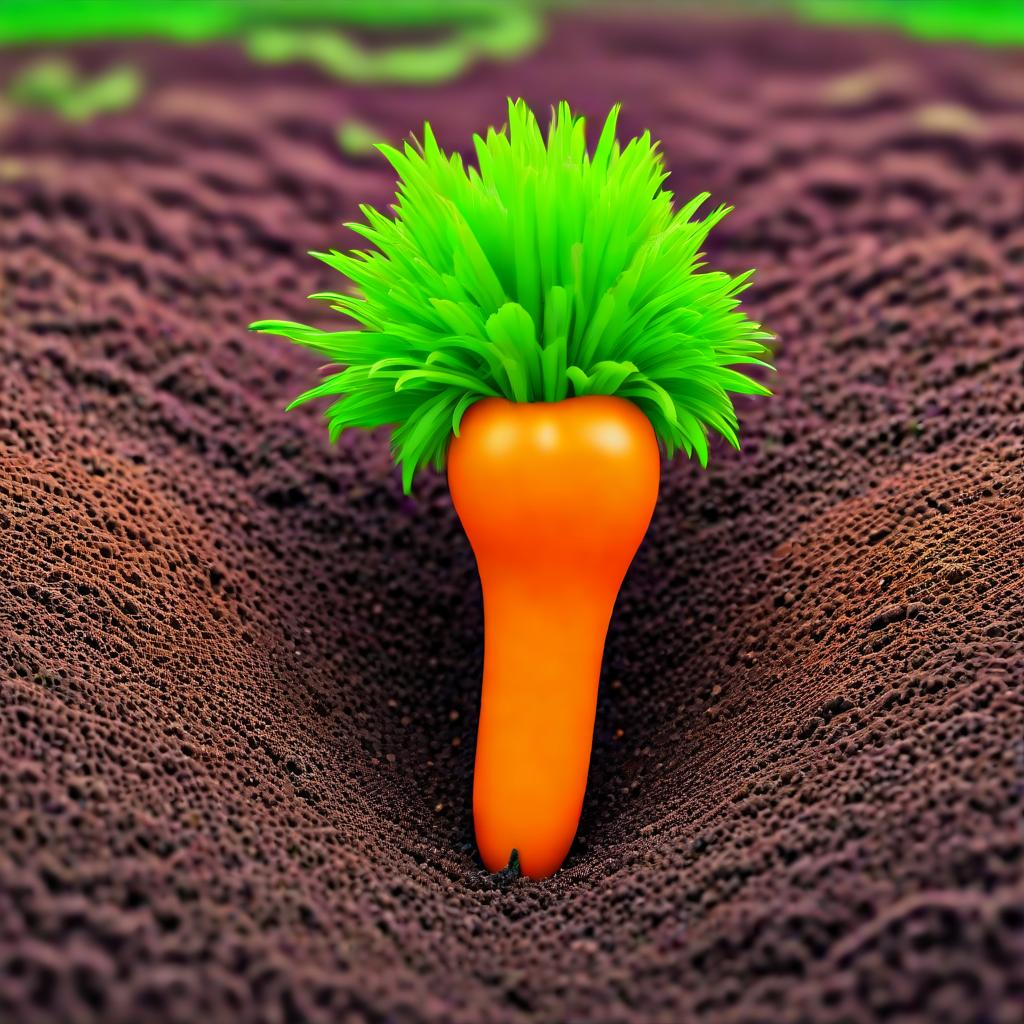} &
        \includegraphics[width=0.147\textwidth,height=0.147\textwidth]{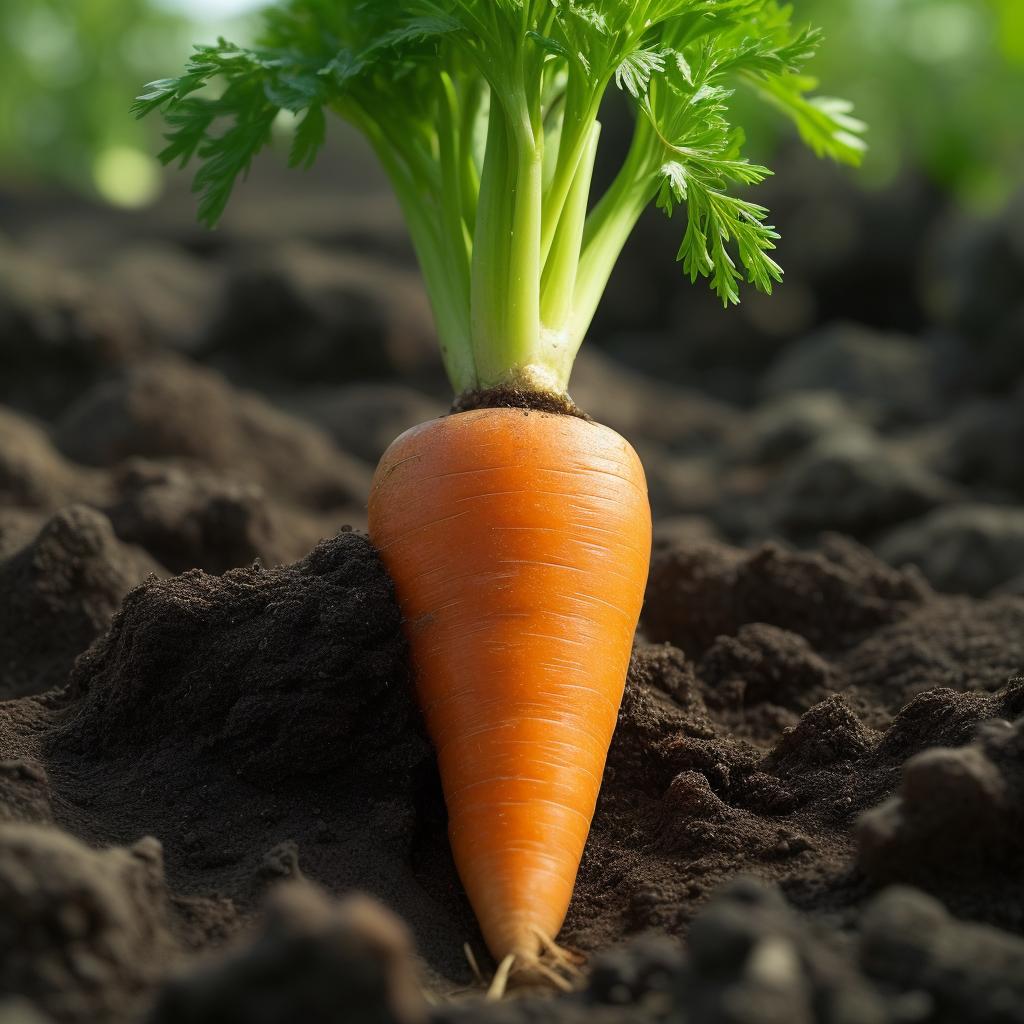} &
        \includegraphics[width=0.147\textwidth,height=0.147\textwidth]{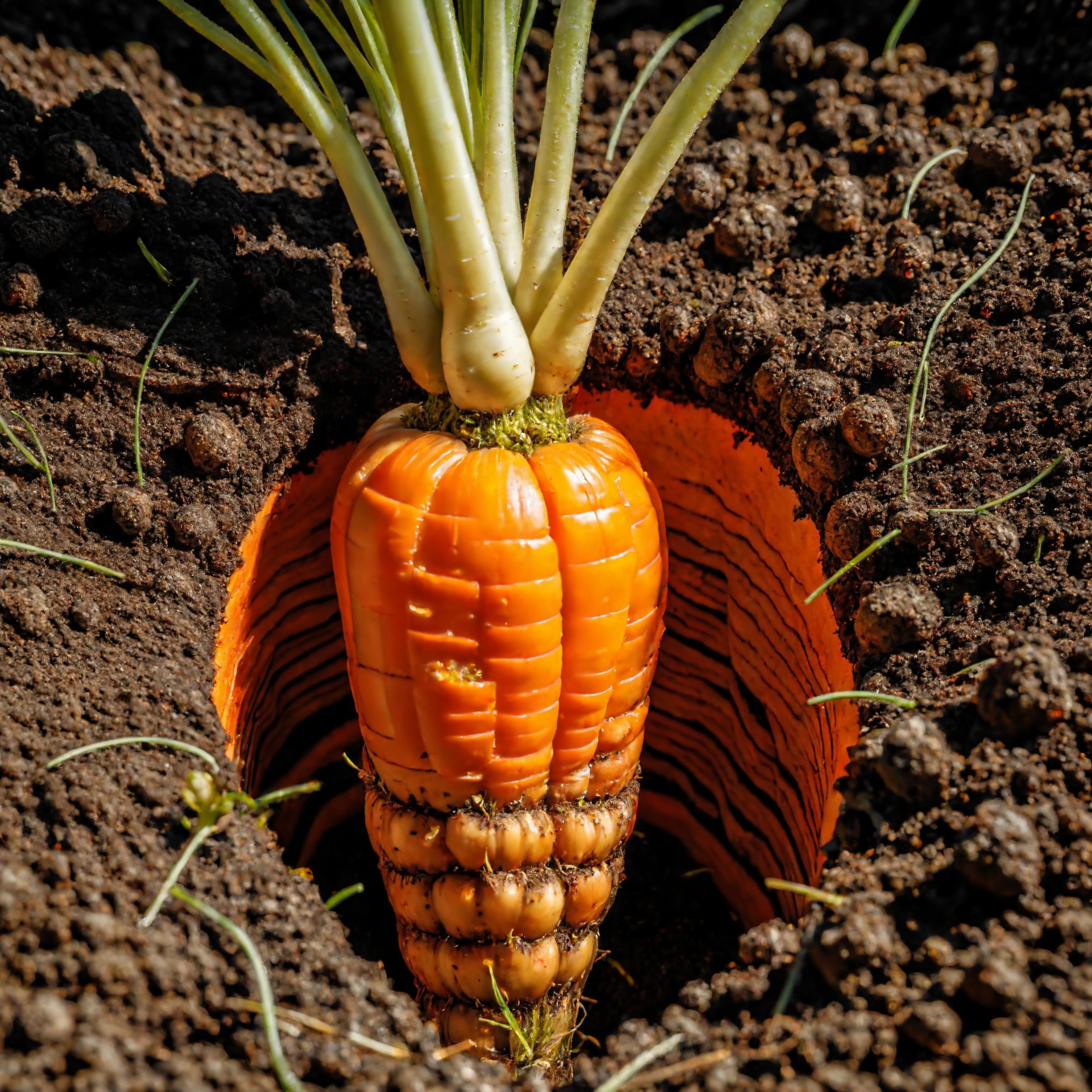} &
        \includegraphics[width=0.147\textwidth,height=0.147\textwidth]{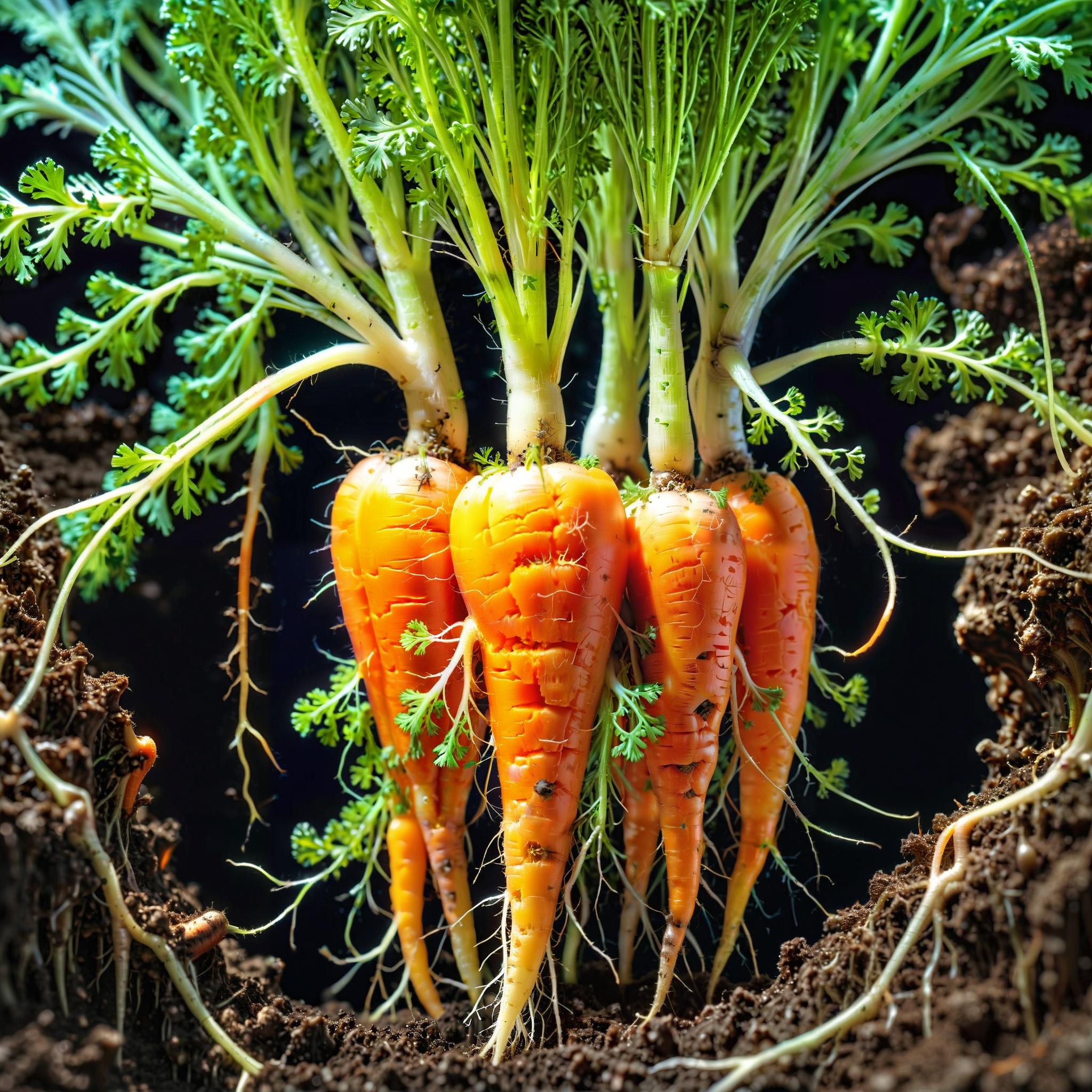}  \\

        \multicolumn{7}{c}{\begin{tabular}{@{}c@{}} \scriptsize ``masterpiece, intricate detail, 8K, HDR, Cross section of a carrot growing in the ground. \\ \scriptsize A scientific perspective on the cross section of a carrot growing" \end{tabular}  } \\

        \includegraphics[width=0.147\textwidth,height=0.147\textwidth]{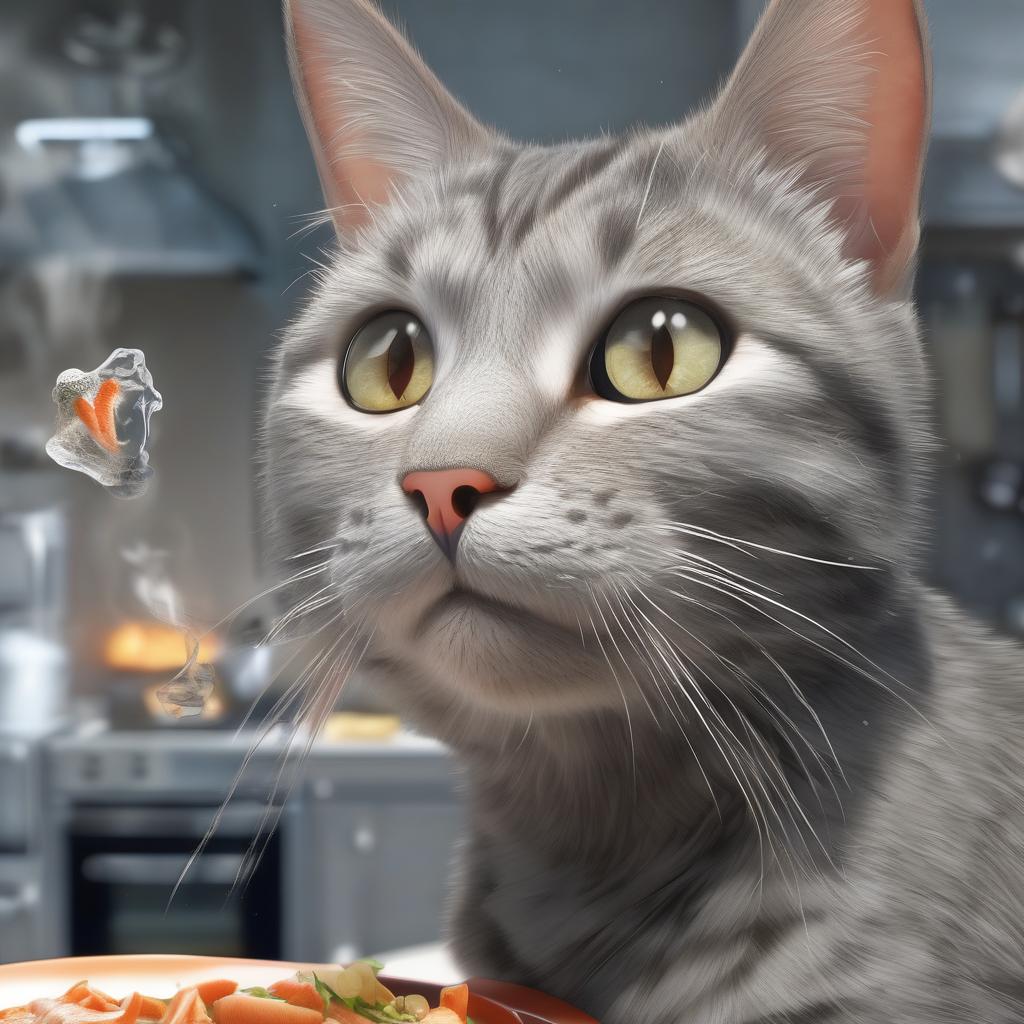} &
        \includegraphics[width=0.147\textwidth,height=0.147\textwidth]{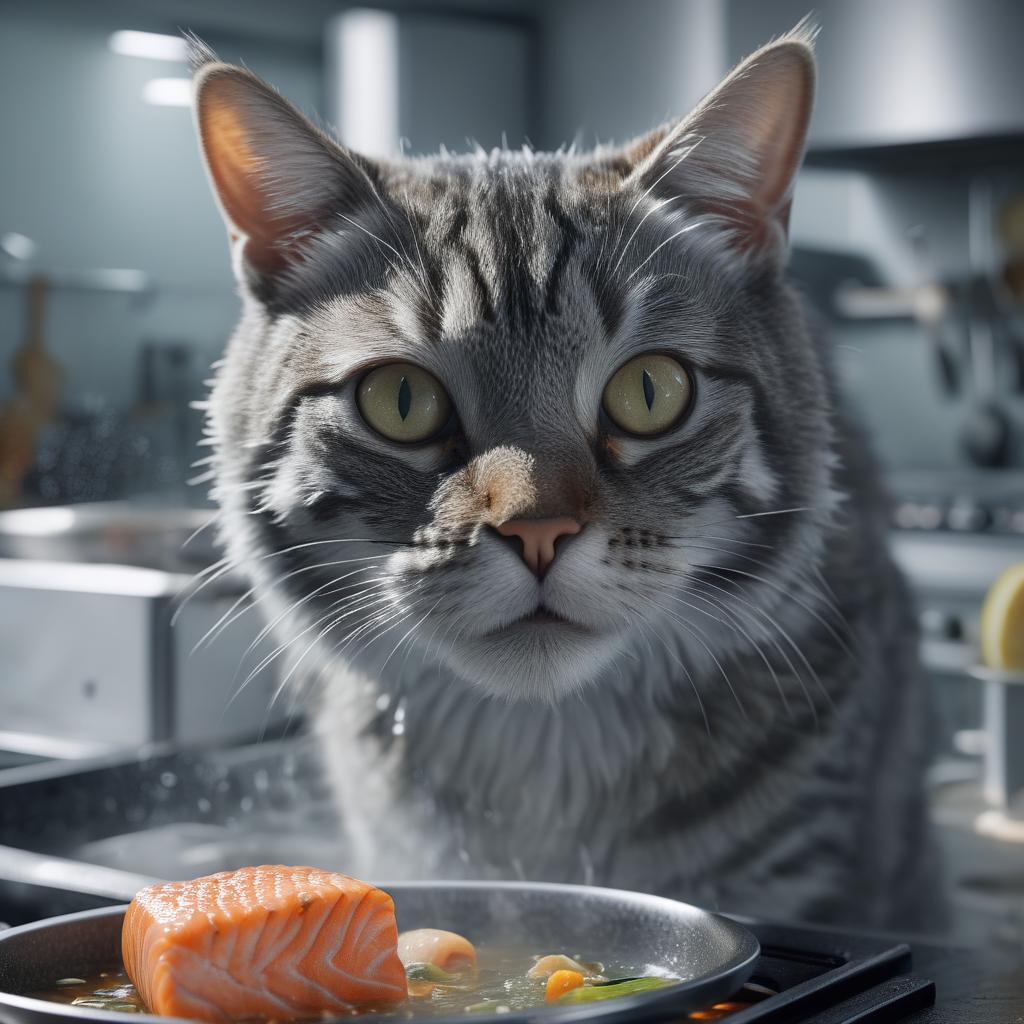} &
        \includegraphics[width=0.147\textwidth,height=0.147\textwidth]{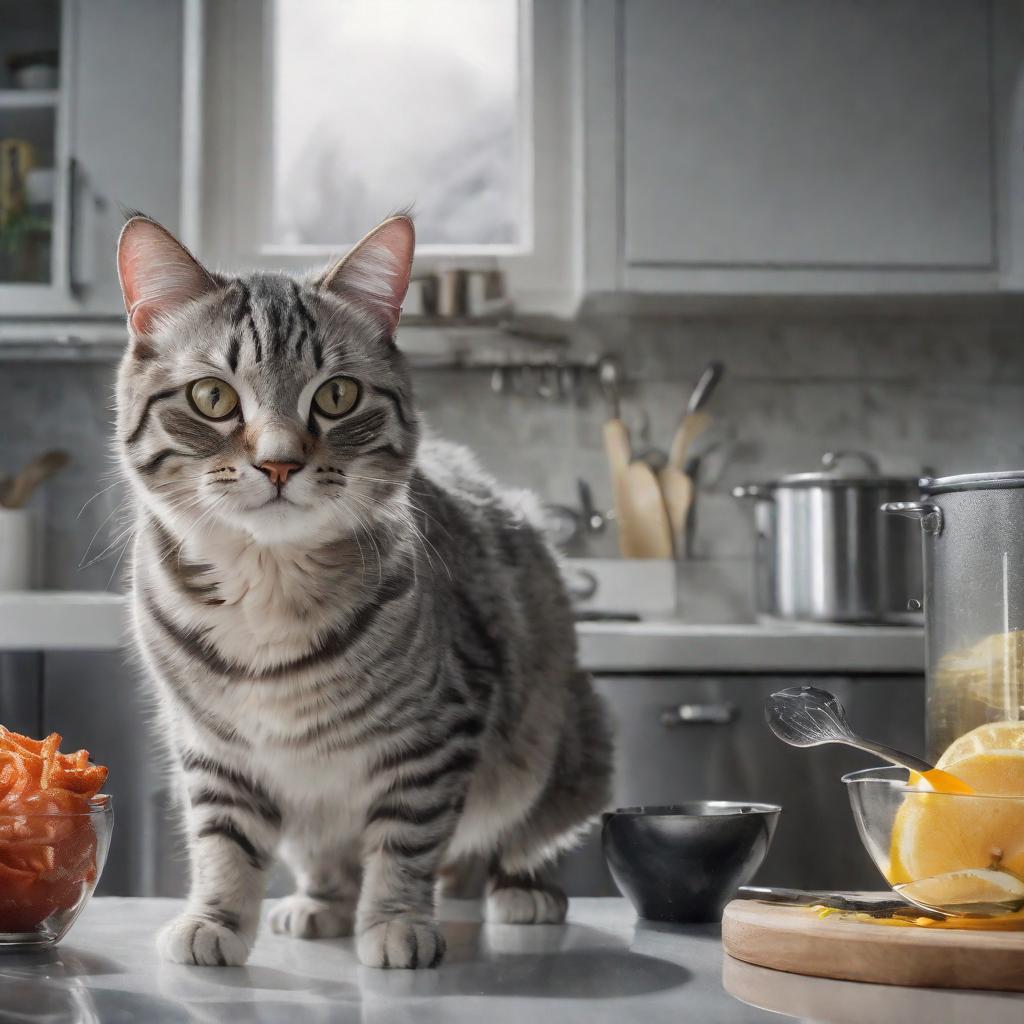} &
        \includegraphics[width=0.147\textwidth,height=0.147\textwidth]{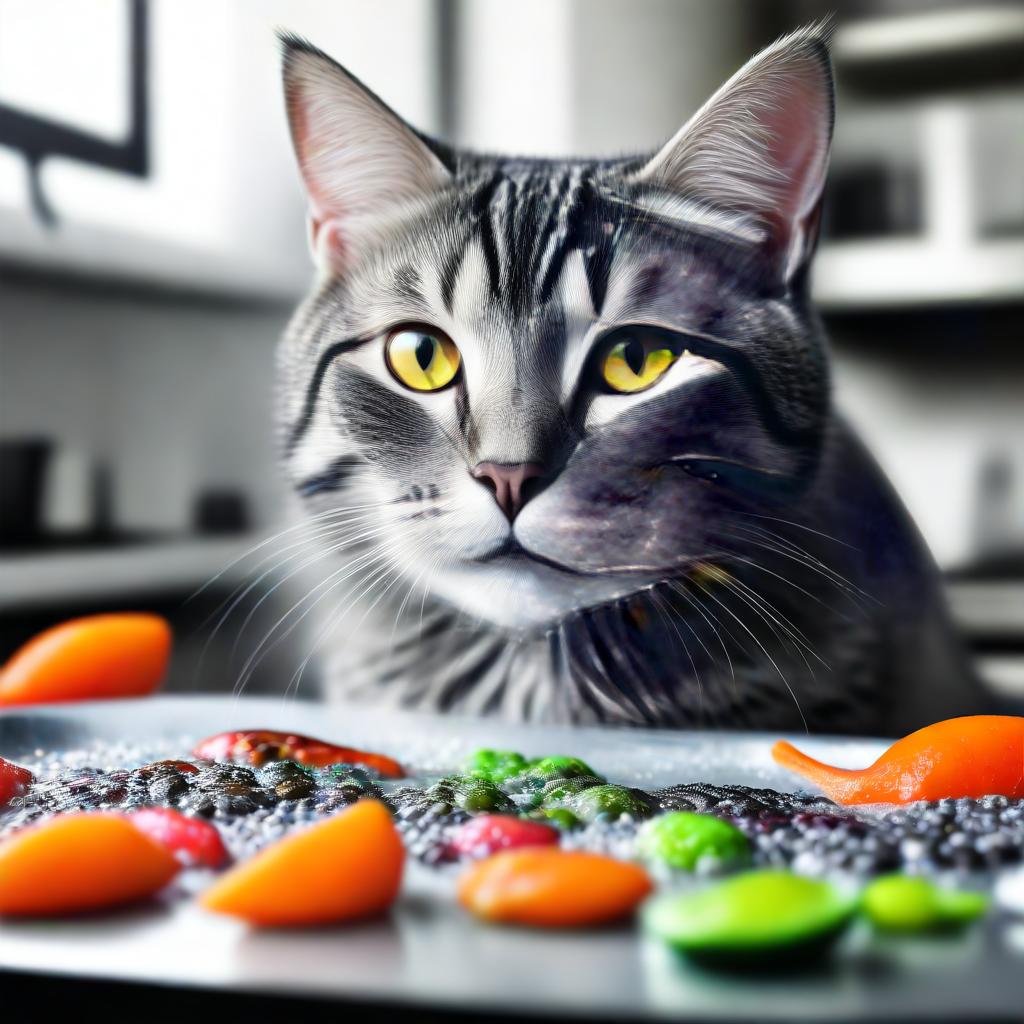} &
        \includegraphics[width=0.147\textwidth,height=0.147\textwidth]{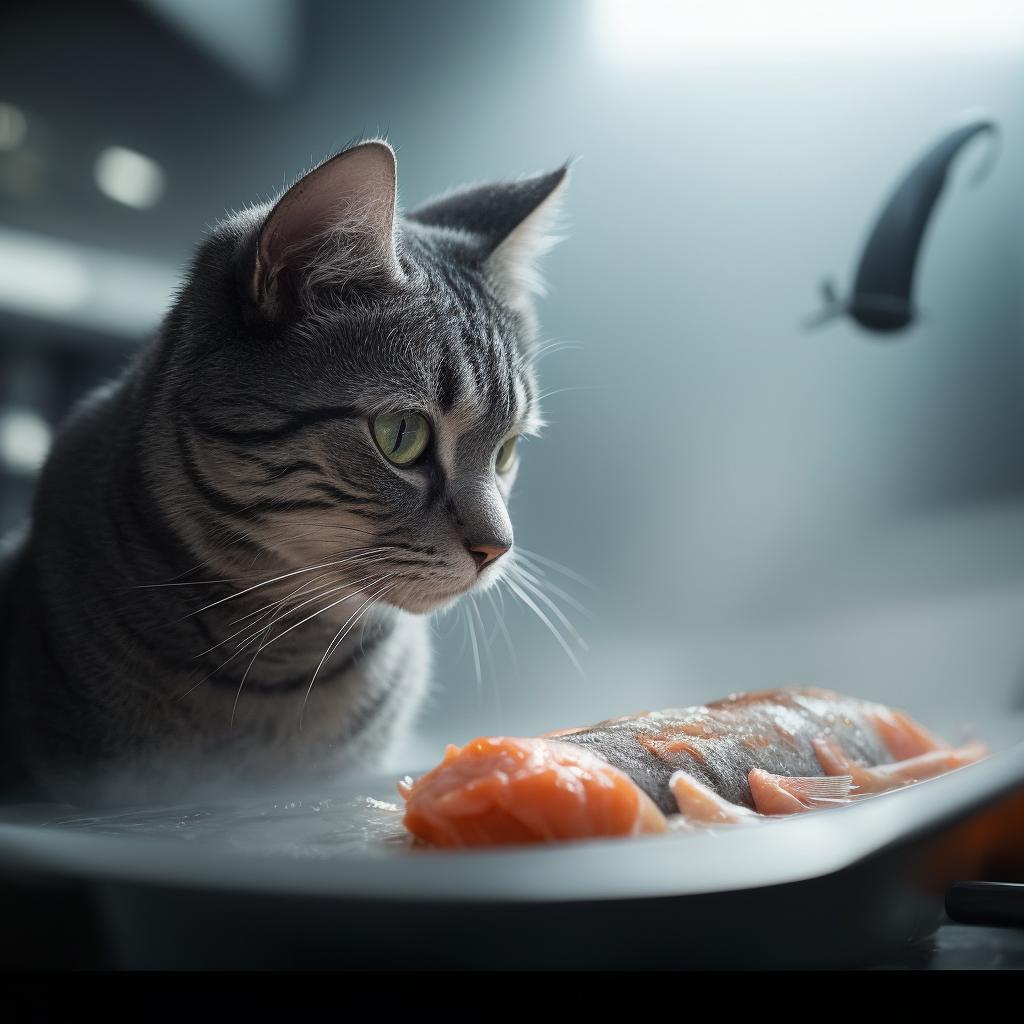} &
        \includegraphics[width=0.147\textwidth,height=0.147\textwidth]{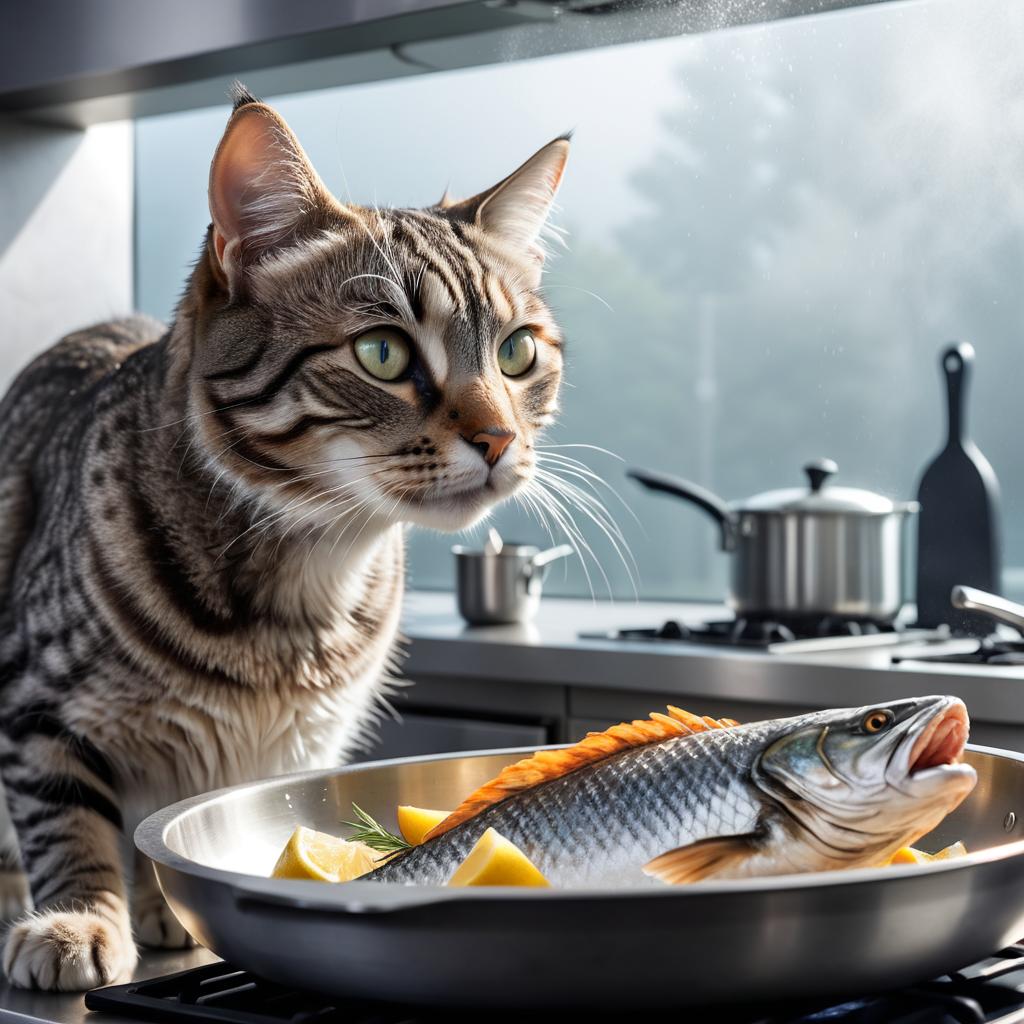} &
        \includegraphics[width=0.147\textwidth,height=0.147\textwidth]{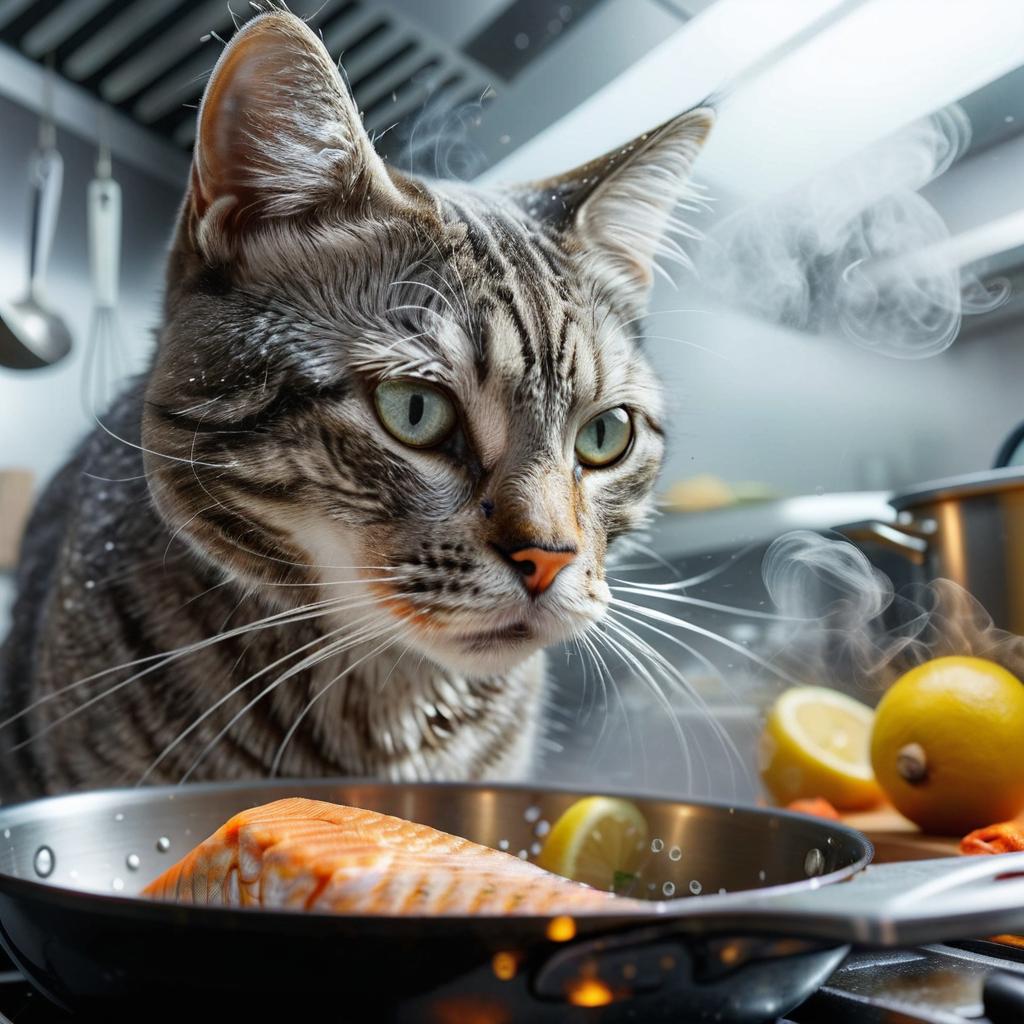}  \\

        \multicolumn{7}{c}{\begin{tabular}{@{}c@{}} \scriptsize ``close-up photography of grey tabby cat, cooking fish, c4ttitude, in glasstech kitchen, hyper realistic, intricate detail, foggy, pov from below" \end{tabular}  } \\

        \includegraphics[width=0.147\textwidth,height=0.147\textwidth]{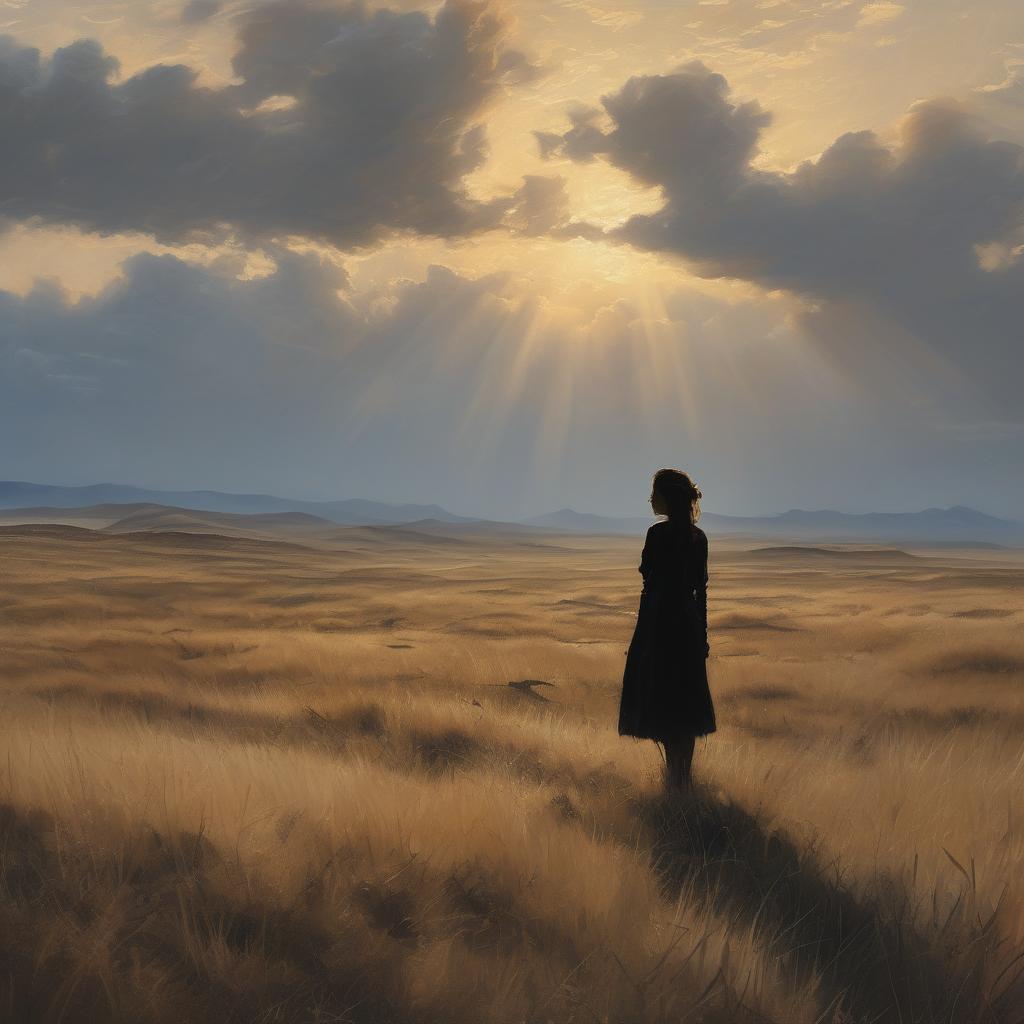} &
        \includegraphics[width=0.147\textwidth,height=0.147\textwidth]{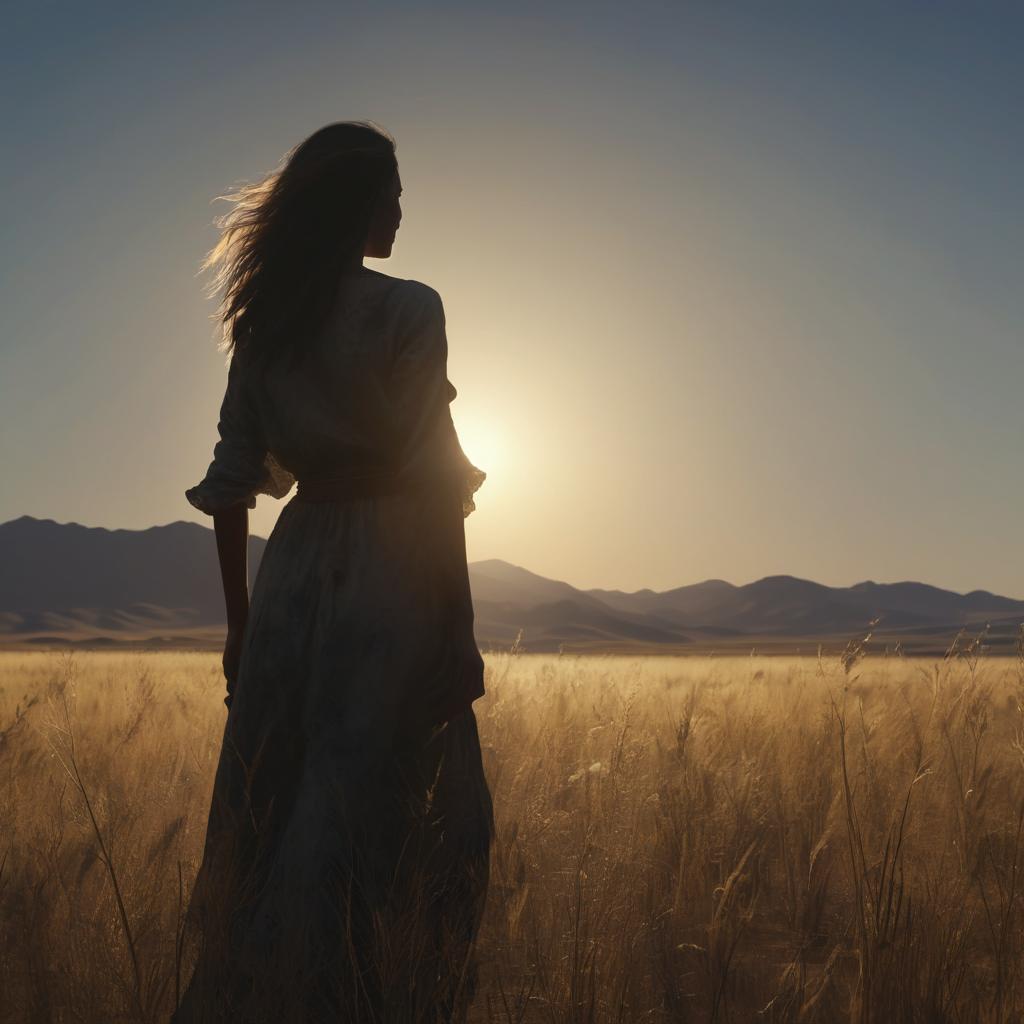} &
        \includegraphics[width=0.147\textwidth,height=0.147\textwidth]{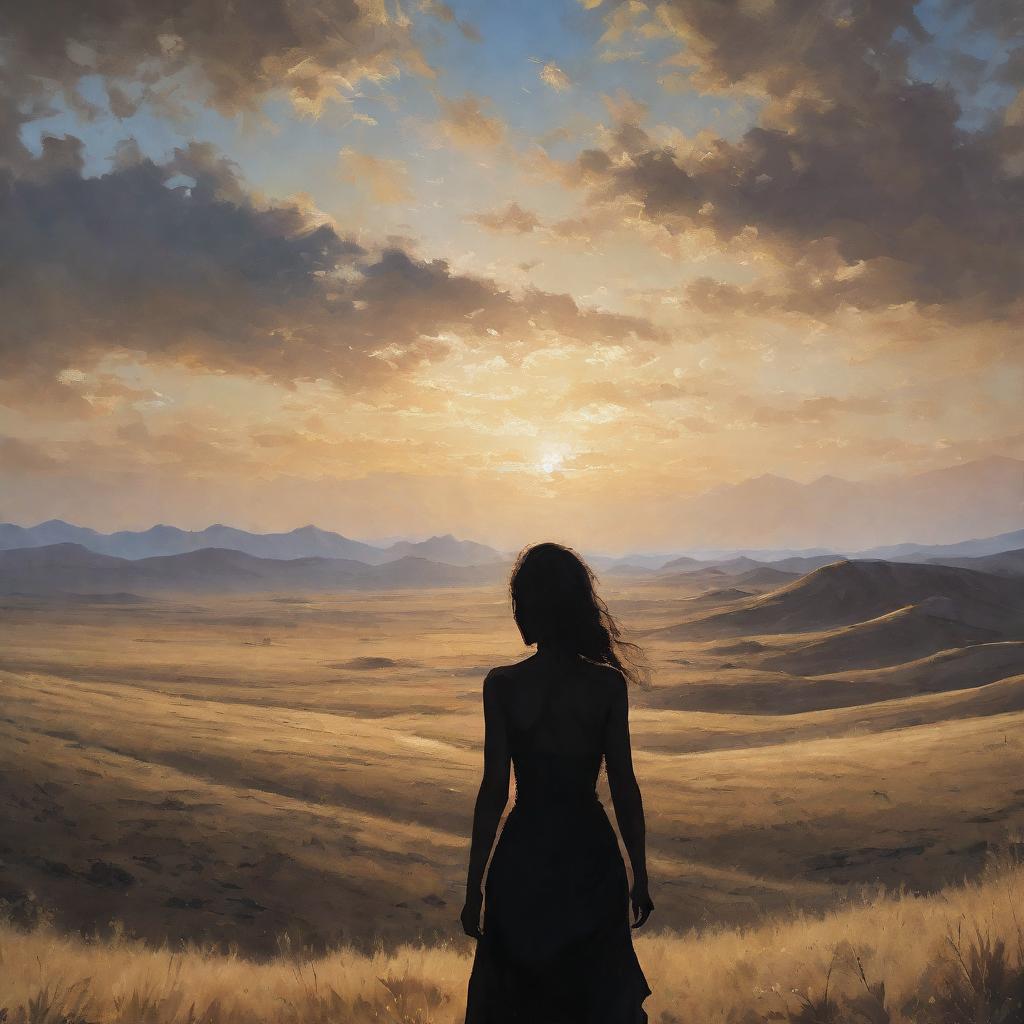} &
        \includegraphics[width=0.147\textwidth,height=0.147\textwidth]{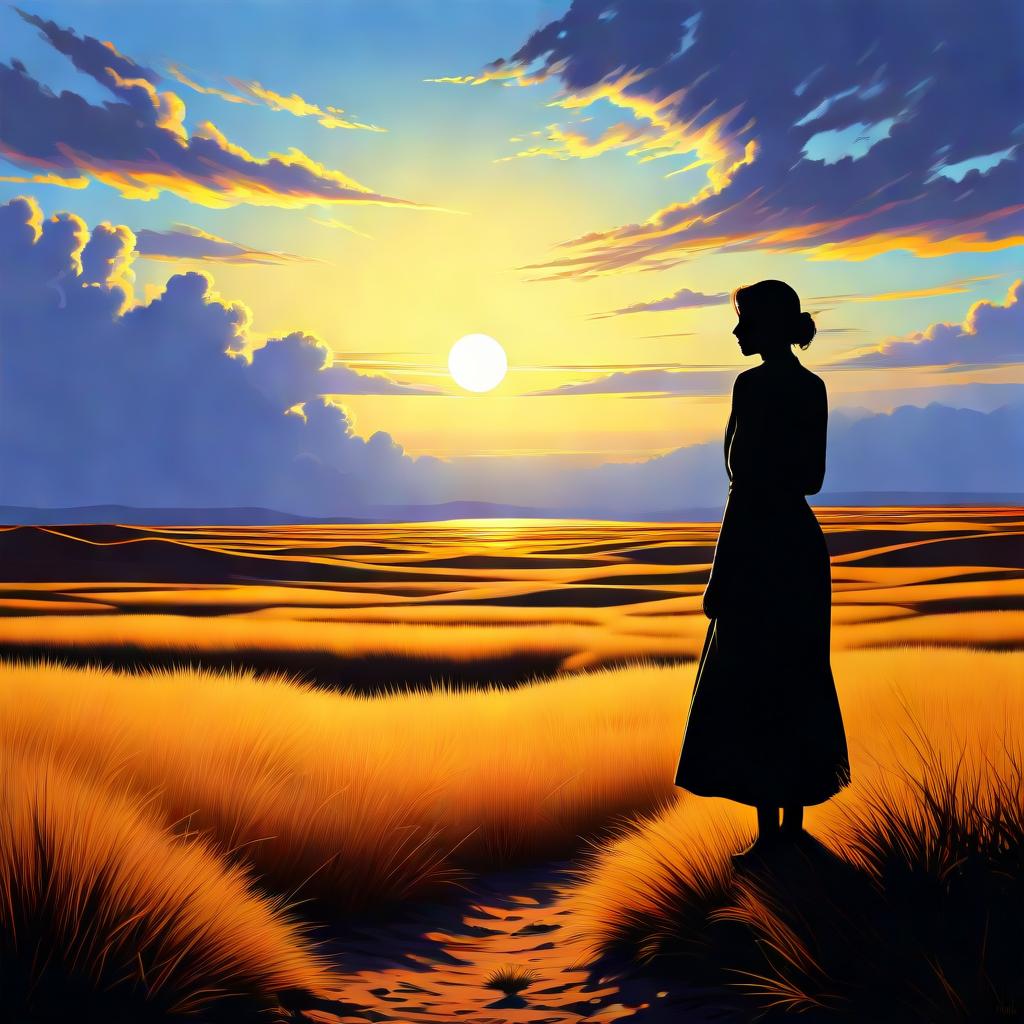} &
        \includegraphics[width=0.147\textwidth,height=0.147\textwidth]{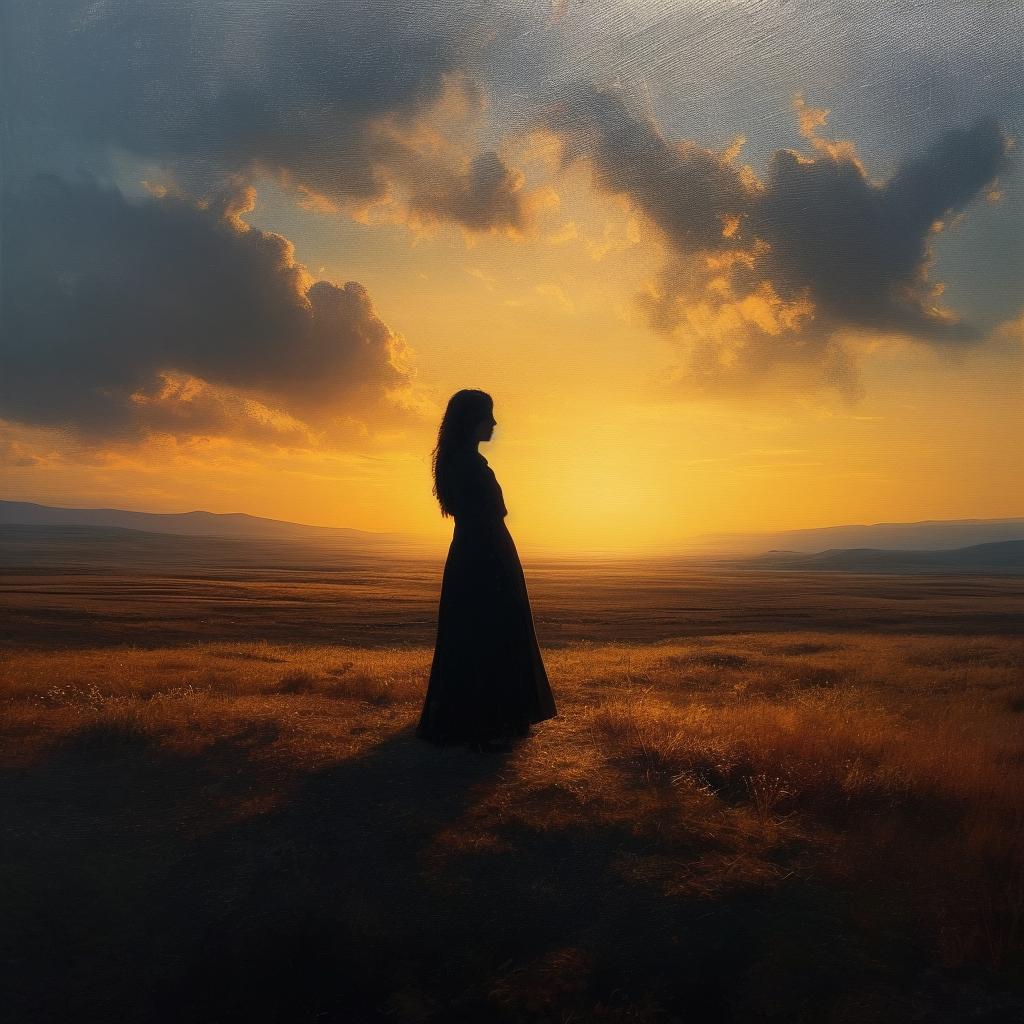} &
        \includegraphics[width=0.147\textwidth,height=0.147\textwidth]{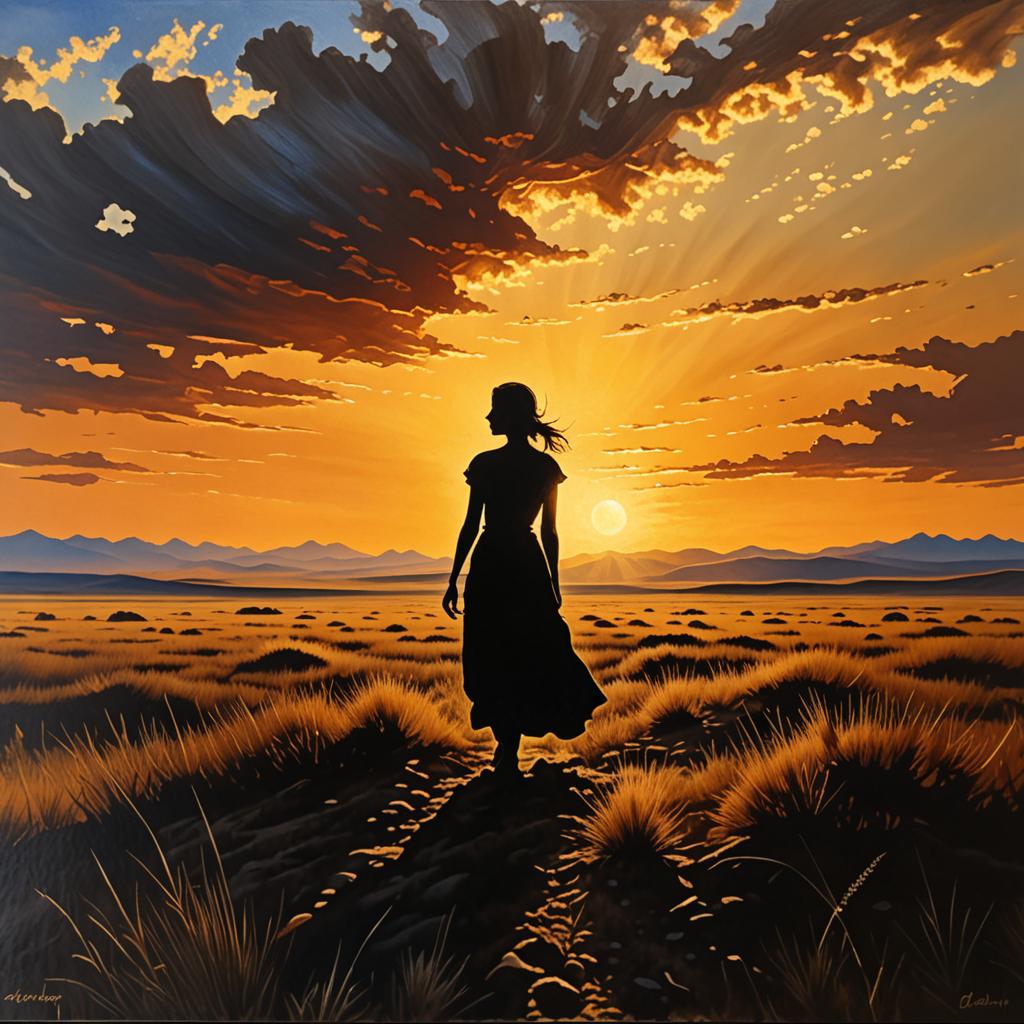} &
        \includegraphics[width=0.147\textwidth,height=0.147\textwidth]{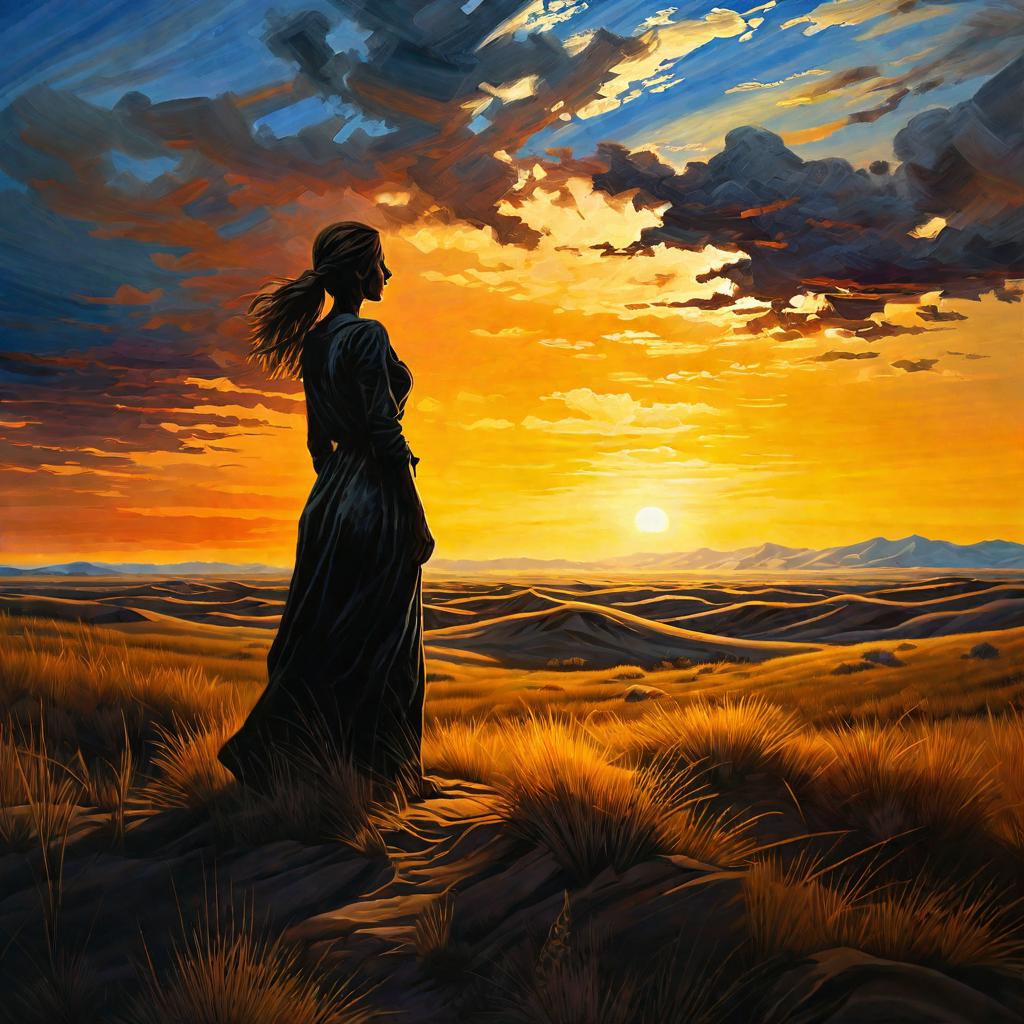}  \\

        \multicolumn{7}{c}{\begin{tabular}{@{}c@{}} \scriptsize ``oil painting, silhouette of a woman in the steppe wilderness, dramatic light, 34K uhd, masterpiece, high detail,  \\ \scriptsize 8k, intricate, detailed, high resolution, high res, high quality, highly detailed, Extremely high-resolution details, fine texture" \end{tabular}  } \\

    \end{tabular}
    
    }
    \caption{Qualitative results on CivitAI prompts. Please zoom in to better appreciate the differences.}\label{fig:civit_qualitative}
\end{figure*}

The results on the CivitAI prompts are shown in \cref{fig:usr_study}, with visual samples for our approach and the $4$ best baselines provided in \cref{fig:civit_qualitative}. 
Both of our approaches outperform all baselines, with notable improvement over simply using the baseline SDXL model. Curiously, we observe 
that some fine-tuned versions of SDXL are competitive with fixed flows, further emphasizing the importance of tailoring flows to specific use cases.

\vspace{-3pt}
\section{Analysis}\label{sec:analysis}
\vspace{-3pt}

Having shown that our approach outperforms existing baselines, we next turn to analyzing its behavior. 
We examine three aspects of \ourmethod{}'s performance: (1) the originality and diversity of the generated flows, (2) whether they show human-interpretable patterns, and (3) the effect of using the target score in the \ourmethod{}-FT prompts. The findings for these aspects are reported below.

\vspace{-3pt}
\subsection{Originality and diversity}
\vspace{-3pt}

We begin by assessing ComfyGen-FT's ability to generate novel flows by comparing its predictions for the $500$ CivitAI prompts to the nearest neighbors in our training corpus. While ComfyGen-IC is retrieval-based and has an expected similarity score of 1.0, ComfyGen-FT achieves 0.9995 similarity, indicating that at the scale of our model, there is little to no generation of unseen flows. This indicates that our fine-tuning approach has also learned to classify flows. However, we note that in contrast to ComfyGen-IC, it has learned so directly from the data with limited ad-hoc choices in the process, and indeed it outperforms ComfyGen-IC in most of our evaluations. Looking ahead, we hope that future methods will also be able to synthesize unseen flows with novel graph structures.

In terms of diversity, we observe that over $500$ prompts, ComfyGen-IC makes use of $41$ unique flows, while ComfyGen-FT uses $79$, indicating a higher diversity. Recall also that our base set contained only $33$ human created templates, which were then augmented through random parameter changes. Hence, both variations identified useful flows which differ from the initial human-created set. This suggests that more data or a more involved search over the input parameter space could yield more diverse outputs and possibly improved performance. 

\vspace{-3pt}
\subsection{Analyzing the chosen flows}
\vspace{-3pt}
Next, we want to see whether we can identify any patterns in the chosen flows that would provide useful information about the strengths of existing models. Towards this goal, we want to see which models are popular across different categories, and which ones are especially prevalent for prompts within a particular category. 

To identify models most strongly associated with specific labels, we parse all flows selected for our $500$-prompt test set. We scan each flow for base models, LoRAs, and upscaling models, appending their names to a to a document corresponding to each label associated with the prompt that generated the flow. 
Then, we use TF-IDF~\citep{sparck1972statistical} to rank the models across these label-documents. In \cref{tab:choice_analysis}, we report the top-scoring model for each of four distinct labels, as well as the most common models across the entire flow set (``General").

We observe that, in many cases, the choices make intuitive sense. For example, the GFPGAN face restoration model is closely tied to the ``People" category. Similarly, ``Anime" prompts make more frequent use of models that better preserve human anatomy, or a LoRA tuned for an anime aesthetic. However, while such patterns exist in the data, the choices are not always intuitively clear. In the future, it may be beneficial to have the LLM explain the reasoning behind its component selections.

\begin{table}[h]
\centering
\setlength{\belowcaptionskip}{-6pt}
\setlength{\tabcolsep}{4pt}
\scriptsize
\begin{tabular}{lccccc}
\toprule
Category & ``People" & ``Nature" & ``Anime" & ``Abstract" & General \\
\midrule
Top Base Model & Proteus v3 & Stable Cascade & JibMixXL v9 ``Better Bodies" & SDVN7-NijiStyleXL & crystalClearXL \\
Top LoRA & SDXL FaeTastic v24 & Add-Detail XL & AnimeTarot & LogoRedmond & MidJourney52 v1.2\\
Top Upscaler & GFPGAN v1.4 & Real-ESRGAN & UltraSharp x4 & None & UltraSharp x4 \\
\bottomrule
\end{tabular}
\caption{Top workflow components by TF-IDF scores for selected categories. In many cases, selections align with human intuition (\eg, a face upscaling model is favored for the ``People" category).}
\label{tab:choice_analysis}
\end{table}

\vspace{-3pt}
\subsection{The effect of target scores}
\vspace{-3pt}

\begin{figure*}
    \centering
    \setlength{\belowcaptionskip}{-4pt}
    \setlength{\tabcolsep}{0.5pt}
    \fontsize{7.5pt}{7.5pt}\selectfont
    \begin{tabular}{@{}m{0.5\textwidth}@{\hspace{0.025\textwidth}}m{0.375\textwidth}@{}}
        \includegraphics[width=\linewidth]{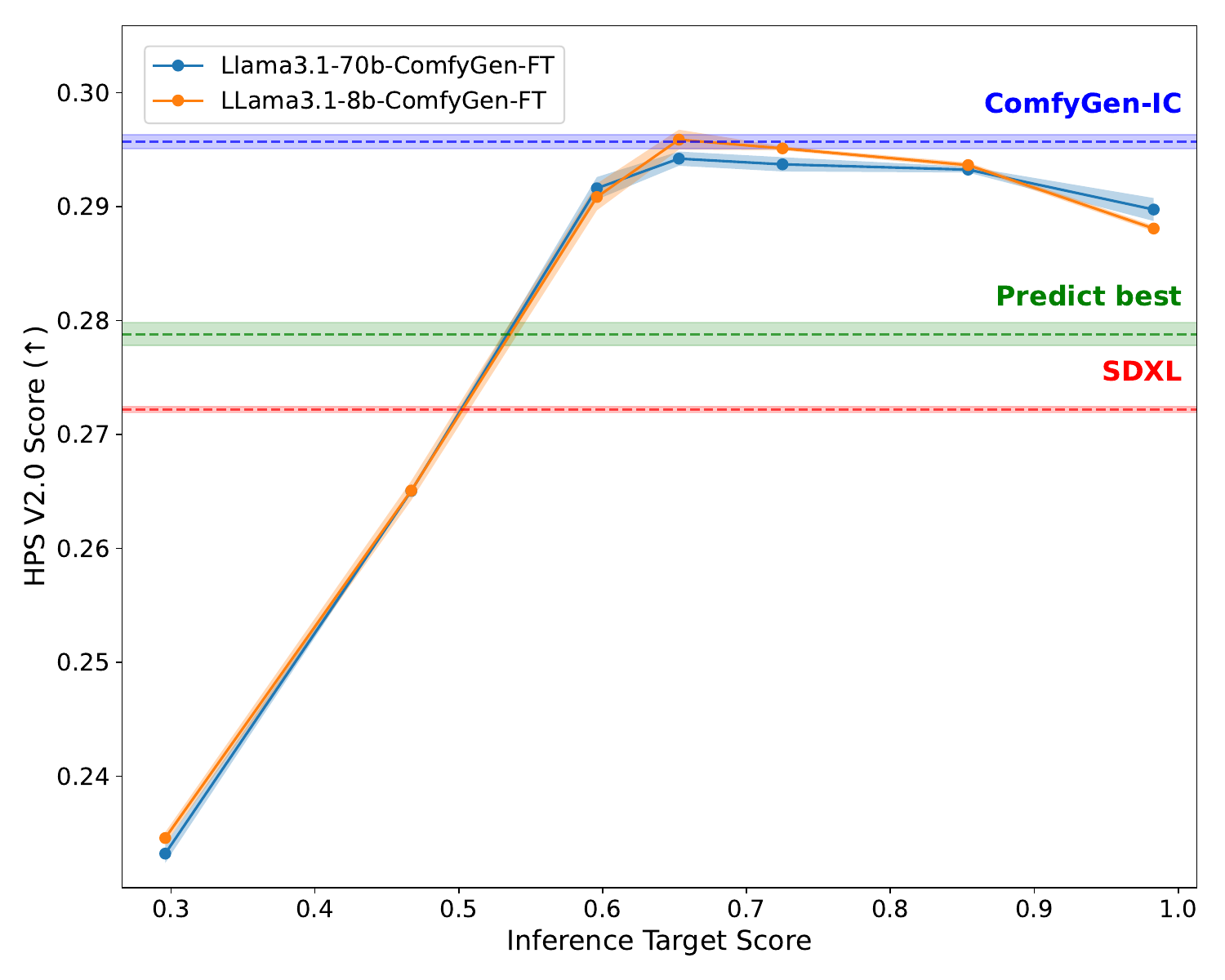} &
        \begin{tabular}{@{}cc@{}}
    
            \multicolumn{2}{c}{\begin{tabular}{@{}c@{}} \tiny ``...red-haired female adventurer in medieval attire \\[-2pt] \tiny standing against a backdrop of a futuristic, geometric, \\[-2pt] \tiny neon-lit landscape, surrealism style, vibrant colors..." \end{tabular}  }
    
            \\
            \includegraphics[width=0.4\linewidth,height=0.4\linewidth]{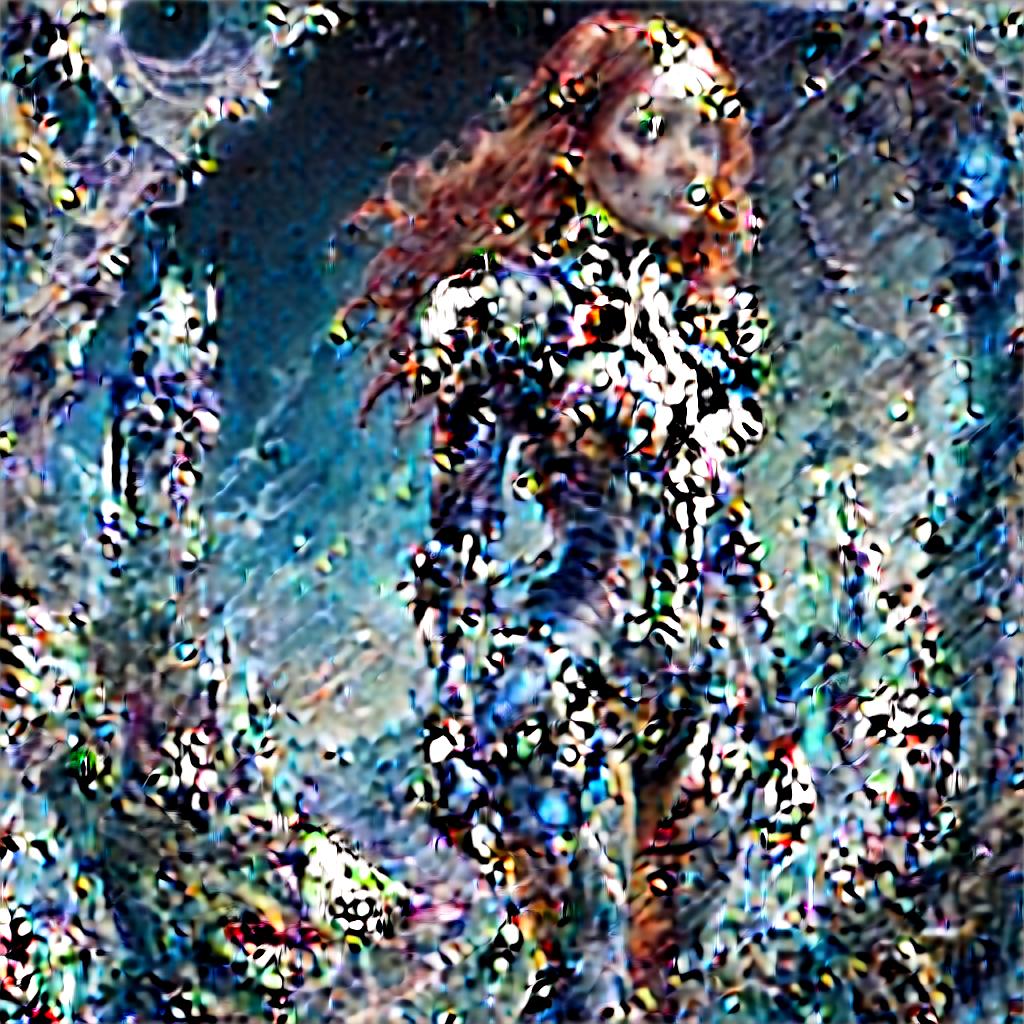} &
            \includegraphics[width=0.4\linewidth,height=0.4\linewidth]{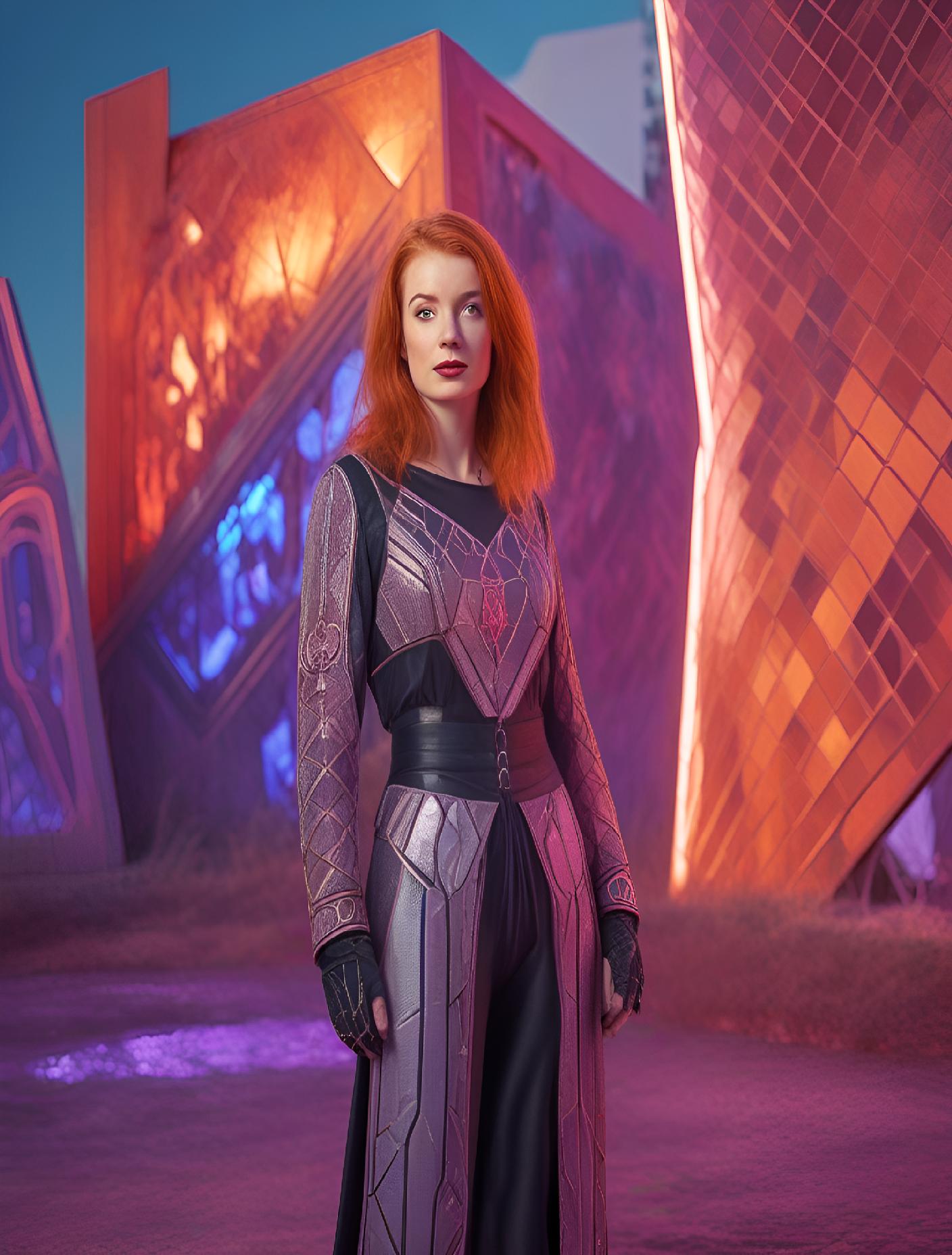} \\
            Target score: 0.296 & Target score: 0.467 \\
            \includegraphics[width=0.4\linewidth,height=0.4\linewidth]{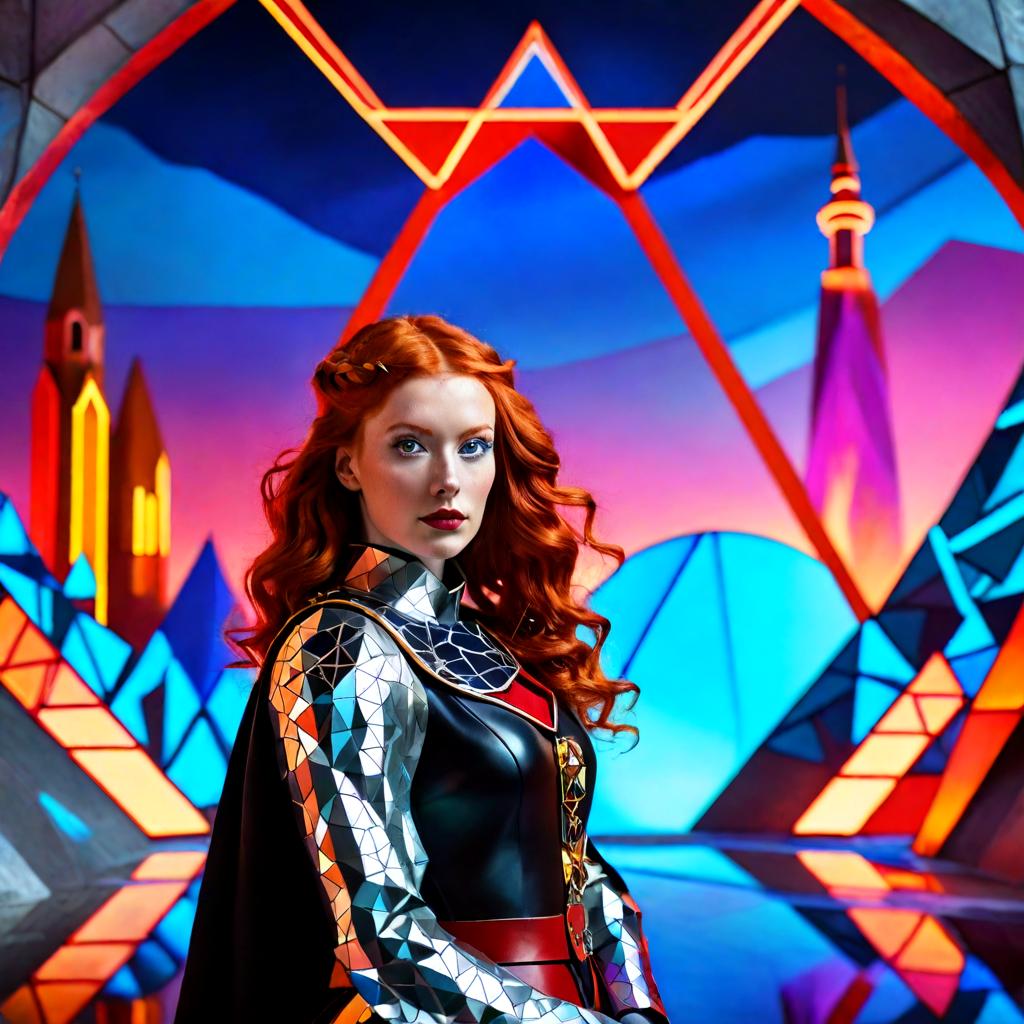} &
            \includegraphics[width=0.4\linewidth,height=0.4\linewidth]{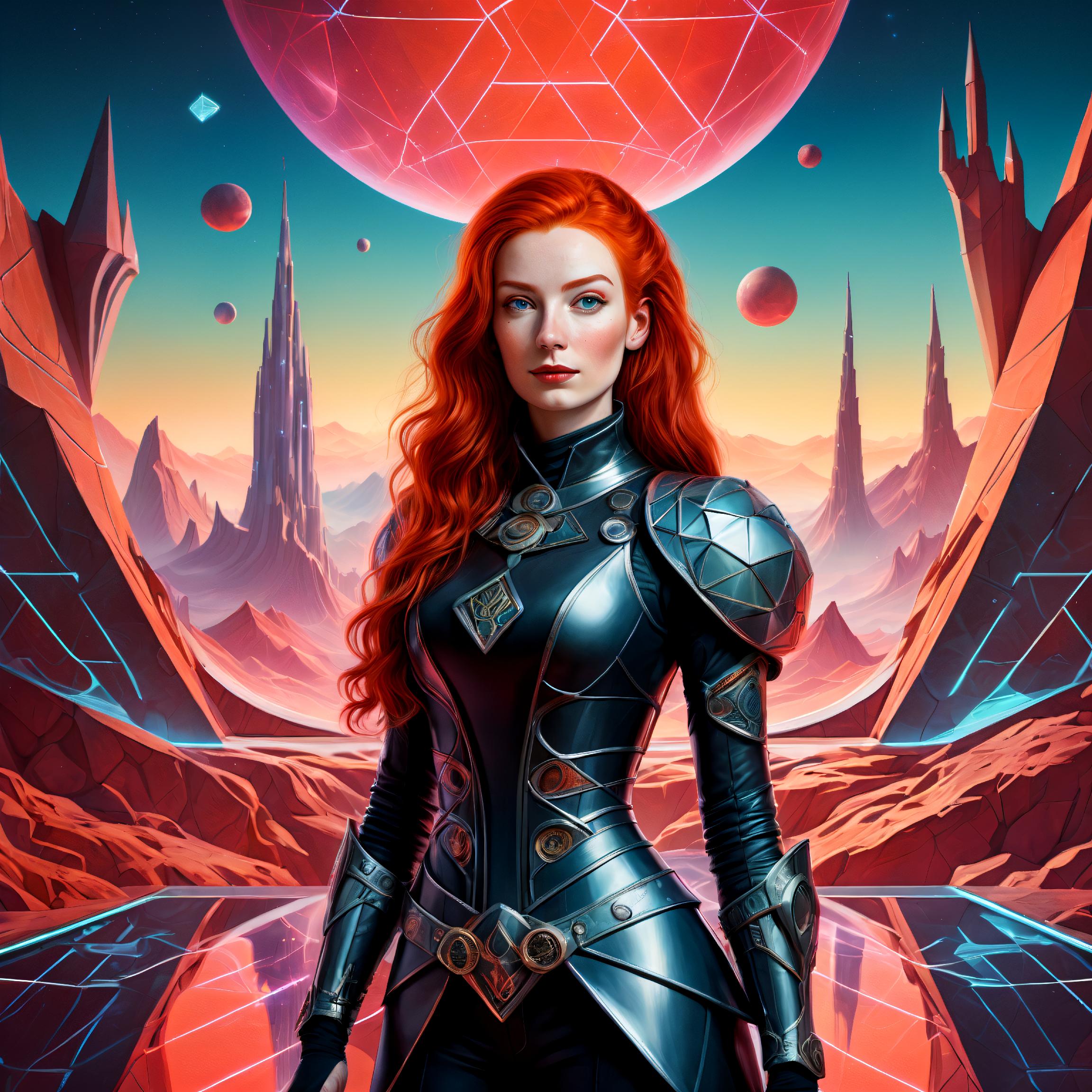} \\
            Target score: 0.596 & Target score: 0.725
            
        \end{tabular}
    \end{tabular}
    \caption{\textbf{(left)} Average HPS V2 score on CivitAI prompts as a function of the inference target score. The 8B and 70B variations of our model perform equally well, and significantly outperform the variation trained to predict the highest scoring flow (green). \textbf{(right)} Model outputs for the same prompt at different target scores.}
    \label{fig:score_analysis}
\end{figure*}

Recall that ComfyGen-FT was fine-tuned to predict a flow based on a given prompt and a target score. Here, we examine the performance of the model according to the target score provided at inference time. 
To do so, we repeat the CivitAI prompt experiments of \cref{sec:experiments}, while adjusting the target score used in our prompts. 
We evaluate both a model tuned from the Llama3.1 8B version and one from Llama3.1 70B. 
The quality of the generated images is assessed using HPS v2.0, and we report the average outcomes. The results are presented in \cref{fig:score_analysis}.
For reference, we provide the scores of the baseline SDXL model, as well as ComfyGEN-IC. We additionally examine a scenario where instead of tuning the model to predict a flow given a prompt and a score, we simply tune it to predict the highest scoring flow ("Predict best"). 

We observe that the ComfyGen-FT model has indeed learned to associate the target scores with flows of varying quality. With an appropriate choice of score (near the top of the training score distribution), ComfyGen-FT achieves comparable performance to ComfyGen-IC. Notably, attempting to predict the best model instead of the score-based tuning leads to greatly diminished performance, highlighting the importance of our approach. We further note that both model sizes achieve comparable performance, hinting that we are far from saturating the capabilities of the models.

Finally, it is important to note that our target scores were calculated using an ensemble of human-preference predictors, excluding HPS v2.0. Therefore, there should be no expectation for alignment between the values on the y-axis and the x-axis.

\vspace{-3pt}
\section{Limitations}
\vspace{-3pt}
While ComfyGen's prompt-dependent workflow approach demonstrates improvements over monolithic models and constant flows, it is not free of limitations. 
Our current model is limited to text-to-image workflows, and cannot address more complex editing or control-based tasks. However, this could potentially be resolved in the future through the use of vision-language models (VLMs).

Similarly, both of our approaches require us to generate images using a large number of flows. With typical generations taking an order of $15$ seconds, even a modest set of $500$ prompts and $300$ flows requires a month of GPU time to create. Therefore, scaling up the approach would likely require significant computational resources or more efficient ways (e.g., Reinforcement Learning) to explore the flow parameter space.

Finally, each of our two methods has its own unique drawbacks. 
The fine-tuning approach cannot easily generalize to new blocks as they become available, requiring retraining with new flows that include these blocks.
On the other hand, the in-context approach can be easily expanded by including the new flows in the score table provided to the LLM. However, this increases the number of input tokens used, making it more expensive to run and eventually saturating the maximum context length. 
We hope that these limitations can be addressed through more advanced retrieval-based approaches or through the use of collaborative agents.

\vspace{-3pt}
\section{Conclusions}\label{sec:conclusions}
\vspace{-3pt}

We introduced the task of prompt-adaptive workflow generation, and presented \ourmethod{} - a set of two approaches that tackle this task. Our experiments demonstrate that such prompt-dependent flows can outperform monolithic models or fixed, user created flows, in a sense providing us with a new path to improving downstream image quality.

In the future, we hope to further explore prompt-dependent workflow creation methods, increasing their originality and expanding their scope to image-to-image or even video-related tasks. Perhaps in the future we could collaborate with the language model on the creation of such flows, providing it feedback through additional instructions or examples of outputs, thereby enabling non-expert users to further push the boundary of content creation.

\section{Acknowledgements} We would like to thank Yotam Nitzan, Ron Mokady, Yael Vinker and Linoy Tsaban for providing feedback on an early version of this manuscript or its figures. We thank Shahar Sarfaty for his assistance in collecting user prompts, Assaf Shocher, Yoad Tewel and Tomer Wolfson for useful discussions, and Yaki Tebeka for his assistance with compute infrastructure.

A very special thanks to Or Patashnik for always lending a willing ear, and for providing moral support throughout the project.

\bibliography{main}
\bibliographystyle{iclr2025_conference}

\end{document}


\maketitle
\vspace{-35pt}
\section{Additional results}\label{sec:supp_additional_results}
\begin{figure*}[b]
    \centering
    \setlength{\tabcolsep}{0.5pt}
    {\normalsize
    \begin{tabular}{c c c}
        \includegraphics[width=0.36\textwidth,height=0.36\textwidth]{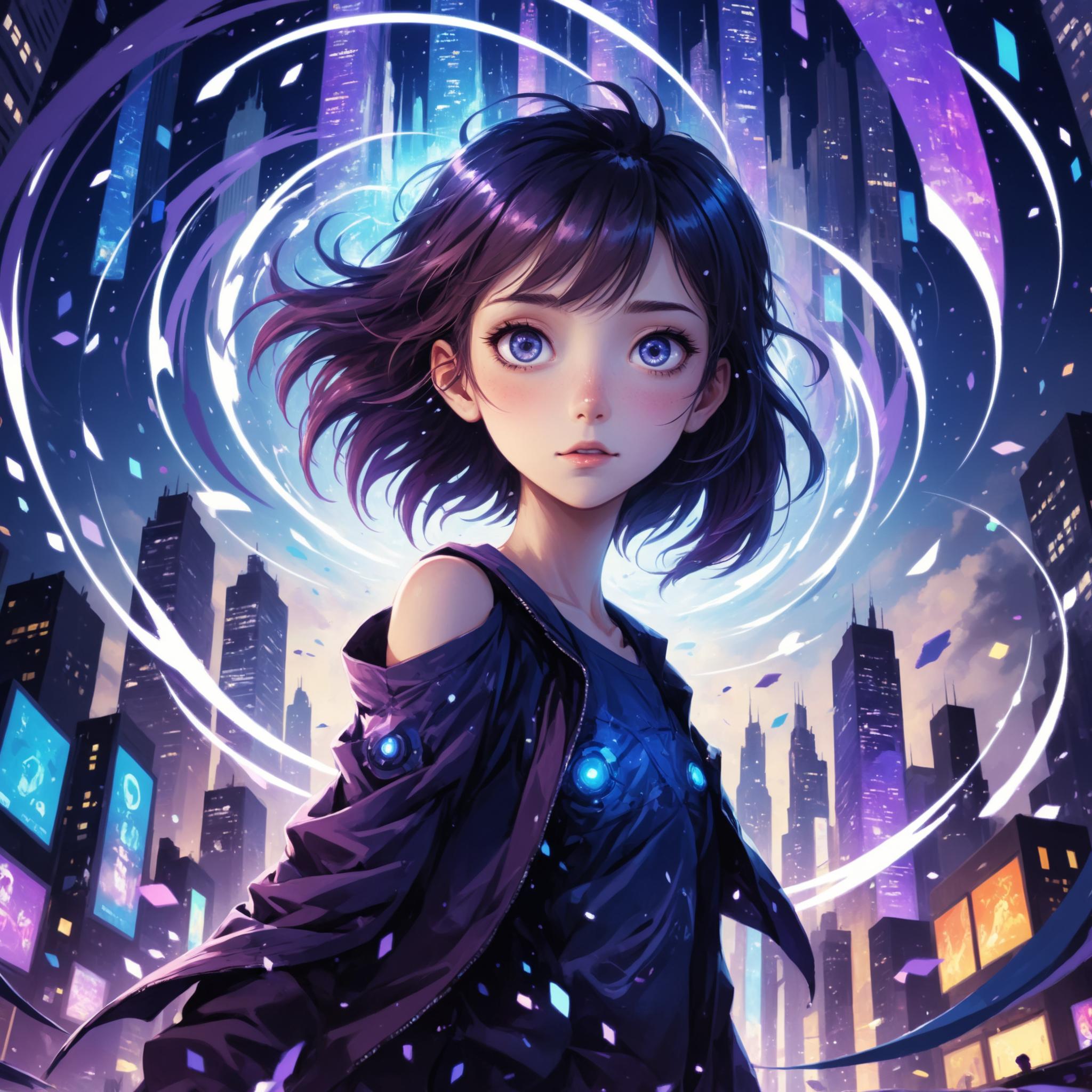} &
        \includegraphics[width=0.36\textwidth,height=0.36\textwidth]{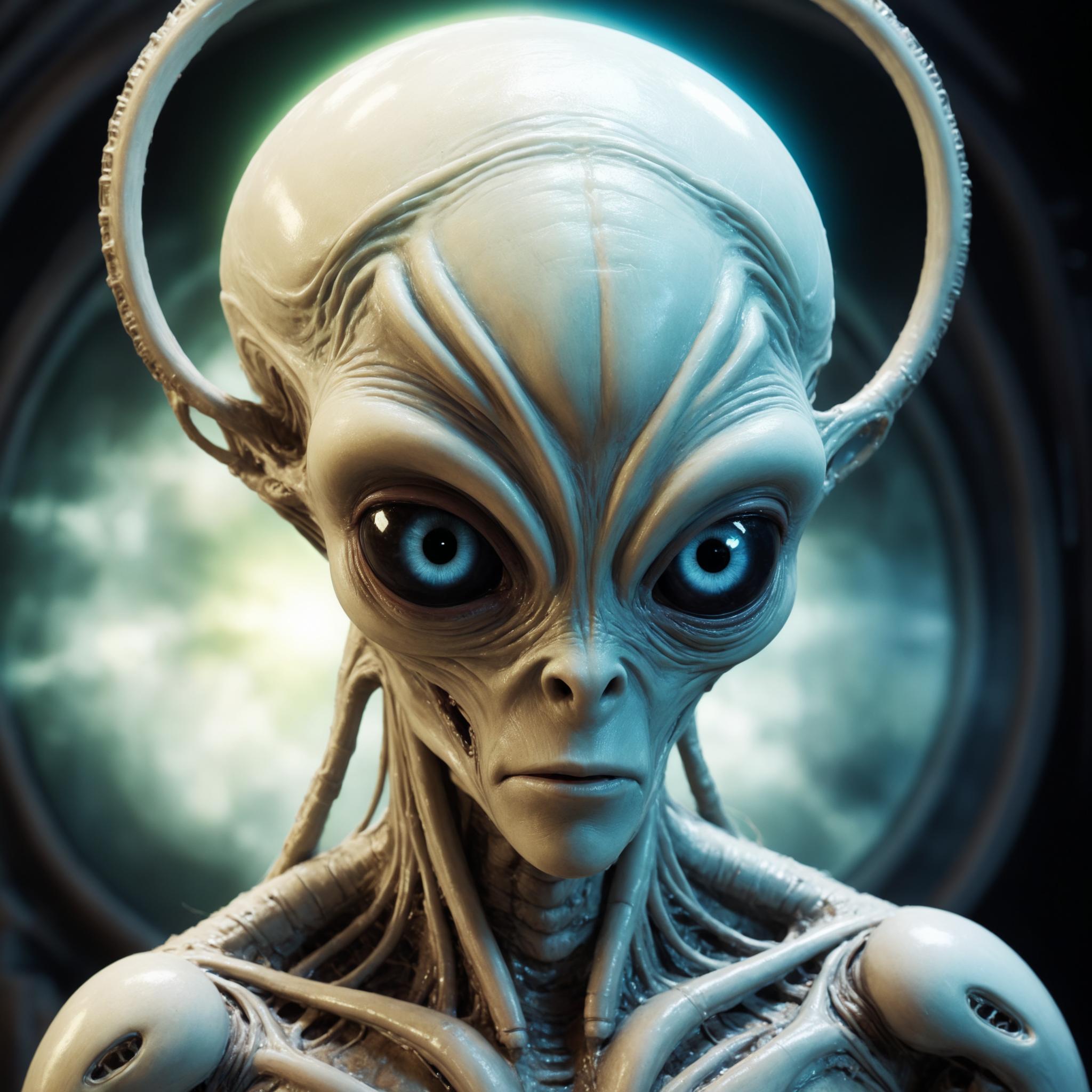} &
        \includegraphics[width=0.36\textwidth,height=0.36\textwidth]{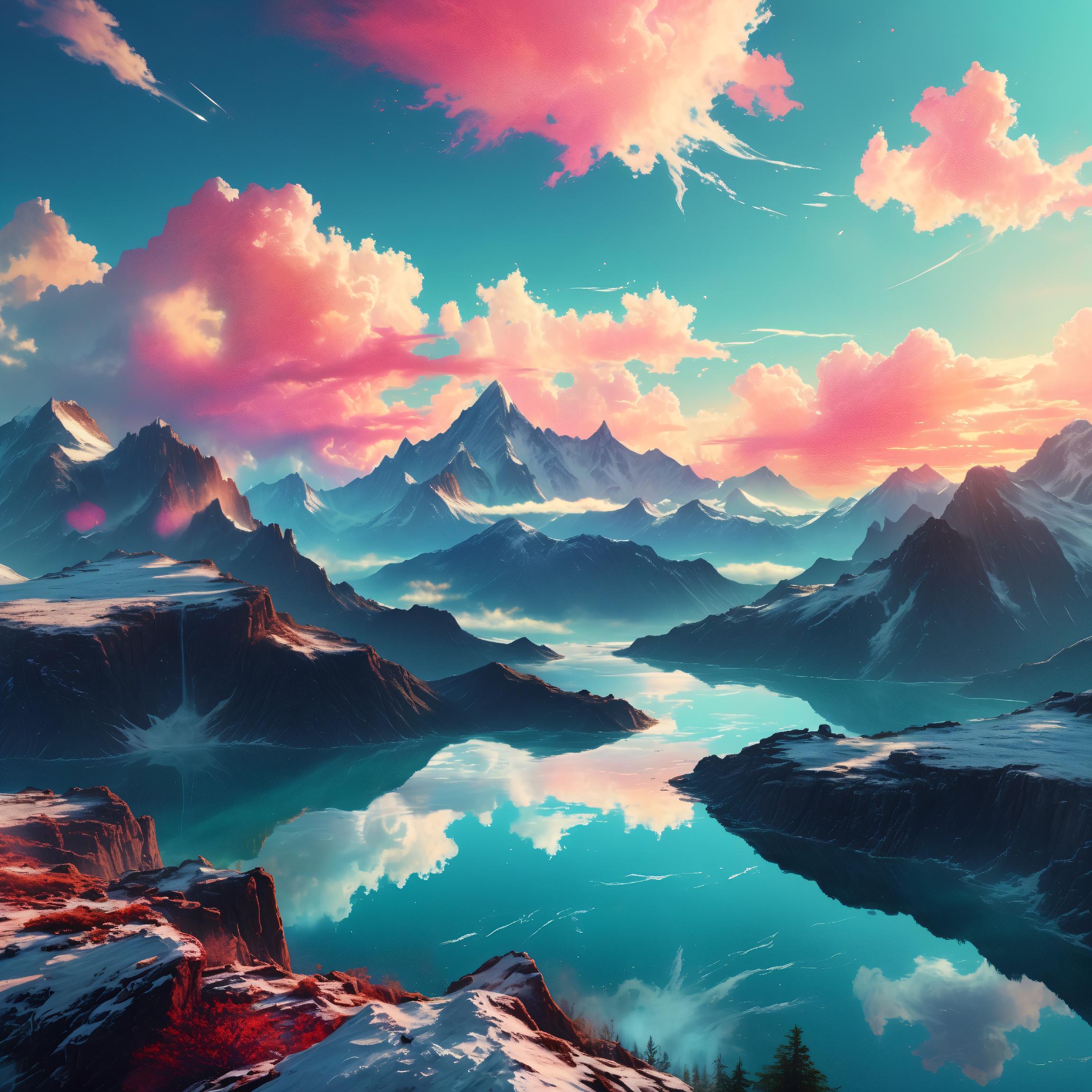} \\

        \includegraphics[width=0.36\textwidth,height=0.36\textwidth]{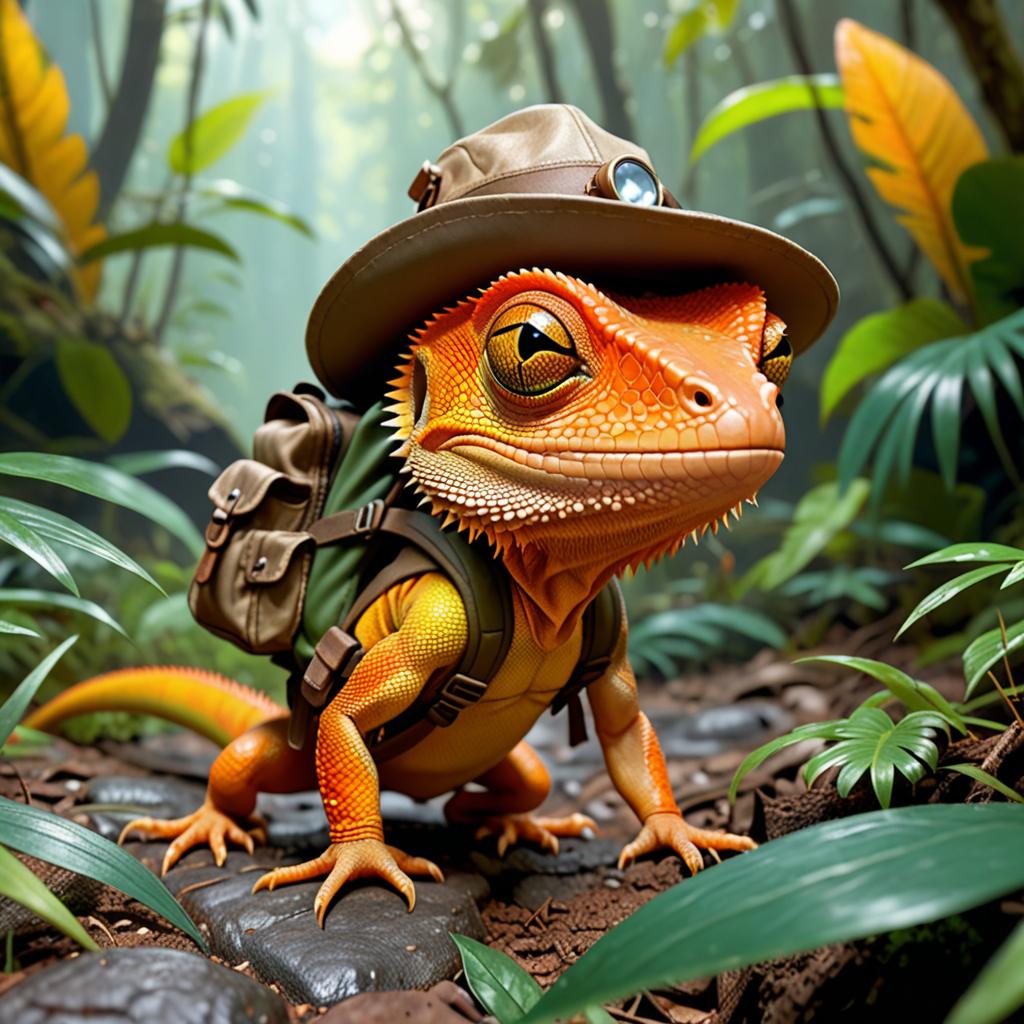} &
        \includegraphics[width=0.36\textwidth,height=0.36\textwidth]{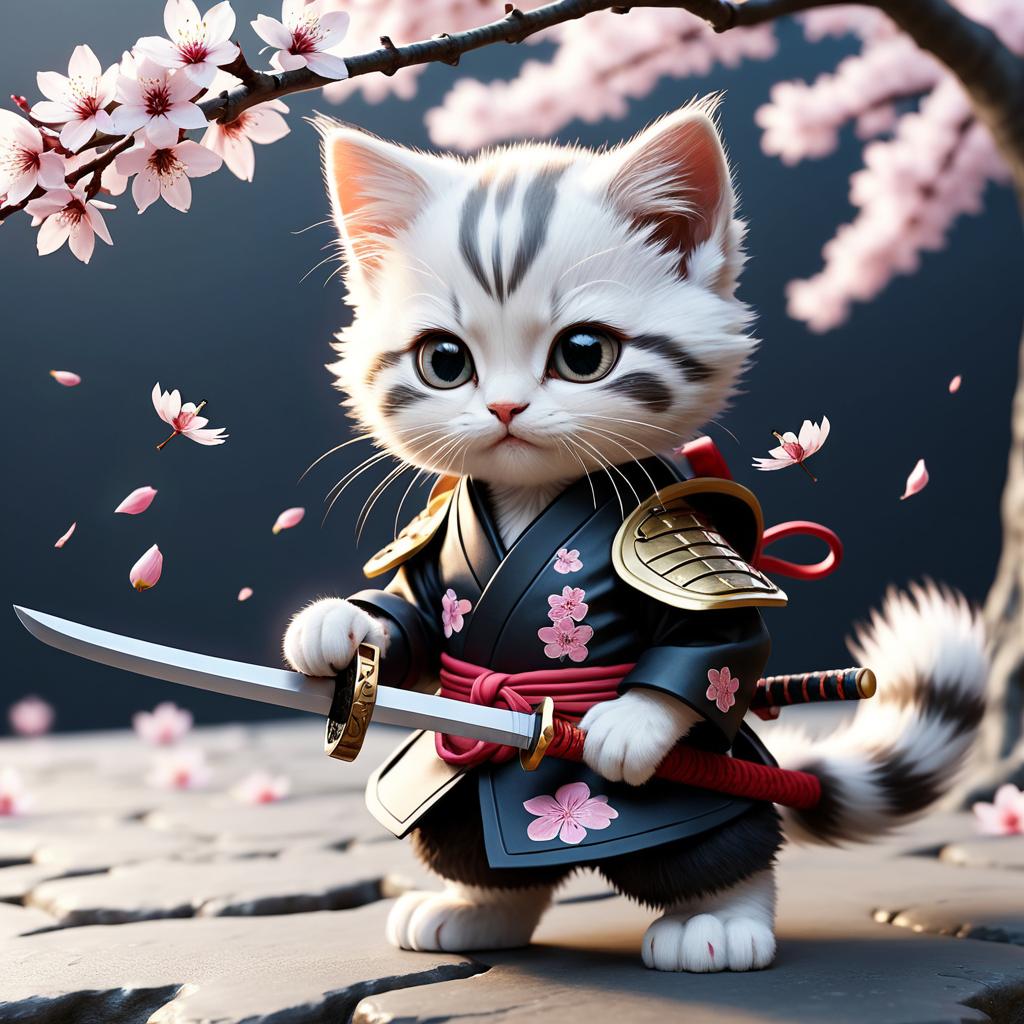} &
        \includegraphics[width=0.36\textwidth,height=0.36\textwidth]{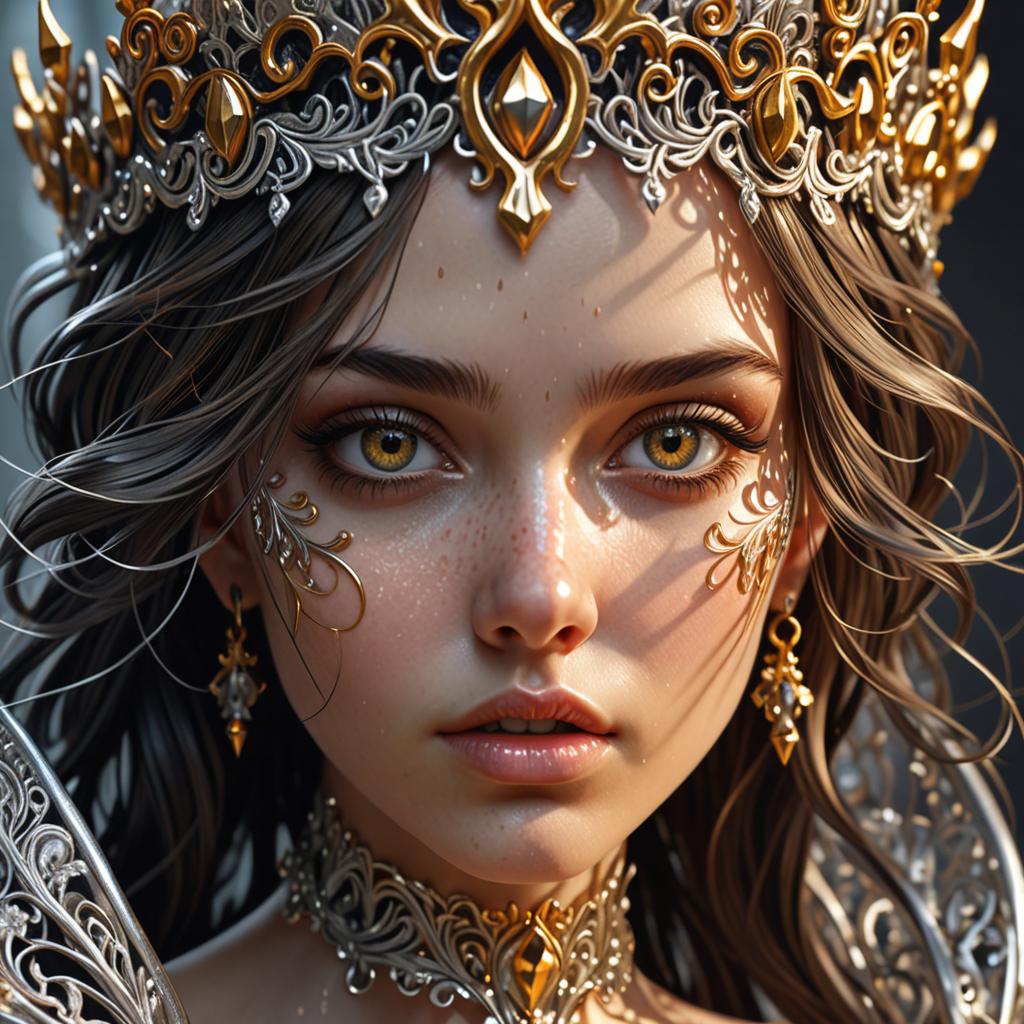} \\
        \includegraphics[width=0.36\textwidth,height=0.36\textwidth]{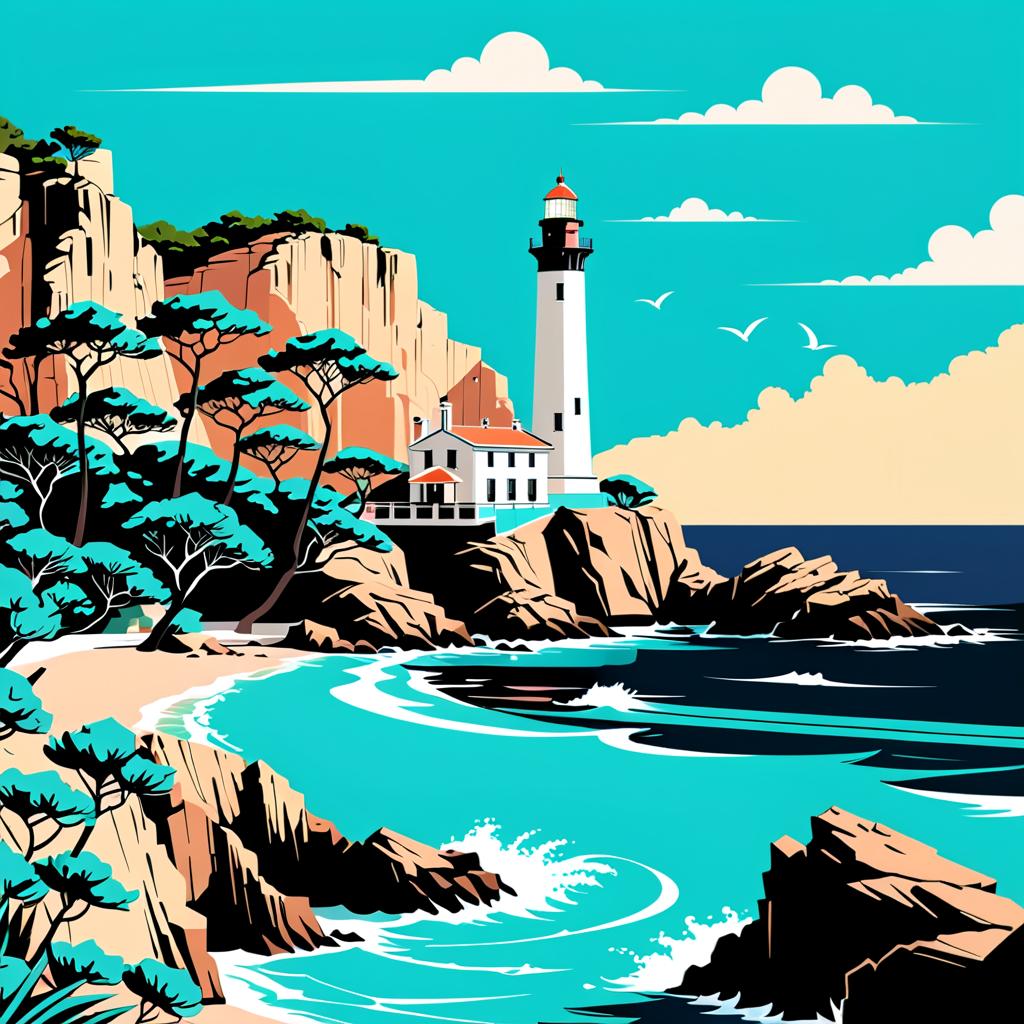} &
        \includegraphics[width=0.36\textwidth,height=0.36\textwidth]{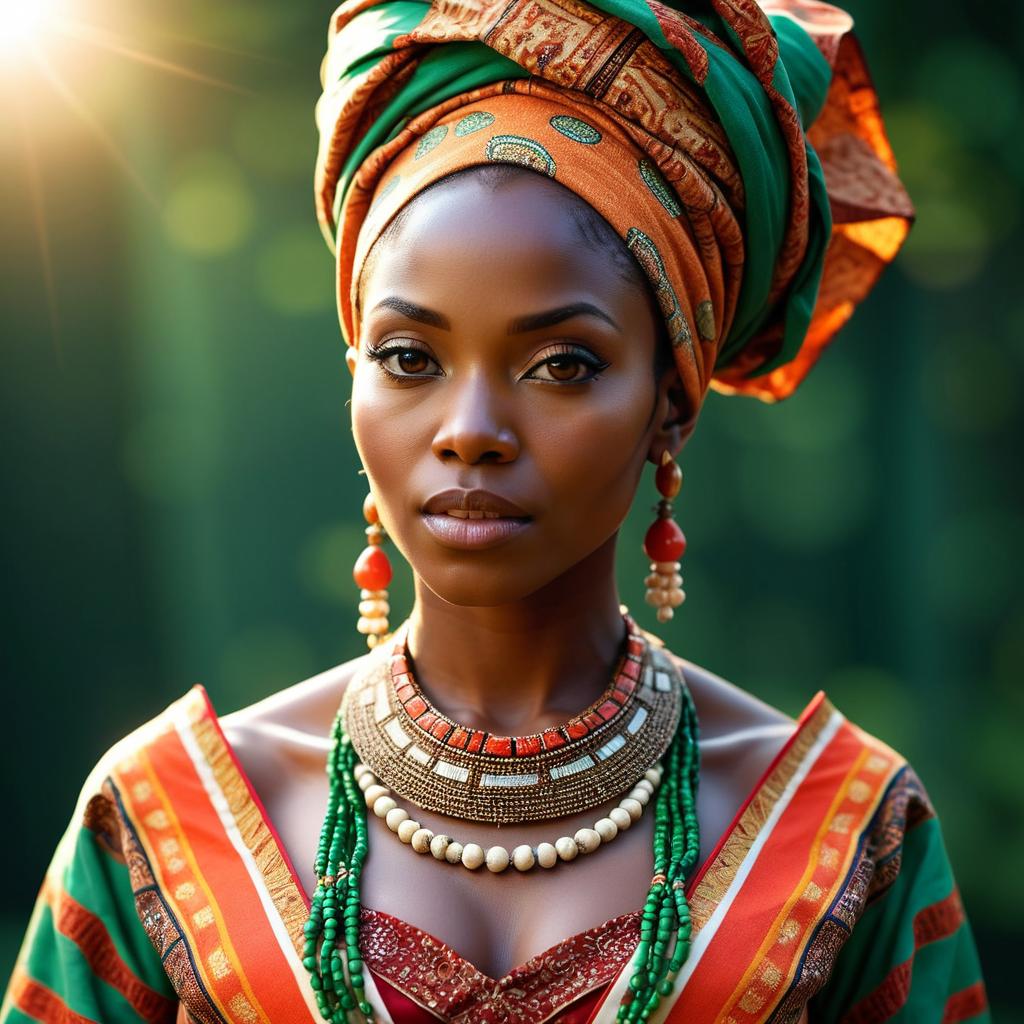} &
        \includegraphics[width=0.36\textwidth,height=0.36\textwidth]{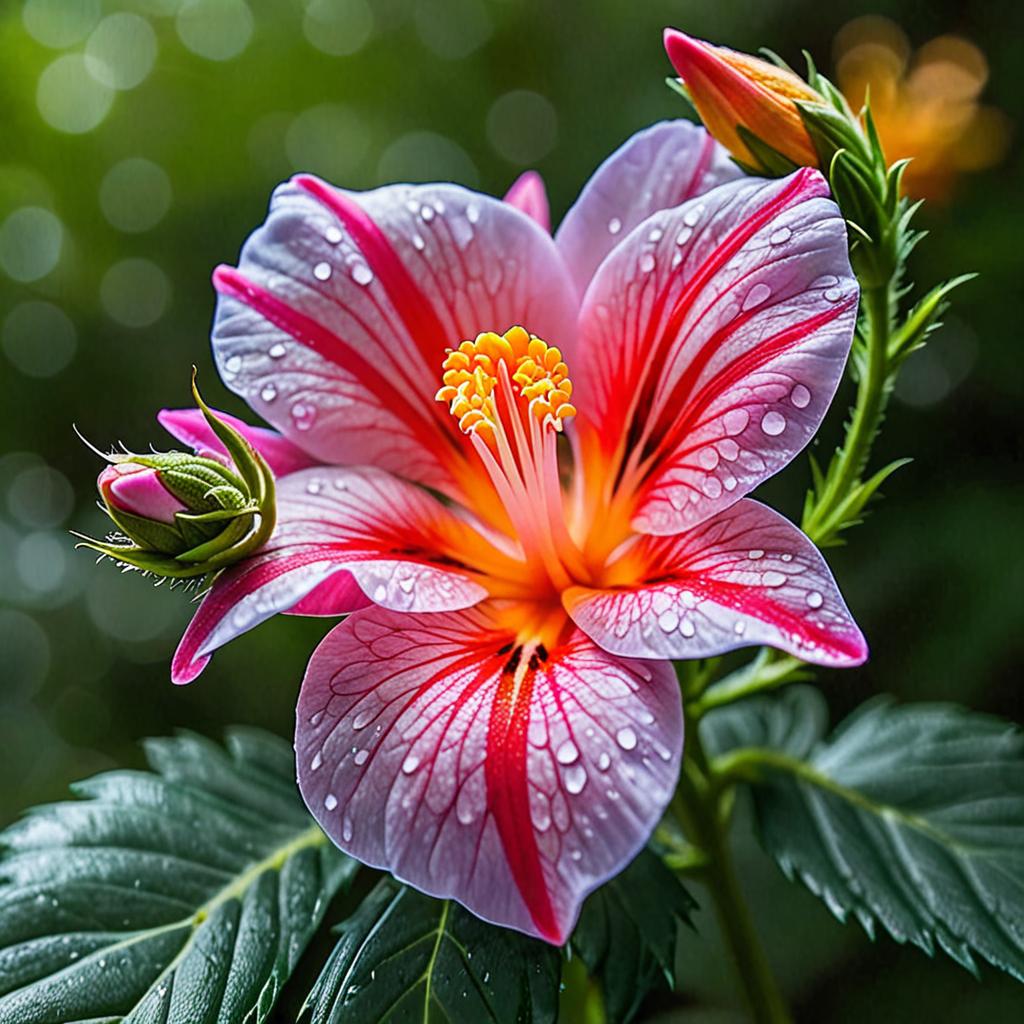} \\

    \end{tabular}
    }
    \caption{Curated images generated using ComfyGen-ICL. The list of prompts is available in the supplementary.}\label{fig:ours_large_supp_icl_1}
\end{figure*}

\begin{figure*}
    \centering
    \setlength{\tabcolsep}{0.5pt}
    {\normalsize
    \begin{tabular}{c c c}
        \includegraphics[width=0.36\textwidth,height=0.36\textwidth]{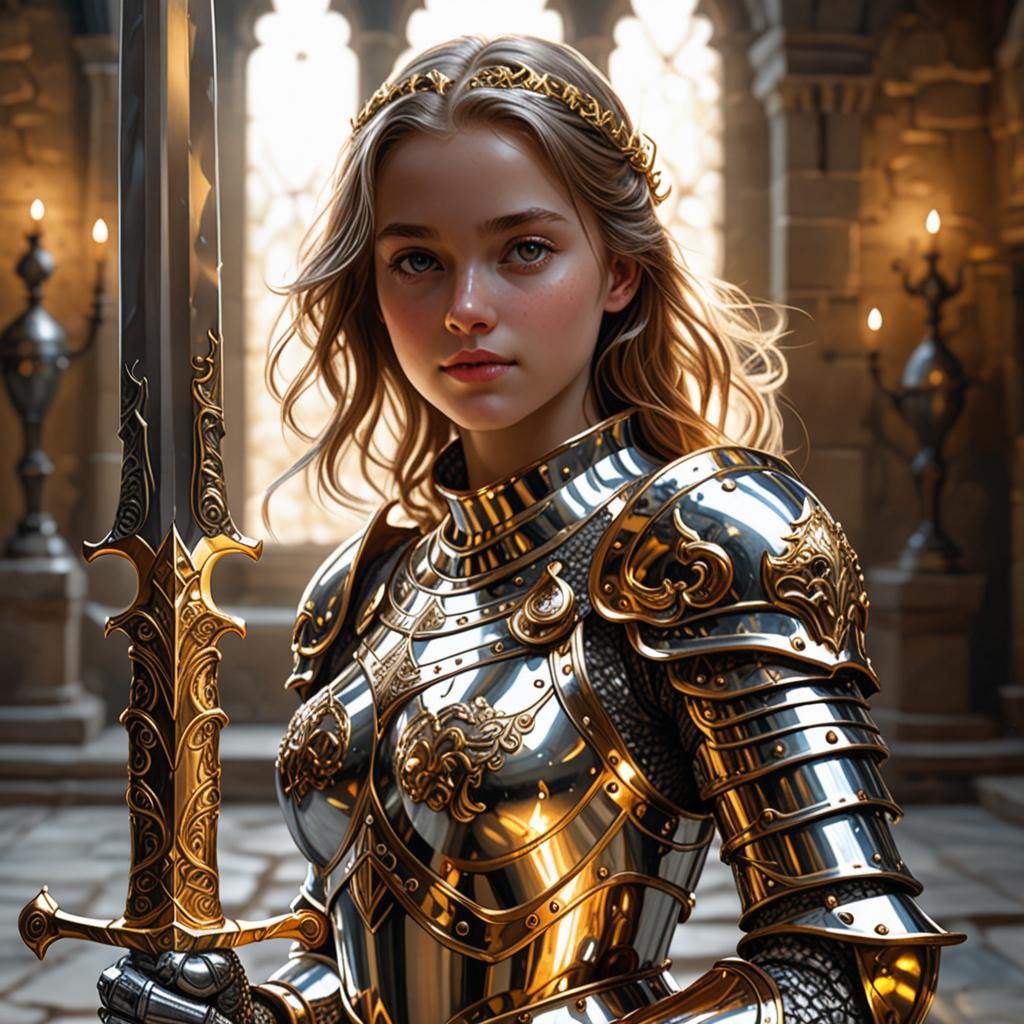} &
        \includegraphics[width=0.36\textwidth,height=0.36\textwidth]{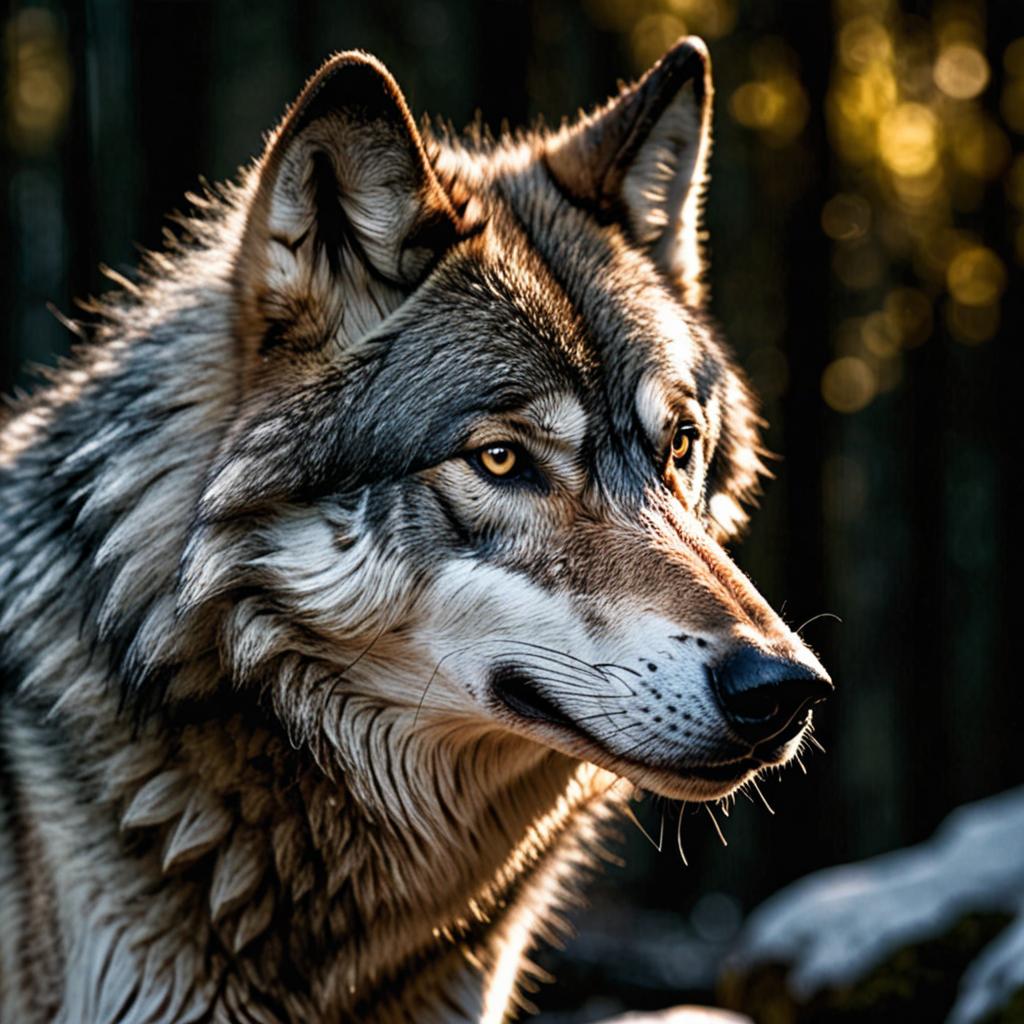} &
        \includegraphics[width=0.36\textwidth,height=0.36\textwidth]{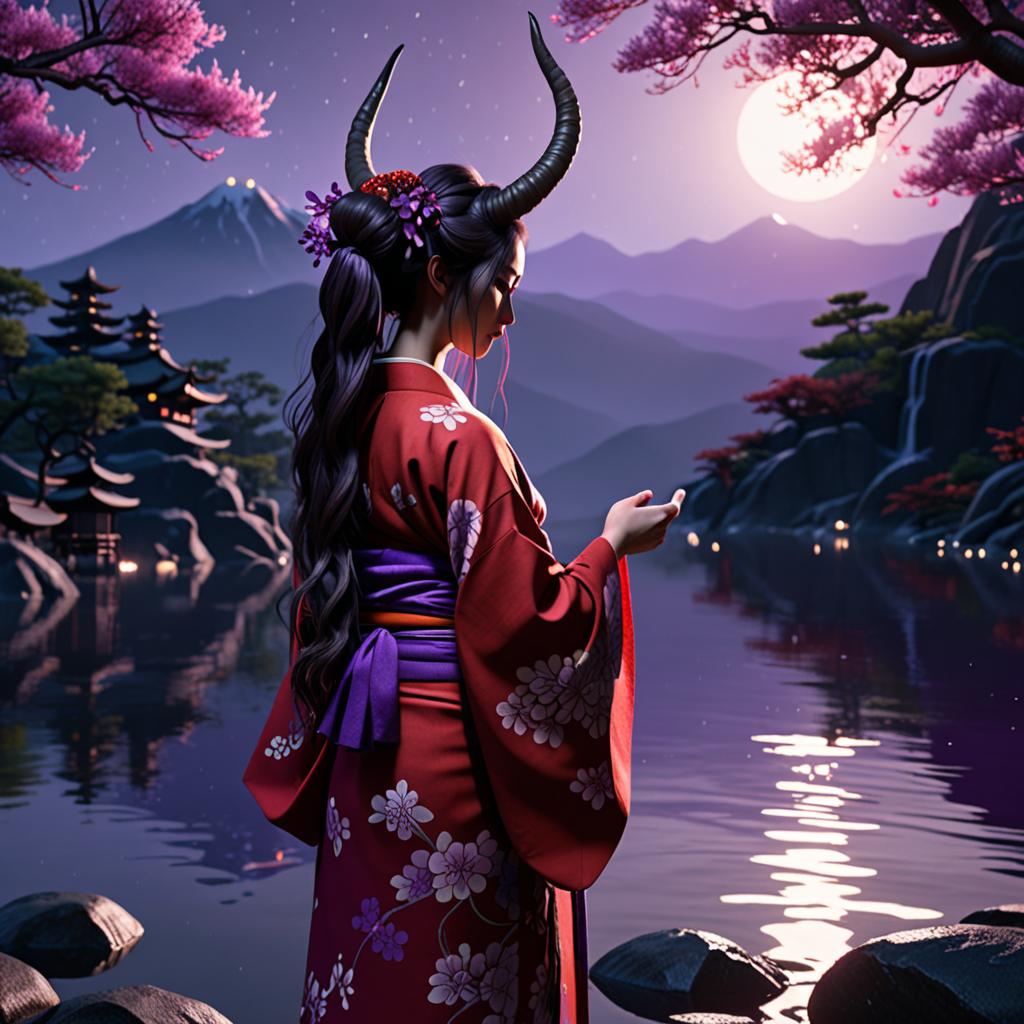} \\

        \includegraphics[width=0.36\textwidth,height=0.36\textwidth]{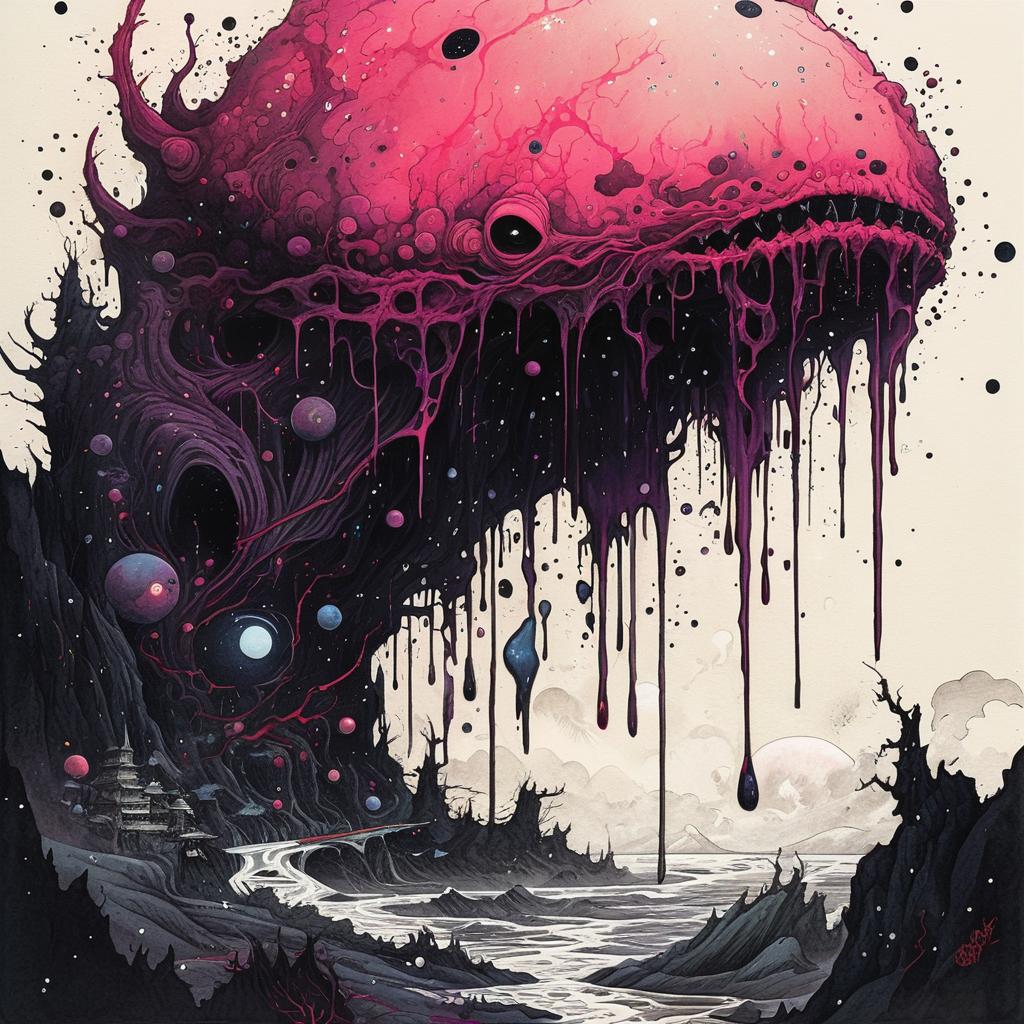} &
        \includegraphics[width=0.36\textwidth,height=0.36\textwidth]{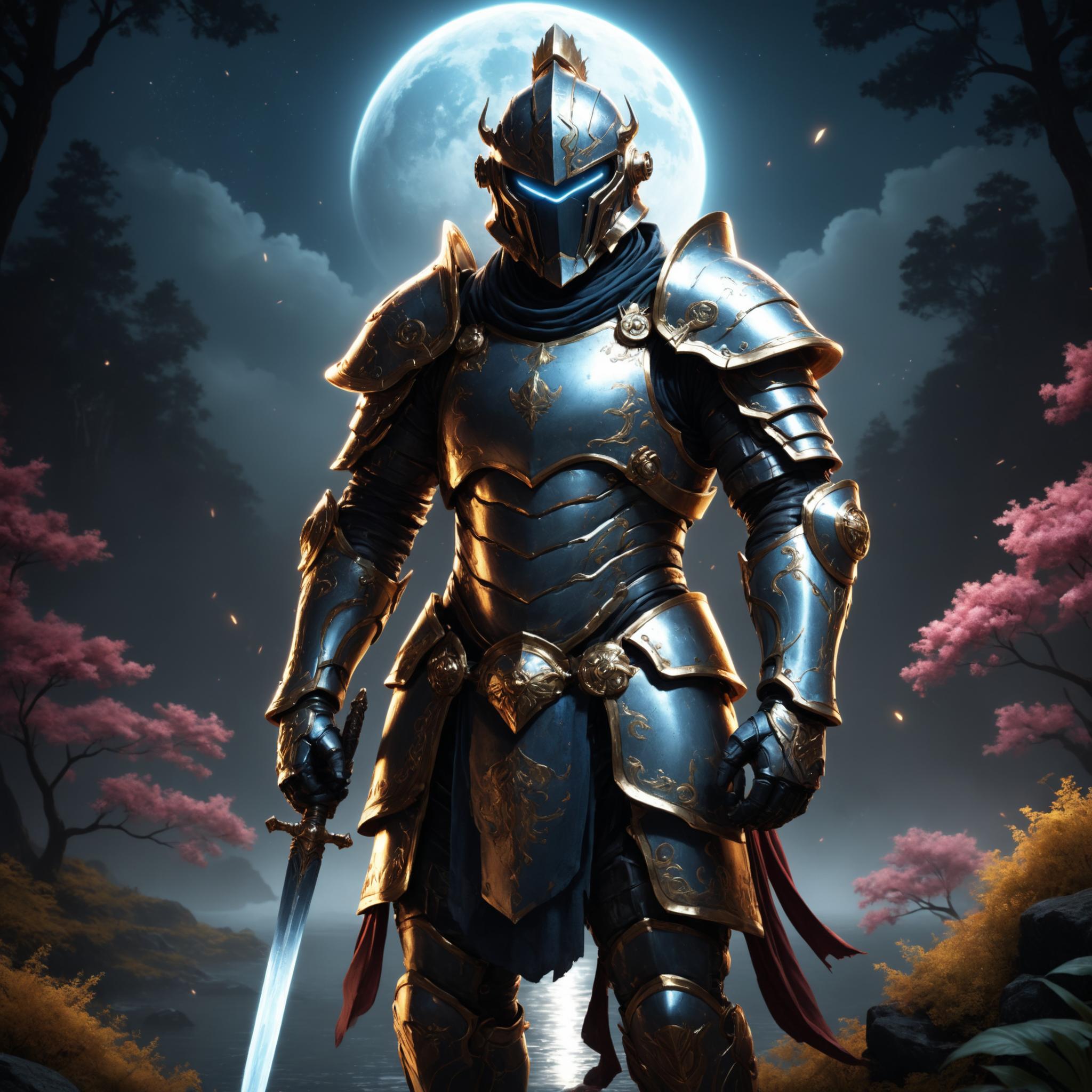} &
        \includegraphics[width=0.36\textwidth,height=0.36\textwidth]{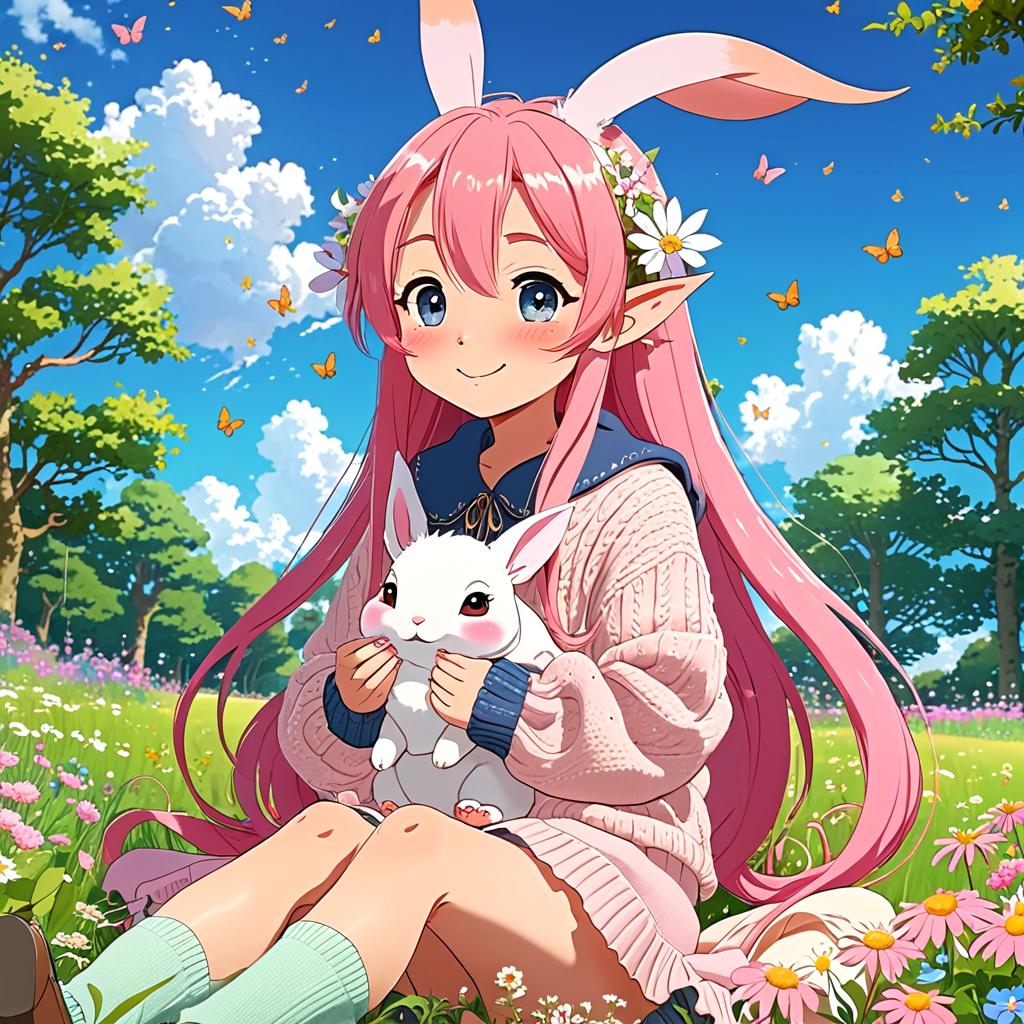} \\

        \includegraphics[width=0.36\textwidth,height=0.36\textwidth]{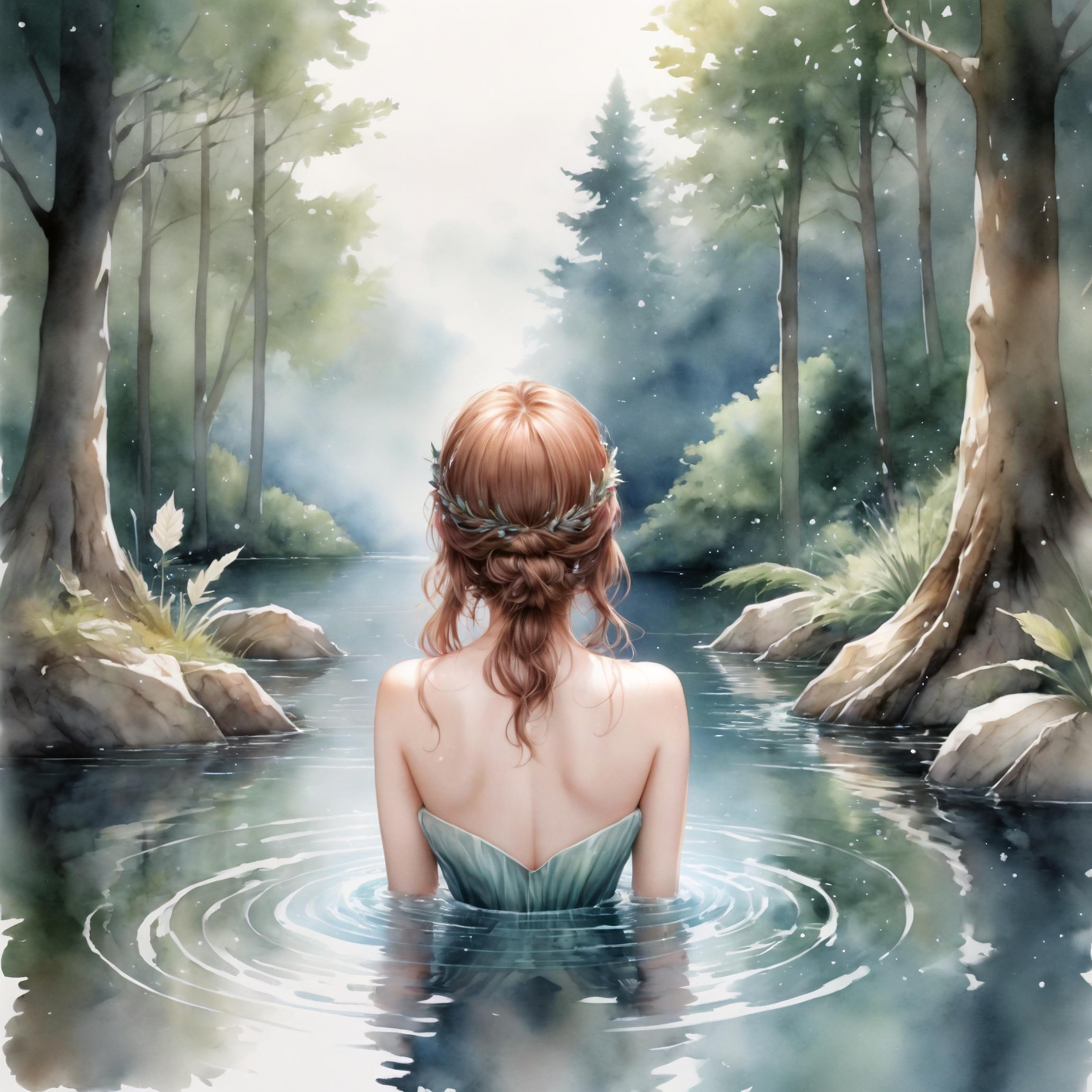} &
        \includegraphics[width=0.36\textwidth,height=0.36\textwidth]{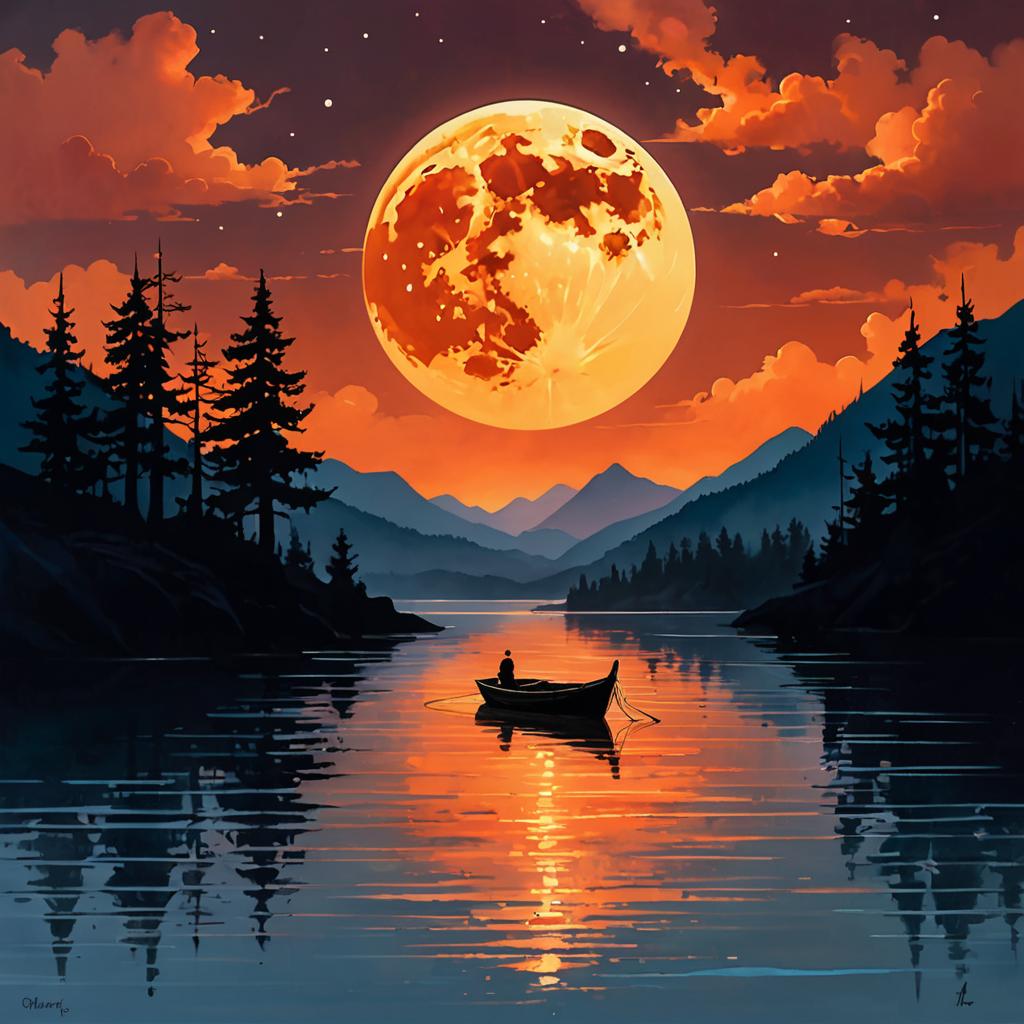} &
        \includegraphics[width=0.36\textwidth,height=0.36\textwidth]{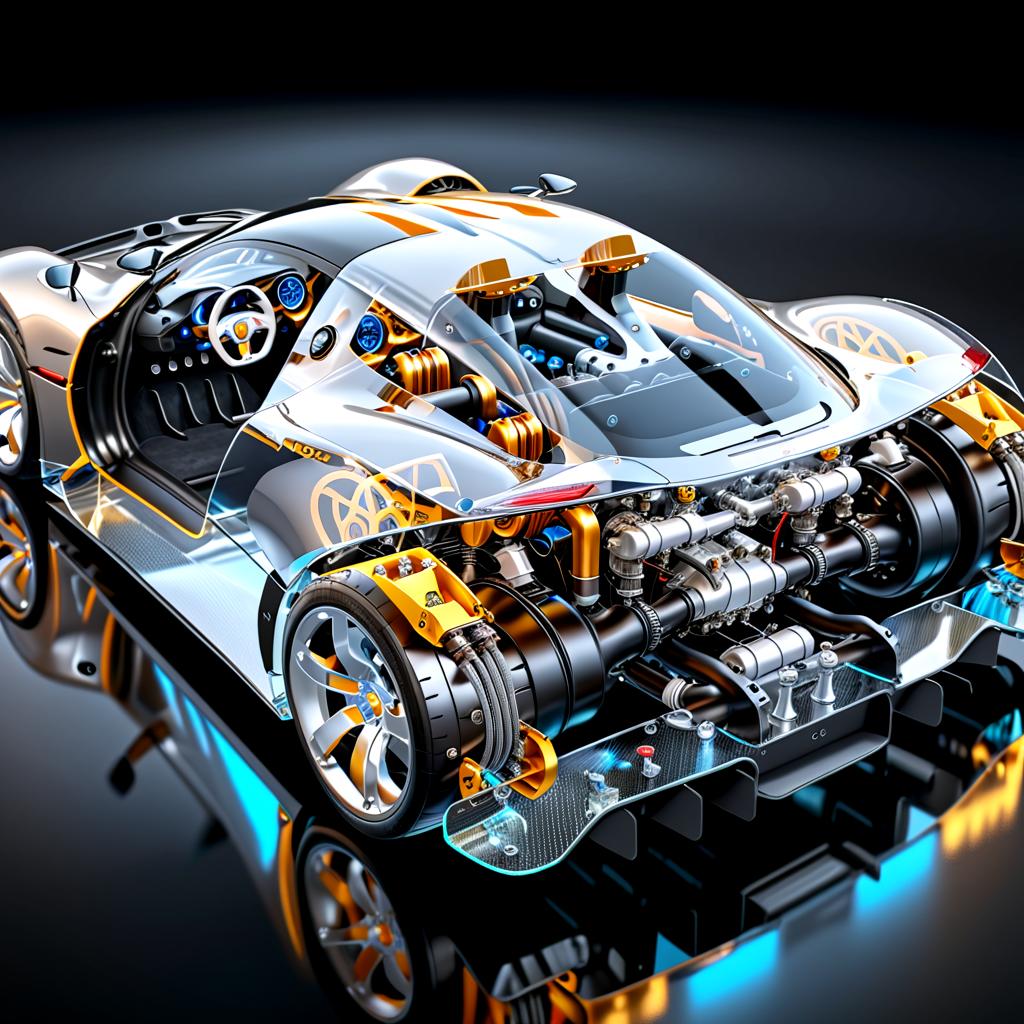} \\
        
    \end{tabular}
    }
    \caption{Curated images generated using ComfyGen-ICL. The list of prompts is available in the supplementary.}\label{fig:ours_large_supp_icl_2}
\end{figure*}

\begin{figure*}
    \centering
    \setlength{\tabcolsep}{0.5pt}
    {\normalsize
    \begin{tabular}{c c c}
        \includegraphics[width=0.36\textwidth,height=0.36\textwidth]{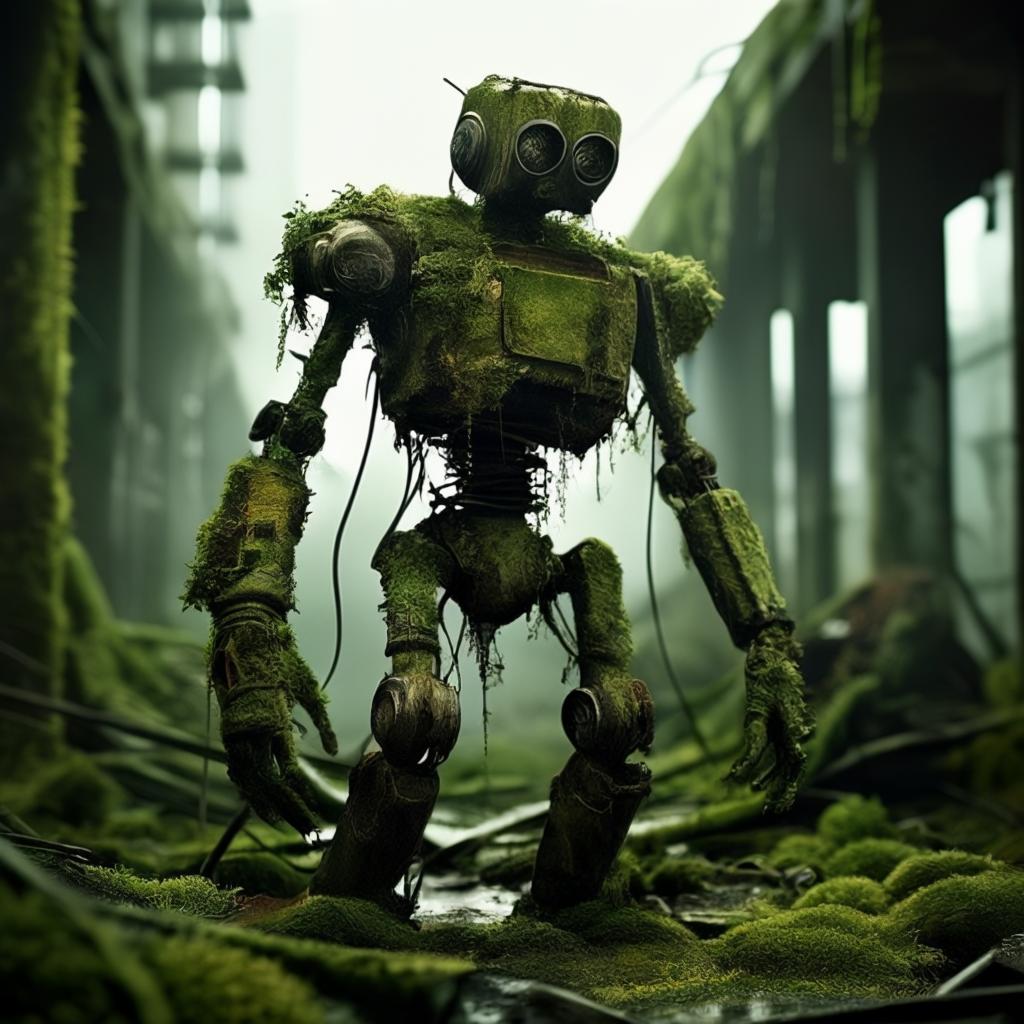} &
        \includegraphics[width=0.36\textwidth,height=0.36\textwidth]{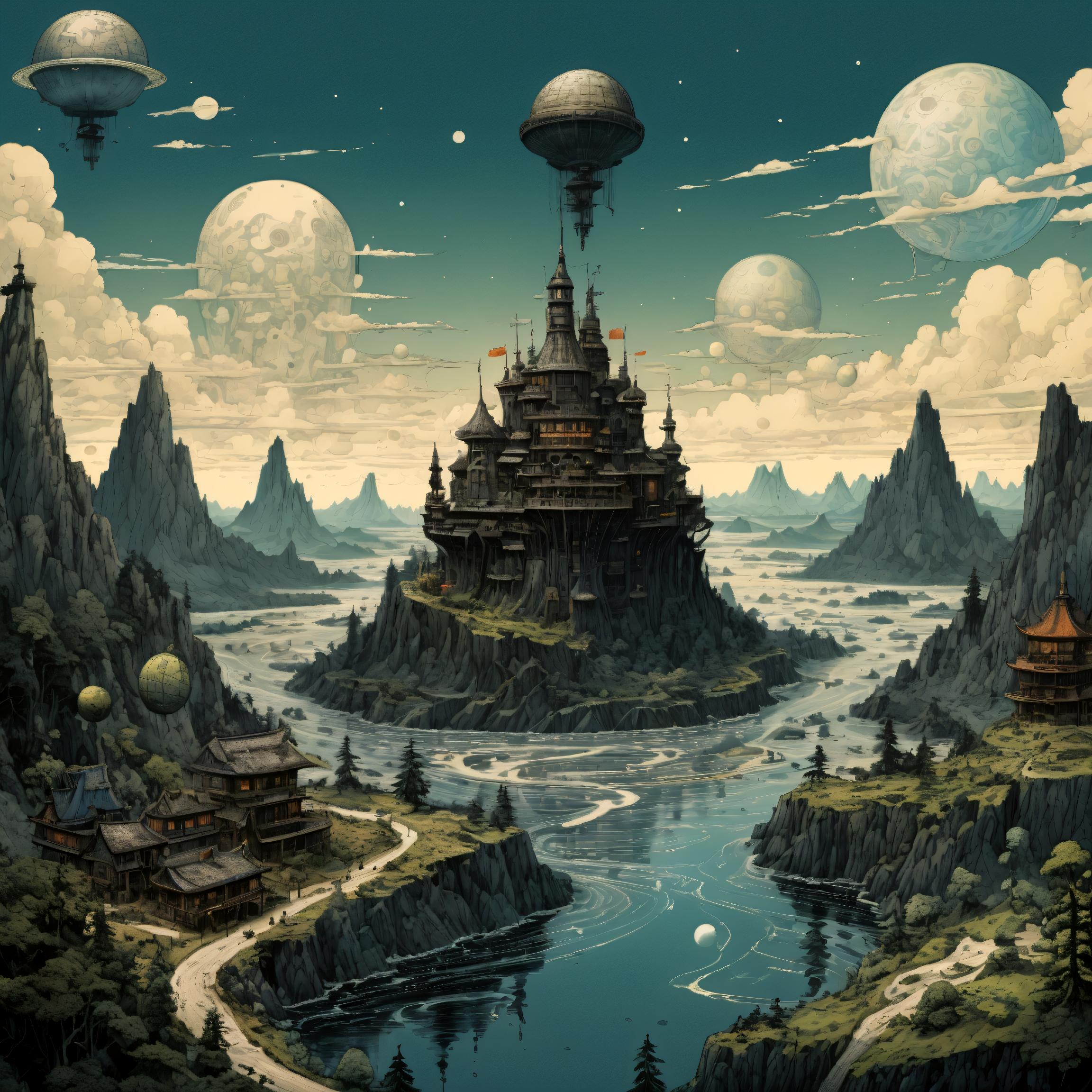} &
        \includegraphics[width=0.36\textwidth,height=0.36\textwidth]{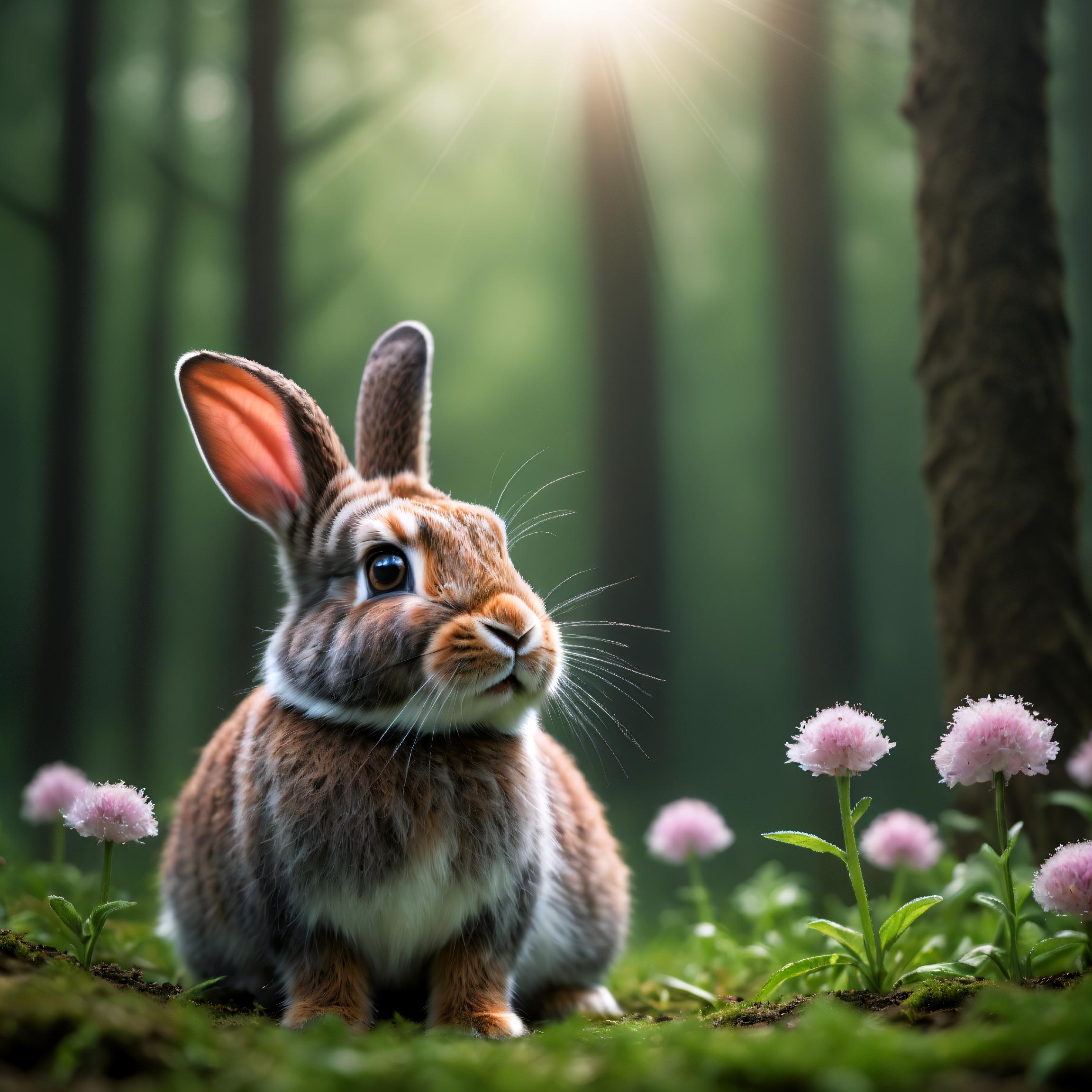} \\

        \includegraphics[width=0.36\textwidth,height=0.36\textwidth]{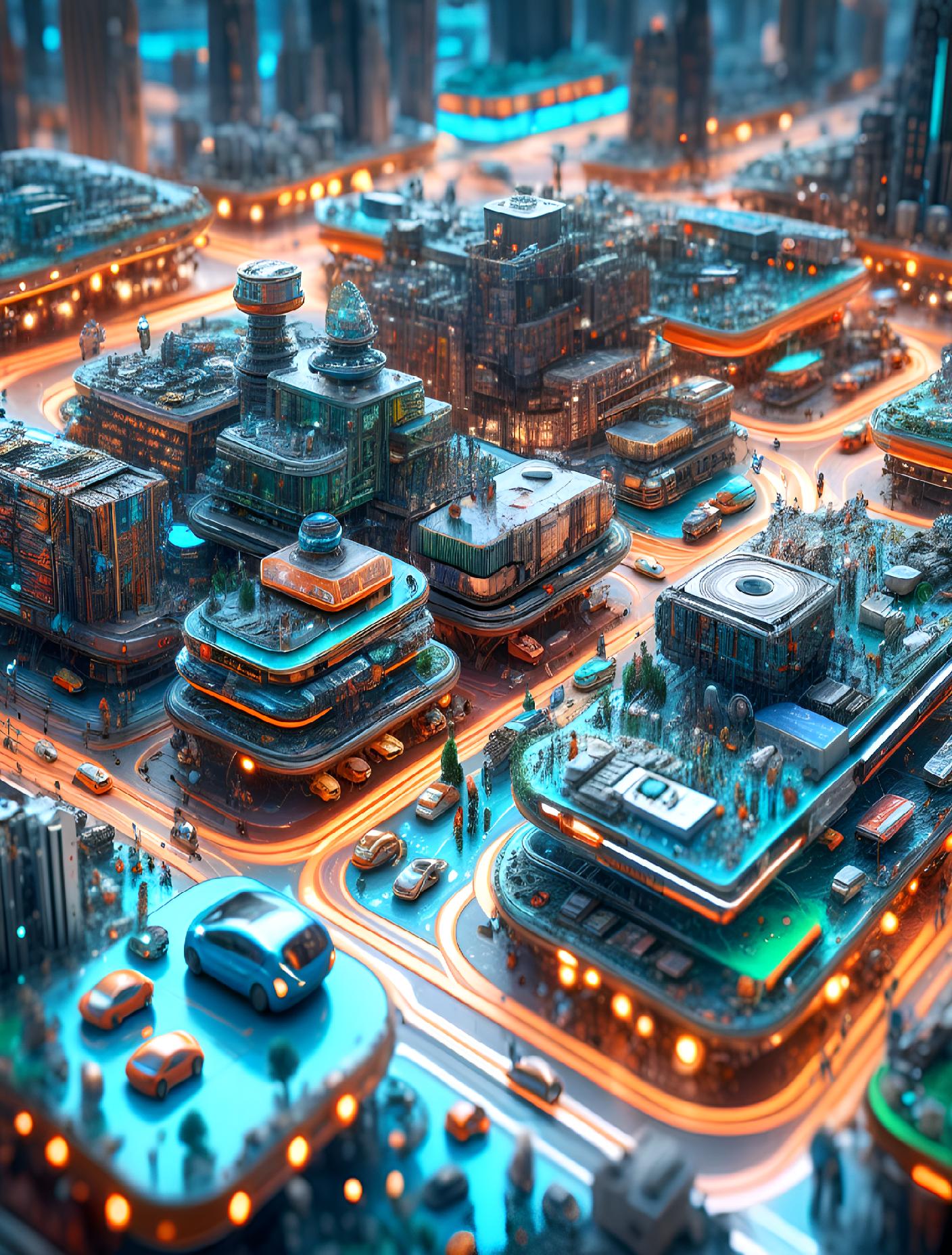} &
        \includegraphics[width=0.36\textwidth,height=0.36\textwidth]{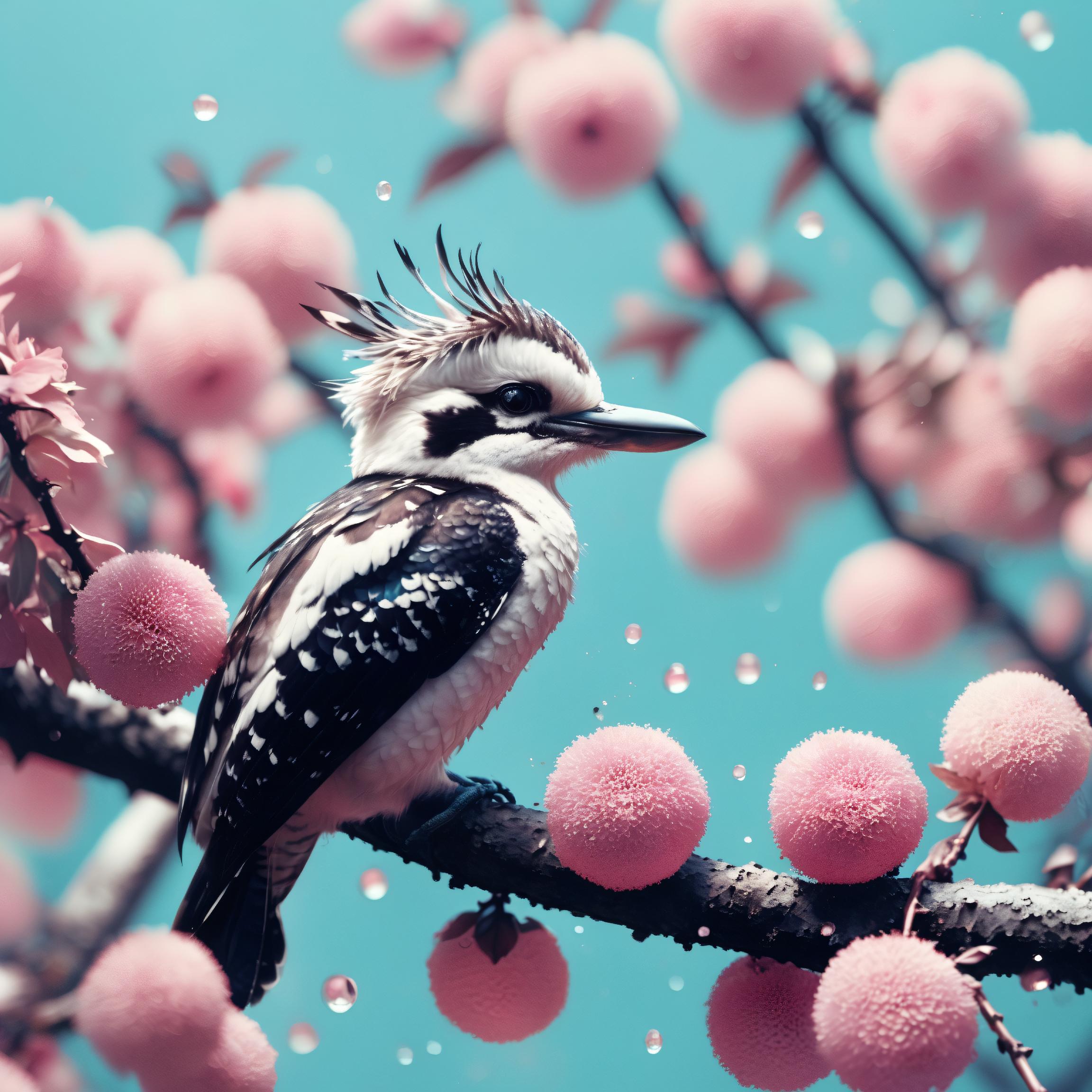} &
        \includegraphics[width=0.36\textwidth,height=0.36\textwidth]{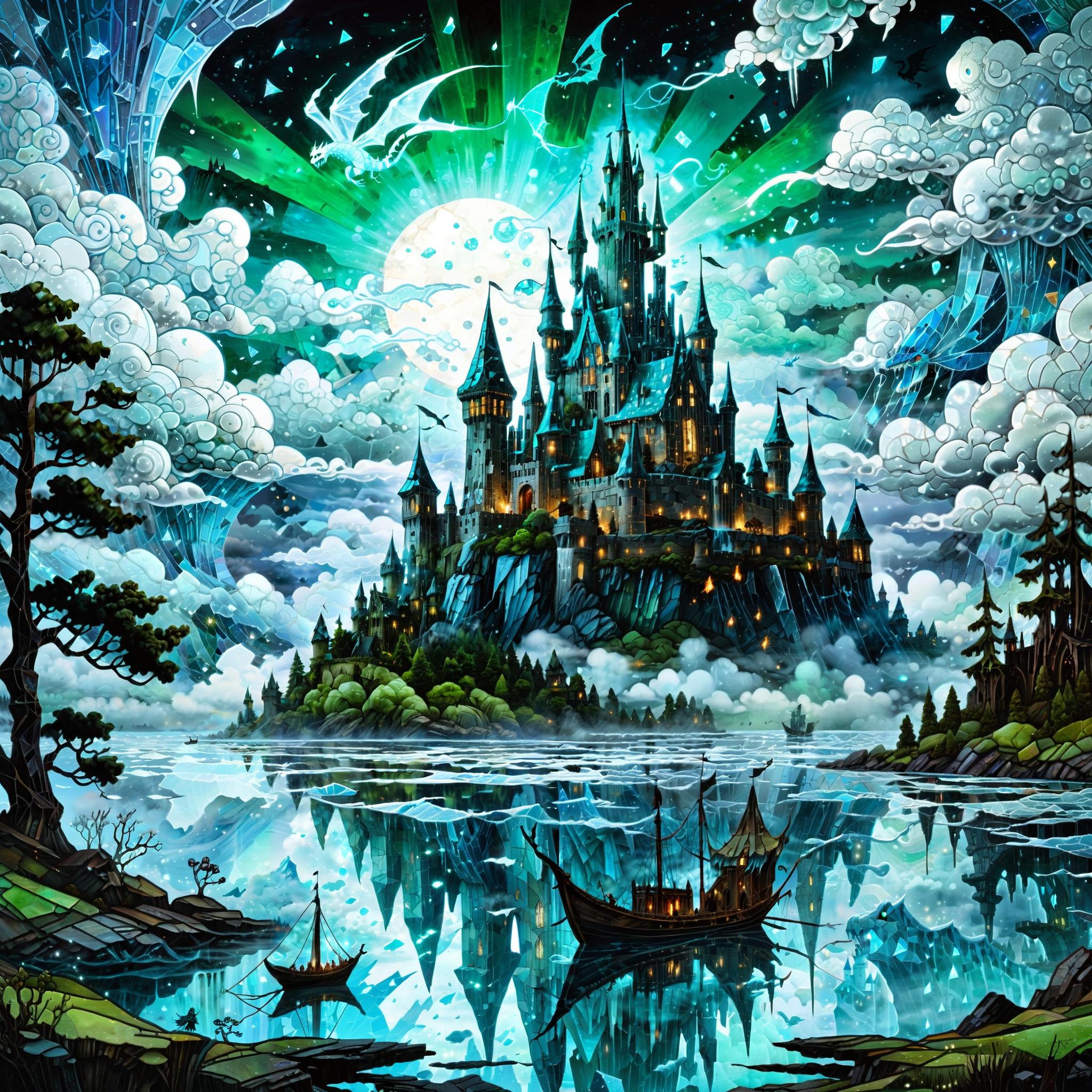} \\
        \includegraphics[width=0.36\textwidth,height=0.36\textwidth]{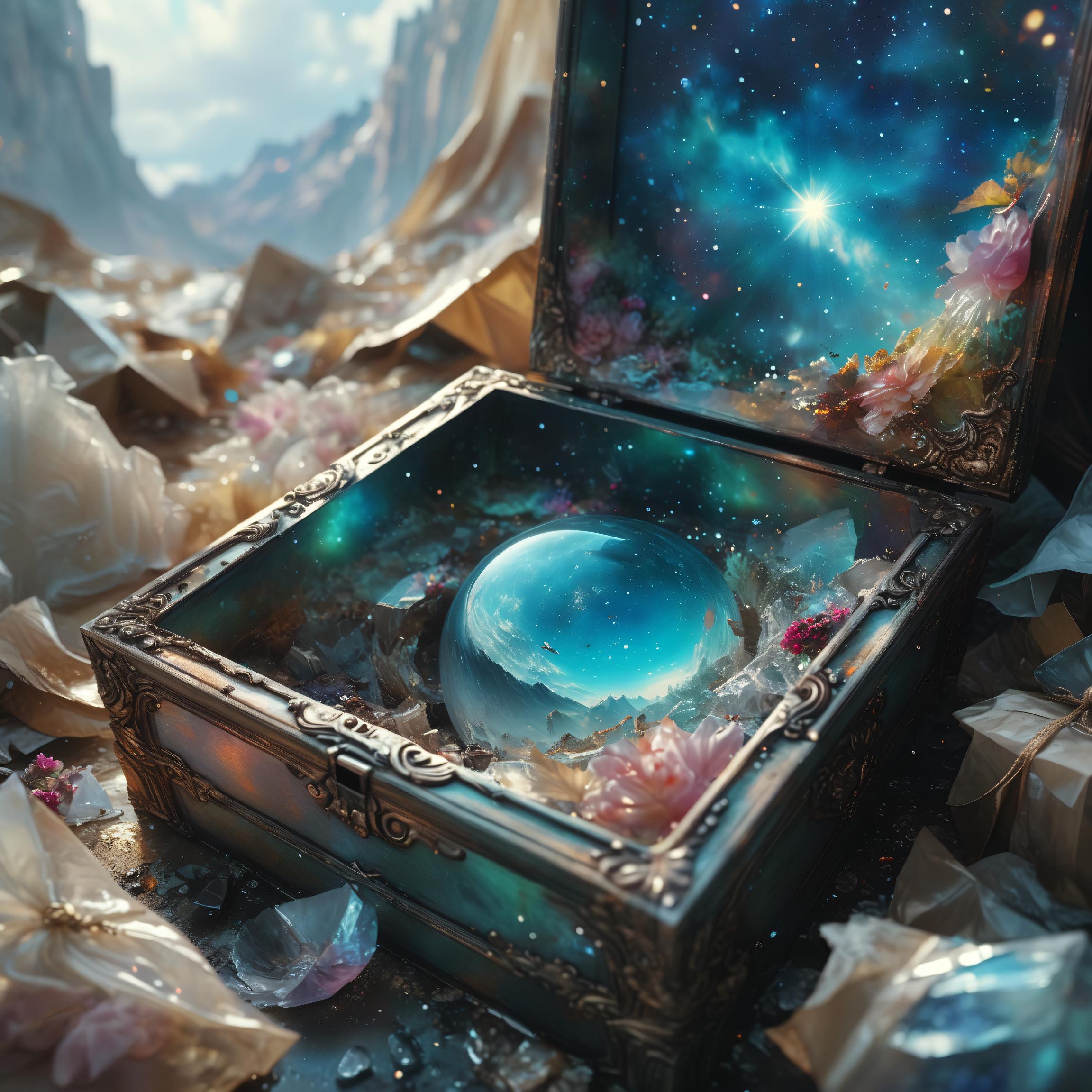} &
        \includegraphics[width=0.36\textwidth,height=0.36\textwidth]{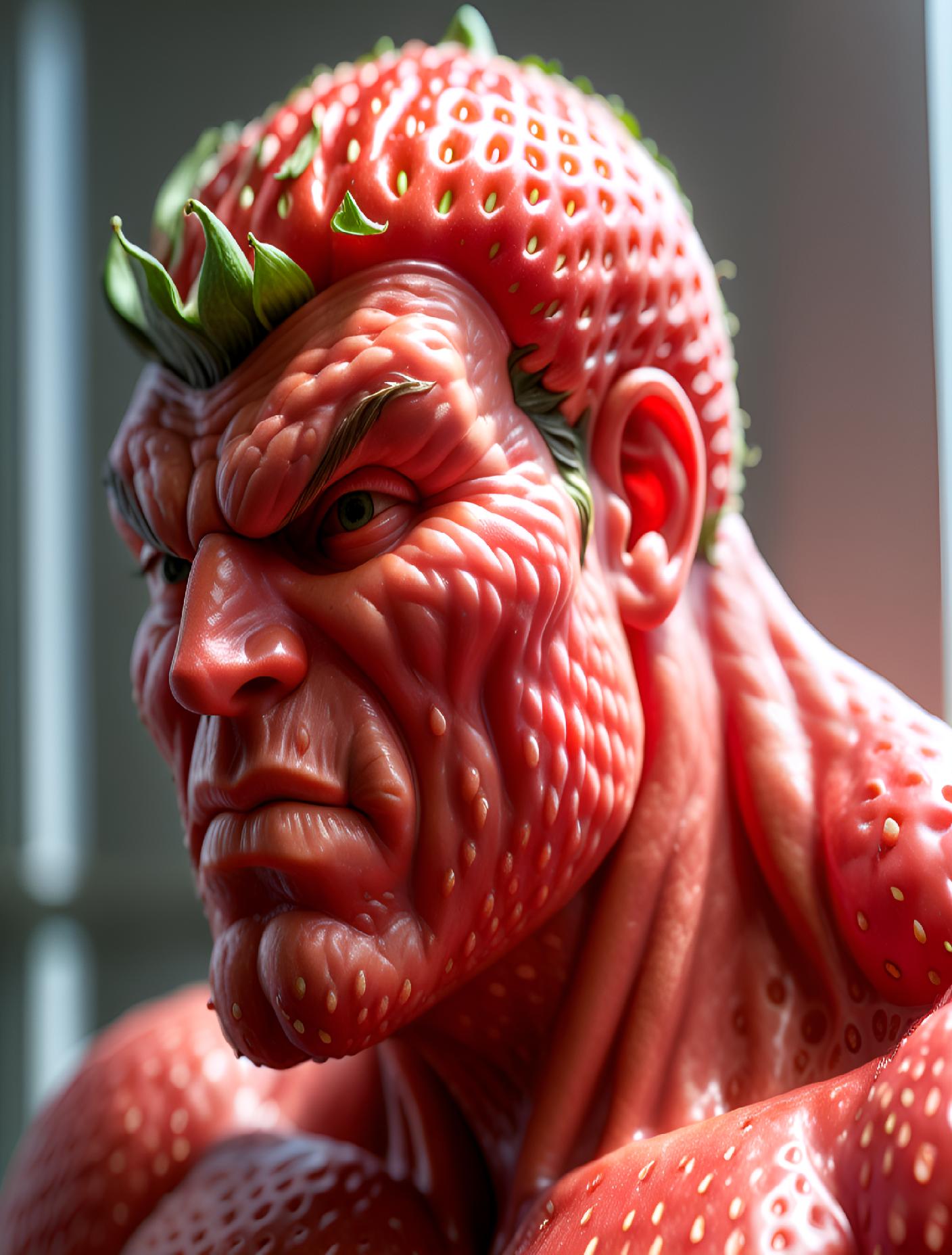} &
        \includegraphics[width=0.36\textwidth,height=0.36\textwidth]{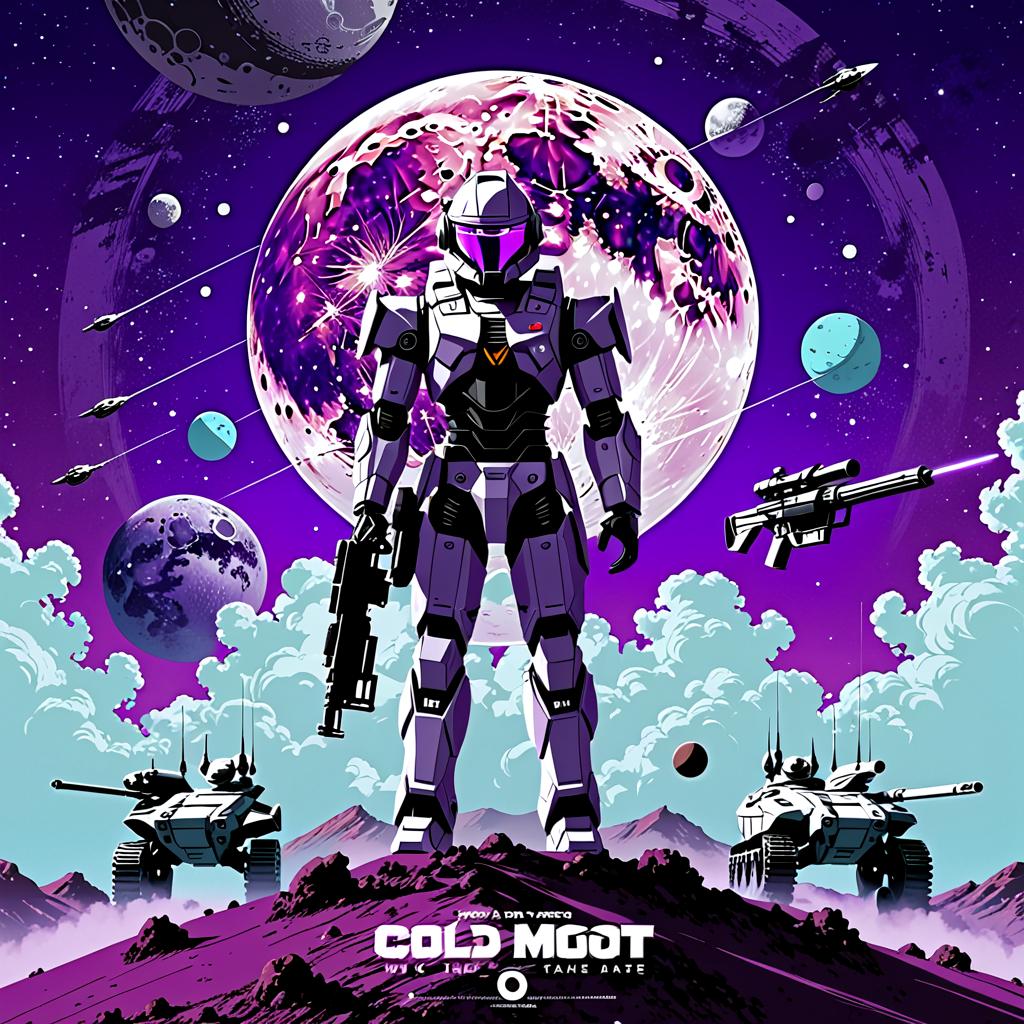} \\

    \end{tabular}
    }
    \caption{Curated images generated using ComfyGen-FT. The list of prompts is available in the supplementary.}\label{fig:ours_large_supp_ft_1}
\end{figure*}

\begin{figure*}
    \centering
    \setlength{\tabcolsep}{0.5pt}
    {\normalsize
    \begin{tabular}{c c c}
        \includegraphics[width=0.36\textwidth,height=0.36\textwidth]{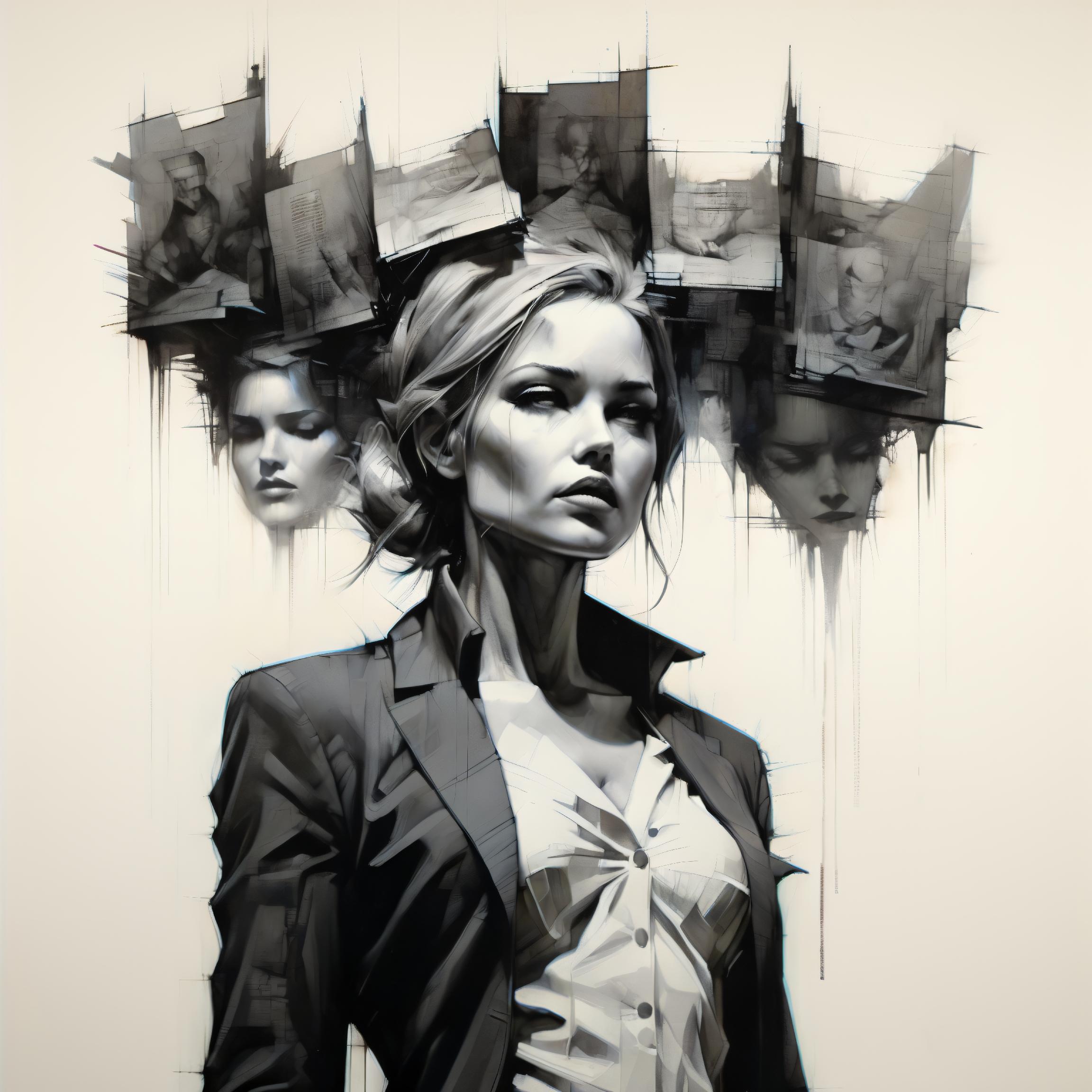} &
        \includegraphics[width=0.36\textwidth,height=0.36\textwidth]{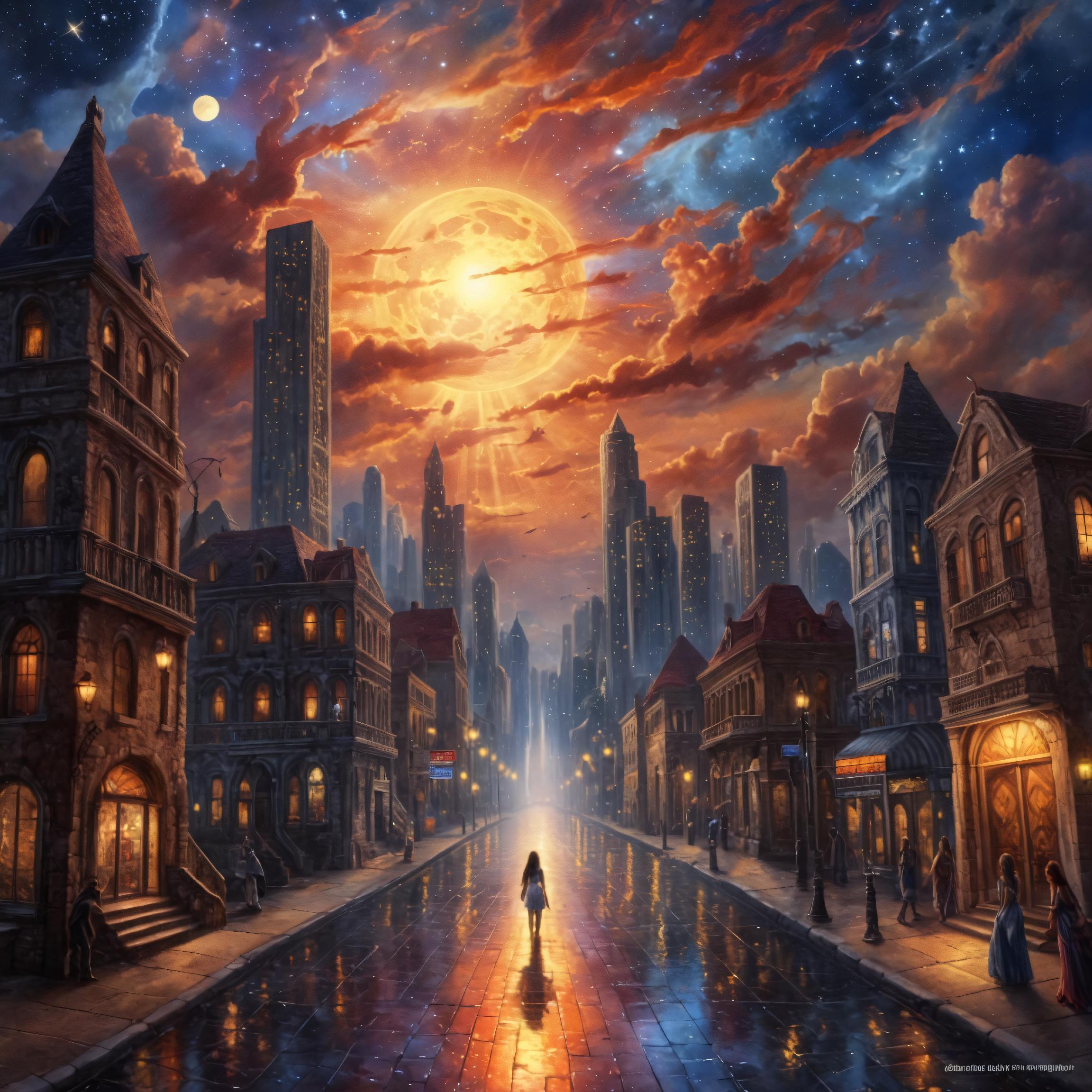} &
        \includegraphics[width=0.36\textwidth,height=0.36\textwidth]{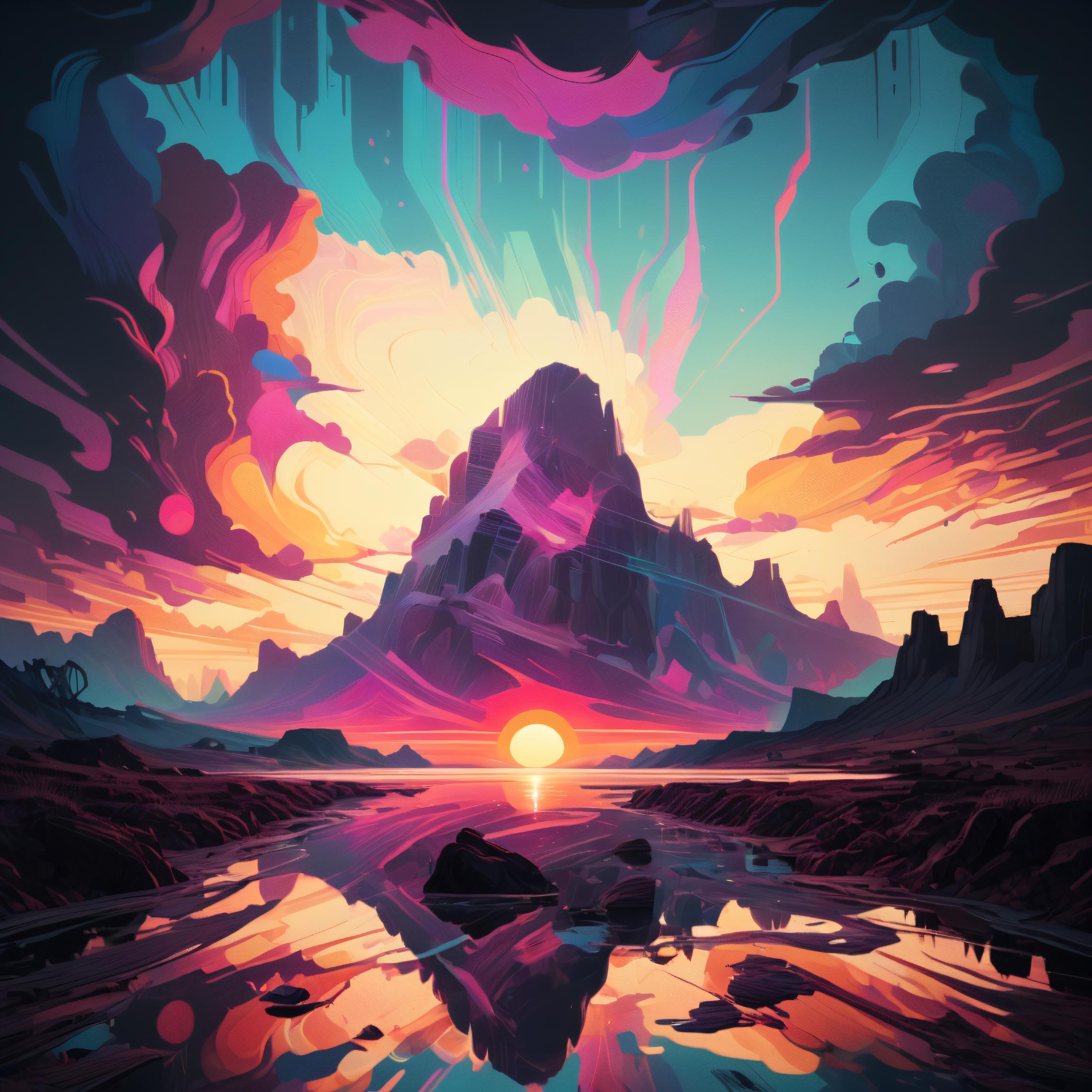} \\

        \includegraphics[width=0.36\textwidth,height=0.36\textwidth]{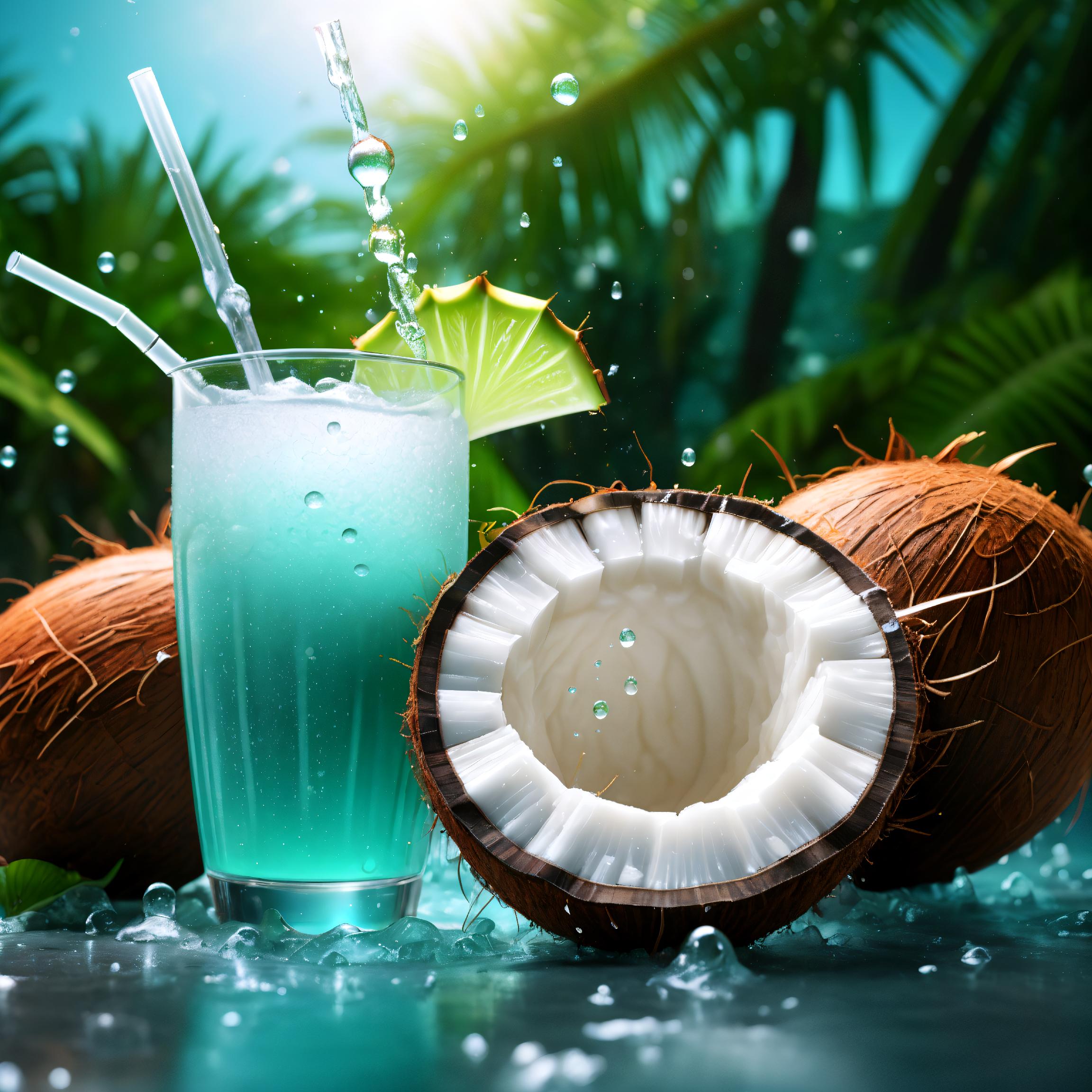} &
        \includegraphics[width=0.36\textwidth,height=0.36\textwidth]{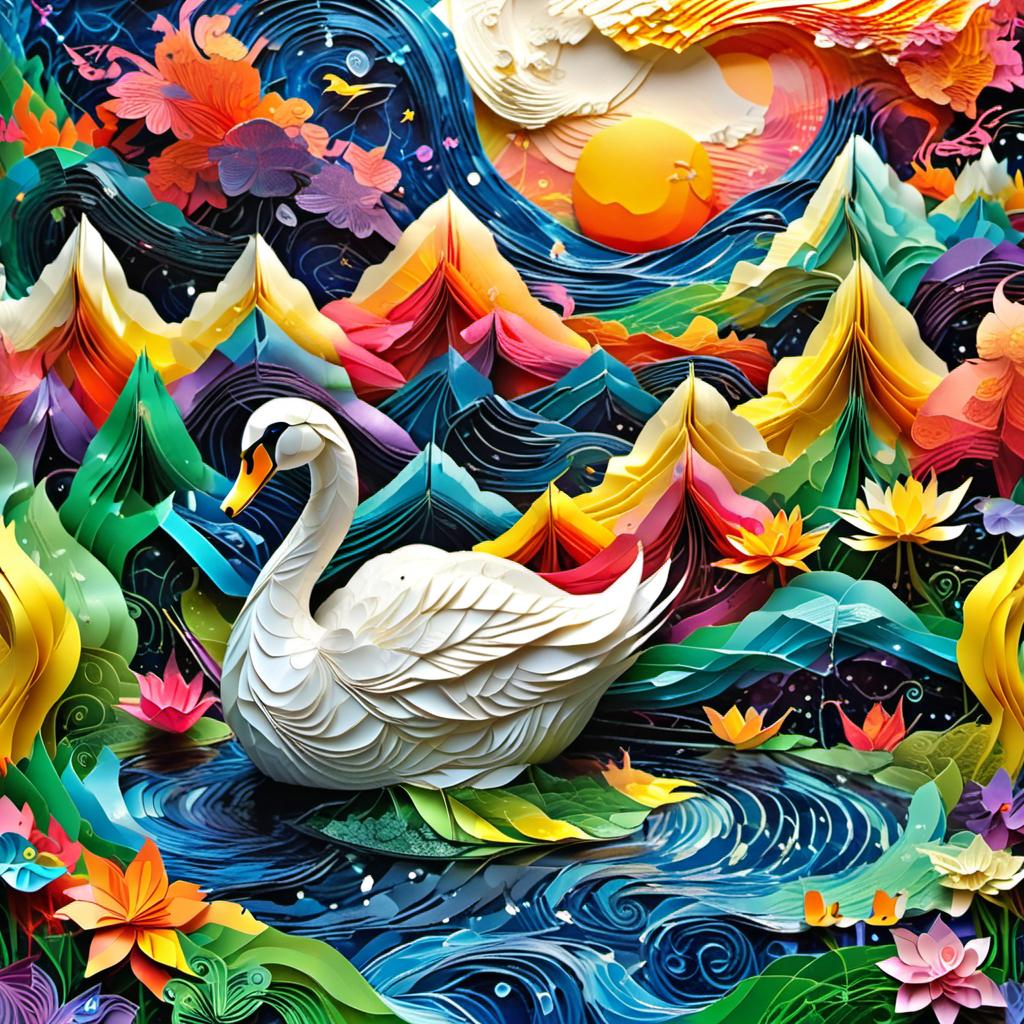} &
        \includegraphics[width=0.36\textwidth,height=0.36\textwidth]{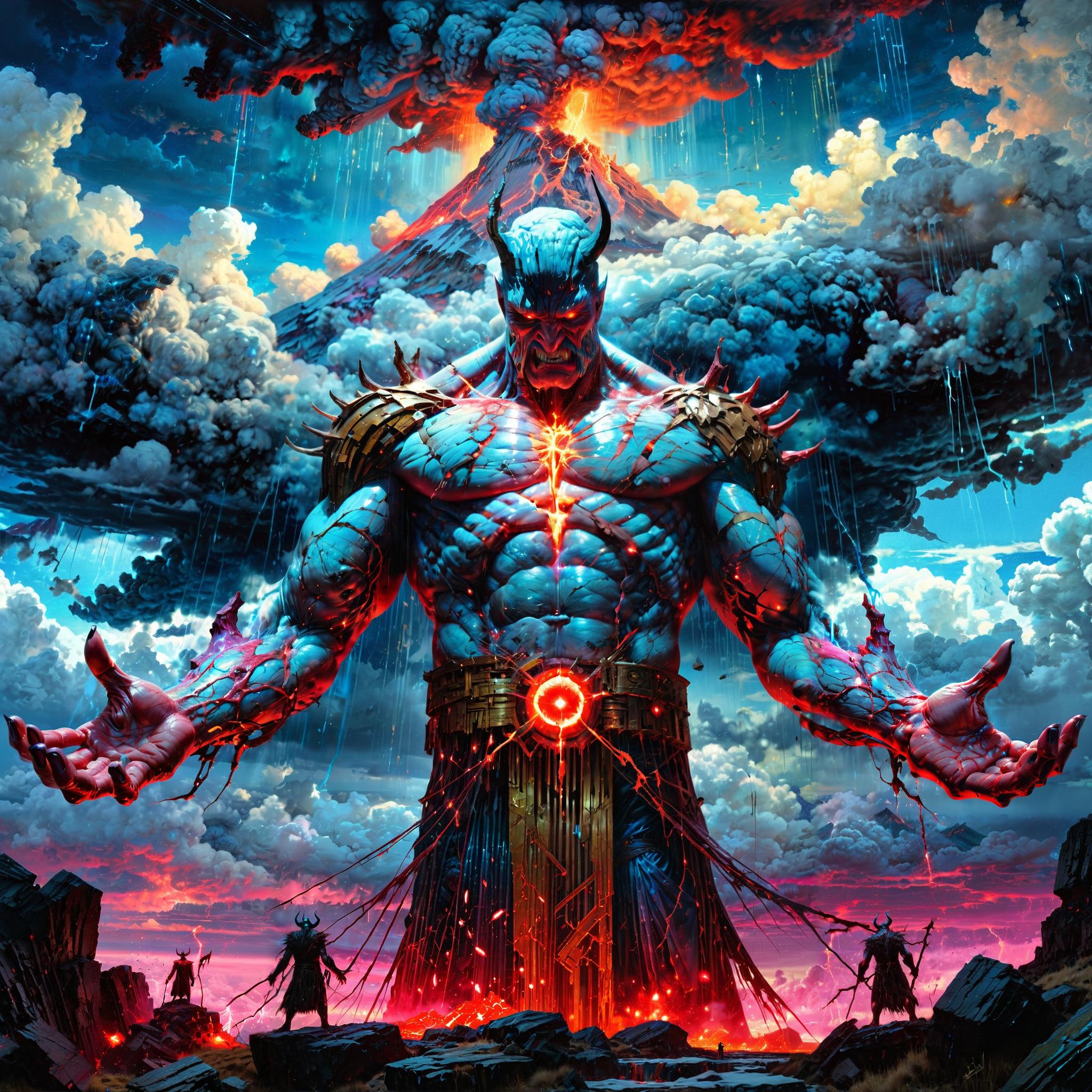} \\

        \includegraphics[width=0.36\textwidth,height=0.36\textwidth]{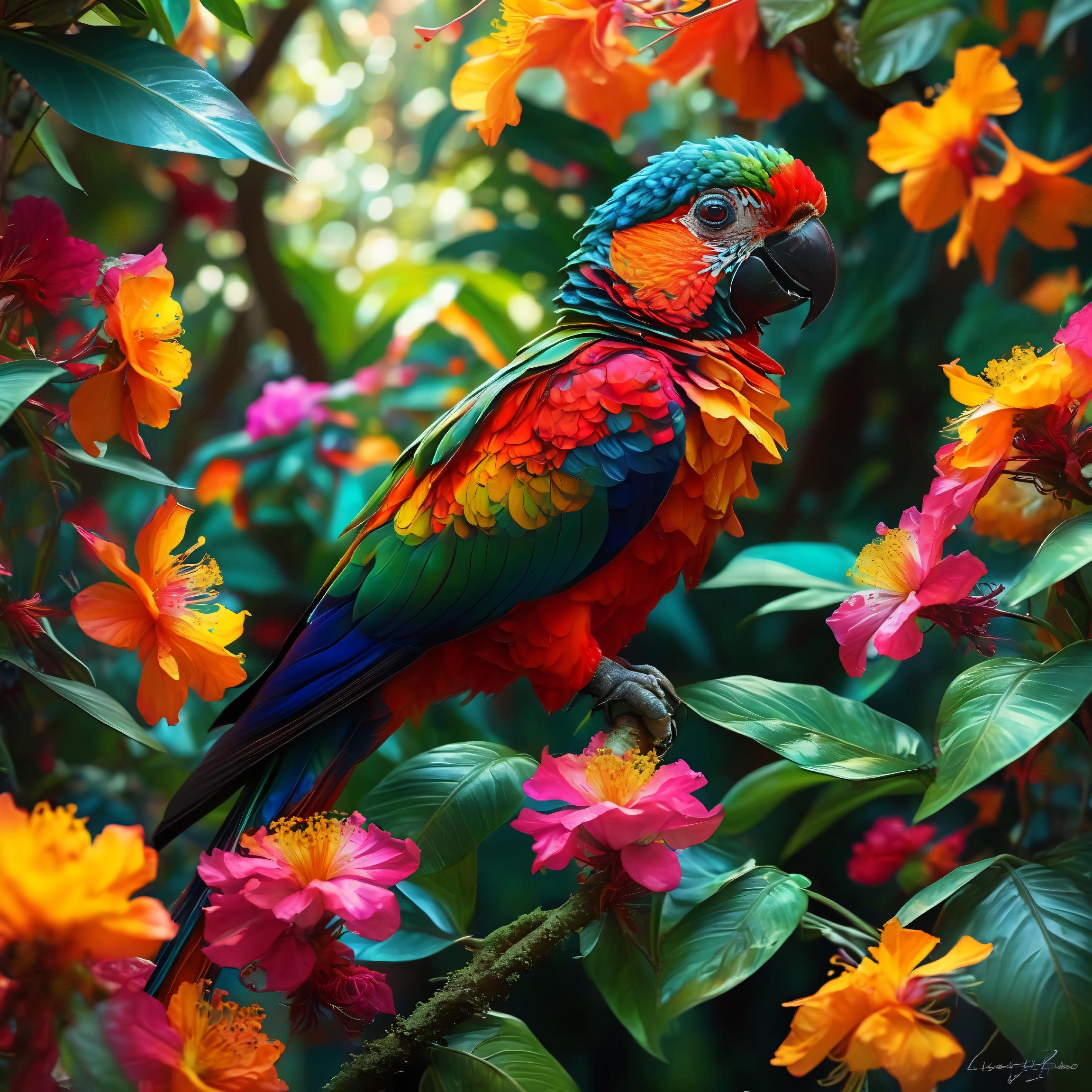} &
        \includegraphics[width=0.36\textwidth,height=0.36\textwidth]{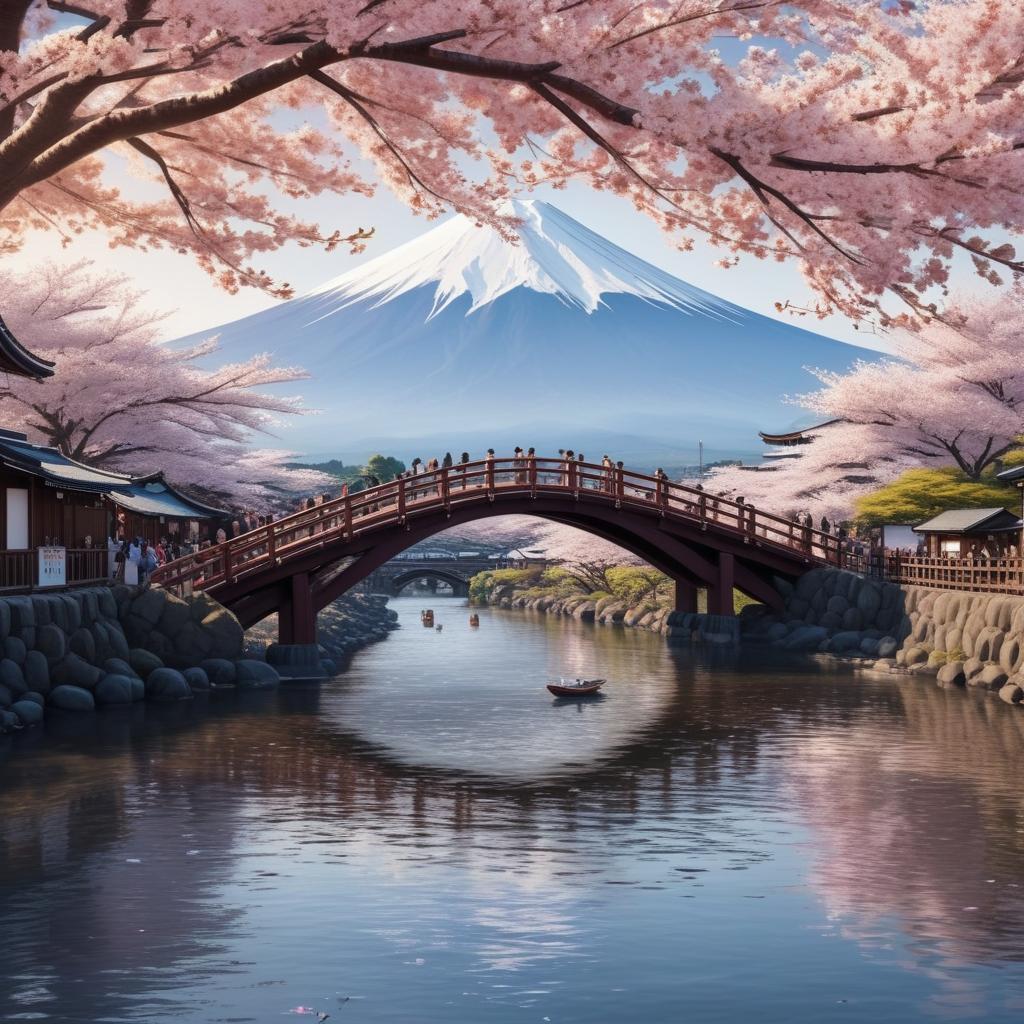} &
        \includegraphics[width=0.36\textwidth,height=0.36\textwidth]{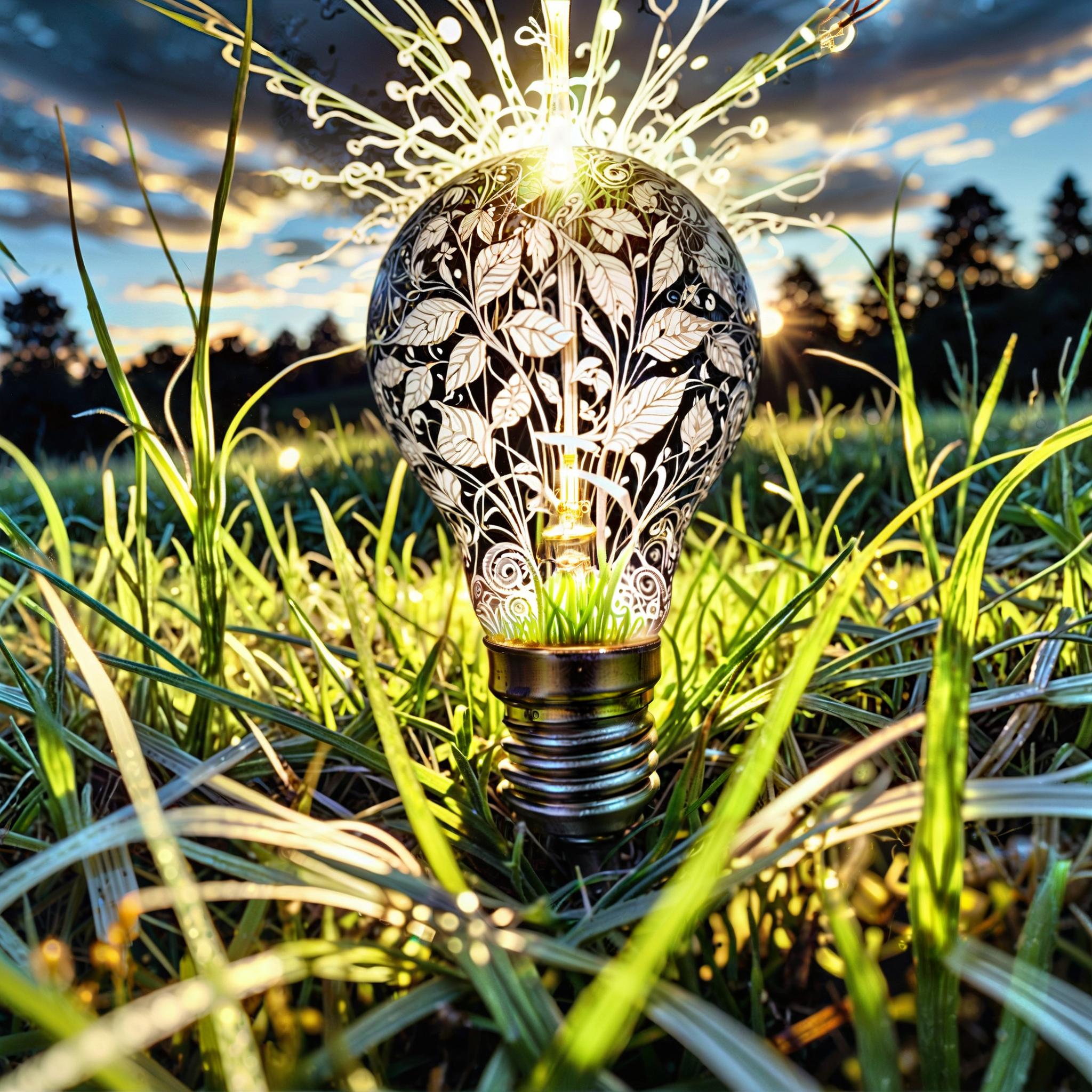} \\

    \end{tabular}
    }
    \caption{Curated images generated using ComfyGen-FT. The list of prompts is available in the supplementary.}\label{fig:ours_large_supp_ft_2}
\end{figure*}

In \cref{fig:ours_large_supp_ft_1,fig:ours_large_supp_ft_2,fig:ours_large_supp_icl_1,fig:ours_large_supp_icl_2} we show additional images generated using our method. \cref{fig:ours_large_supp_icl_1,fig:ours_large_supp_icl_2} show images generated using ComfyGen-IC for workflow selection, while \cref{fig:ours_large_supp_ft_1,fig:ours_large_supp_ft_2} show images created with flows from ComfyGen-FT.

These results showcase the generalizability of our approach to a wide range of prompts, from portraying unique artistic styles to portraying photo-realistic and imagined scenes.

\section{Additional implementation details}
\subsection{ComfyGen-IC}
Here, we provide additional details on our ComfyGen-IC approach, including the prompts used to create the class labels, and those used for workflow selection.

To generate the set of category labels, we prompted Claude Sonnet 3.5 with the following:

\fbox{\parbox{\linewidth}{``Given the following list of image prompts, generate a list of 20 labels that describe the key elements, styles, or themes of the images these prompts might produce. \\ Focus on general descriptors like 'people', 'photo-realistic', 'photo-artistic', 'nature', 'abstract', etc. \\ \\ Here are the prompts: [prompts]"}}

This produced the following list of category labels: 'People', 'Photo-realistic', 'Photo-artistic',  'Fantasy', 'Sci-fi', 'Horror', 'Anime', 'Abstract',  'Surreal', 'Cyberpunk', 'Steampunk', 'Gothic',  'Digital art',  'Portrait', 'Nature', 'Landscape', 'Wildlife', 'Urban', 'Cosmic', 'Underwater'. 

To assign labels to each text-to-image training prompt, we prompted the LLM with the following:

\fbox{\parbox{\linewidth}{
"Given the following image prompt and list of labels, select the most relevant labels that describe the key elements, styles, or themes of the image this prompt might produce. Provide only the selected labels, separated by commas. \\ \\
Image prompt: [prompt] \\ \\
Available labels: [labels] \\ \\
Selected labels:"
}}

The LLM assigned an average of $4.96 \pm 1.33$ labels per prompt, with a maximum of 10. For 2 prompts ($0.4\%$ of the training set), the LLM failed to assign any label and so they were discarded.

The full table of flows and scores amounted to roughly $80,000$ tokens. To further reduce this, we filter out all flows which achieved a score below the median for every single category. This filters out a total of $125$ (or roughly $40\%$ of the flows).

The filtered table serves at the context which we provide to the LLM with the following flow selection prompt:

\fbox{\parbox{\linewidth}{``[context] \\

Please classify the following prompt into one of the flows mentioned above:\\

[prompt]
\\

Provide the flow ID and a brief explanation for your classification."}}

As part of our investigation, we also ablated the use of $30$ labels and different LLM labeling prompts, but were unable to improve results over this baseline.

\subsection{ComfyGen-FT}
Our ComfyGen-FT models were finetuned using 4-bit quantization, with a batch size of $8$ and $4$ gradient accumulation steps. We used an AdamW optimizer with weight decay of $0.01$. We further use 5 warmup steps, and a linear LR scheduler. 

All experiments were conducted on a single NVIDIA H100 GPU.

During training and inference, we use the following instruction when prompting the LLM:

\fbox{\parbox{\linewidth}{``Below is a prompt that describes an image a user wants to generate, and a numerical score describing the quality of an image. Please output a ComfyUI workflow in json format that will create an image with this score when given the prompt.\\

\textgreater\textgreater\textgreater{} Prompt:\\

[prompt] \\

\textgreater\textgreater\textgreater{} Score: \\

[score] \\

\textgreater\textgreater\textgreater{} Flow:"}}

\section{Image prompts}

Below, we provide the prompts used to synthesize all the large-figure images in the core paper and in \cref{sec:supp_additional_results}.
In each table, we point to the image location using the figure number and a grid position. The grids are 0 indexed, and list the image location from top-to-bottom and left-to-right.

\section{List of models appearing in our workflows}
Recall that we augment human created flows by randomly swapping the models that they use. Below is a list of all models that appear in our dataset. All base models are on the scale of SDXL and below.

\section{Additional User Study details}

\begin{figure}[!h]
    \centering
   \setlength{\fboxsep}{0pt}  %
\setlength{\fboxrule}{1pt}  %
     \fbox{\includegraphics[width=\linewidth]{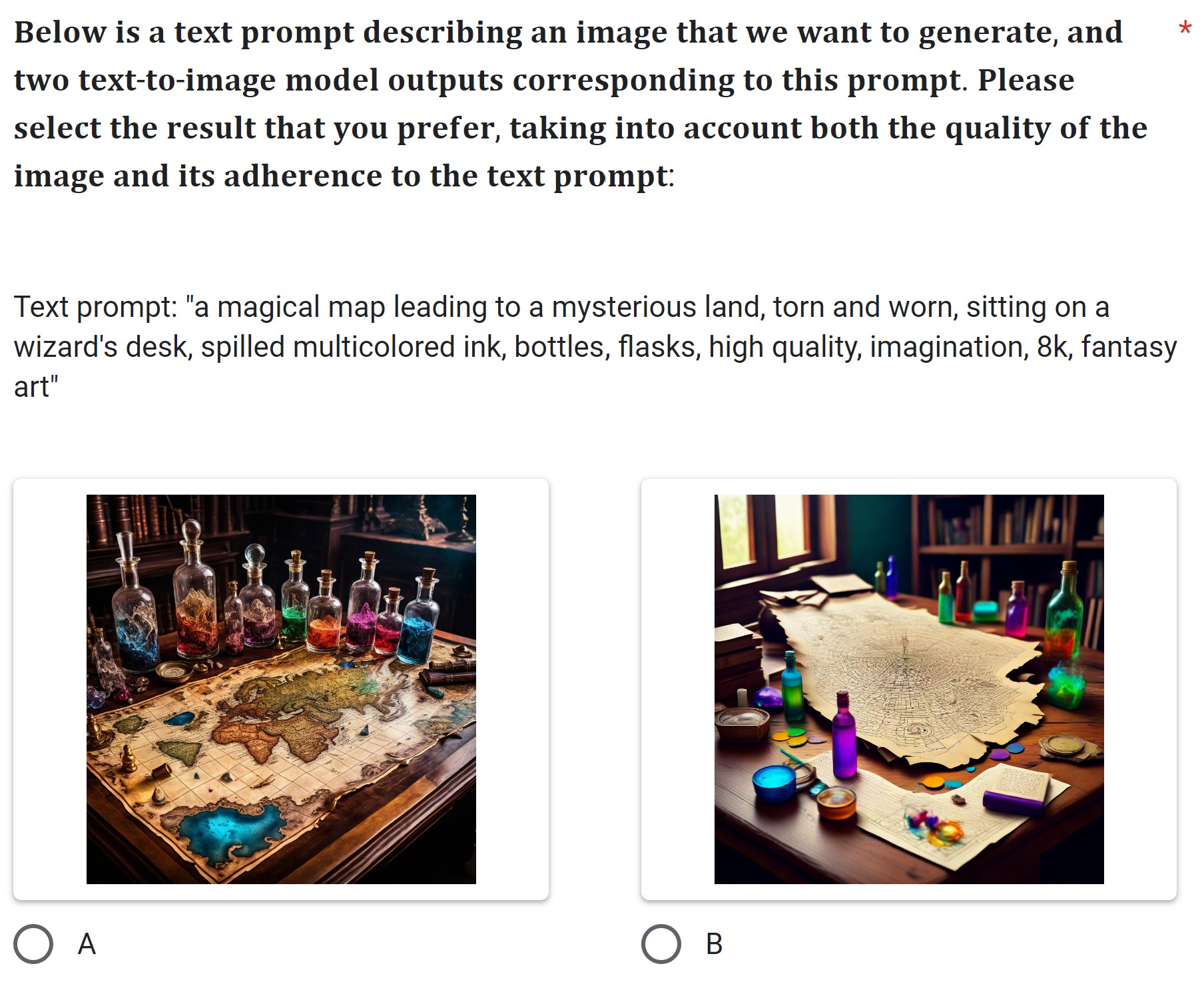}}
    \caption{An example question from our user study.}
    \label{fig:user_study_example}
\end{figure}

In \cref{fig:user_study_example} we show an example question from our user study form. To create the forms, we randomly sampled $120$ prompts for ComfyGen-FT and $120$ prompts for ComfyGen-IC. We divided each group of $120$ questions into $6$ blocks of $20$, and matched each block to a baseline. From this matching, we created $240$ questions that pit one of our methods against one of the baselines. We discarded all questions for which at least one of the images contained nudity or overly sexualized content, leaving us with a total of $231$ questions. Each question was initially answered by at least $3$ users, but we discarded all responses from users that picked at least $2$ answers where the chosen image had no relation to the input prompt (\eg, a scenario where the prompt describes a cigar, one image shows a cigar, and the user picked an image that shows an empty street).

\clearpage

\begin{table}[h]
\centering
\small
\begin{tabularx}{\textwidth}{|l|>{\raggedright\arraybackslash}X|}
\hline
\textbf{Figure} & \textbf{Prompt} \\
\hline
Fig. 3 $(0, 0)$ & ``The entire observable universe in a single bottle, Dreamlike, Surreal landscapes, Mystical creatures, Twisted reality, Surreal still life, Extremely Detailed Oil Painting, glow effects, god rays, Hand drawn, render, 8k, cartoon, octane render, cinema 4d, blender, dark, atmospheric 4k ultra detailed, cinematic sensual, Sharp focus, humorous illustration, big depth of field, Masterpiece, colors, 3d octane render, 4k, concept art, trending on artstation, hyper realistic, Vivid colors, extremely detailed CG unity 8k wallpaper, trending on ArtStation, trending on CGSociety, Intricate, High Detail, dramatic, masterpiece, best quality, ultra-detailed, unreal engine, octane render, HDR"  \\
\hline
Fig. 3 $(0, 1)$ & ``cottoncandy, In a whimsical candy kingdom, there stands a unicorn crafted from marshmallows, exuding childlike wonder and an air of mystery. Its body is as white and fluffy as a cloud, embodying the spirit of a playful sprite emerging from sweet dreams, Its head features a radiant rainbow candy gem that forms the iconic horn, shining with seven-colored brilliance, seemingly illuminating every innocent aspiration. The eyes are crafted from translucent hard candies, filled with tender yet inquisitive expressions, The mane consists of multi-hued strands of marshmallow fluff, each one dazzling like a rainbow, swaying gently in the breeze, whispering enchanting tales. Its tail resembles a billowy marshmallow cloud, soft and dreamy; when it gallops joyfully, it leaves behind a trail of vibrant marshmallow magic, Its hooves resemble those cast from creamy white chocolate, solid yet sweet, firmly planted on the ground made of colorful candies. This marshmallow unicorn stands tall, serving as both a tangible representation of children's naive fantasies and a mystical presence brimming with magical allure within the fairy tale world, UHD, Extreme detail, natural light, volume light, fantasism, professional color, professional composition"  \\
\hline
Fig. 3 $(0, 2)$ & ``Masterpiece double exposure of a girl silhouette blending with monochrome apocalypse aftermath and a colorful natural landscape in the underlying backdrop, sharp contrast, detailed crisp lines, in focus"  \\
\hline
Fig. 3 $(1, 0)$ & ``A photo of a large man with beard riding a vespa wearing loriseg armor and helmet, red tunic"  \\
\hline
Fig. 3 $(1, 1)$ & ``A hyper-realistic, ultra-detailed painting featuring a fantastical landscape with a village at the end of time. The scene has a perfect composition and chiaroscuro inspired by Rembrandt, with concept art elements by Mariusz Lewandowski, Louis Aston Knight, Karine Eibatova, John Howe, Jessica Rossier, JarosÅaw JaÅnikowski, Inessa Garmash, HenriÃtte Ronner-Knip, and Harumi Hironaka. The painting is vibrant, beautiful, painterly, detailed, and textural. Watercolor painting, masterpiece, best quality, hyper detailed, ultra realistic, 32k, RAW photo, landscape, fantasy, village at the end of time, perfect composition, chiaroscuro by Rembrandt, concept art, vibrant, beautiful, painterly, detailed, textural, artistic, Rembrandt, Mariusz Lewandowski, Louis Aston Knight, Karine Eibatova, John Howe, Jessica Rossier, JarosÅaw JaÅnikowski, Inessa Garmash, HenriÃtte Ronner-Knip, Harumi Hironaka., extremely detailed, swirling ink"  \\
\hline
Fig. 3 $(1, 2)$ & ``true masterpiece, masterpiece cinematic lighting, cinematic shot from below, extremely detailed, high detail, hires textures, incredibly detailed, intricate details, photorealism, intricately designed, sea storm up inside a large transparent glass ball, drops outside the ball rainfall on floor background, HD32k, focus"  \\
\hline
Fig. 3 $(2, 0)$ & ``no humans, scenery, closeup, branch, tree, leaf, nature, rain, outdoors, depth of field, droplets"  \\
\hline
Fig. 3 $(2, 1)$ & ``fine art, oil painting, best quality, dark tales, illustration, each color adds depth, and the entire piece comes together to create a breathtaking spectacle of motion and tranquility., while the ball is adorned with an array of stripes in various hues. the figurine, while her right hand delicately holds a small, epic splash cover art in the van gogh style, starry sky, dan mumford, andy kehoe, 2d, flat, delightful, vintage, art on a cracked paper, patchwork, stained glass, fairytale, storybook detailed illustration, cinematic, ultra highly detailed, tiny details, beautiful details, mystical, luminism, vibrant colors, complex background"  \\
\hline
Fig. 3 $(2, 2)$ & ``a ral-sun flying flower, dusty background, epic, heroic, bokeh, sharp detailed, hyperrealistic, amazing, macro photo, god rays, volumetric"  \\
\hline
\end{tabularx}
\caption{List of prompts used for Fig. 3 in the core paper.}
\label{tab:fig3_prompts}
\end{table}

\begin{table}[h]
\centering
\small
\begin{tabularx}{\textwidth}{|l|>{\raggedright\arraybackslash}X|}
\hline
\textbf{Figure} & \textbf{Prompt} \\
\hline
Supp. Fig. 1 $(0, 0)$ & ``2D anime style, elegant girl floating in a vortex of city lights and digital elements, deep and expressive eyes, vibrant blue and purple colors, intricate details, magical and futuristic ambiance, highly detailed background with swirling buildings and lights, complex and dynamic scene"  \\
\hline
Supp. Fig. 1 $(0, 1)$ & ``A close-up portrait of an alien being, with luminescent skin and eyes that hold centuries of wisdom, set against the backdrop of their advanced, technology-filled habitat., 90th photos from photo album, film, vintage style, flash from camera, hyper realistic, fine textures, high quality textures of materials, volumetric textures, natural textures, natural colors, correct white balance, color correction, dehaze, clarity"  \\
\hline
Supp. Fig. 1 $(0, 2)$ & ``masterpiece, best quality, high quality, intricate, absurdres, very aesthetic, no humans, landscape, outdoors, mountain tops, wind, windy, wind lines, clouds, above clouds, cliff, wind magic, aurora, ultra wide angle shot, cinematic style, highly detailed, extremely detailed, sharp detail, majestic, shallow depth of field, movie still, soft light, circular polarizer, colorful, wallpaper, professional illustration, anime"  \\
\hline
Supp. Fig. 1 $(1, 0)$ & ``adventurous cute funny lizard, donned a tiny explorer's hat and carried a mini backpack, walk in wild colorfully jungle, detailed scales, by Jean-Baptiste Monge, Gilles Beloeil, Tyler Edlin, Marek Okon, Pixar, 8k, album art, comic style, golden ratio, perfect composition, a masterpiece, trending on artstation, extreme close up, shot from below . High dynamic range, vivid, rich details, clear shadows and highlights, realistic, intense, enhanced contrast, highly detailed"  \\
\hline
Supp. Fig. 1 $(1, 1)$ & ``Amazing detailed photography of a cute adorable samurai kitten holding Katana with 2 paws, Cherry Blossom Tree petals floating in air, high resolution, piercing eyes, lifelike fur, Anti-Aliasing, FXAA, De-Noise, Post-Production, SFX, insanely detailed and intricate, hypermaximalist, elegant, ornate, hyper realistic, super detailed, noir coloration, serene, 16k resolution, full body"  \\
\hline
Supp. Fig. 1 $(1, 2)$ & ``bg\_imgs, portrait, wallpaper, colorful, highres, absurdres, huge filesize, fantasy, foreshortening, black dress, Extremely gorgeous metal style, Metal crown with ornate stripes, Various metals background, Sputtered molten iron, floating hair, Hair like melted metal, Clothes made of silver, Clothes with gold lace, flowing gold and silver, everything flowing and melt, flowing iron, flowing silver, lace flowing and melt, best quality, masterpiece, illustration, an extremely delicate and beautiful, extremely detailed, CG, unity, 8k wallpaper, Amazing, finely detail, masterpiece, best quality, official art, extremely detailed CG unity 8k wallpaper, absurdres, incredibly absurdres, huge filesize, ultra-detailed, highres, extremely detailed, beautiful detailed girl, extremely detailed eyes and face, beautiful detailed eyes, light on face"  \\
\hline
Supp. Fig. 1 $(2, 0)$ & ``4n1v3rs3,  4n1v3rs3, 2, Create an art deco inspired illustration of A scenic lighthouse perched on a rocky coastline, overlooking the turquoise waters of Cala Ratjada. Include vibrant pastel colors, sleek lines, and a retro summer atmosphere. The style should be reminiscent of vintage travel posters with a modern twist"  \\
\hline
Supp. Fig. 1 $(2, 1)$ & ``lighting Style, dim light, low light, dramatic light, partially covered in shadow, award winning photography, RAW photo, Hyperrealistic, beautiful African woman, wearing traditional outfit, vibrant colors, intricate patterns, detailed textures, soft natural lighting, ethereal glow, participating in a traditional ceremony, serene and dignified expression, traditional headwrap, detailed jewelry, lush green background, realistic shadows, fine details in fabric and skin, ultra-quality, cultural richness, ceremonial setting, glowing skin, traditional makeup, colorful beads, elegant pose, cultural heritage, ceremonial attire Detailed natural skin"  \\
\hline
Supp. Fig. 1 $(2, 2)$ & ``perfect flower in the nature"  \\
\hline
\end{tabularx}
\caption{List of prompts used for Fig. 1 in the supplementary.}
\label{tab:fig1_supp_prompts}
\end{table}

\begin{table}[h]
\centering
\small
\begin{tabularx}{\textwidth}{|l|>{\raggedright\arraybackslash}X|}
\hline
\textbf{Figure} & \textbf{Prompt} \\
\hline
Supp. Fig. 2 $(0, 0)$ & ``amazing quality, masterpiece, best quality, hyper detailed, ultra detailed, UHD, perfect anatomy, in castle, girl knight, holding a sword, dazzling, transparent, polishing, 1 arm up, waving sword, gold armor, glowing armor, glowing eyes, full armor, shine armor, dazzling armor, extremely detailed"  \\
\hline
Supp. Fig. 2 $(0, 1)$ & ``closeup award winning photo of wolf, perfect environment, extremely detailed, dark shot"  \\
\hline
Supp. Fig. 2 $(0, 2)$ & ``detailed, aesthetic, 8k unreal engine photorealism, ethereal lighting, purple, nighttime, darkness, surreal art, fantasy, glowing, night, dark environment, AyameNewYears, horns, long hair, side, red kimono, floral print, hair flower, sash, haori, scenery, ink, mountains, water, trees, full body, reflection, light, arm behind back, sandals, from side"  \\
\hline
Supp. Fig. 2 $(1, 0)$ & ``Japanese Ink Drawing, Ink Dripping Drawing, horror-themed space-themed Surrealism, cartoon style, concept art surreal, anatomy, anatomical drawings, human body, fantastical organs, space demon, sharp teeth, wide grimace, masterpiece, best quality, highly detailed, sharp focus, dynamic lighting, vivid colors, texture detail, particle effects, storytelling elements, narrative flair, 16k, UE5, HDR, subject-background isolation . digital artwork, illustrative, painterly, matte painting, highly detailed, expressive, dramatic, organic lines and forms, dreamlike and mysterious, Surrealism . cosmic, celestial, stars, galaxies, nebulas, planets, science fiction, highly detailed . eerie, unsettling, dark, spooky, suspenseful, grim, highly detailed, ink drawing, dripping ink, ink drawing, inkwash, Japanese cartoon style, japanese torii"  \\
\hline
Supp. Fig. 2 $(1, 1)$ & ``a mystical, and high-resolution image of an anthropomorphic Elysium Knight in Radiant Darkness, The character should be depicted in a manga cover style with wealthy portraiture and poster art elements. Volumetric lighting, The image should feature rich colors and high contrast, focusing on the best quality, official art, and a beautiful and aesthetic appearance. wearing Sci-fi clothes and enhanced Astronaut suit, and standing in a cowboy shot pose. The character should have tattoos, muscular abs, shoulder armor, mystic Ethereal moon forest lake in the background. Emphasize a samurai theme with a detailed background and floral elements."  \\
\hline
Supp. Fig. 2 $(1, 2)$ & ``ova, background 2D anime, anime screencap, a pastel-colored character with long, flowing pink hair, large horns, and elf-like ears, wearing a cozy cable-knit sweater. The character is sitting in a beautiful, flower-filled meadow, surrounded by butterflies and small woodland creatures. She is gently petting a small bunny on her lap, smiling softly. The background is detailed with various types of flowers, trees, and a bright blue sky. The lighting is natural and bright, enhancing the cheerful and peaceful atmosphere. The camera shot is a medium shot, capturing the character and the detailed meadow."  \\
\hline
Supp. Fig. 2 $(2, 0)$ & ``watercolor art painting, watercolor, 1 forest queen, solo, sitting in the water, vintage anime aesthetic, back view, illustration, turn back her face, mysital, detailed eyes"  \\
\hline
Supp. Fig. 2 $(2, 1)$ & ``the image portrays a tranquil scene of a boat floating gently on the water, surrounded by an expansive landscape. the moon, full and glowing with a warm, reddish orange hue, casts a mystical ambiance over the entire scene. its reflection shimmers off the surface of the water, adding to the serene atmosphere. in the distance, mountains loom under the moon's soft glow, their peaks partially obscured by the low hanging clouds. they appear majestic yet gentle, as if watching over the peaceful night below. trees line the shore in the foreground, their silhouettes faintly visible against the darkening sky. this picturesque setting evokes a sense of calm and tranquility, inviting viewers to take a moment and appreciate the beauty of nature. it is a symphony of colors and shapes, each element working harmoniously together to create a visually captivating and emotionally soothing composition."  \\
\hline
Supp. Fig. 2 $(2, 2)$ & ``spectacular digital rendering of a rear view of a transparent hyper car revealing internal mechanical components such as engine, car chassis, suspension, and internal wiring, detailed textures, accurate lighting and shadows, 8k quality, intricate patterns, high-definition, glossy finish, vivid reflections, perfect lighting, showroom"  \\
\hline
\end{tabularx}
\caption{List of prompts used for Fig. 2 in the supplementary.}
\label{tab:fig2_supp_prompts}
\end{table}

\begin{table}[h]
\centering
\small
\begin{tabularx}{\textwidth}{|l|>{\raggedright\arraybackslash}X|}
\hline
\textbf{Figure} & \textbf{Prompt} \\
\hline
Supp. Fig. 3 $(0, 0)$ & ``A photo of a abyssal destroyed robot covered in moss, post apocalyptic city, lush overgrowth, by Luis Royo, by Greg Rutkowski, dark, gritty, intricate, volumetric lighting, volumetric atmosphere, concept art, cover illustration, octane render, trending on artstation, 8k, dynamic pose"  \\
\hline
Supp. Fig. 3 $(0, 1)$ & ``by Mattias Adolfsson and Satoshi Kon, hyper realistic medium full shot photo of a otherworldly landscape, oppulent, glamourous, masterful, poster art, bold lines, hyper detailed, expressive, award winning, dark limited color palette, high contrast, depth of field, intricate details, masterpiece, best quality, rim lighting, looking at viewer, dynamic pose, wide angle panoramic view"  \\
\hline
Supp. Fig. 3 $(0, 2)$ & ``close up photo of a rabbit, forest, haze, halation, bloom, dramatic atmosphere, centred, rule of thirds, 200mm 1.4f macro shot"  \\
\hline
Supp. Fig. 3 $(1, 0)$ & ``closeup of a circuit board city with very small futuristic cars, small flying vehicles, small robots and small humans, ultra hd, realistic, vivid colors, highly detailed, UHD drawing, pen and ink, perfect composition, beautiful detailed intricate insanely detailed octane render trending on artstation, 8k artistic photography, photorealistic concept art, soft natural volumetric cinematic perfect light"  \\
\hline
Supp. Fig. 3 $(1, 1)$ & ``Cross-Processing by Artur Amijewski and Rodelio Astudillo, award winning, kookaburra, aesthetic of symbolism with bubbling atmosphere, skybox / skydome, well-defined edges, creative tour de force with meticulous details, morganite pink and mauve colors"  \\
\hline
Supp. Fig. 3 $(1, 2)$ & ``detailed ink, pen and ink, mail art, best quality, detailed epic ice transparent ethereal otherworldly ghost castle in the blue sky, clouds, smoke, fog, detailed landscape, ghost figures, lake, boat, green forest, detailed flying dragon at the sky, detailed scales, warm lights, glittering, Craola, Dan Mumford, Andy Kehoe, 2d, flat, art on a cracked paper, patchwork, stained glass, cute, adorable, fairytale, storybook detailed illustration, cinematic, ultra highly detailed, tiny details, beautiful details, mystical, luminism, vibrant colors, complex background"  \\
\hline
Supp. Fig. 3 $(2, 0)$ & ``ethereal fantasy concept art of masterpiece, best quality, RAW macro photo of just some garbage that someone put in a box/frame . magnificent, celestial, ethereal, painterly, epic, majestic, magical, fantasy art, cover art, dreamy"  \\
\hline
Supp. Fig. 3 $(2, 1)$ & ``detailed realistic close up of a strawberry shaped like a muscular man, sitting, natural light"  \\
\hline
Supp. Fig. 3 $(2, 2)$ & ``minimalist, cinematic, movie poster, cold colors, purple theme,  standing, weapon, outdoors, sky, cloud, gun, moon, helmet, robot, ground vehicle, motor vehicle, multiple planets, planet, spacecraft"  \\
\hline
\end{tabularx}
\caption{List of prompts used for Fig. 3 in the supplementary.}
\label{tab:fig3_supp_prompts}
\end{table}

\begin{table}[h]
\centering
\small
\begin{tabularx}{\textwidth}{|l|>{\raggedright\arraybackslash}X|}
\hline
\textbf{Figure} & \textbf{Prompt} \\
\hline
Supp. Fig. 4 $(0, 0)$ & ``Once upon a time, organized theory, systematic scientific papers, elegant, figuration, sharp lines, black and white colors, beautiful, trending on artstation, volumetric lighting, by Guy Denning, no text, colorized drawings, black and white, artistic"  \\
\hline
Supp. Fig. 4 $(0, 1)$ & ``celestial promenade, the horror of stars, twilight sunshine, heaven help us, cityscape masterpiece, realistic, best quality, cosmic horror, bright horror"  \\
\hline
Supp. Fig. 4 $(0, 2)$ & ``highly detailed and hyper realistic photo, by Alena Aenami, by Archibald Thorburn, by Daniele Afferni, a breathtaking otherworldly landscape with psychedelic colors, in the style of monochromatic silhouette reflection, limited dark palette, unusual dark colors, faded colors, atmospheric haze, highly dramatic cinematic lighting, motion blur, film grain, professional, excellent composition, finest details, maximized details, ultimate detail level, masterpiece, best quality"  \\
\hline
Supp. Fig. 4 $(1, 0)$ & ``refreshing, vibrant glowing coconut juice drink, dew drops, refreshing, in the style of a product hero shot in motion, dynamic magazine ad image, photorealism, sleep and mystical elements around the background"  \\
\hline
Supp. Fig. 4 $(1, 1)$ & ``The art of Origami, Paper folding, Swan on a lake, Amazing colours, Intricate details, Painstaking Attention to Details, UHD"  \\
\hline
Supp. Fig. 4 $(1, 2)$ & ``The devil with glowing eyes ingenious opus magnum by Michael Vincent Manalo and Elvira Vigna, pastel oil painting, high key with glistening light, cosmic anddireful and dramatic atmosphere, zestful quinacridone nickel azo gold and electricblue color swatch, raw details, crisp details, bokeh lights, analog photo, full body, tall height, against the background volcano mouth in the background, blue sky, epic clouds, oil on canvas, art by Jeremy Mann, god rays, dramatic light, art by Eduard Wilhelm Pose and Master of the Holy Blood and Ivan Bilibin, oxygen-rich air and sheltered atmosphere, snowy, aesthetic of hard - edge painting, symmetry and balance, pioneering unparalleled masterwork with superior details, skin texture, masterpiece, top quality, best quality, official art, highest detailed, atmospheric lighting, cinematic composition, complex multiple subjects, 4k HDRvaporwave style, cyberpunk, vibrant, neon colors, highly detailed, Leica Q2 with Summilux 35mm f/1.2 ASPH, clear face, Ultra High Resolution, wallpaper, 8K, Rich texture details, hyper detailed, detailed eyes, detailed background, dramatic angle"  \\
\hline
Supp. Fig. 4 $(2, 0)$ & ``In a lush, tropical rainforest, a vibrant, exotic bird like a parrot perches amidst a kaleidoscope of flowers, with dappled sunlight filtering through the dense canopy. This scene is captured in the style of Marco Lumiere, known for his vivid and lively color palettes and a slightly impressionistic touch, highlighting the vitality of nature and the majestic beauty of the bird in its natural habitat."  \\
\hline
Supp. Fig. 4 $(2, 1)$ & ``wooden arch bridge, river, crowd, mount fuji, boats, cherry tree, scenery, outdoors, best quality, highly detailed"  \\
\hline
Supp. Fig. 4 $(2, 2)$ & ``zentangle Flickering grass beneath a lone smart light bulb that casts an eerie glow in the otherwise serene sunlit meadow, as if waiting for something to emerge from the shadows. . intricate, abstract, monochrome, patterns, meditative, highly detailed"  \\
\hline
\end{tabularx}
\caption{List of prompts used for Fig. 4 in the supplementary.}
\label{tab:fig4_supp_prompts}
\end{table}

\clearpage

\textbf{Base models and refiners:}
\begin{itemize}
    \itemsep-1em 
    \item AetherverseLightning v10 \\
    \item AlbedobaseXL v13 \\
    \item AnimagineXL v30 \\
    \item AnythingXL \\
    \item crystalClearXL ccXL \\
    \item DreamshaperXL turboDpmppSDEKarras \\
    \item EnvyhyperdriveXL v10 \\
    \item GleipnirV0.3 \\
    \item JibMixXL v9 ``BetterBodies" \\
    \item JuggernautXL v9 Rdphoto2Lightning \\
    \item LeosamsHelloworldXL v70 \\
    \item Proteus v03 \\
    \item RealismEngineSDXL v10 \\
    \item RealvisXL v40 BakedVAE \\
    \item RealvisXL v40 LightningBakedVAE \\
    \item SDXL Base 1.0 0.9VAE \\
    \item SDXL Base 1.0 \\
    \item SDXL Refiner 1.0 0.9VAE \\
    \item SDXL Refiner 1.0 \\
    \item SDVN7 - NijiStyleXL v1 \\
    \item SSD-1B \\ 
    \item TurbovisionXL SuperFastXL V431BakedVAE \\
    \item Stable Cascade \\
    \item Pixart-$\Sigma$ \\
\end{itemize}

\textbf{LoRAs and embeddings:}
\begin{itemize}
    \itemsep-1em 
    \item easynegative \\
    \item bad-hands-5 \\
    \item nfixer \\
    \item Add-Detail XL \\
    \item EpicF4nta5yXL \\
    \item AnimeTarot \\
    \item JuggerCineXL2 \\
    \item LCM LoRA SSD-1B \\
    \item LCM LoRA SDXL \\
    \item LogoRedmond \\
    \item MJ52 v2.0 \\
    \item MJ52 \\
    \item PerfectEyesXL \\
    \item Pixel-Art-XL v1.1 \\
    \item Ral-Dissolve-SDXL \\
    \item SDXL Glass \\
    \item SDXLFaetastic v24 \\
    \item Sinfully Stylish SDXL \\
    \item Werewolf SDXL \\
    \item WowifierXL v2 \\
    \item XL more art-full-beta1 \\
\end{itemize}

\textbf{Super resolution and face restoration models:}
\begin{itemize}
    \itemsep-1em 
    \item 4x NMKD Superscale - SP 178000 G \\
    \item 4x UltraSharp \\
    \item RealESRGAN x2 plus \\
    \item codeformer \\
    \item GFPGAN v1.4 \\
\end{itemize}

\textbf{VAEs:}
\begin{itemize}
    \itemsep-1em 
    \item SharpSpectrum VAEXL \\
    \item SDXL VAE fp16 fix \\
    \item SDXL VAE
\end{itemize}

